\documentclass[letterpaper,10pt]{article}

\usepackage{probml}

\ShortHeadings{CogFormer}

\usepackage{blindtext}
\usepackage{wrapfig}
\usepackage{booktabs}
\usepackage{multirow}
\usepackage{tabularx}
\usepackage{algorithm}
\usepackage{hyperref}
\usepackage{algpseudocode}
\newcolumntype{C}{>{\centering\arraybackslash}X}

\usepackage{bm,amssymb,amsmath}

\makeatletter
\def\th@plain{%
  \thm@notefont{}
  \itshape 
}
\def\th@definition{%
  \thm@notefont{}
  \normalfont 
}
\makeatother

\def\1{\bm{1}}

\def\rvtheta{{\boldsymbol{\theta}}}

\def\rvs{{\mathbf{s}}}

\def\rvw{{\mathbf{w}}}
\def\rvx{{\mathbf{x}}}
\def\rvy{{\mathbf{y}}}

\def\rmB{{\mathbf{B}}}

\def\rmM{{\mathbf{M}}}

\def\rmS{{\mathbf{S}}}

\def\rmX{{\mathbf{X}}}
\def\rmY{{\mathbf{Y}}}
\def\rmZ{{\mathbf{Z}}}

\DeclareMathAlphabet{\mathsfit}{\encodingdefault}{\sfdefault}{m}{sl}
\SetMathAlphabet{\mathsfit}{bold}{\encodingdefault}{\sfdefault}{bx}{n}

\begin{document}

\title{
  CogFormer: Learn All Your Models Once
}

\author[1,$\dagger$]{Jerry M. Huang}
\author[2]{Lukas Schumacher}
\author[3]{Niek Stevenson}
\author[1]{Stefan T. Radev}

\affil[1]{
    Center for Modeling, Simulation, and Imaging in Medicine (CeMSIM)\\
    Rensselaer Polytechnic Institute\\
    Troy, New York, United States
}

\affil[2]{
    Center for Economic Psychology\\
    University of Basel\\
    Basel, Switzerland
}

\affil[3]{
    Amsterdam Mathematical Psychology Lab\\
    University of Amsterdam \\
    Amsterdam, Netherlands
}

\affil[$\dagger$]{Correspondence to \url{huangm13@rpi.edu}}

\maketitle

\begin{abstract}
Simulation-based inference (SBI) with neural networks has accelerated and transformed cognitive modeling workflows.
SBI enables modelers to fit complex models that were previously difficult or impossible to estimate, while also allowing rapid estimation across large numbers of datasets.
However, the utility of SBI for iterating over varying modeling assumptions remains limited: changes to parameterizations, generative functions, priors, and design variables all necessitate model retraining, thereby diminishing the benefits of amortization.
To address these issues, we pilot the CogFormer, a meta-amortized framework for cognitive modeling.
Our framework trains a transformer-based architecture that remains valid across a combinatorial number of structurally similar models, allowing for changing data types, parameters, design matrices, and sample sizes.
We present promising quantitative results across families of decision-making models for binary, multi-alternative, and continuous responses. Our evaluation suggests that CogFormer can accurately estimate parameters across model families with minimal amortization offset, making it a potentially powerful engine that catalyzes cognitive modeling workflows.
\end{abstract}

\section{Introduction}
\label{sec:intro}

Neural simulation-based inference \citep[SBI;][]{cranmer2020frontier, deistler2025simulation} employs neural networks trained on simulated data to avoid iterative estimation. In particular, \textit{amortized methods} \citep{burkner2023some} learn from simulated parameter–observation pairs $(\rvtheta, \rvy)$, enabling rapid sampling from the posterior $p(\rvtheta \mid \rvy^*)$ for any new observation $\rvy^*$ without retraining.
This advantage has led to a plethora of SBI applications in cognitive modeling \citep[e.g.,][]{radev2020amortized, wieschen2020jumping, fengler2021likelihood, boelts2022flexible, von2022mental, ghaderi2023general, sokratous2023ask, schumacher2023superstat, schumacher2024validation, rmus2024artificial, hato2025, schaefer2025conflictModels, scholten2026diffusionLens, kvam2024using, kvam2025comparing, belov2026supervised, wu2026testing}.

Cognitive modeling is concerned with building probabilistic models that connect latent constructs $\rvtheta$ (e.g., processing speed) to observable responses $\rvy$ (e.g., reaction times).
These models are typically \textit{generative}; that is, they define a recipe for generating synthetic data from parameters in the form of a conditional distribution $p(\rvy \mid \rvtheta)$. However, the generative mapping does not typically preserve all information about $\rvtheta$, so inference needs to represent uncertainty via the posterior $p(\rvtheta \mid \rvy)$. Since posteriors of cognitive models are rarely tractable, inference often relies on time-consuming iterative Monte Carlo methods \citep[e.g.,][]{stricklandRacingRemember2018, mileticParameterRecovery2017}, further underscoring the advantage of amortized SBI in its efficiency and flexibility. 

In many ways, cognitive models are ideal generators of synthetic data: they are easy to implement, fast to simulate (often in the millisecond range), and low-dimensional.
Yet they present unique challenges for SBI.
In practice, inference rarely involves fitting only a single model.
Instead, we fit multiple candidate models, run many iterative cycles of free vs.~fixed parameter configurations, regress parameters on covariates, and generally reconfigure the model as theory and data negotiate \citep{voss2019sequential, evans2019assessing, stevensonBayesianHierarchical2026}.
This dynamic process can be packaged into principled \textit{Bayesian workflows} \citep{gelman2020bayesian, schad2021toward, li2024amortized}.

For example, when modeling response times with Sequential Sampling Models \citep[SSM;][see also~\autoref{app:decision-models}]{smith2025ssmBible}, a researcher might compare variants with a fixed decision threshold to ones with a collapsing threshold that decreases over time, repeatedly refit the model while constraining or freeing parameters, or test whether an experimental manipulation affects a single parameter (e.g., drift rate) or multiple components of the decision process \citep{von2022mental}.
This creates a dilemma for SBI workflows, since model changes typically demand re-training the network, steadily eroding the very amortization benefits that made SBI attractive to begin with.

The current work builds on a stream in SBI that extends amortization \citep{chang2025amortized, gloeckler2024all, elsemuller2023sensitivity, schroder2023simultaneous}: 
training a single inference network that remains valid across many, potentially infinitely many, structurally similar models. 
We dub our architecture the \textbf{CogFormer}, directly inspired by the Simformer \citep{gloeckler2024all}.
CogFormer can adapt to any design configurations within a model's combinatorial parameter space and infer latent parameters accordingly, without substantially compromising inference quality compared to single-model posterior approximation. 
In this work, we taxonomize the scope of amortization in cognitive modeling and demonstrate compelling results from the first meta-amortized inference engine designed for amortizing cognitive models.

\section{Methods}
\label{sec:math}

\subsection{Amortization scopes: from models to classes}
\label{ssec:scopes}

\begin{figure}[t]
    \centering
    \includegraphics[width=\linewidth]{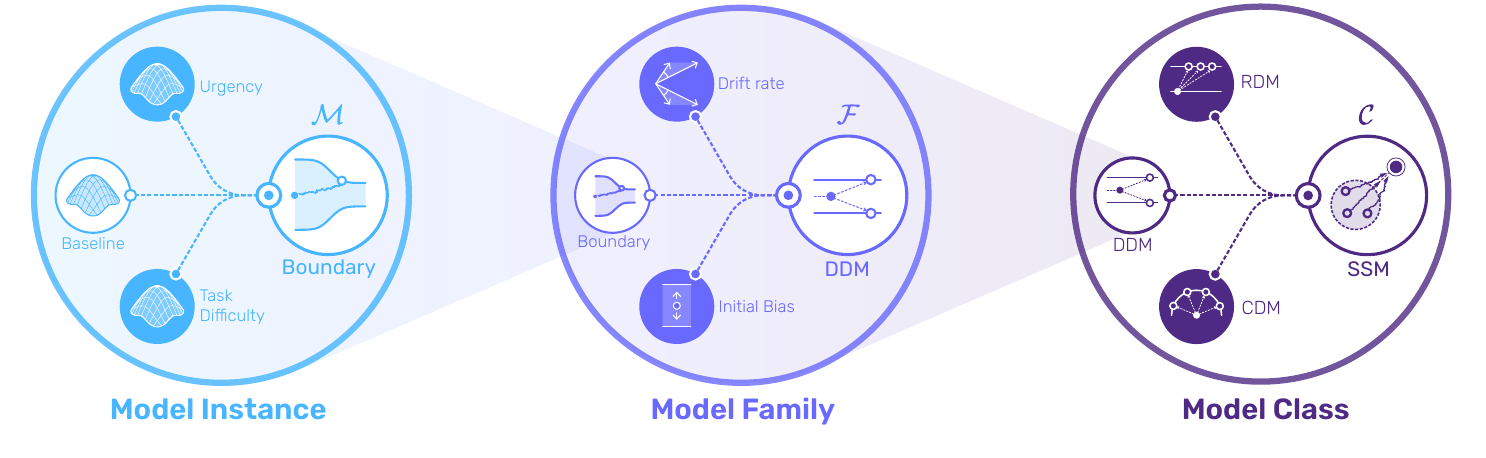}
    \caption{\textbf{Taxonomy of amortization scopes}. In cognitive modeling, one fits concrete model instances $\mathcal{M}$ nested within model families $\mathcal{F}$ as part of a larger model class $\mathcal{C}$. Here, we train neural estimators that generalizes across all model families within a given model class (in this paper, Sequential Sampling Models).}
    \label{fig:levels}
\end{figure}

To formalize the scope of amortization in cognitive modeling, we introduce a hierarchical taxonomy that organizes models by structural similarity.
This taxonomy specifies the level at which inference can be amortized and clarifies the scope of generalization for a given neural estimator.
We distinguish three nested levels: \textbf{model instance}, \textbf{model family}, and \textbf{model class} (see also \autoref{fig:levels}), where each level contains multiple instances of the level below.

A \textbf{model instance} $\mathcal{M}$ is at the most specific level, fully specified with fixed parameters and likelihood. All structural aspects, including the number of parameters, their interpretation, and the form of the generative process, are fixed.
This has been the standard setting for SBI in cognitive modeling, for which a neural estimator is trained to perform amortized inference for a single $\mathcal{M}$.

A \textbf{model family} $\mathcal{F}$ comprises multiple specifications sharing the same computational structure but differing in parameterization or auxiliary assumptions.
For example, variants of the classic diffusion decision model \citep[DDM;][]{ratcliff1978diffusionModel, ratcliffDiffusionModel2004} may differ in collapsing vs.~fixed boundaries \citep{bogaczPhysicsOptimal2006, ditterichEvidenceTimevariant2006, rasanan2025collapsing}, starting-point variability, non-decision time structure, or regressors on parameters.
Importantly, these specifications are typically nested: simpler variants can be recovered from more general ones by fixing certain parameters to zero or other constant values.
In theory, a single network can thus perform inference across all such variants.

A \textbf{model class} $\mathcal{C}$ groups distinct model families that implement a shared computational principle.
For instance, in decision-making, families such as the DDM, racing diffusion model \citep[RDM;][]{tillman2020racingDiffusion, zandbeltResponseTimes2014}, and circular diffusion model \citep[CDM;][]{smith2016circularDiffusion, rasananDiscretechoiceOptions2024} form a broader class of sequential sampling models (SSMs).
These models are generally not nested and cannot be obtained from one another through simple parameter constraints.
Therefore, amortization at this level requires generalization across structurally distinct families while still exploiting shared features, such as common target observables.

This paper demonstrates family- and class-level amortization, with future work aimed at even broader amortization scopes. Accordingly, we fix a model class $\mathcal{C}$ as the root and consider the following generative meta-model, sampled top-down:
\begin{equation}\label{eq:generative}
  p(\mathcal{F},\mathcal{M},\mathcal{D},\theta_{\mathcal{F}},\rmY \mid \mathcal{C})
  = p(\mathcal{F}\mid\mathcal{C})\,
      p(\mathcal{M},\mathcal{D}\mid\mathcal{F})\,
      p(\theta_{\mathcal{F}}\mid\mathcal{M},\mathcal{D})\,
      p(\rmY\mid\theta_{\mathcal{F}},\mathcal{M},\mathcal{D}).
\end{equation}
\noindent where $\mathcal{D}$ denotes a design configuration for the set of intrinsic parameters $\theta_\mathcal{F}$ and $\rmY$ is model observables. The next sections describe how the abstract model ($\mathcal{M}$) and design ($\mathcal{D}$) variables are represented numerically through (embeddings of) positional encodings and design matrices and processed by our CogFormer architecture to arrive at an amortized posterior $p(\theta_{\mathcal{F}} \mid \rmY, \mathcal{M}, \mathcal{D})$.

\subsection{Meta-simulator and generative hierarchy}
\label{sec:meta-simulator}

\begin{figure}
    \centering
    \includegraphics[width=0.99\linewidth]{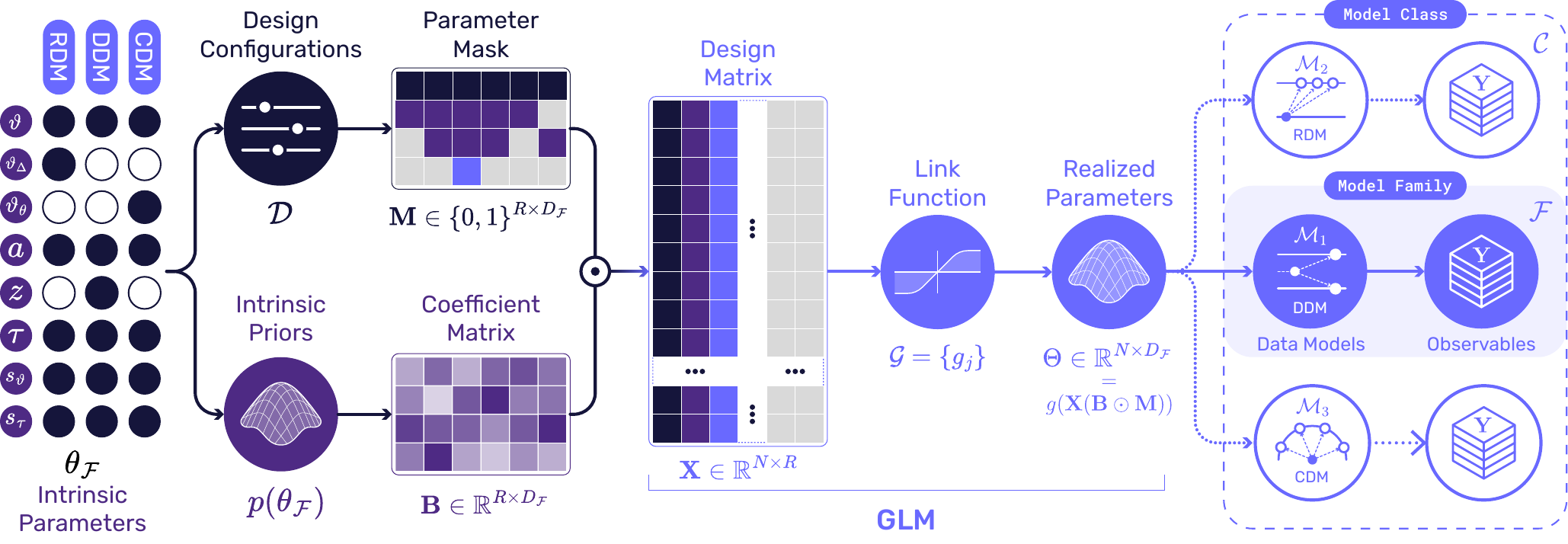}
    \caption{\textbf{Meta-simulator architecture.} For a single model family (\textit{middle}), a coefficient matrix $\rmB$ is constructed under a flexible generalized linear model (GLM) configuration, and configured through a random binary parameter mask $\rmM$. Together with a sampled design matrix $\rmX$ and element-wise link functions $g$, this construction induces the trial-level parameter matrix $\Theta$, which is passed to the observation model $\mathcal{M}$ to generate observables $\rmY$. The same generative process applies to every model in the class $\mathcal{C}$.}
    \label{fig:meta-simulator}
\end{figure}

To realize the generative hierarchy in \cref{eq:generative}, we implement a meta-simulator that subsumes a broad class of cognitive models under a flexible generalized linear model (GLM) parameterization (see~\autoref{fig:meta-simulator}; see also~\autoref{alg:model-family} and~\autoref{alg:model-class} for implementation details). 
Each model family $\mathcal{F}$ is characterized by $D_{\mathcal{F}}$ intrinsic parameters $\theta_{\mathcal{F}}$. 
For example, the DDM has six intrinsic parameters: drift rate $\vartheta$, threshold $a$, bias $z$, non-decision time $\tau$, and variabilities $s_{\vartheta}, s_\tau$. We express each intrinsic parameter as a linear function of covariates $x_c$, for example, $\vartheta = \beta_{0, \vartheta} + \beta_{1, \vartheta}\,x_1 + \beta_{2, \vartheta}\,x_2$. 
All regression weights are assembled into a coefficient matrix $\rmB \in \mathbb{R}^{I \times D_{\mathcal{F}}}$, where each column corresponds to one intrinsic parameter and each row to one column of the design matrix (intercept first, followed by regression weights).
The intercepts are drawn from the intrinsic prior $p(\theta_{\mathcal{F}})$ while the regression weights are drawn from $\mathcal{N}(0, 1)$. 
A set of link functions $\mathcal{G} = \{g_j\}$ (e.g., sigmoid) maps the resulting linear predictors to the valid domain of each intrinsic parameter.        

The generative process for a single dataset $\rmY \in \mathbb{R}^{N \times O_{\mathcal{F}}}$ of $N$ trials and $O_{\mathcal{F}}$ observables is thus:                                                   
  \begin{equation}
  \rmY \sim \mathrm{Model}(\Theta;\,\mathrm{RNG}), \quad
  \Theta = g\!\left(\rmX\,(\rmM \odot \rmB)\right)
  \end{equation}
where $\Theta \in \mathbb{R}^{N \times D_{\mathcal{F}}}$ is the realized parameter matrix, distinct from the intrinsic parameter vector $\theta_{\mathcal{F}}$, and $\mathrm{RNG}$ denotes the source of stochastic variation in the model outputs.
The design matrix $\rmX \in \mathbb{R}^{N \times I}$ encodes the experimental structure across $N$ observations and $I$ design columns (i.e., regressors).
Each $\rmX$ is sampled from a prior over design configurations. 
Its first column is an intercept (all ones), followed by one block of $C_\mathrm{max} - 1$ columns per regressor: continuous regressors occupy only the first column of their block; categorical regressors use all columns as dummy codes. 
Interaction terms are generated via column-wise multiplication. The parameter mask $\rmM \in \{0,1\}^{I \times D_{\mathcal{F}}}$, sharing the shape of $\rmB$, distinguishes active ($1$) from fixed ($0$) coefficients.
During training, each entry $M_{i, j}$ is sampled from a Bernoulli distribution with a random probability parameter $p \sim \mathrm{Beta}(2, 2)$, enabling the model to generalize to a power set of nested design configurations; during validation, $\rmM$ is fixed to pre-defined benchmark designs (see~\autoref{app:design-config}). 
  
Since $N$ and $I$ vary across simulations, $\rmX$ is zero-padded to a maximum capacity $I_\mathrm{max}$, reflecting the pragmatic limit on regressors in a realistic analysis. To avoid static padding, the number of trials $N$ is held constant within a batch but varies across batches. Crucially, the parameter mask $\rmM$, design matrix $\rmX$, and coefficients $\rmB$ are all resampled for each individual simulation within a batch. Finally, to achieve amortization over model families, a model family index $f \sim p(f \mid \mathcal{C})$ is sampled for each simulation.  

\vspace{-1em}
\subsection{Generative encoder-decoder architecture}
\label{sec:backbone}

\paragraph{Flow matching for expressive posterior estimation}
\label{par:flow-matching}

Let $\mathcal{A} = (\rmY, \rmX, \rmM, f)$ be a placeholder for all conditioning variables defining the amortization scope; equivalently, $\mathcal{A}$ collects the data $\rmY$, the design $\mathcal{D} = (\rmX, \rmM)$, and the model family index $f$.
The goal of our framework is to estimate the full joint posterior $p(\rmB \mid \mathcal{A})$ without making restrictive parametric or autoregressive assumptions. To achieve this flexibility, we leverage \textit{conditional flow matching} \citep{lipman2023flow, liu2023flow}. This choice is motivated by recent benchmarks in SBI suggesting that flow matching is a highly competitive free-form neural sampler \citep{arruda2025diffusion}. Notably, our architecture is fully compatible with other free-form inference methods, such as score-based diffusion.

Intuitively, flow matching learns to transport a simple distribution, typically a Gaussian latent $\rmZ_1 \sim \mathcal{N}(\mathbf{0}, \mathbf{I})$, into the complex target posterior $\rmZ_0 \equiv \rmB \sim p(\rmB \mid \mathcal{A})$ by learning a time-dependent conditional velocity field $\hat{u}(\rmZ_t; \mathcal{A}, t)$. The velocity field defines an ordinary differential equation (ODE) with boundary conditions so that $\rmZ_1$ is noise and $\rmZ_0$ is the target coefficient matrix. The model then learns an approximate velocity field $\hat{u}$ by minimizing the flow matching objective:
\begin{equation}\label{eq:flow-matching-loss}
  \mathcal{L}_{\text{FM}} =
  \mathbb{E}_{p(t)\,p(\rmB, \mathcal{A})\,p(\rmZ_1 \,\mid\,\rmZ_0)}
  \left[\, \omega_t \, \Vert \rmM \odot \bigl(\hat{u}(\rmZ_t; \mathcal{A}, t) - (\rmZ_1 - \rmB)\bigr) \Vert_{2}^2 \,\right]
\end{equation}
where $\omega_t$ is an optional weight and $\rmZ_t = (1-t)\rmB + t\rmZ_1$ defines a linear probability path \citep{liu2023flow}. The expectation is taken over a uniform time distribution, the meta-simulator, and the Gaussian latent distribution of $\rmZ_1$; the mask $\rmM$ in the norm operator $\Vert \cdot \Vert$ ensures that only active coefficients contribute to the loss. At inference time, we generate samples from the approximate posterior $q(\rmB \mid \mathcal{A}, \hat{u})$ by integrating the learned velocity field from $t=1$ to $t=0$ using any off-the-shelf ODE solver. In our experiments, we use a simple Euler scheme with $\Delta t = 10^{-3}$, but various error-correcting schemes are also viable \citep{arruda2025diffusion}.

\begin{figure}[t]
    \centering
    \includegraphics[width=\linewidth]{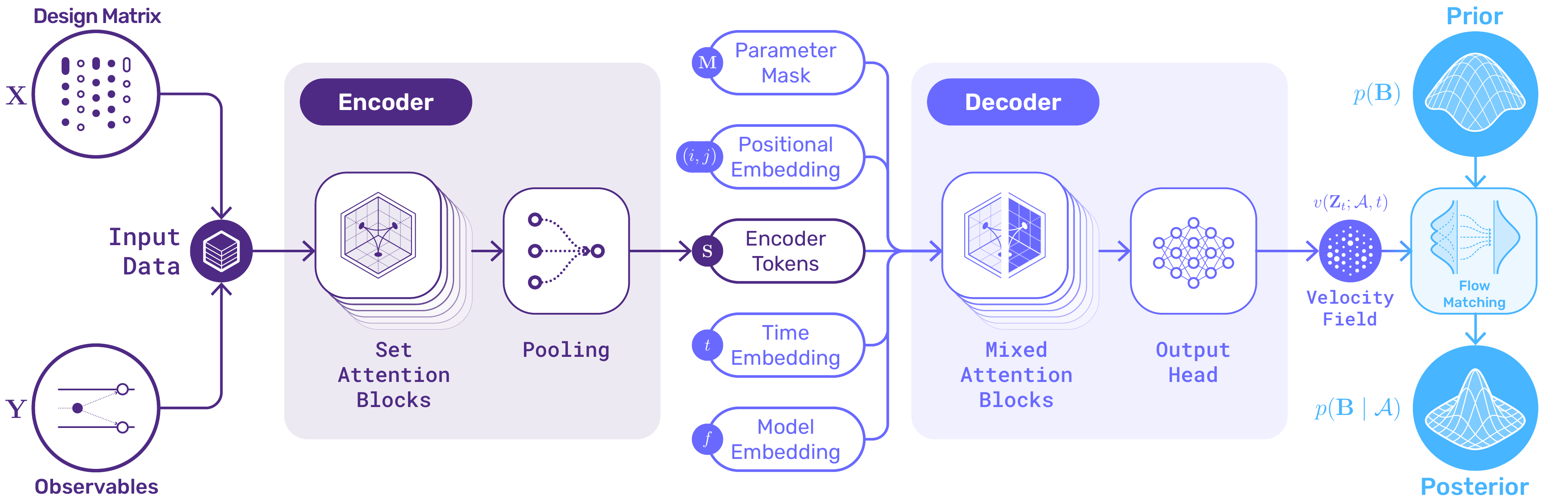}
    \caption{\textbf{CogFormer architecture.} CogFormer is an encoder-decoder architecture in which the encoder is a Set Transformer that converts a design matrix $\rmX$ and model observables $\rmY$ into summary representations $\rmS$, and the decoder is a stack of mixed-attention layers that produce a learned velocity field $\hat{u}$. The latter parameterizes the learned velocity field $\hat{u}$. Training minimizes a conditional flow-matching objective, yielding a procedure for sampling from the resulting ensemble posterior.}
    \label{fig:transformer}
\end{figure}

\paragraph{Transformer backbone}
\label{par:transformer}

To realize the conditional velocity field $\hat{u}$, we develop a transformer-based encoder–decoder architecture (see~\autoref{fig:transformer}; further details about the individual attention blocks can be found in~\autoref{fig:cogformer-attentions}). The architecture is dictated by an asymmetry in our problem: the observations are exchangeable, while the parameters are not. CogFormer treats each side accordingly.
  
The \textbf{encoder} is a modernized version of the Set Transformer \citep{lee2019set} that encodes the design matrix $\rmX$ and the model observables $\rmY$ into a sequence of $L$ learned summary representations, $\rmS = (\rvs_1,\dots,\rvs_L)$ with $\rvs_{\ell} \in \mathbb{R}^S$. Since the trial-level pairs $(\rvx_n,\rvy_n)$ are exchangeable, the posterior is invariant to joint permutations of the rows of $(\rmX,\rmY)$; a set-based encoder bakes this invariance in by construction rather than forcing the network to learn it. As a result, the permutation-invariant encoder can handle varying trial counts and experimental designs.
  
The \textbf{decoder} maps the encoder summaries $\rmS$ to the velocity field $\hat{u}$ of the parameters $\rmB$. Unlike the observations, the parameters are not exchangeable: each entry of $\rmB \in \mathbb{R}^{I \times D_{\mathcal{F}}}$ plays a distinct role, and their posterior correlations are themselves objects of inference. This calls for structured sequence attention rather than a set operation. A unique feature of our architecture is that it unrolls the coefficient matrix $\rmB$ into a 1D sequence of $I \times D_{\mathcal{F}}$ tokens, which is what allows us to work with both varying numbers of columns (intrinsic parameters) and rows (regressors).

To restore the spatial layout of the GLM coefficient matrix from the unrolled $\rmB$, each token is augmented with sinusoidal embeddings of the positional encoding triplet $(i, j, f)$: the regressor index $i$, the intrinsic parameter index $j$, and the model family index $f$. The first two coordinates re-impose the grid structure a flat sequence would destroy, while $f$ tells each parameter which model family it belongs to, so that semantically identical positions are not conflated across families. The binary parameter mask $\rmM$ is also provided as a query mask, allowing the network to distinguish active from fixed parameters.

Information in the decoder flows through mixed-attention blocks, each of which alternates between a cross-attention layer over the encoder summaries $\rmS$ and a self-attention layer over the tokens formed by concatenating $\rmZ_t$ with the positional embeddings. The use of both types of attention layers addresses the two things a posterior must do at once: cross-attention conditions each parameter on the data summary, while self-attention propagates information across the parameter grid to capture their joint dependencies. Within each decoder block, we use a residual network as the output head to fuse these inputs. Although FiLM \citep{perez2018film} was found to improve conditional flow matching in previous SBI work \citep{wildberger2023flow, arruda2025diffusion}, it did not result in significant improvements in our meta-amortized setting (see results in~\autoref{app:ablations}). 

\begin{figure}[t]
    \centering
\includegraphics[width=\linewidth]{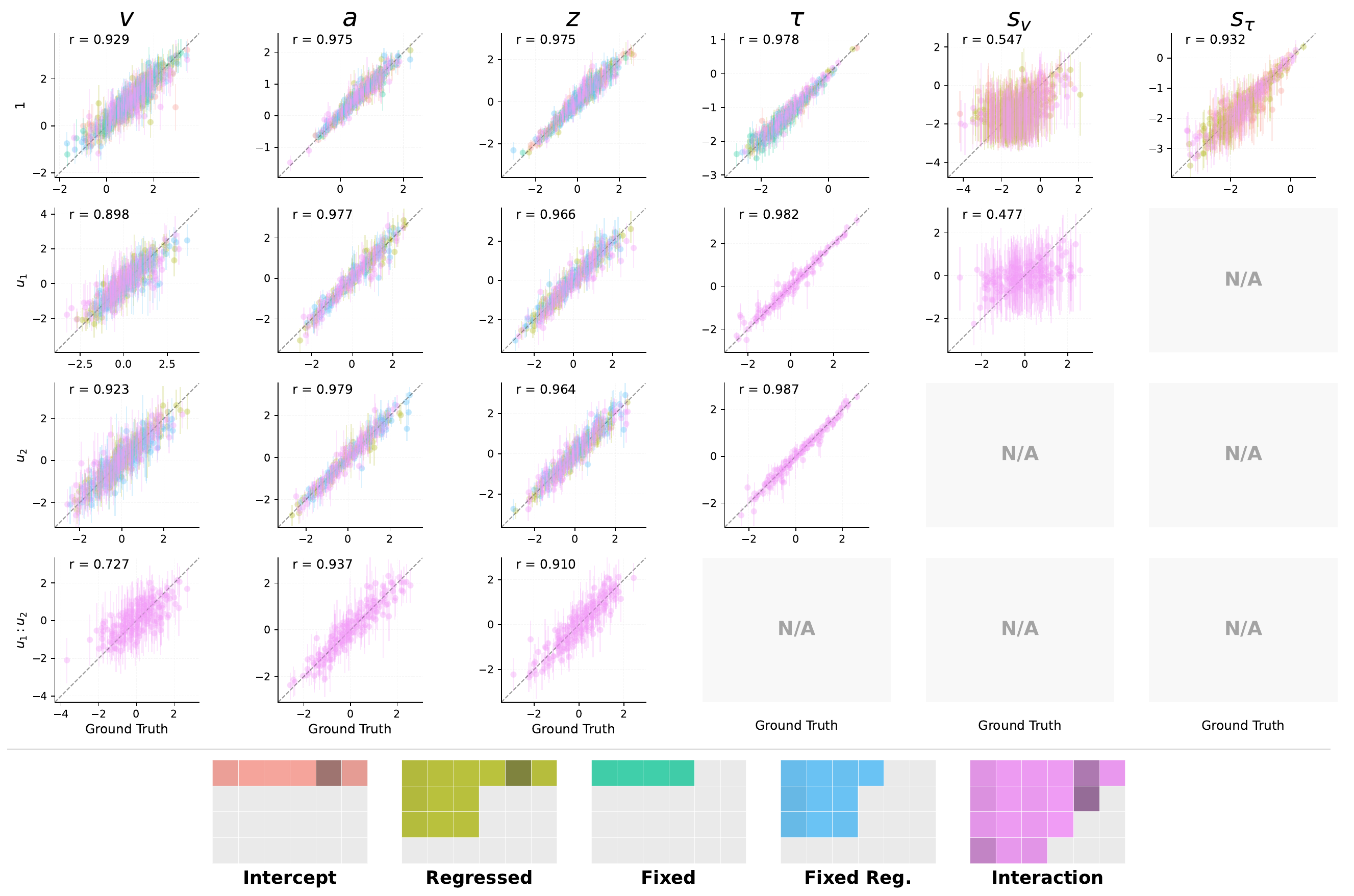}
    \caption{Ensemble parameter recoveries for CogFormer-$\mathcal{F}$ applied to the five design configurations for the DDM benchmark cases overlaid in their corresponding mapping to the shared coefficient matrix. The pixel thumbnails indicate the active estimation targets for each design configuration, with lighter shades of color depicting higher Pearson correlations $r$. The per-cell correlations are averaged across configurations.}
    \label{fig:mf-ensemble-recovery}
\end{figure}

\section{Related work}
\label{sec:related}

Meta-amortization has earlier roots in variational settings. For example, MetaVAE \citep{wu2020meta} proposed doubly amortized inference across a family of related probabilistic models via a MetaELBO objective, but it was not designed for flexible posteriors in SBI. Within SBI, earlier work considered simultaneous inference both over discrete model indices \citep{radev2020amortized} and over discrete models and continuous parameters for compositional simulators \citep{schroder2023simultaneous}.

A separate line of work moves toward multi-query SBI, in which a single amortized artifact can answer many conditional inference queries at test time. For instance, SimFormer \citep{gloeckler2024all} trains a transformer-based diffusion model on the joint distribution of parameters and observations and uses masking and attention structure to sample arbitrary conditionals, including posteriors and likelihoods. OneFlowSBI \citep{nautiyal2026oneflowsbi} pursues a similar goal with flow matching: it learns a vector field over the joint state space and realizes posterior, likelihood, and mixed conditionals. \citet{reuter2025can} amortizes over different statistical model families (GLMs, latent factor models) within a single in-context learning framework. These primarily expand query flexibility within a single learned joint model and serve as inspiration for our masking approach. In contrast, we shift the emphasis from many conditionals of a single joint distribution to a flexible posterior estimator that spans a power set of structural model configurations.

A complementary approach pursues amortization across different tasks, such as predictive and posterior estimation. For instance, ACE \citep{chang2025amortized} uses transformers to flexibly reinterpret latent and observed variables at training and test time, enabling both runtime prior injection and continuous and discrete targets via parametric heads. Closest in spirit is SA-ABI \citep{elsemuller2023sensitivity}, which formalizes sources of sensitivity, i.e., priors and data model assumptions, as context variables and uses weight sharing and deep ensembles to reduce refitting in sensitivity analysis. Similarly, the Distribution Transformer \citep{whittle2025distribution} maps prior distributions directly to (time-varying) posteriors via learnable Gaussian mixtures. In our IID cognitive-modeling settings, we take meta-amortization one step further by substantially expanding generalization to combinatorial design spaces and learning an expressive neural estimator without parametric assumptions. Furthermore, we conduct a systematic and multidimensional analysis of amortization gaps.

\begin{figure}[t]
    \centering
    \includegraphics[width=\linewidth]{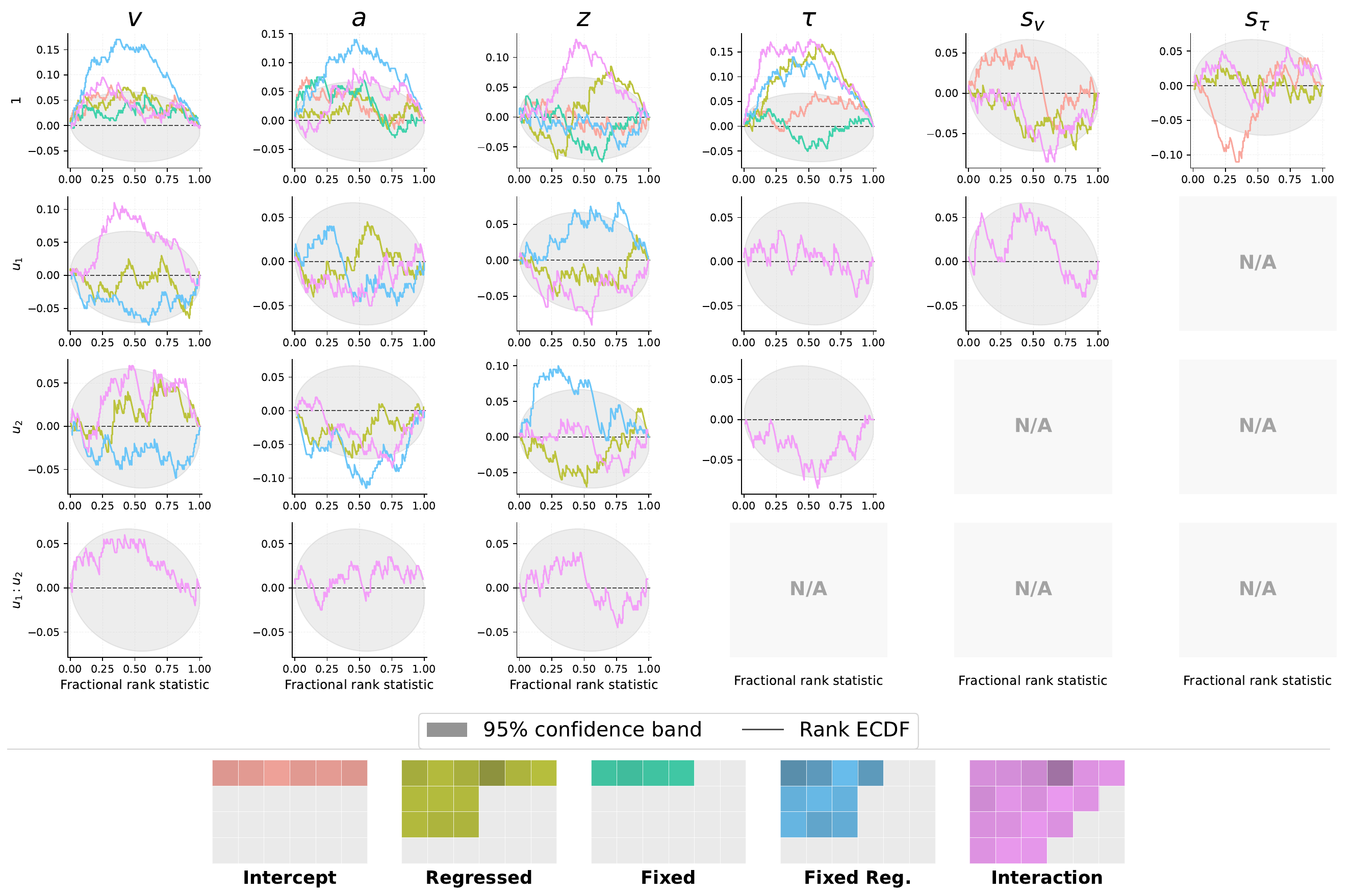}
    \caption{Ensemble calibration ECDFs for CogFormer-$\mathcal{F}$ applied to five design configurations for the DDM benchmark cases overlaid in their corresponding mapping to the coefficient matrix. The pixel thumbnails (bottom row) indicate the active estimation targets for each design configuration in the shared coefficient matrix, with lighter shades depicting lower calibration errors (ECEs).}
    \label{fig:mf-ensemble-calibration}
\end{figure}

\vspace{-1em}
\section{Experiments}
\label{sec:experiments}

\paragraph{Meta-model}
\label{sec:model-families}
Our experiments focus on SSMs, a widely used class of decision-making models that capture the dynamics of information processing underlying both choices and response times via stochastic differential equations \citep{smith2025ssmBible}.
Within this model class, we evaluate our framework on three representative model families: the DDM, the RDM, and the CDM. Detailed formulations, parameterizations, and priors for each model are provided in \autoref{app:decision-models}. We train CogFormer with meta-simulators both for each model family (CogFormer-$\mathcal{F}$) and for the model class as a whole (CogFormer-$\mathcal{C}$).

These models were selected for several reasons.
First, they have been extensively validated in cognitive neuroscience and psychology \citep{ratcliffDiffusionDecision2008, voss2004validation, forstmann2016sequential, smith2020, tillman2020racingDiffusion}, providing a well-established benchmark for inference methods.
Second, despite differences in their generative structure, they share core parameters such as drift rates, boundary separation, and non-decision times.

At the same time, the three models differ in key aspects, including response format, dimensionality, and internal dynamics, making them sufficiently diverse to test generalization.
Further, they span the spectrum of response types commonly studied in decision-making: the DDM models binary choices, the RDM multi-alternative responses, and the CDM continuous responses.

\begin{figure}[t]
    \centering
    \includegraphics[width=\linewidth]{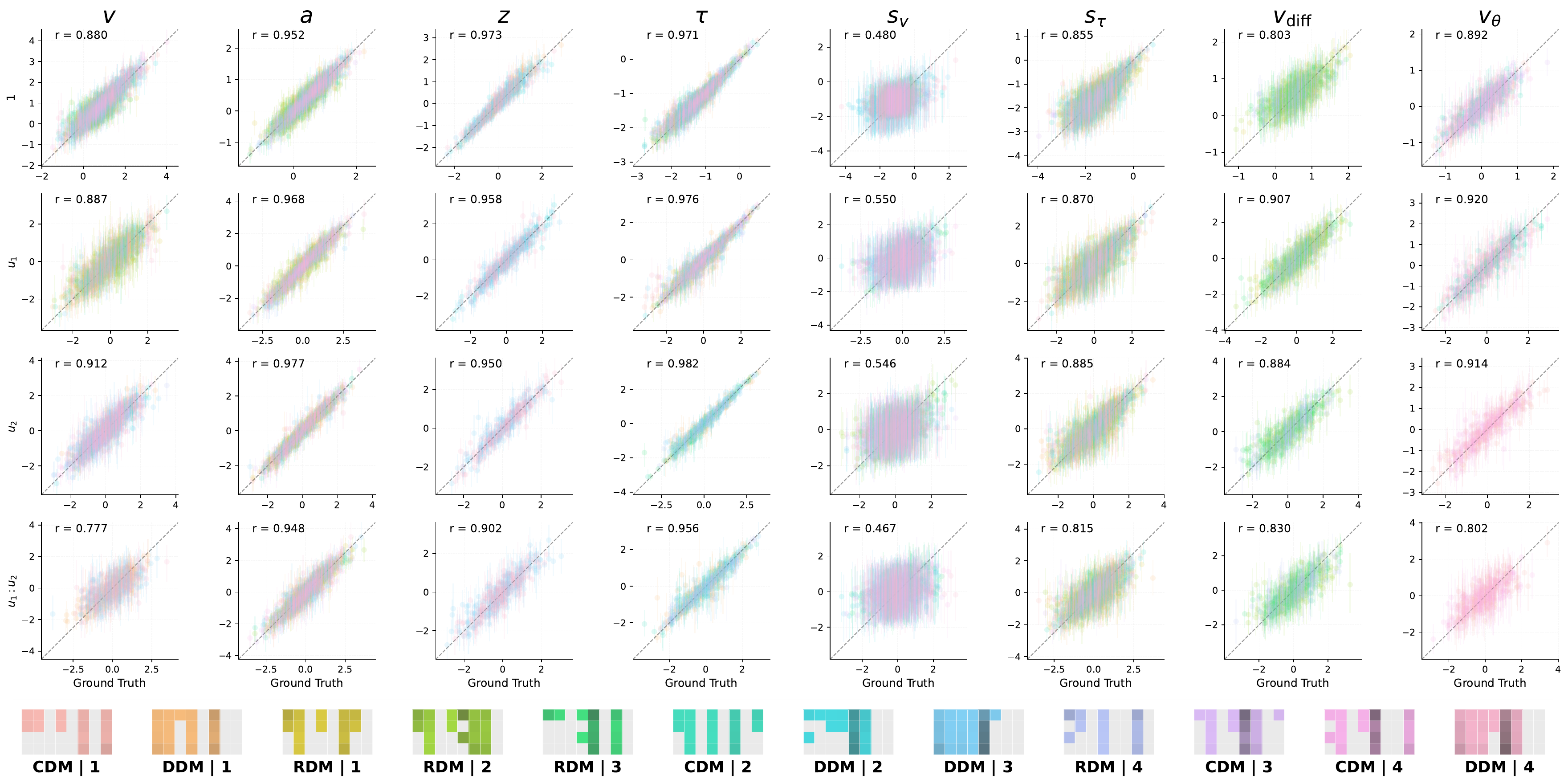}
    \caption{Ensemble parameter recoveries for CogFormer-$\mathcal{C}$ for 2,400 simulated datasets with 500 trials each over 12 randomly generated design configurations. The pixel thumbnails at the bottom indicate the active estimation targets within each configuration, with lighter shades of color indicate higher Pearson correlation $r$. Results are overlaid on top of each other, and the per-cell values are averaged across configurations.}
    \label{fig:mc-ensemble}
\end{figure}

\paragraph{Evaluation metrics} 
\label{par:eval-matrics}
We evaluate the quality of the trained amortized estimators using multiple complementary metrics (see also~\autoref{app:eval-metrics} for more information).
In particular, we assess posterior quality and the resulting amortization gap across progressively broader levels of data model amortization: instance ($\mathcal{M}$; baseline), family ($\mathcal{F}$), and class ($\mathcal{C}$).
Posterior quality is evaluated using comprehensive diagnostics from the Bayesian workflow \citep{schad2021toward}.
Specifically, we report (i) normalized root mean square error (NRMSE) between true data-generating parameters and posterior samples, (ii) expected calibration error (ECE), and (iii) posterior contraction (PC). (i) and (iii) serve as a proxy for posterior sharpness \citep{gneiting2007probabilistic}, while (ii) reflects computational faithfulness.
Finally, to quantify distributional differences between posterior estimates obtained under different amortization levels, we report results from classifier two-sample tests between samples from approximate joint posteriors \citep[C2ST;][]{lopez2016revisiting}. 

\paragraph{Conservative evaluation} To ensure comparative fairness, we first defined five practically relevant design configurations as testing points.
Details of these design configurations can be found in \autoref{app:design-config}.
For our baseline, we used the \texttt{BayesFlow} software \citep{kuhmichel2026bayesflow} to train separate flow matching architectures for each of the five design configurations over 1000 epochs, with 100 steps per epoch and 64 simulations per training step (i.e., batch size). These architectures also use a Set Transformer \citep{lee2019set} as a summary network. Additionally, we train CogFormer for each of the three model families ($\mathcal{F}$) and the model class ($\mathcal{C}$) over 5000 epochs, with 100 steps per epoch and 64 simulations per training step. Thus, \textit{the total training budgets for all levels of amortization are the same}: 32,000,000 simulations. We evaluate sharpness metrics on 200 held-out simulations per configuration and compute the C2ST on pairs of posteriors obtained from a subset of 10 datasets. All network hyperparameters are listed in \autoref{app:cogformer-details}. 

\paragraph{Ablation studies} To comprehensively examine CogFormer's architecture, we also conducted a series of ablation studies (see results in~\autoref{app:ablations}).

\vspace{-1em}
\subsection{Generalization within model families}
\label{model-families-results}

To demonstrate generalization across model families, we use CogFormer-$\mathcal{F}$ to evaluate parameter estimation across the pre-specified benchmark design configuration ensemble. Our recovery results for DDM are shown in \autoref{fig:mf-ensemble-recovery}, with further results for RDM and CDM, along with their calibration ECDF available in \autoref{app:additional-benchmarks}. We observe that, across all configurations, all coefficients $\rmB$ show comparable recovery to the baseline. This suggests that CogFormer can perform as well as single-model amortized frameworks, while covering more design configurations under the same simulation budget. Further, we evaluate the calibration ECDF for these benchmark configurations (see \autoref{fig:mf-ensemble-calibration}). The results suggest CogFormer-$\mathcal{F}$ is generally well-calibrated across all benchmark configurations, with calibration curves largely within the 95\% confidence bands for each parameter.

\vspace{-1em}
\subsection{Generalization within a model class}
\label{sec:model-class-results}

To demonstrate generalizability across model classes and beyond the benchmarks above, we use CogFormer-$\mathcal{C}$ to evaluate parameter recovery over a random ensemble of model configurations. This ensemble is generated by 1) randomly selecting a model family, 2) selecting the corresponding subset of parameters within the model class, and 3) specifying a randomized parameter mask given a defined free and fixed parameter configurations. To ensure minimal bias towards any specific model family, we evenly distribute model family selection so that the ensemble includes 4 configurations from each of the DDM, RDM, and CDM families (a total of 12 configurations). 

The random ensemble parameter recovery results are shown in \autoref{fig:mc-ensemble}, with more results presented in \autoref{app:additional-ensembles}, including excellent results for each of the five test configurations shown in the previous section. As can be observed, all estimation targets are well recovered and generally well-calibrated across all model families within the model class.

\begin{figure}[t]
    \centering
    \includegraphics[width=\linewidth]{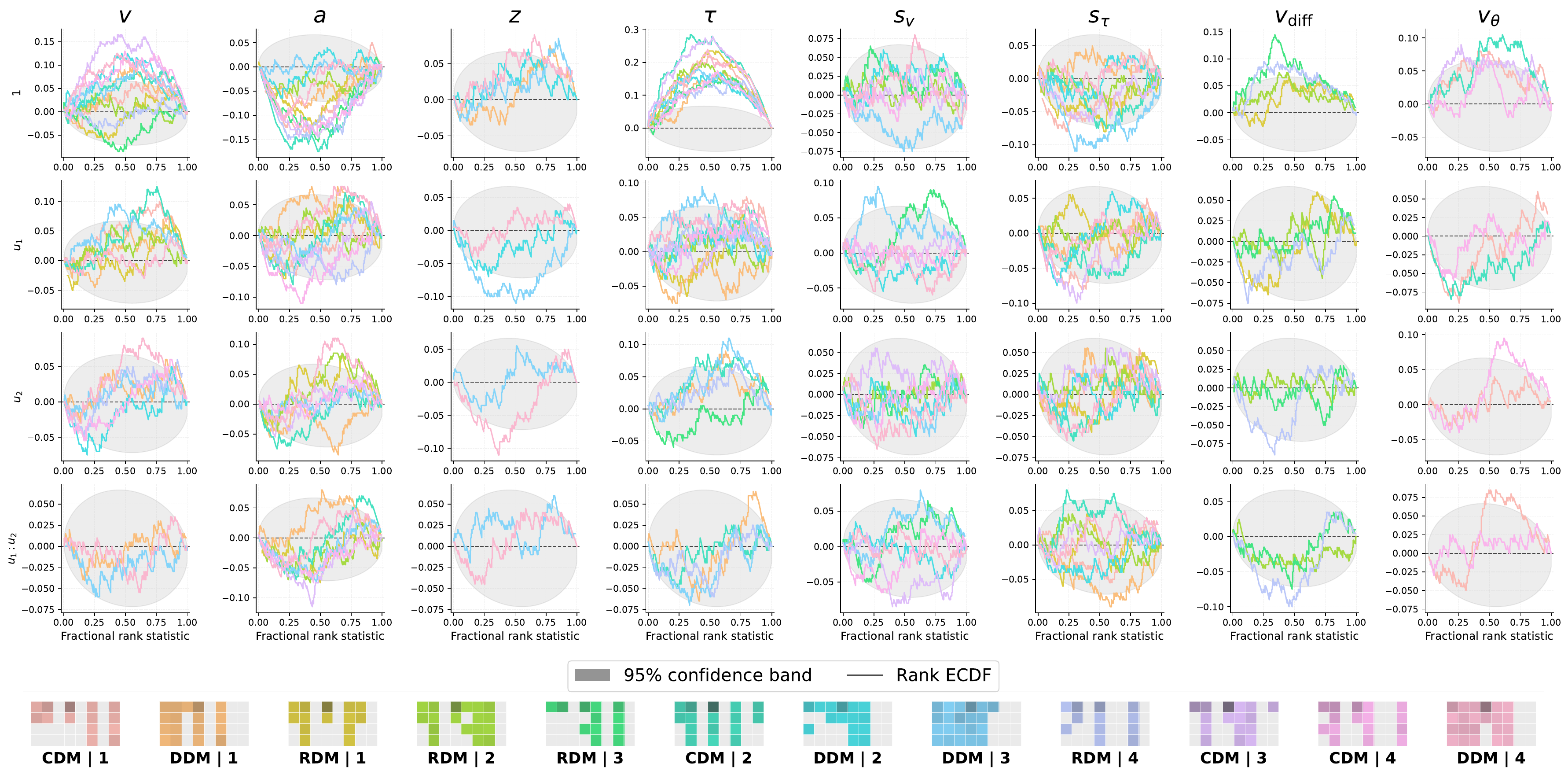}
    \caption{Ensemble calibration ECDFs for CogFormer-$\mathcal{C}$ for 2,400 simulated datasets with 500 trials each over 12 model configurations. The pixel thumbnails at the bottom indicate the active estimation targets within each configuration, with lighter shades indicating lower calibration errors (ECEs). Results are plotted corresponding to these thumbnails at their corresponding locations within the shared coefficient matrix.}
    \label{fig:mc-ensemble-calibration}
\end{figure}

\vspace{-1em}
\subsection{Amortization gaps}
\label{par:amortization-gaps}

\begin{table}[t]
\centering
\small
\caption{Absolute metric values (median $\pm$ SEM) aggregated across the five benchmark design configurations and all parameters in each configuration.}
\label{tab:absolute-metrics}
\begin{tabularx}{\textwidth}{llCCC}
\toprule
\textbf{Model} & \textbf{Method} & \textbf{NRMSE} ($\downarrow$) & \textbf{Cal. Error} ($\downarrow$) & \textbf{Post. Contr.} ($\uparrow$) \\
\midrule
\textbf{DDM} & \textbf{Baseline}& $0.078 \pm 0.007$ & $0.022 \pm 0.002$ & $0.902 \pm 0.023$ \\
 & \textbf{CogFormer}-$\mathcal{F}$ & $0.087 \pm 0.007$ & $0.021 \pm 0.002$ & $0.881 \pm 0.025$ \\
 & \textbf{CogFormer}-$\mathcal{C}$& $0.091 \pm 0.007$ & $0.029 \pm 0.003$ & $0.862 \pm 0.026$ \\
\midrule
\textbf{RDM} &\textbf{Baseline} & $0.112 \pm 0.008$ & $0.061 \pm 0.010$ & $0.803 \pm 0.032$ \\
 & \textbf{CogFormer}-$\mathcal{F}$ & $0.112 \pm 0.009$ & $0.026 \pm 0.004$ & $0.786 \pm 0.033$ \\
 & \textbf{CogFormer}-$\mathcal{C}$ & $0.117 \pm 0.009$ & $0.029 \pm 0.003$ & $0.750 \pm 0.036$ \\
\midrule
\textbf{CDM} & \textbf{Baseline} & $0.075 \pm 0.008$ & $0.020 \pm 0.001$ & $0.896 \pm 0.030$ \\
 & \textbf{CogFormer}-$\mathcal{F}$ & $0.083 \pm 0.008$ & $0.024 \pm 0.002$ & $0.874 \pm 0.032$ \\
 & \textbf{CogFormer}-$\mathcal{C}$ & $0.087 \pm 0.008$ & $0.028 \pm 0.002$ & $0.859 \pm 0.032$ \\
\bottomrule
\end{tabularx}
\end{table}

To summarize performance at each amortization level, we aggregate results across all model configurations and parameters, and report absolute values in \autoref{tab:absolute-metrics}. Across 
all three model families, CogFormer-$\mathcal{F}$ and CogFormer-$\mathcal{C}$ achieve NRMSE within ${\sim}15\%$ of the baseline, with posterior contraction declining by only $2$–$4\%$ in absolute terms---a consistent and modest cost of vastly expanded amortization. Calibration error in absolute values is consistently minimal across all three model families.
Finally, posterior contraction tends to decrease with broader amortization scope.

\begin{table}[t]
\centering
\small
\caption{Joint C2ST (median $\pm $SEM) aggregated across 10 posteriors for each configuration.}
\label{tab:c2st}
\begin{tabularx}{\textwidth}{lCCC}
\toprule
\textbf{Comparison} & \textbf{DDM} & \textbf{RDM} & \textbf{CDM} \\
\midrule
\textbf{CogFormer}-$\mathcal{F}$ vs.\ \textbf{Baseline} &  $0.693 \pm 0.067$& $0.702 \pm 0.076$  & $0.668 \pm 0.068$\\
\textbf{CogFormer}-$\mathcal{C}$ vs.\ \textbf{CogFormer}-$\mathcal{F}$ &  $0.640 \pm 0.049$& $0.600 \pm 0.032$ & $0.635 \pm 0.041$\\
\bottomrule
\end{tabularx}
\end{table}

The C2ST results in~\autoref{tab:c2st} show that while the marginal posteriors of CogFormer-$\mathcal{F}$, CogFormer-$\mathcal{C}$, and the single-posterior baseline are highly similar, there are detectable differences in their joint posteriors. This suggests that, when simulation budgets are matched, meta-amortization induces an amortization gap in capturing the joint structure. The gap can be reduced by increasing the simulation budget for meta-amortization, as deemed typical in practice (see results in~\autoref{app:simulation-budget}).

\section{Conclusions and future work}
\label{sec:conclusions}

In this work, we presented CogFormer, a meta-amortized Bayesian framework for cognitive modeling. Leveraging a transformer-based architecture that flexibly adapts to varying configurations of structurally similar cognitive models, we can use CogFormer to learn all your models once. We demonstrated that, at a minimal amortization cost, we can reliably estimate model parameters for any given design configuration within a combinatorial design space, offering a powerful approach to further accelerate cognitive modeling workflows. 

CogFormer reveals endless possibilities for further developments and improvements. For example, one of its current limitations is that it only deals with exchangeable models, and is not yet capable of amortizing more complex probabilistic symmetries, such as hierarchical models \citep{elsemuller2023sensitivity} and dynamic models \citep{schumacher2023superstat}. These generative landscapes are what we intend to tackle in our future work. Further, adding entirely new models to a class would currently necessitate retraining or fine-tuning; the latter is what we intend to explore in future work, along with unsupervised continual learning \citep{mishra2026unsupervised}.
We hope that all results inspire cognitive modelers about the potential for a general, shareable, and extensive end-to-end inference framework.

\section*{Acknowledgments}
This work was funded by the National Science Foundation under Grant No.~2448380. We would like to thank Niels L. Bracher for his insightful feedback on the transformer backbone and the manuscript. We would also like to thank Mischa von Krause on his helpful suggestions on sequential sampling model implementation. 

\section*{Code Availability}
The code and interactive demo associated with this paper is available as a GitHub repository: \href{https://github.com/bayesflow-org/CogFormer}{https://github.com/bayesflow-org/CogFormer}.

\bibliography{references}

\clearpage

\appendix

\section{Summary of notations}

A summary of all notations presented in the main manuscript is shown in~\autoref{tab:notation}.

\begin{table}[h]
\centering
\caption{Table of notations}
\renewcommand{\arraystretch}{1.1}
\begin{tabularx}{\linewidth}{@{}p{2.4cm} l >{\raggedright\arraybackslash}X@{}}
\toprule
\textbf{Category} & \textbf{Symbol} &\textbf{Meaning} \\
\midrule
\multirow{4}{=}{\textit{Amortization scopes}}
 & $\mathcal{C}$ & Model class (root; e.g.\ sequential sampling models) \\
 & $\mathcal{F}$ & Model family (shared computational structure) \\
 & $\mathcal{M}$ & Model instance / observation model (fully specified) \\
 & $\mathcal{D}$ & Design configuration over the intrinsic parameters \\
\midrule
\multirow{7}{=}{\textit{Generative objects}}
 & $\theta_{\mathcal{F}}$ & Intrinsic parameters of family $\mathcal{F}$ \\
 & $\rmB$ & Coefficient matrix, $\rmB \in \mathbb{R}^{R \times D_{\mathcal{F}}}$ (inference target) \\
 & $\rmM$ & Binary parameter mask, $\rmM \in \{0,1\}^{R \times D_{\mathcal{F}}}$; entries $M_{i,j}$ \\
 & $\rmX$ & Design matrix, $\rmX \in \mathbb{R}^{N \times R}$ \\
 & $\Theta$ & Realized parameter matrix, $\Theta = g(\rmX(\rmM \odot \rmB)) \in \mathbb{R}^{N \times D_{\mathcal{F}}}$ \\
 & $\rmY$ & Model observables, $\rmY \in \mathbb{R}^{N \times O_{\mathcal{F}}}$ \\
 & $\mathrm{RNG}$ & Source of stochastic variation in the observation model \\
\midrule
\multirow{9}{=}{\textit{Dimensions and indices}}
 & $N$ & Number of trials (observations) per dataset \\
 & $D_{\mathcal{F}}$ & Number of intrinsic parameters in family $\mathcal{F}$ \\
 & $O_{\mathcal{F}}$ & Number of observable dimensions in family $\mathcal{F}$ \\
 & $I$ & Number of regressors \\
 & $R$ & Number of design-matrix columns, $R = I\,(C_{\mathrm{max}}-1)+1$ \\
 & $R_{\mathrm{max}}$ & Maximum column capacity (zero-padding target) \\
 & $C_{\mathrm{max}}$ & Maximum number of categories per regressor \\
 & $i,\,j$ & Row (regressor/design column) and column (parameter) indices of $\rmB,\rmM$ \\
\midrule
\multirow{3}{=}{\textit{Functions and parameters}}
 & $\mathcal{G}=\{g_j\}$ & Per-parameter link functions; $g$ denotes column-wise application \\
 & $\beta_{i,j}$ & Regression weight (entry of $\rmB$); $\beta_{0,j}$ is the intercept \\
 & $x_c$ & Covariate $c$ \\
\midrule
\multirow{9}{=}{\textit{Flow matching and inference}}
 & $\mathcal{A}$ & Conditioning set, $\mathcal{A}=(\rmY,\rmX,\rmM,\mathcal{F})$ \\
 & $\hat{u}$ & Learned conditional velocity field, $\hat{u}(\rmZ_t;\mathcal{A},t)$ \\
 & $\rmZ_t$ & Flow state at time $t$, $\rmZ_t=(1-t)\rmB+t\rmZ_1$ \\
 & $\rmZ_0$ & Target endpoint, $\rmZ_0 \equiv \rmB$ \\
 & $\rmZ_1$ & Gaussian latent, $\rmZ_1\sim\mathcal{N}(\mathbf{0},\mathbf{I})$ \\
 & $t$ & Flow time, $t\in[0,1]$ \\
 & $\omega_t$ & Optional time-dependent loss weight \\
 & $\mathcal{L}_{\mathrm{FM}}$ & Conditional flow matching loss \\
 & $q(\rmB\mid\mathcal{A},\hat{u})$ & Approximate posterior induced by $\hat{u}$ \\
\midrule
\multirow{6}{=}{\textit{Architecture}}
 & $\rmS$ & Encoder summary representations, $\rmS=(\rvs_1,\dots,\rvs_L)$ \\
 & $\rvs_{\ell}$ & Summary vector, $\rvs_{\ell}\in\mathbb{R}^{S}$ \\
 & $L$ & Number of summary tokens (encoder seeds) \\
 & $S$ & Summary-representation dimension \\
 & $B_{i,j}$ & Decoder token for coefficient cell $(i,j)$ \\
 & $f$ & Integer index of family $\mathcal{F}$ (model-family embedding) \\
\bottomrule
\end{tabularx}
\label{tab:notation}
\end{table}

\section{Sequential sampling models and priors}
\label{app:decision-models}

\subsection{Cognitive model formulations}

\paragraph{Diffusion decision model (DDM)}

The DDM describes binary choices as a process of noisy evidence accumulation toward one of two decision boundaries.
Let $x(t)$ denote the accumulated evidence at time $t$.
The evolution of the decision variable follows the stochastic differential equation
\begin{equation}
dx(t) = \vartheta \,dt + s\,dW_t,
\end{equation}
where $\vartheta$ denotes the drift rate, reflecting the average rate of evidence accumulation, $s$ is the diffusion coefficient, and $W_t$ denotes a Wiener process.
The process starts at an initial state $x(0) = z\,a$, where the relative start point $z \in (-1, 1)$ controls the initial bias between the two absorbing boundaries at $-a$ and $+a$. A value of $z = 0$ corresponds to an unbiased start, whereas positive (negative) values bias the process toward the upper (lower) boundary. 

A decision is made when the diffusion trajectory first reaches one of the boundaries.
Let $T_{\text{dec}}$ denote the corresponding first-passage time.
The observed decision $D$ is determined by the boundary reached:

\begin{equation}
D =
\begin{cases}
1 & \text{if } x(T_{\text{dec}}) = +a, \\
0 & \text{if } x(T_{\text{dec}}) = -a .
\end{cases}
\end{equation}

The observed response time is modeled as the sum of the decision time and a non-decision component,

\begin{equation}
RT = T_{\text{dec}} + \tau,
\end{equation}

where $\tau$ represents non-decision time capturing processes such as stimulus  encoding and motor execution.

\paragraph{Racing diffusion model (RDM)}

The RDM generalizes evidence accumulation to multiple alternatives by assuming independent accumulators that race toward their respective decision thresholds. For $K$ alternatives, each option $k$ is associated with a diffusion process
\begin{equation}
dx_k(t) = \vartheta_k\,dt + s\,dW_{k,t},
\end{equation}
where $\vartheta_k$ denotes the drift rate of accumulator $k$ and $W_{k,t}$ are independent Wiener processes.
Each process starts at $x_k(0)=z_k$ and evolves toward a threshold $a_k$.

A decision occurs when the first accumulator reaches its threshold. Let $T_k$ denote the first-passage time of accumulator $k$.
The chosen alternative corresponds to the winning accumulator
\begin{equation}
k^* = \arg\min_k T_k.
\end{equation}
Response time is given by the winning finishing time plus the non-decision component $\tau$.

In this paper, we employ a simplified version of the RDM.
First, the model is restricted to two alternatives ($K=2$).
Second, both accumulators start at zero,
\begin{equation}
x_0(0) = x_1(0) = 0,
\end{equation}
and share a common decision threshold,
\begin{equation}
a_0 = a_1 = a.
\end{equation}
Drift rates are parameterized using response coding rather than stimulus coding.
That is, drift rates are specified directly for each accumulator without reference to stimulus correctness. Let $\vartheta$ denote the base drift rate for accumulator $0$, and let $\vartheta_\Delta$ denote the drift difference between the two accumulators.
The drift rates are defined as

\begin{equation}
\vartheta_0 = \vartheta, \qquad
\vartheta_1 = \vartheta + \vartheta_\Delta.
\end{equation}

Under this parameterization, positive values of $\vartheta_\Delta$ favor response $1$, whereas negative values favor response $0$.

\paragraph{Circular diffusion model (CDM)}

The CDM extends the diffusion framework to continuous response spaces, such as directional decisions.
Evidence accumulation occurs in two dimensions:

\begin{equation}
d\mathbf{x}(t) =
\begin{pmatrix}
dx_1(t) \\
dx_2(t)
\end{pmatrix}
=
\boldsymbol{\mu}\,dt + s\,d\mathbf{W}_t,
\end{equation}

where $\boldsymbol{\mu}$ is a two-dimensional drift vector and $\mathbf{W}_t$ is a two-dimensional Wiener process.

Rather than parameterizing the drift directly in Cartesian coordinates, we express it in polar coordinates using a drift magnitude $\vartheta$ and a drift angle $\vartheta_\theta$.
The Cartesian drift components are then obtained via

\begin{equation}
\vartheta_x = \vartheta \cos(\vartheta_\theta), \qquad
\vartheta_y = \vartheta \sin(\vartheta_\theta).
\end{equation}

The decision space is bounded by a circular absorbing boundary with radius $a$.
A response is generated when the diffusion trajectory first reaches this boundary.
The angular position of the boundary crossing determines the response direction, while the hitting time determines the decision time.

In this paper, the model is formulated using error coding.
Specifically, the observed response variable corresponds to the angular deviation between the true stimulus direction and the participant's response.
Consequently, the response variable is centered at zero, corresponding to unbiased responses.
Under this parameterization, the drift angle $\vartheta_\theta$ reflects a systematic error bias: values different from zero indicate that responses tend to deviate from the correct direction.
The drift magnitude $\vartheta$ controls the speed of evidence accumulation and is therefore directly comparable to the drift rate parameter in the other diffusion models.
Again, response time is determined by the first-passage time plus the non-decision component $\tau$.

\paragraph{Shared model assumptions}

The following assumptions were applied to all three models.
First, diffusion noise was fixed to unit variance ($s=1$), which is a standard identifiability constraint in diffusion models.

Second, we allowed for across-trial variability in both the drift rate and the non-decision time.
Specifically, the drift rate on trial $i$ was assumed to vary according to

\begin{equation}
\vartheta_i \sim \mathcal{N}(\vartheta, s_\vartheta),
\end{equation}

where $\vartheta$ denotes the mean drift rate and $s_\vartheta$ controls across-trial variability in drift.
Similarly, non-decision time was allowed to vary across trials according to

\begin{equation}
\tau_i = \tau + U(0, s_\tau),
\end{equation}

where $\tau$ denotes the minimum non-decision time and $s_\tau$ determines the range of the uniform variability component.

The following priors were used for simulating the respective models.
All priors are Gaussian distributions defined in an unconstrained latent space.
Before simulation, latent parameter draws are transformed using link functions to enforce the appropriate parameter constraints.

\paragraph{Diffusion decision model (DDM) priors}

\begin{align}
    \vartheta &\sim \mathcal{N}(1.0, 0.8), &
    a &\sim \mathcal{N}(0.5, 0.5), \notag\\
    z &\sim \mathcal{N}(0.0, 0.8), &
    \tau &\sim \mathcal{N}(-1.2, 0.5), \notag\\
    s_\vartheta &\sim \mathcal{N}(-1.2, 1.0), &
    s_\tau &\sim \mathcal{N}(-1.5, 0.7)
\end{align}

\paragraph{Racing diffusion model (RDM) priors}

\begin{align}
    \vartheta &\sim \mathcal{N}(0.6, 0.5), &
    \vartheta_\Delta &\sim \mathcal{N}(0.6, 0.5), \notag\\
    a &\sim \mathcal{N}(0.25, 0.5), &
    \tau &\sim \mathcal{N}(-1.2, 0.5), \notag\\
    s_\vartheta &\sim \mathcal{N}(-1.0, 0.6), &
    s_\tau &\sim \mathcal{N}(-1.5, 0.7)
\end{align}

\paragraph{Circular diffusion model (CDM) priors}:

\begin{align}
    \vartheta &\sim \mathcal{N}(1.0, 0.8), &
    \vartheta_\theta &\sim \mathcal{N}(0.0, 0.5), \notag\\
    a &\sim \mathcal{N}(0.5, 0.5), &
    \tau &\sim \mathcal{N}(-1.2, 0.5), \notag\\
    s_\vartheta &\sim \mathcal{N}(-1.0, 0.6), &
    s_\tau &\sim \mathcal{N}(-1.5, 0.7)
\end{align}

To ensure that parameters respect their natural constraints, all models employed differentiable link functions that transform unconstrained latent values into their respective domains prior to simulation. Across all three models, drift rates, decision thresholds, non-decision times, and across-trial drift variability were mapped using a \emph{softplus} function,
\begin{equation}
    f(x) = \log(1 + \exp(x)),
\end{equation}
which guarantees positive outputs. In the RDM, the drift-difference parameter $\vartheta_\Delta$ was transformed using the same softplus link. Parameters with bounded support, namely the starting point $z$ in the DDM and the drift angle $\vartheta_\theta$ in the CDM, were transformed using a \emph{scaled sigmoid} function,
\begin{equation}
    f(x) = \text{lower} + (\text{upper} - \text{lower}) \frac{1}{1 + \exp(-x)}.
\end{equation}
These link functions allow estimation on an unconstrained space while ensuring that simulated parameters remain within their admissible ranges.

\newpage
\section{Meta-simulator algorithms}
\label{app:meta-simulator}

Below, we present the detailed algorithm of our meta-simulator for a model family $\mathcal{F}$ (\autoref{alg:model-family}) and a model class $\mathcal{C}$ (\autoref{alg:model-class}).

  \begin{algorithm}[H]
  \caption{Model Family Simulator}
  \label{alg:model-family}
  \begin{algorithmic}[1] 
  \Require Model family $\mathcal{F}$ with observation model $\mathcal{M}$, coefficient prior $p(\rmB)$,
      intrinsic parameters $\theta_{\mathcal{F}}$, link $g$
  \Require Number of observations $N$, number of regressors $I$

  \State \textbf{Sample design configuration} $\mathcal{D}$:
      assign each regressor (intercept $+$ main effects $+$ interactions)
      to a random subset of intrinsic parameters
  
  \State \textbf{Build parameter mask} $\rmM \in \{0,1\}^{R \times D_{\mathcal{F}}}$
      from $\mathcal{D}$, where $R$ = number of design columns, $D_{\mathcal{F}}$ = number of intrinsic parameters

  \State \textbf{Build design matrix} $\rmX \in \mathbb{R}^{N \times R}$:
      sample continuous or discrete regressor values per trial

  \State \textbf{Sample coefficient matrix} $\rmB \sim p(\rmB)$, with $\rmB \in \mathbb{R}^{R \times D_{\mathcal{F}}}$
  
  \State \textbf{Compute per-trial parameters}:
  \[
      \Theta = g\!\left(\rmX\,(\rmM \odot \rmB)\right) \in \mathbb{R}^{N \times D_{\mathcal{F}}}
  \]
  
  \State \textbf{Simulate trial data} $\rmY \sim \mathcal{M}(\Theta)$
  
  \State \Return $\{\rmX,\, \rmB,\, \rmM,\, \Theta,\, \rmY\}$
  \end{algorithmic}   
  \end{algorithm}

  \begin{algorithm}[H]
  \caption{Model Class Meta-Simulator}
  \label{alg:model-class}
  \begin{algorithmic}[1]
  \Require Model families $\{\mathcal{F}_f\}_{f=1}^{K}$ (constituting model class $\mathcal{C}$), sampling weights $\rvw$
  \Require Global parameter space $\Omega = \bigcup_f \theta_{\mathcal{F}_f}$
  \Require Batch size $N_{\mathrm{batch}}$, observation range $[N_{\min}, N_{\max}]$, regressor range $[I_{\min}, I_{\max}]$

  \State Sample $N \sim \mathrm{Uniform}(N_{\min}, N_{\max})$ 
  
  \For{$i = 1, \ldots, N_{\mathrm{batch}}$}
      \State Sample model family index $f_i \sim \mathrm{Categorical}(\rvw)$
      \State Sample number of regressors $I_i \sim \mathrm{Uniform}(I_{\min}, I_{\max})$
      \State Run \textbf{Algorithm~\ref{alg:model-family}} with $\mathcal{F}_{f_i}$, $N$, $I_i$
          to obtain $\{\rmX_i, \rmB_i, \rmM_i, \Theta_i, \rmY_i\}$
      \State \textbf{Lift} $\rmB_i$, $\rmM_i$ to global space $\Omega$:
          insert local parameter columns at their global indices,
          pad with zeros for parameters absent in $\mathcal{F}_{f_i}$
  \EndFor

  \State \textbf{Collate} batch: zero-pad all arrays to maximum dimensions
  
  \State \Return $\bigl\{\rmX_i,\, \rmB_i^{\Omega},\, \rmM_i^{\Omega},\, \rmY_i,\, f_i\bigr\}_{i=1}^{N_{\mathrm{batch}}}$
  \end{algorithmic}   
  \end{algorithm}

\newpage
\section{Further details on CogFormer architecture}
\label{app:cogformer-details}

\subsection{Attention blocks}
\label{app:cogformer-attentions}

\begin{figure}[h]
    \centering
    \includegraphics[width=0.75\linewidth]{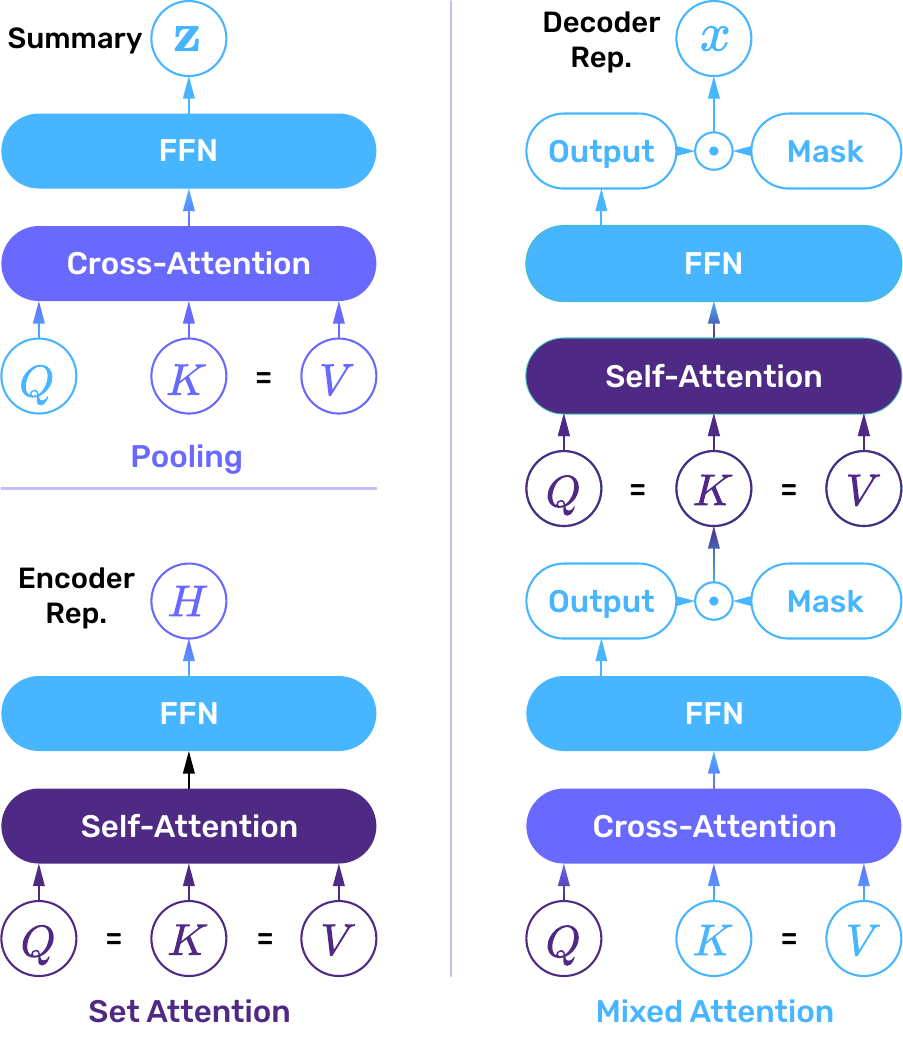}
    \caption{\textbf{Breakdown of the attention blocks in the CogFormer architecture.} For set attention, the input $QKV$ matrices are identical linear layer outputs of the input data. Its output $H$ is the encoder representation, which serves as the input $KV$ matrices for pooling along with a learnable seed vector as the query $Q$. The mixed attention later takes the encoder summary $\textbf{z}$ as query, along with the constructed decoder embeddings listed in \autoref{fig:transformer} as the $KV$ matrices. The mixed-attention architecture places the cross-attention layer first, followed by the self-attention layer, and further alternates between the two. The layer-wise output is element-wise multiplied by the parameter mask $M$ before passing to the next layer. The resulting output $x$ serves as the decoder representation that is used to generate a velocity field $v$.}
    \label{fig:cogformer-attentions}
\end{figure}

A detailed breakdown of the attention blocks in CogFormer architecture can be found in \autoref{fig:cogformer-attentions}. There are three attention blocks at play: 1) self-attention block for the encoder, 2) multi-head cross-attention block as the pooling layer for the encoder, and 3) mixed attention block that consists of an undulated pattern between self-attention and cross-attention.

\clearpage
\subsection{Hyperparameters}
\label{app:hypers}

\begin{table}[h]
\centering
\small
\caption{CogFormer hyperparameters for the family-level models (DDM, RDM, CDM) and the model-class model. QKV dim per head = projection dim $\div$ heads.}
\label{tab:hyperparams}
\begin{tabular}{lcc}
\toprule
\textbf{Hyperparameter} & \textbf{CogFormer-$\mathcal{F}$} & \textbf{CogFormer-$\mathcal{C}$} \\
\midrule
\addlinespace[2pt]
\textbf{Layers} & & \\
\quad Encoder layers & 8 & 8 \\
\quad Decoder layers & 8 & 8 \\
\addlinespace[2pt]
\textbf{Attention} & & \\
\quad Heads (encoder \& decoder) & 8 & 8 \\
\quad Projection dim $d_{\text{model}}$ & 256 & 256 \\
\quad QKV dim per head & 32 & 32 \\
\addlinespace[2pt]
\textbf{Embeddings} & & \\
\quad Time embedding dim & 32 & 32 \\
\quad Positional embedding dim & 32 & 32 \\
\quad Seed tokens & 32 & 32 \\
\quad Seed dim & 64 & 128 \\
\quad Model embedding dim & $-$ & 8 \\
\bottomrule
\end{tabular}
\end{table}

\section{Formal definition of evaluation metrics}
\label{app:eval-metrics}

\paragraph{Parameter recovery} Given model observables $\rmY$ and a discrepancy (loss) metric $\delta$ (e.g., RMSE), parameter recovery examines whether the trained ABI framework can recover the true data-generating parameter $\theta$ from its posterior estimates $\tilde{\theta}$:

\begin{equation}
  \mathrm{Recovery}(\delta) := \mathbb{E}_{(\theta,\, \rmY) \, \sim \, p(\theta, \rmY)}
  \left[ \int \delta(\theta, \tilde{\theta}) \, q(\tilde{\theta} \mid \rmY)\,\mathrm{d}\tilde{\theta} \right].
\end{equation}
  
Parameter recovery is a significant step in model verification because it informs the accuracy of parameter estimation of trained ABI frameworks \citep{kuhmichel2026bayesflow}.

\paragraph{Simulation-based calibration (SBC)} \citet{talts2018validating} introduced SBC for checking the consistency of the data-averaged posterior with respect to the prior via their respective rank statistics. For a scalar parameter $\theta$, given $S$ simulations each with $M$ posterior draws, we compute the rank statistic as
\begin{equation}
      \mathrm{rank}^{(n)} = \sum_{m=1}^{M} \mathbf{1}\!\left[\tilde{\theta}_m^{(n)} < \theta^{(n)}\right],
      \quad \tilde{\theta}_m^{(n)} \sim q(\theta \mid \rmY^{(n)}),
      \quad \theta^{(n)} \sim p(\theta)
\end{equation}

so $\mathrm{rank}^{(n)} \in \{0, 1, \ldots, M\}$. Under a well-calibrated posterior, $\mathrm{rank}^{(n)} \sim \mathrm{Uniform}\{0, \ldots, M\}$. The ECDF is defined over the rank statistic:
\begin{equation}
      \hat{F}(\mathrm{rank}) = \frac{1}{S}\sum_{n=1}^{S} \mathbf{1}\!\left[\mathrm{rank}^{(n)} \leq \mathrm{rank}\right].
\end{equation}

\paragraph{Single-value metrics} Three scalar metrics are used for our evaluation: 1) normalized root mean square error (NRMSE), 2) empirical calibration error (ECE), and 3) posterior contraction. These metrics are well-defined in Appendix A of \citet{arruda2025diffusion}. 

\newpage
\section{Experiment benchmark designs}
\label{app:design-config}

For our experiments, we use the design configurations listed in~\autoref{tab:benchmark-configs} as our benchmarks to evaluate all relevant metrics. Taken as a whole, these configurations cover a comprehensive set of scenarios in which intrinsic parameters are free, fixed, and regressed, and allow us to isolate individual conditions for model validation.  

\begin{table}[h]
\centering
\small
\caption{Test design configurations. Each cell lists the parameters assigned to the corresponding effect type. A dash ($-$) indicates no parameters are assigned. $\vartheta_\Delta$ abbreviates $\vartheta_\Delta$ (RDM).}
\label{tab:benchmark-configs}
\begin{tabular}{llllll}
\toprule
\textbf{Configuration} & \textbf{Model} & \textbf{Intercept} ($\mathbf{1}$) & \textbf{Slope} ($u_1$) & \textbf{Slope} ($u_2$) & \textbf{Interaction} ($u_1{\times}u_2$) \\
\midrule
\textbf{Intercept Only} & DDM & $\vartheta, a, z, \tau, s_{\vartheta}, s_\tau$ & $-$ & $-$ & $-$ \\
 & RDM & $\vartheta, \vartheta_\Delta, a, \tau, s_{\vartheta}, s_\tau$ & $-$ & $-$ & $-$ \\
 & CDM & $\vartheta, \vartheta_\theta, a, \tau, s_{\vartheta}, s_\tau$ & $-$ & $-$ & $-$ \\
\midrule
\textbf{Fixed Variability} & DDM & $\vartheta, a, z, \tau$ & $-$ & $-$ & $-$ \\
 & RDM & $\vartheta, \vartheta_\Delta, a, \tau$ & $-$ & $-$ & $-$ \\
 & CDM & $\vartheta, \vartheta_\theta, a, \tau$ & $-$ & $-$ & $-$ \\
\midrule
\textbf{Regressed} & DDM & $\vartheta, a, z, \tau, s_{\vartheta}, s_\tau$ & $\vartheta, a, z$ & $\vartheta, a, z$ & $-$ \\
 & RDM & $\vartheta, \vartheta_\Delta, a, \tau, s_{\vartheta}, s_\tau$ & $\vartheta_\Delta, a$ & $\vartheta_\Delta, a$ & $-$ \\
 & CDM & $\vartheta, \vartheta_\theta, a, \tau, s_{\vartheta}, s_\tau$ & $\vartheta, a$ & $\vartheta, a$ & $-$ \\
\midrule
\textbf{Fixed + Regressed} & DDM & $\vartheta, a, z, \tau$ & $\vartheta, a, z$ & $\vartheta, a, z$ & $-$ \\
 & RDM & $\vartheta, \vartheta_\Delta, a, \tau$ & $\vartheta_\Delta, a$ & $\vartheta_\Delta, a$ & $-$ \\
 & CDM & $\vartheta, \vartheta_\theta, a, \tau$ & $\vartheta, a$ & $\vartheta, a$ & $-$ \\
\midrule
\textbf{With Interaction} & DDM & $\vartheta, a, z, \tau, s_{\vartheta}, s_\tau$ & $\vartheta, a, z, \tau, s_{\vartheta}$ & $\vartheta, a, z, \tau$ & $\vartheta, a, z$ \\
 & RDM & $\vartheta, \vartheta_\Delta, a, \tau, s_{\vartheta}, s_\tau$ & $\vartheta, \vartheta_\Delta, a, \tau, s_{\vartheta}$ & $\vartheta, \vartheta_\Delta, a, \tau$ & $\vartheta, \vartheta_\Delta, a$ \\
 & CDM & $\vartheta, \vartheta_\theta, a, \tau, s_{\vartheta}, s_\tau$ & $\vartheta, \vartheta_\theta, a, \tau, s_{\vartheta}$ & $\vartheta, \vartheta_\theta, a, \tau$ & $\vartheta, \vartheta_\theta, a$ \\
\bottomrule
\end{tabular}
\end{table}

\section{Ablation studies}
\label{app:ablations}

\subsection{Architectural changes}

We investigated CogFormer-$\mathcal{F}$'s performance under several architectural changes outlined in \autoref{tab:component_ablation_summary}, with their corresponding single-value metrics listed in \autoref{tab:component_ablation_raw}. 

\begin{table}[!h]
\centering
\caption{Summary of component ablation variants and the architectural role of each
ablated component.}
\label{tab:component_ablation_summary}
\begin{tabularx}{\textwidth}{>{\raggedright\arraybackslash}l >{\raggedright\arraybackslash}X}
\toprule
\textbf{Variant} & \textbf{Component Role} \\
\midrule
No SAB
  & Self-attention across parameter tokens; allows parameters to coordinate
    with one another during the denoising pass. \\[4pt]
No MAB
  & Cross-attention from parameter tokens to the encoder summary; the primary
    pathway through which the decoder reads the observed data. \\[4pt]
Baseline
  & Feature-wise linear modulation (FiLM) of the output projection,
    conditioned on the current diffusion timestep; scales and shifts the
    final representation before predicting velocity. \\[4pt]
No FE
  & Learnable Fourier embedding (FE) of the diffusion timestep; projects the scalar $t$ into a
    high-dimensional frequency space before concatenation. \\
With FiLM
  & Feature-wise linear modulation (FiLM) of the output projection,
    conditioned on the current diffusion timestep; scales and shifts the
    final representation before predicting velocity. \\
\bottomrule
\end{tabularx}
\end{table}

\begin{table}[h]
\centering
\caption{Component ablation metrics across all design conditions, ordered by number of
active parameters. NRMSE and Calibration Error: lower is better ($\downarrow$).
Posterior Contraction: higher is better ($\uparrow$). Bold indicates best per metric
within each condition group. $^\dagger$No MAB collapses Posterior Contraction
across all conditions; its Cal.\ Error optimum is not meaningful, since the prior itself is perfectly calibrated.}
\label{tab:component_ablation_raw}
\begin{tabularx}{\textwidth}{>{\raggedright\arraybackslash}X >{\raggedright\arraybackslash}l >{\centering\arraybackslash}X >{\centering\arraybackslash}X >{\centering\arraybackslash}X}
\toprule
\textbf{Condition} & \textbf{Variant} & \textbf{NRMSE} $\downarrow$ & \textbf{Cal.\ Error} $\downarrow$ & \textbf{Contraction} $\uparrow$ \\
\midrule
\multirow{5}{*}{\shortstack[l]{Fixed\\{\small ($n=4$)}}}
 & Baseline    & 0.055 & 0.021          & \textbf{0.966} \\
 & No SAB     & 0.058 & 0.039          & 0.944 \\
 & No MAB     & 0.255 & 0.017          & 0.037 \\
 & No FE & \textbf{0.053} & 0.023 & 0.965 \\
 & With FiLM   & 0.058 & \textbf{0.017} & 0.966 \\
\midrule
\multirow{5}{*}{\shortstack[l]{Intercept Only\\{\small ($n=6$)}}}
 & Baseline    & \textbf{0.087} & 0.022 & \textbf{0.845} \\
 & No SAB     & 0.103 & 0.022          & 0.825 \\
 & No MAB     & 0.249 & 0.026          & 0.077 \\
 & No FE & 0.090 & \textbf{0.019} & 0.820 \\
 & With FiLM   & 0.096 & 0.025          & 0.833 \\
\midrule
\multirow{5}{*}{\shortstack[l]{Fixed + Regressed\\{\small ($n=10$)}}}
 & Baseline    & 0.071 & 0.023          & 0.946 \\
 & No SAB     & 0.077 & 0.018          & 0.925 \\
 & No MAB$^\dagger$ & 0.263 & 0.016   & 0.030 \\
 & No FE & \textbf{0.067} & 0.027 & \textbf{0.947} \\
 & With FiLM   & 0.072 & 0.020          & 0.938 \\
\midrule
\multirow{5}{*}{\shortstack[l]{Regressed\\{\small ($n=12$)}}}
 & Baseline    & \textbf{0.073} & 0.021 & 0.903 \\
 & No SAB     & 0.082 & 0.021          & 0.891 \\
 & No MAB     & 0.262 & 0.027          & 0.086 \\
 & No FE & 0.076 & \textbf{0.020} & \textbf{0.904} \\
& With FiLM   & 0.081 & 0.023          & 0.894 \\
\midrule
\multirow{5}{*}{\shortstack[l]{Interaction\\{\small ($n=18$)}}}
 & Baseline    & 0.099 & 0.020          & 0.838 \\
 & No SAB     & 0.103 & 0.028          & 0.834 \\
 & No MAB     & 0.268 & 0.018          & 0.029 \\
 & No FE & \textbf{0.094} & 0.025 & \textbf{0.849} \\
 & With FiLM   & 0.103 & \textbf{0.017} & 0.822 \\
\bottomrule
\end{tabularx}
\end{table}

\newpage
\subsection{Model embedding for CogFormer-$\mathcal{C}$}

To demonstrate the utility of model embedding for CogFormer-$\mathcal{C}$, we performed a comparative study with a similar architecture without model embedding. The metric results are shown in \autoref{tab:ddm_embedding_ablation}. From these results, we can observe that in the majority of designs, CogFormer-$\mathcal{C}$ can estimate better with embedding, at a small expense of calibration errors. 

\begin{table}[h]
\centering
\small
\caption{Model embedding ablation for the DDM across all five design conditions.
Mean metrics across all active parameters.
$\Delta$ (in parentheses) is the change relative to the with-embedding baseline.
NRMSE and Calibration Error: lower is better ($\downarrow$).
Posterior Contraction: higher is better ($\uparrow$).
Best result per condition per metric is \textbf{bolded}.}
\label{tab:ddm_embedding_ablation}
\begin{tabular}{l l ccc}
\toprule
\textbf{Design} & \textbf{Condition} & \textbf{NRMSE} $\downarrow$ & \textbf{Cal.\ Error} $\downarrow$ & \textbf{Contraction} $\uparrow$ \\
\midrule
\multirow{2}{*}{Intercept Only}
 & Without Embedding & 0.104 \small{(+0.005)} & \textbf{0.033} \small{(-0.009)} & 0.794 \small{(-0.033)} \\
 & With Embedding    & \textbf{0.099}           & 0.042                            & \textbf{0.827}           \\
\midrule
\multirow{2}{*}{Fixed}
 & Without Embedding & 0.068 \small{(+0.008)} & \textbf{0.044} \small{(-0.019)} & 0.947 \small{(-0.006)} \\
 & With Embedding    & \textbf{0.060}           & 0.063                            & \textbf{0.954}           \\
\midrule
\multirow{2}{*}{Regressed}
 & Without Embedding & \textbf{0.082} \small{(-0.002)} & 0.031 \small{(+0.000)} & 0.884 \small{(-0.002)} \\
 & With Embedding    & 0.084                            & \textbf{0.031}          & \textbf{0.886}           \\
\midrule
\multirow{2}{*}{Fixed Regressed}
 & Without Embedding & 0.079 \small{(+0.001)} & \textbf{0.025} \small{(-0.010)} & 0.919 \small{(-0.003)} \\
 & With Embedding    & \textbf{0.078}           & 0.035                            & \textbf{0.922}           \\
\midrule
\multirow{2}{*}{Interaction}
 & Without Embedding & \textbf{0.105} \small{(-0.002)} & \textbf{0.024} \small{(-0.002)} & \textbf{0.811} \small{(+0.006)} \\
 & With Embedding    & 0.107                            & 0.026                            & 0.806                            \\
\bottomrule
\end{tabular}
\end{table}


\section{Size comparison study}

We ran four versions of CogFormer-$\mathcal{F}$ with different hyperparameter sets -- small (S), medium (M), large (L), extra-large (XL) -- over 1000 epochs with varying number of attention layers, with the following aggregated results showing how CogFormer performs with increasing level of expressiveness (see \autoref{tab:size_comparison}). This result for hyperparameter sweep suggests that deeper architecture is not necessarily better, with our baseline setting  (8 layers and 8 heads) performing the best among all candidate hyperparameter sets. 

\begin{table}[h]
\centering
\caption{Mean posterior approximation metrics across all parameters and conditions for each CogFormer-$\mathcal{F}$ size variant over the number of attention layers (L) and the number of attention heads (H). NRMSE and Calibration Error: lower is better ($\downarrow$). Posterior Contraction: higher is better ($\uparrow$). Best result per column is \textbf{bolded}.}
\label{tab:size_comparison}
\begin{tabular}{lcccc}
\toprule
 & S (2L, 2H) & M (4L, 4H) & \textbf{L (8L, 8H; baseline)} & XL (16L, 16H) \\
\midrule
NRMSE $\downarrow$        & 0.179 & 0.170 & \textbf{0.124} & 0.127 \\
Cal.\ Error $\downarrow$  & 0.024 & 0.032 & \textbf{0.022} & 0.045 \\
Post.\ Contr.\ $\uparrow$ & 0.472 & 0.566 & \textbf{0.746} & 0.701 \\
\bottomrule
\end{tabular}
\end{table}

\newpage
\section{Baseline comparison study}
\label{app:baseline-comparison}

Our baseline uses Set Transformer \citep{lee2019set} and flow matching implemented in \texttt{BayesFlow} \citep{kuhmichel2026bayesflow}, which is closely aligned with the encoder-decoder architecture of CogFormer. The choice of flow matching as inference network is justified by the most comprehensive benchmark of free-form models to date \citep{arruda2025diffusion}, which demonstrated that flow matching with said settings achieves the best results on established SBI benchmarks \citep{lueckmann2021benchmarking} and high-dimensional problems of genuine scientific interest. 

\begin{table}[h]
\centering
\caption{Mean posterior approximation metrics across all DDM parameters for each inference network. NRMSE and Calibration Error: lower is better ($\downarrow$). Posterior Contraction: higher is better ($\uparrow$). Best result per column is \textbf{bolded}.}
\label{tab:baseline_comparison}
\begin{tabular}{lcccc}
\toprule
 & Coupling Flow& Diffusion Model & \textbf{Flow Matching} & Consistency Model \\
\midrule
NRMSE $\downarrow$        & 0.128 & 0.131 & \textbf{0.117} & 0.129 \\
Cal.\ Error $\downarrow$  & 0.020 & 0.025 & \textbf{0.017} & 0.067 \\
Post.\ Contr.\ $\uparrow$ & 0.793 & 0.746 & \textbf{0.820} & 0.706 \\
\bottomrule
\end{tabular}
\end{table}

To confirm our justification, we compare the baseline against 3 other neural estimators with different inference algorithms: coupling flows, diffusion model, and consistency model. Training them over one of the benchmark cases (Interaction) for the DDM model family yields aggregated results shown in \autoref{tab:baseline_comparison}. We can observe that, across all evaluated metrics, flow matching outperforms all other inference networks.

\clearpage
\section{Simulation budget study}
\label{app:simulation-budget}

\begin{figure}[h]
    \centering
    \includegraphics[width=0.9\linewidth]{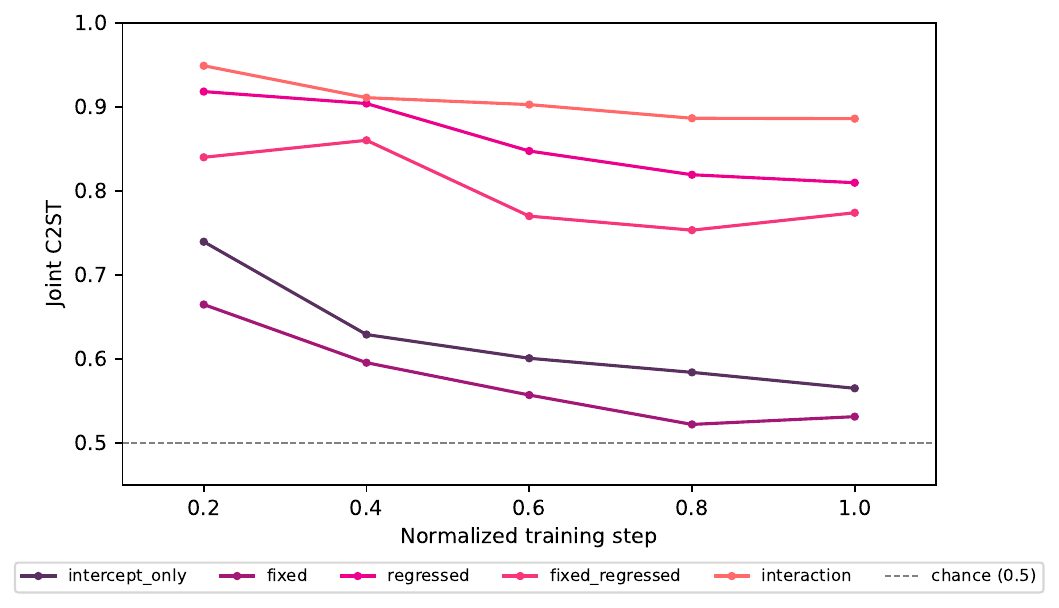}
    \includegraphics[width=0.9\linewidth]{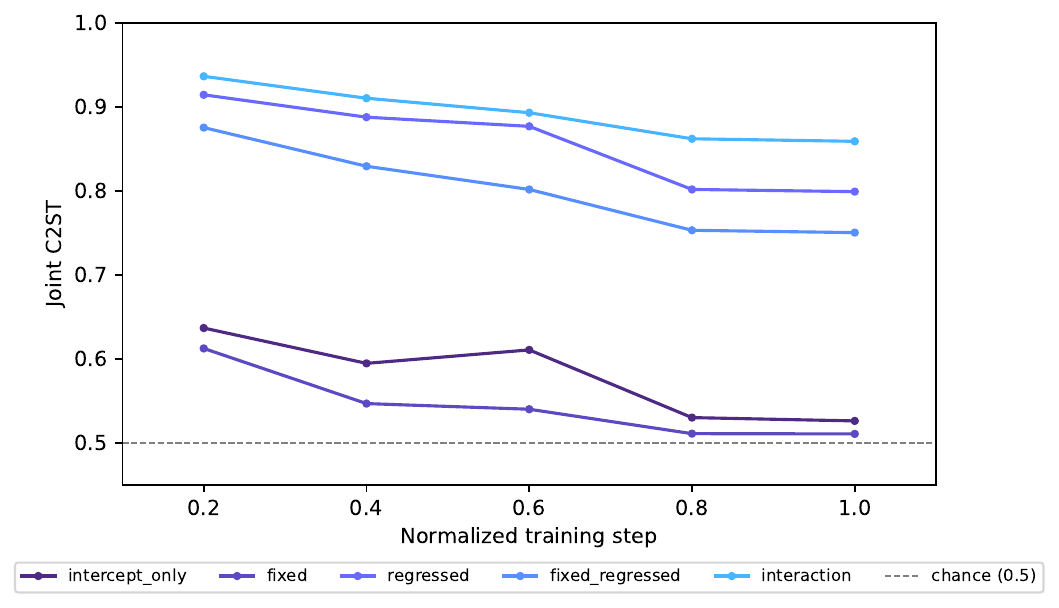}
    \caption{Joint C2ST aggregated across 10 posteriors for each benchmark design configuration per increasing simulation budget for CogFormer-$\mathcal{F}$ (\emph{top}) and CogFormer-$\mathcal{C}$ (\emph{bottom}). The training epochs are normalized to between 0 and 1, with 0.2 indicating Epoch 1000 and 1 indicating Epoch 5000.}
    \label{fig:amortization-gap-sim-budget}
\end{figure}

To investigate how the amortization gap closes as simulation budget increases, we trained CogFormer-$\mathcal{F}$ and CogFormer-$\mathcal{C}$ under the standard training protocol (5000 epochs, 100 steps per epoch, and a batch size of 64), and evaluated the joint C2ST for each benchmark design every 1000 epochs. The result is shown in~\autoref{fig:amortization-gap-sim-budget}. In general, the result suggests that as the simulation budget increases, the amortization gap closes correspondingly. The gaps are generally larger with more parameters to estimate, with regressed cases having larger gaps than non-regressed cases. 

\newpage
\section{Additional results for ensemble evaluations}
\label{app:additional-ensembles}

\begin{figure}[!h]
    \centering
    \includegraphics[width=0.96\linewidth]{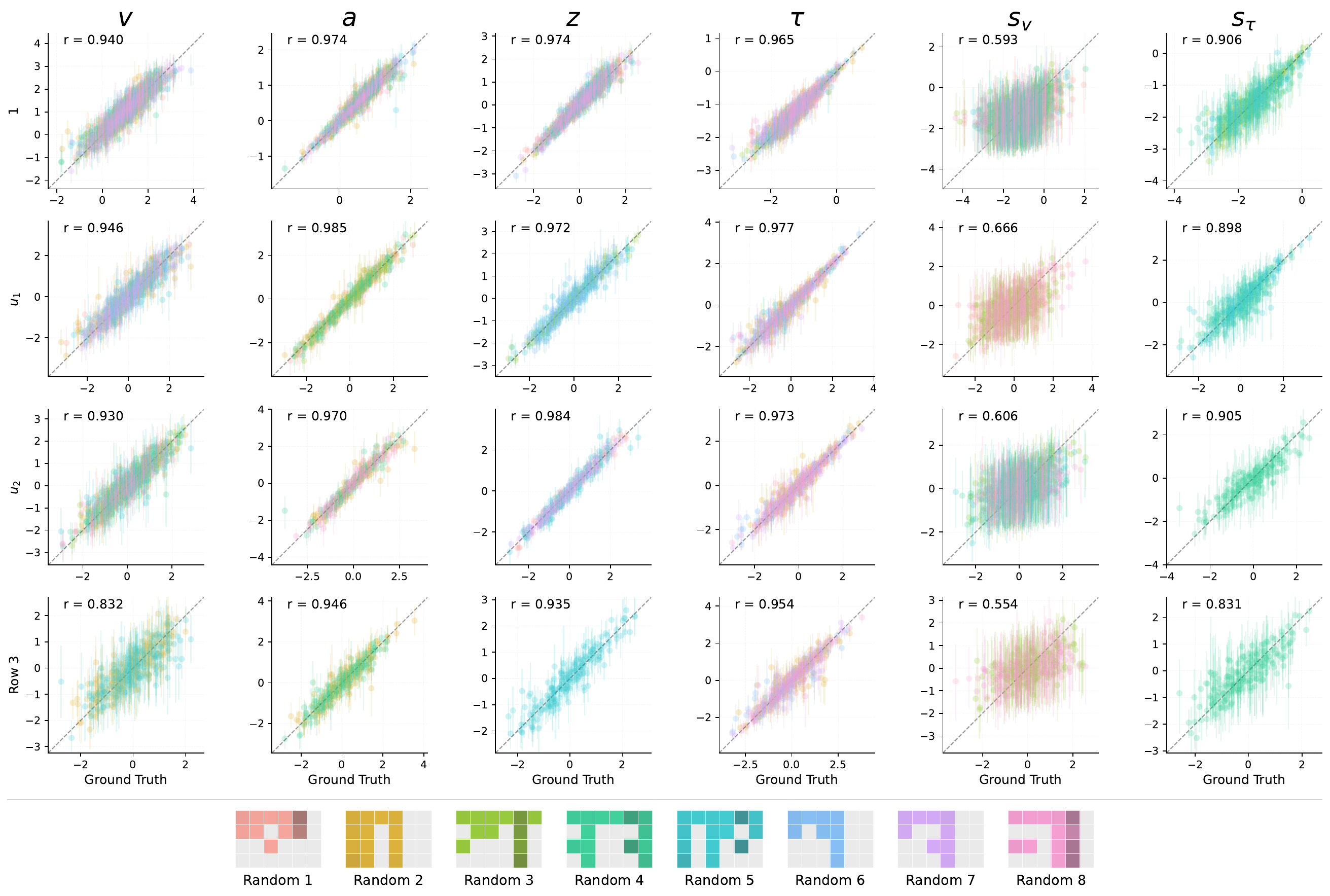}
    \includegraphics[width=0.96\linewidth]{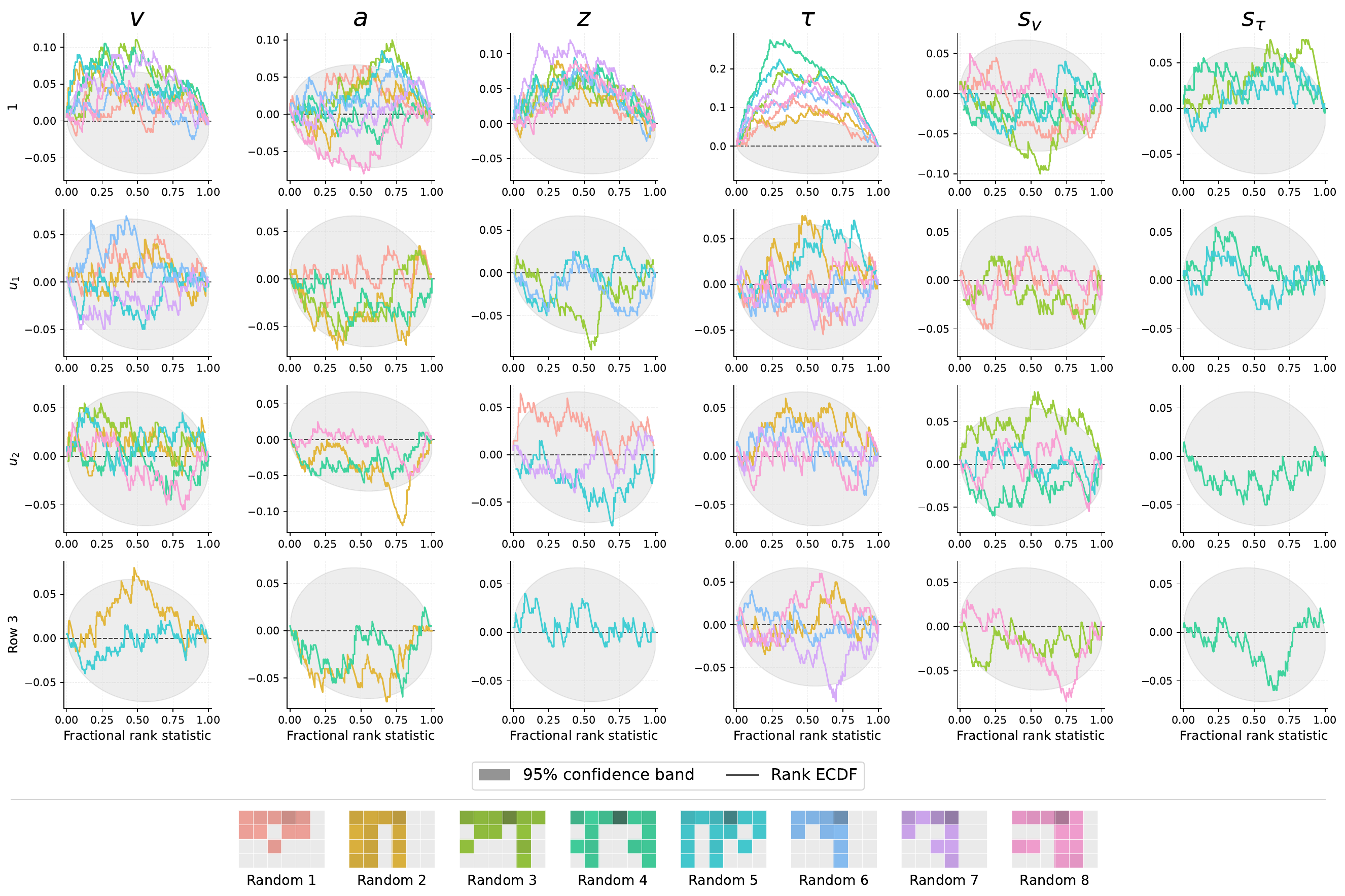}
    \caption{Random ensemble parameter recovery (\emph{top}) and calibration ECDF (\emph{bottom}) for meta-amortized model family for DDM. The pixel legends at the bottom indicates the parameter mask for each benchmark design configurations.}
    \label{fig:ddm-ensemble-r}
\end{figure}

\begin{figure}[!h]
    \centering
    \includegraphics[width=0.96\linewidth]{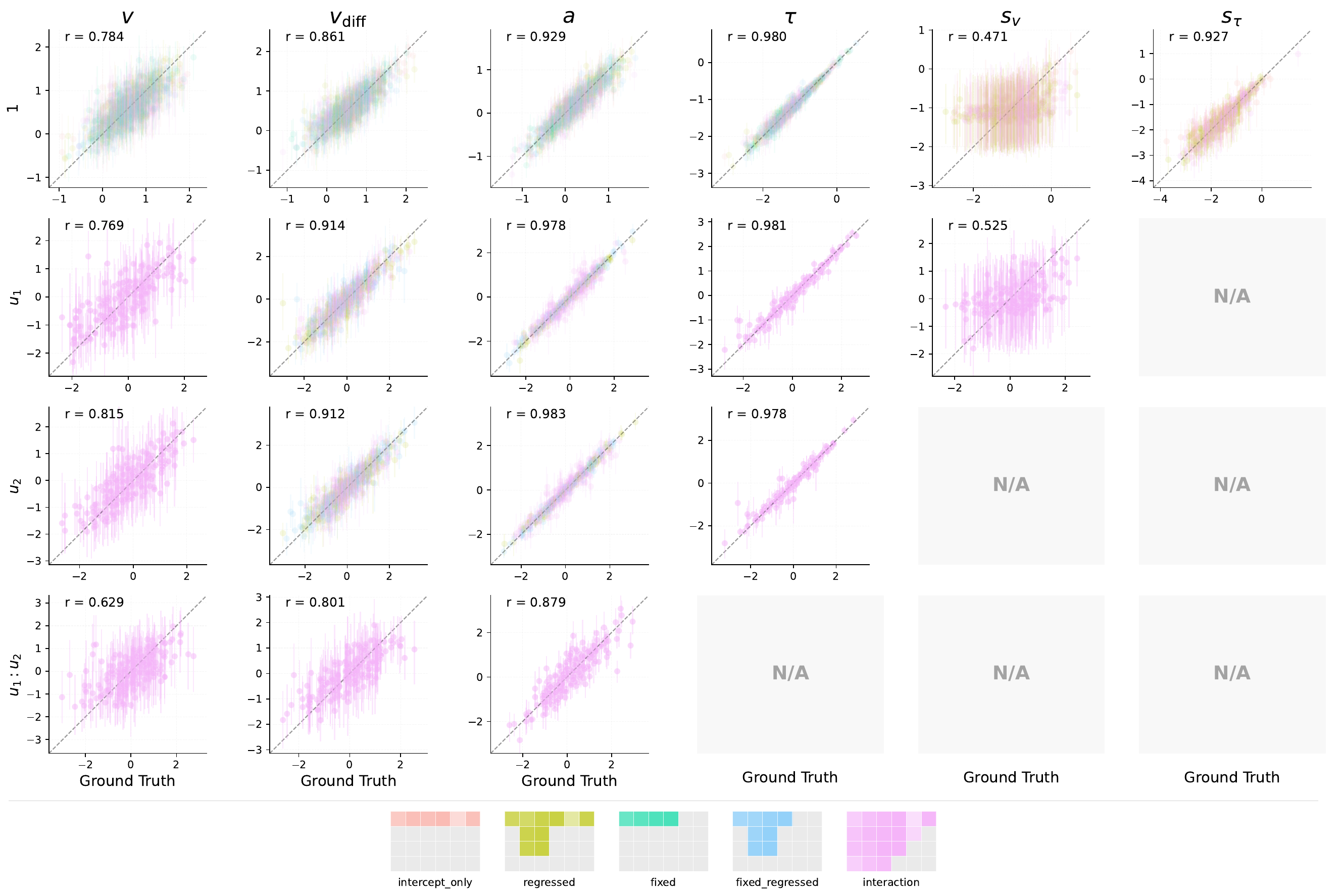}
    \includegraphics[width=0.96\linewidth]{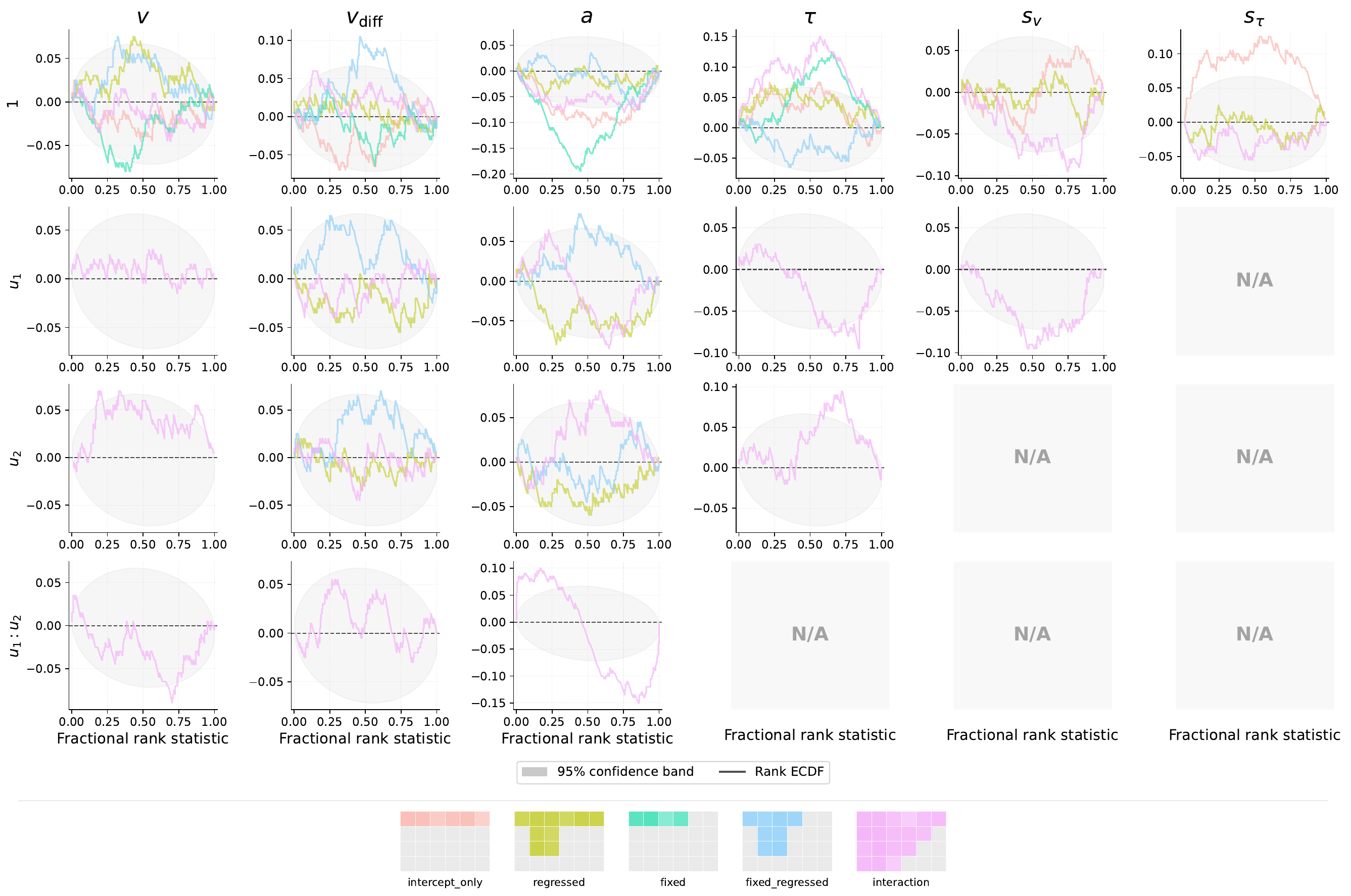}
    \caption{Ensemble parameter recoveries for CogFormer-$\mathcal{F}$ applied to the five design configurations for the RDM benchmark cases overlayed in their corresponding mapping to the shared coefficient matrix. The pixel thumbnails indicates the active estimation targets for each design configuration, with lighter shades of color depicting higher correlations $r$. The per-cell $r$-s are averaged across cases.}
    \label{fig:rdm-ensemble}
\end{figure}

\begin{figure}[!h]
    \centering
    \includegraphics[width=0.96\linewidth]{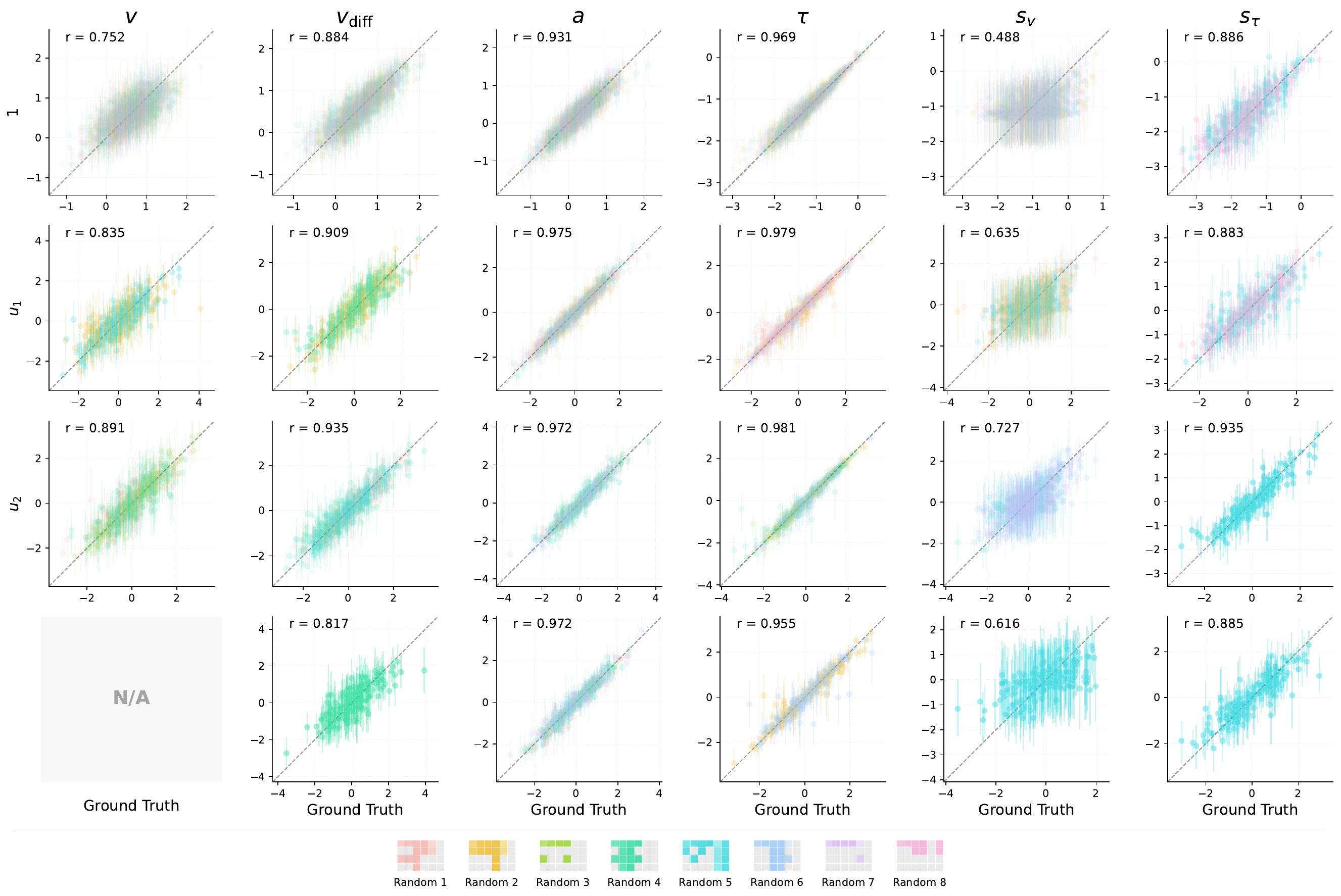}
    \includegraphics[width=0.96\linewidth]{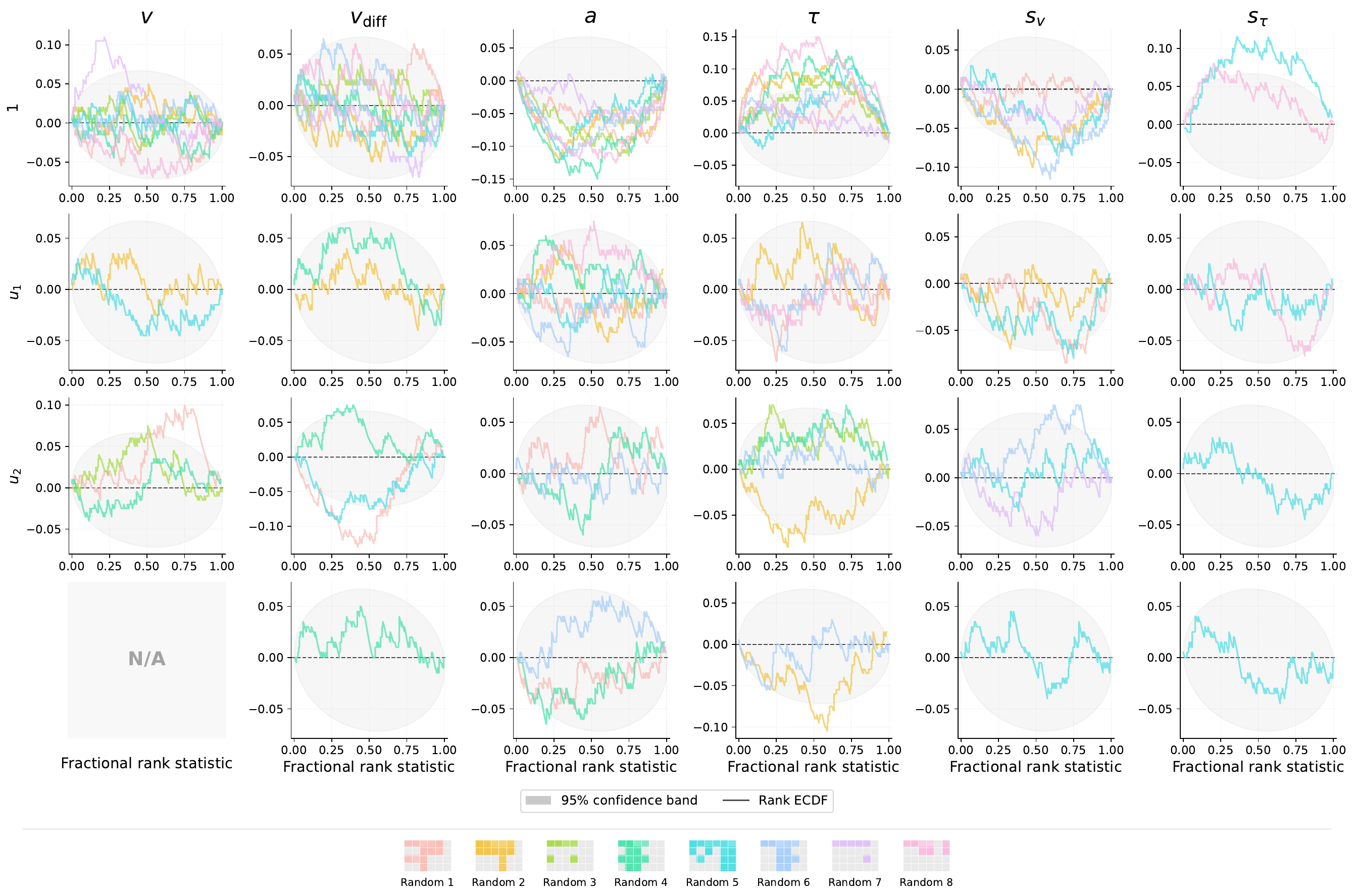}
    \caption{Random ensemble parameter recovery (\emph{top}) and calibration ECDF (\emph{bottom}) for meta-amortized model family for RDM. The pixel legends at the bottom indicates the parameter mask for each benchmark design configurations.}
    \label{fig:rdm-ensemble-r}
\end{figure}

\begin{figure}[!h]
    \centering
    \includegraphics[width=0.96\linewidth]{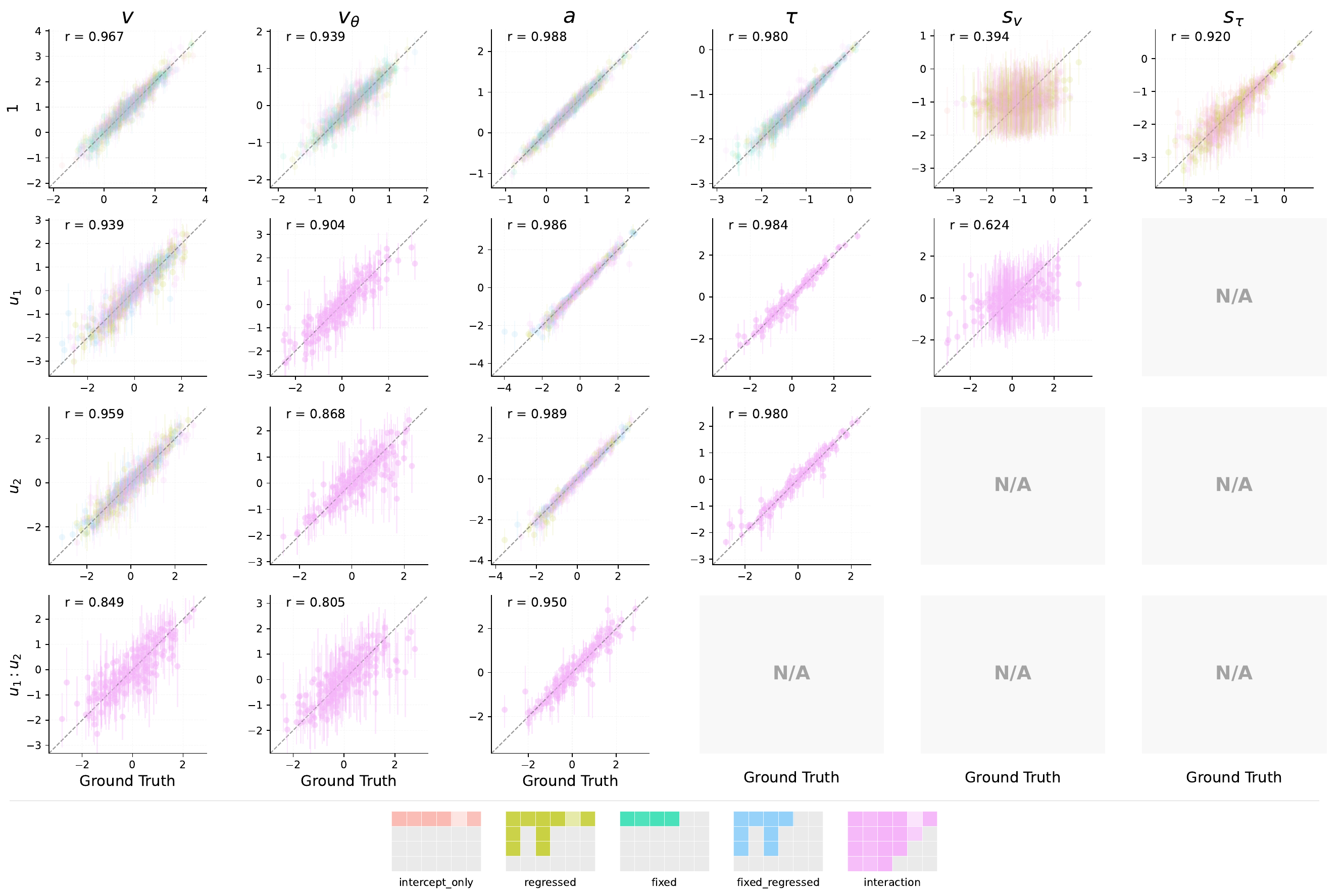}
    \includegraphics[width=0.96\linewidth]{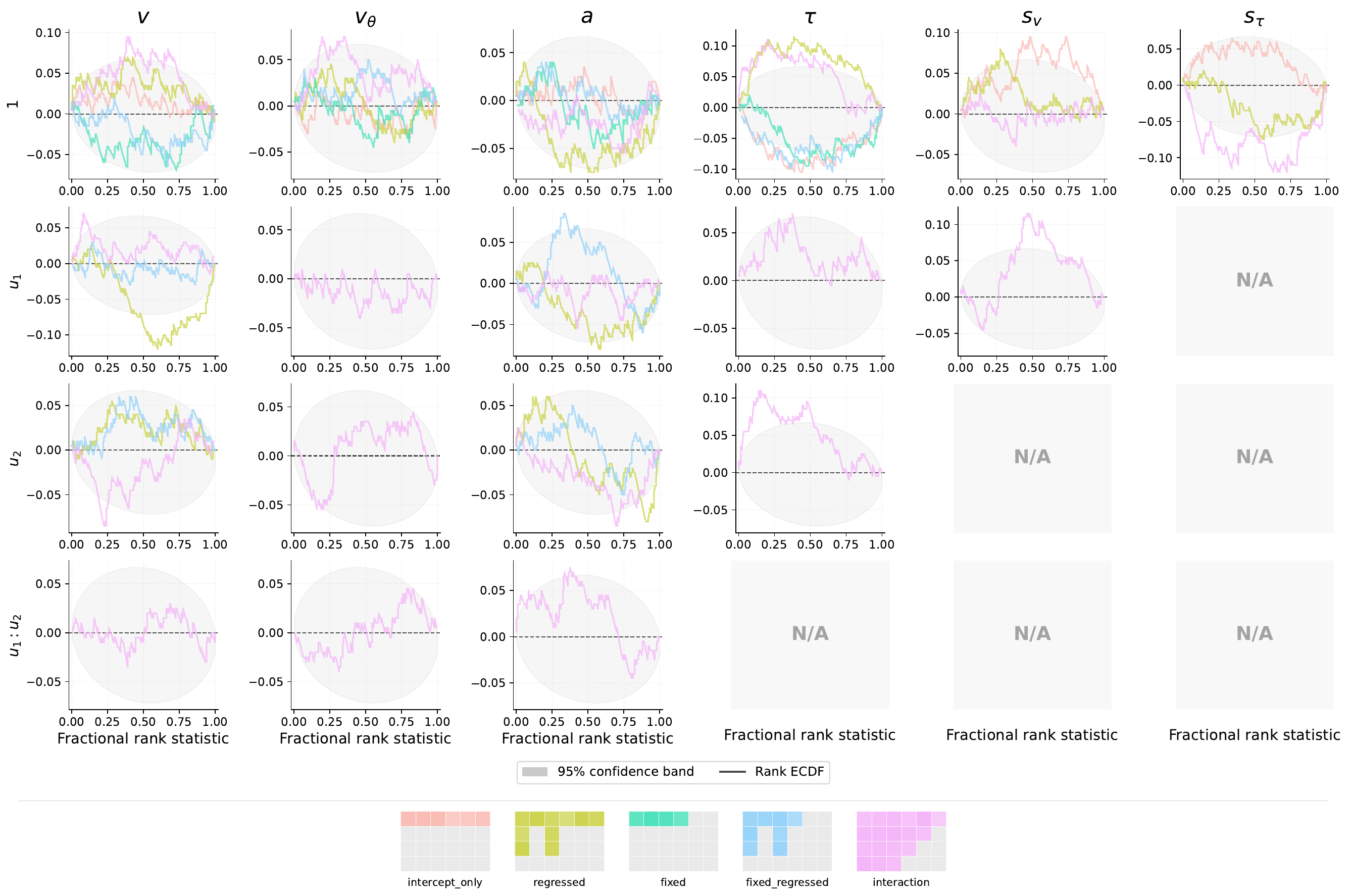}
    \caption{Ensemble parameter recoveries for CogFormer-$\mathcal{F}$ applied to the five design configurations for the CDM benchmark cases overlayed in their corresponding mapping to the shared coefficient matrix. The pixel thumbnails indicates the active estimation targets for each design configuration, with lighter shades of color depicting higher correlations $r$. The per-cell $r$-s are averaged across cases.}
    \label{fig:cdm-ensemble}
\end{figure}

\begin{figure}[!h]
    \centering
    \includegraphics[width=0.96\linewidth]{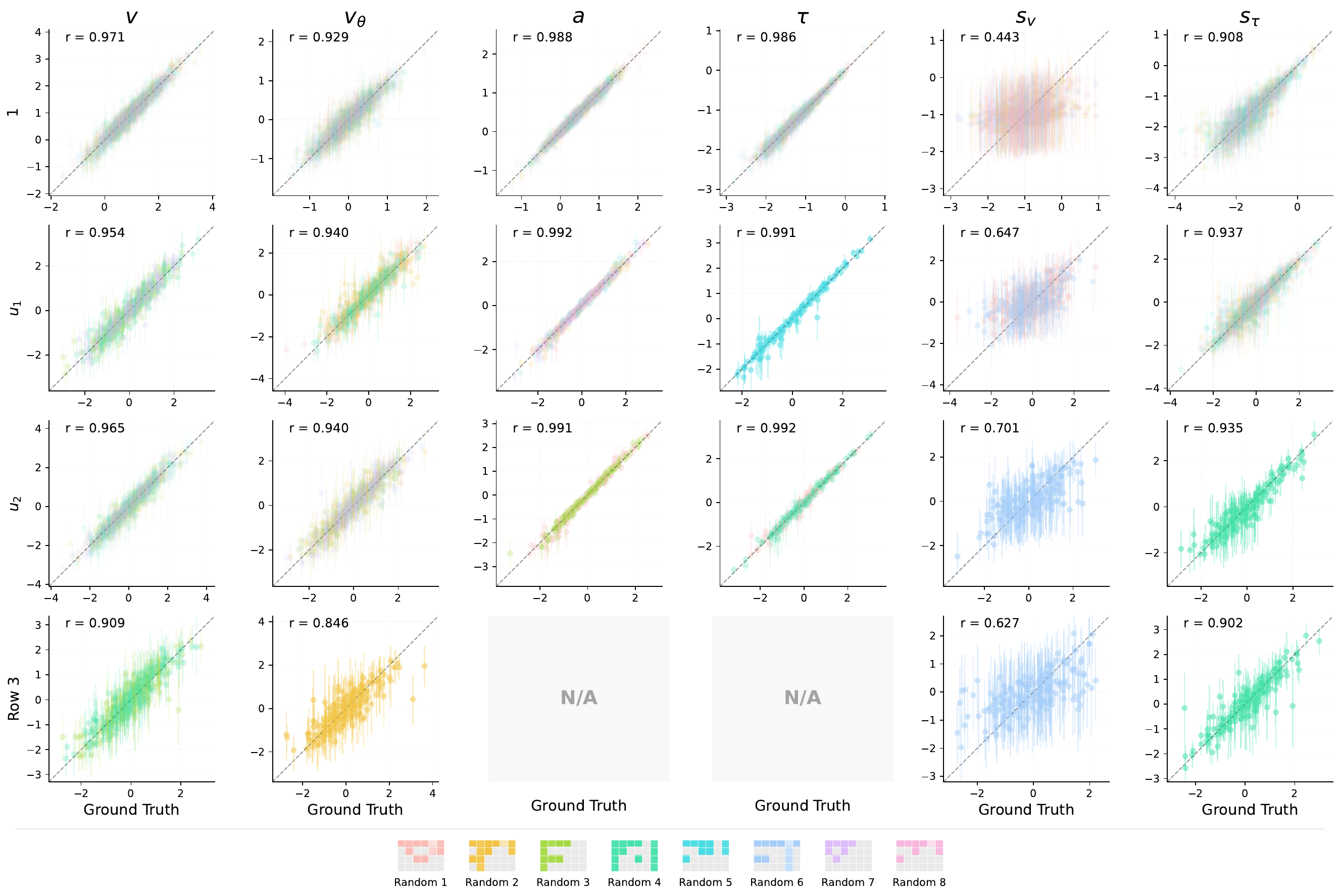}
    \includegraphics[width=0.96\linewidth]{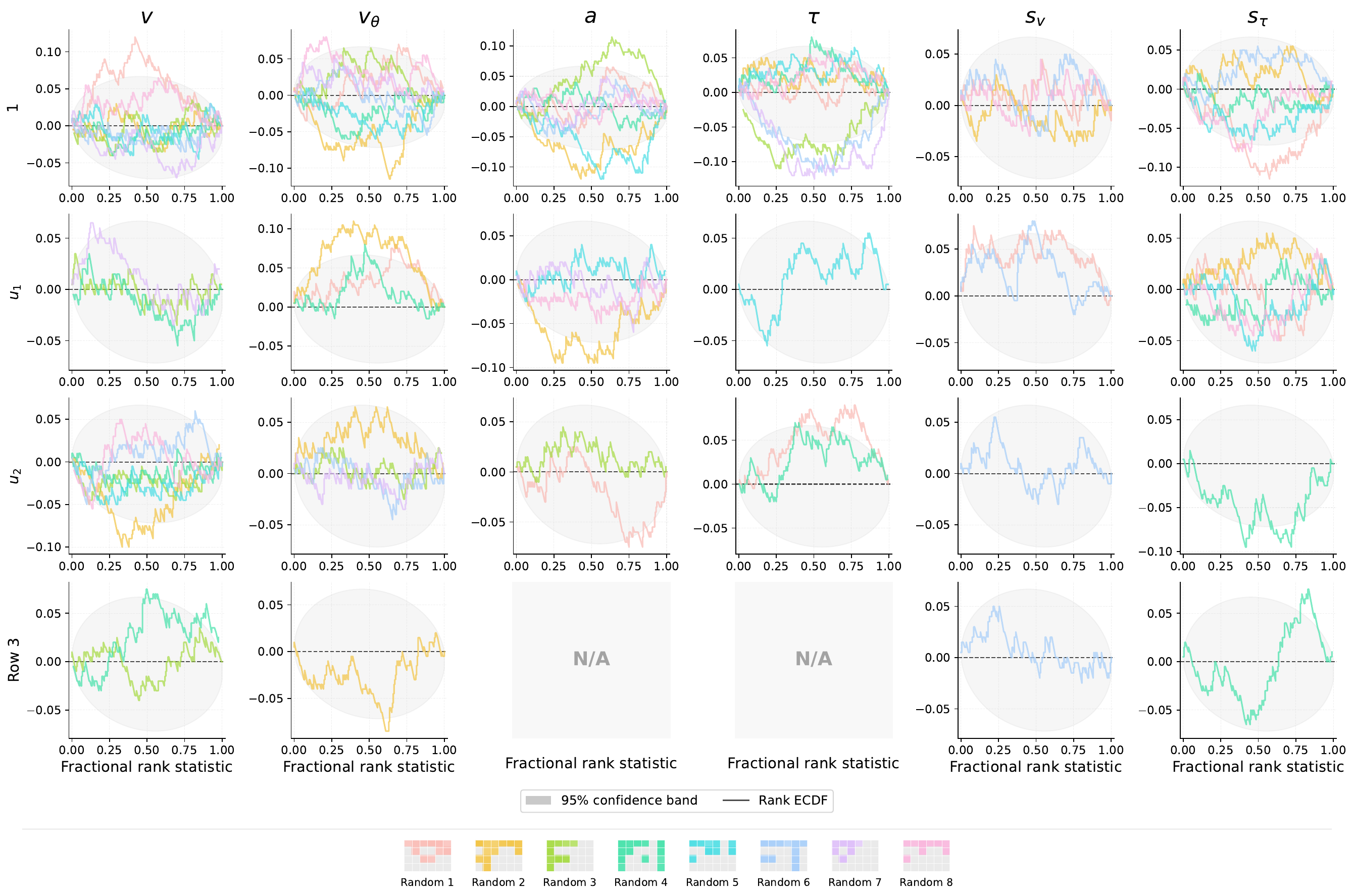}
    \caption{Random ensemble parameter recovery (\emph{top}) and calibration ECDF (\emph{bottom}) for meta-amortized model family for CDM. The pixel legends at the bottom indicates the parameter mask for each benchmark design configurations.}
    \label{fig:cdm-ensemble-r}
\end{figure}

 \clearpage
\section{Additional results for benchmarking}
\label{app:additional-benchmarks}

\begin{table}[h]
    \centering
    \caption{Corresponding frameworks, figures, and colors for the additional results for benchmarking.}
    \begin{tabular}{lcc}
    \toprule
    \textbf{Frameworks}&\textbf{Figures}&\textbf{Color Scheme}\\
    \midrule
    Baseline&15-19 (DDM) / 30-34 (RDM) / 45-49 (CDM) &Green\\
    CogFormer-$\mathcal{F}$&20-24 (RDM) / 35-39 (CDM) / 50-54 (DDM)&Red\\
    CogFormer-$\mathcal{C}$&25-29 (RDM) / 40-44 (CDM) / 55-59 (DDM)&Blue\\
    \bottomrule
    \end{tabular}
    \label{tab:benchmark-guide}
\end{table}

Below, we present results for all benchmarking cases. To guide readers through these results, we represent~\autoref{tab:benchmark-guide} that outlines the corresponding figures and visual aids for the individual benchmark cases and their corresponding frameworks. All figures preserve the layout of their shared coefficient matrix $\rmB$ to demonstrate that they are structured and masked inputs to their respective frameworks.


\begin{figure}[!h]
    \centering
    \includegraphics[width=0.96\linewidth]{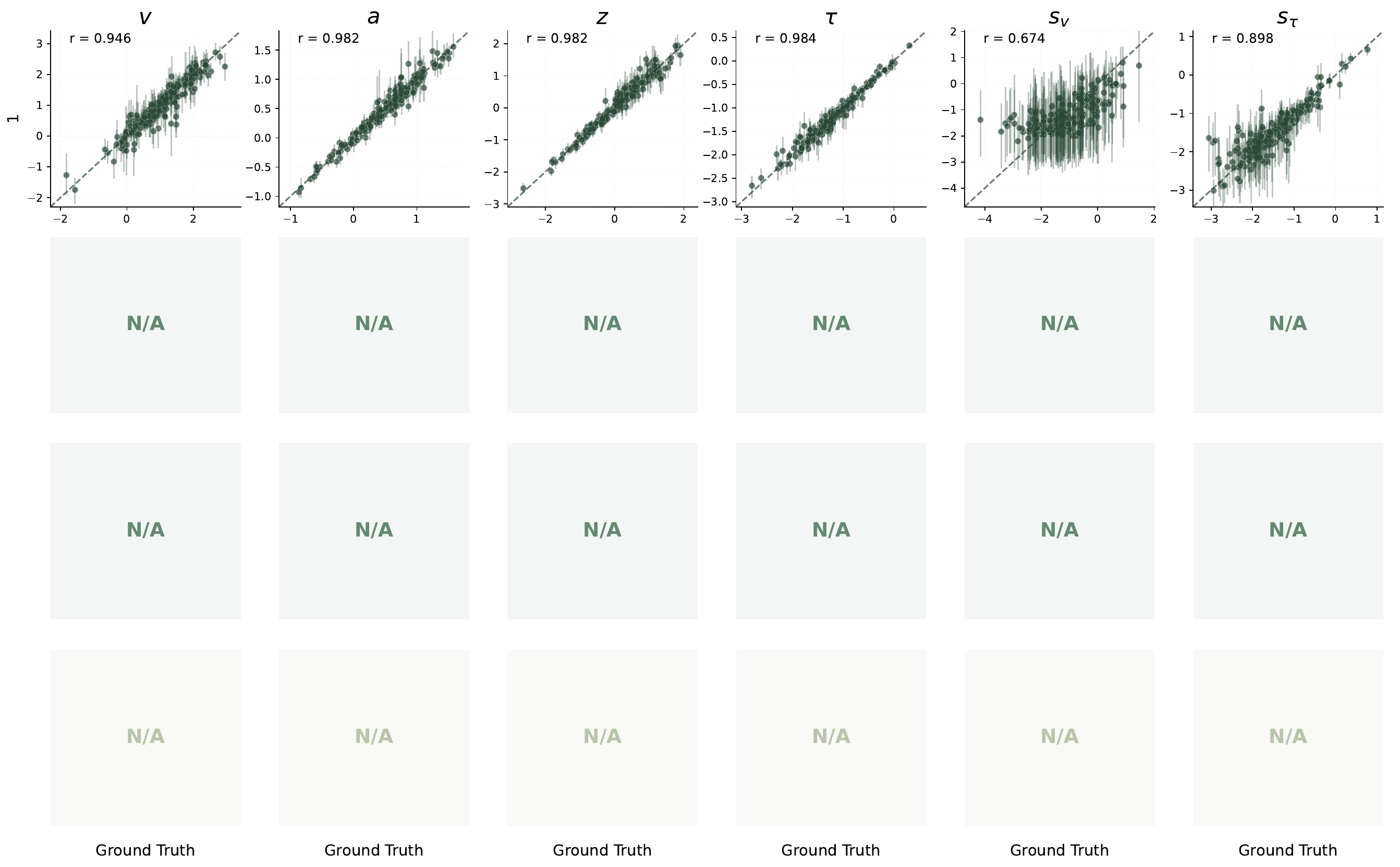}
    \includegraphics[width=0.96\linewidth]{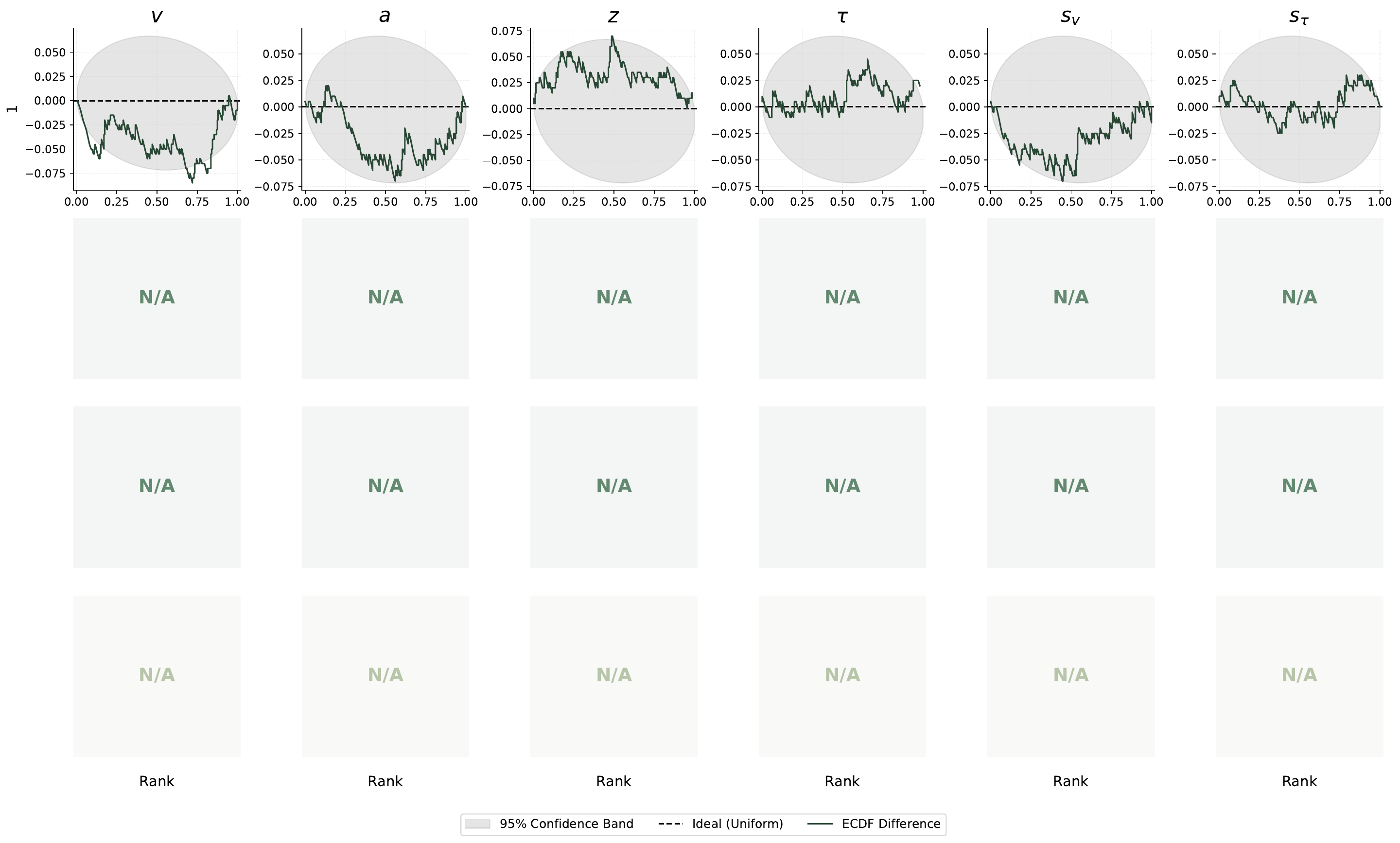}
    \includegraphics[width=0.96\linewidth]{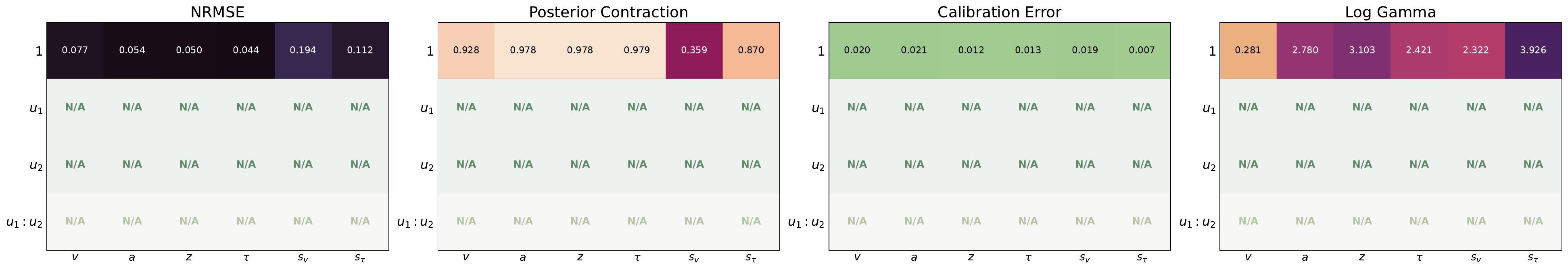}
    \caption{Parameter recovery (\emph{top}), calibration ECDF (\emph{middle}), and validation metrics (NRMSE, calibration error, and posterior contraction) for DDM baseline (Case \textbf{intercept\_only}).}
    \label{fig:ddm-bf-intercept-only}
\end{figure}

\begin{figure}[!h]
    \centering
    \includegraphics[width=0.97\linewidth]{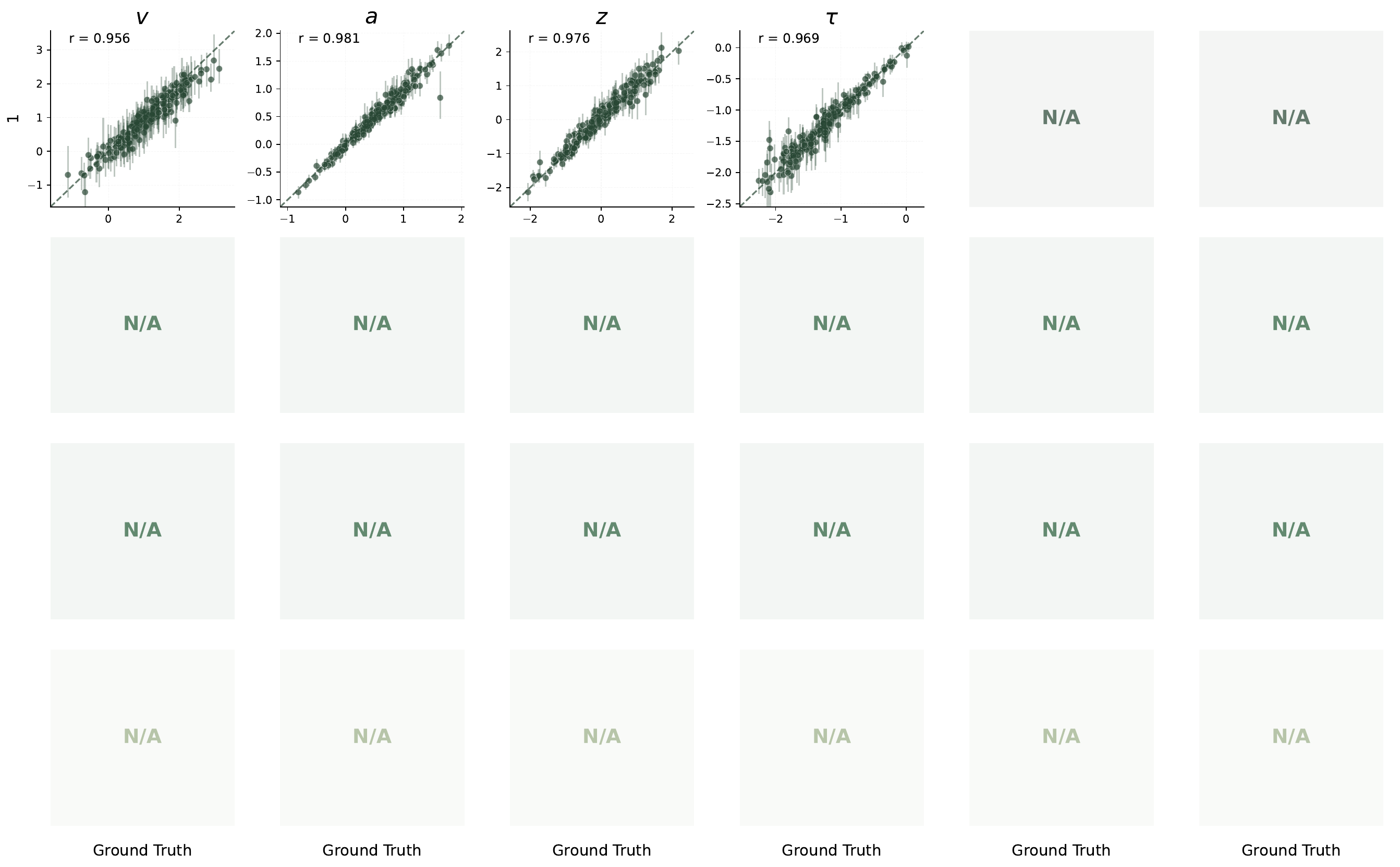}
    \includegraphics[width=0.97\linewidth]{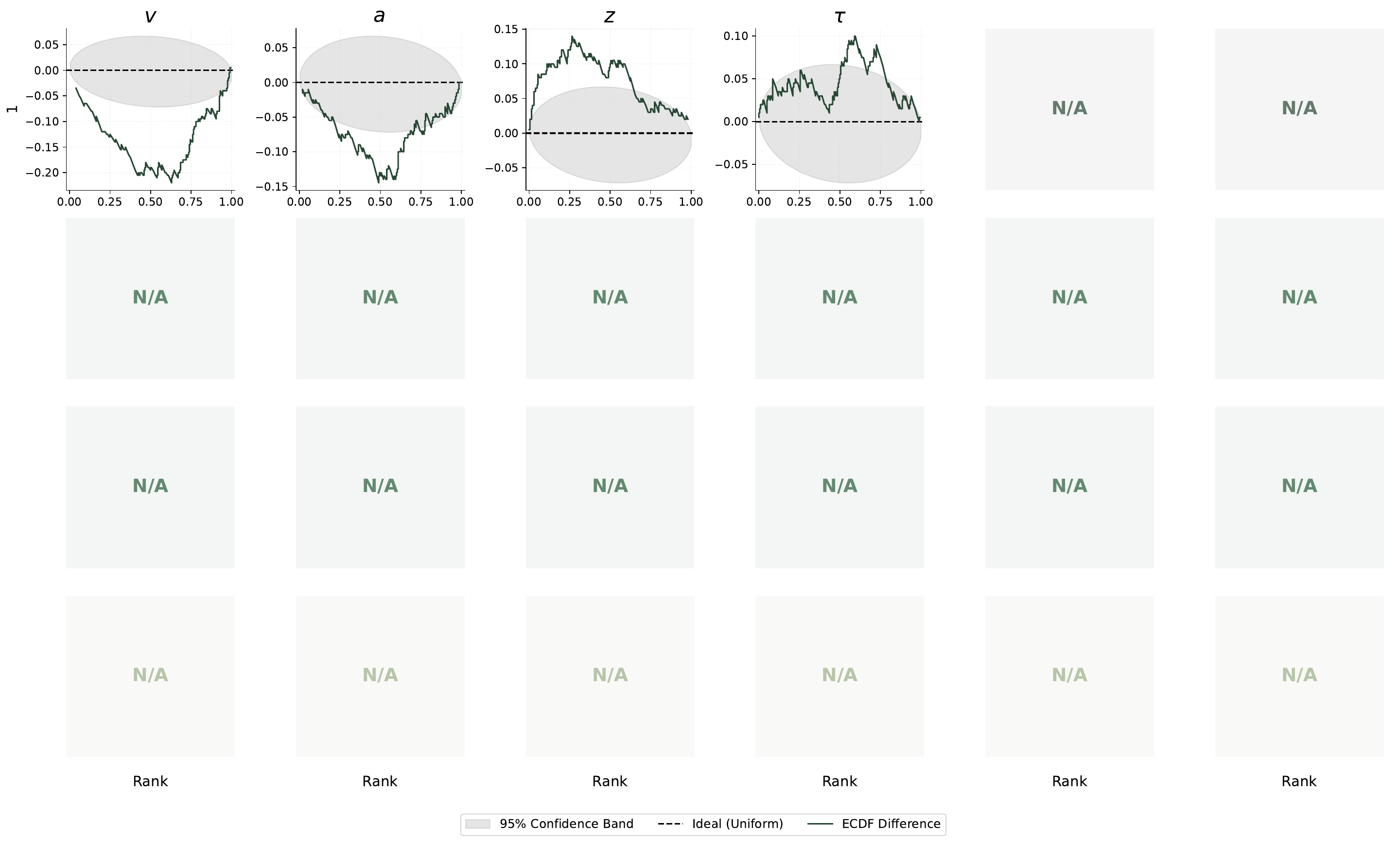}
    \includegraphics[width=0.97\linewidth]{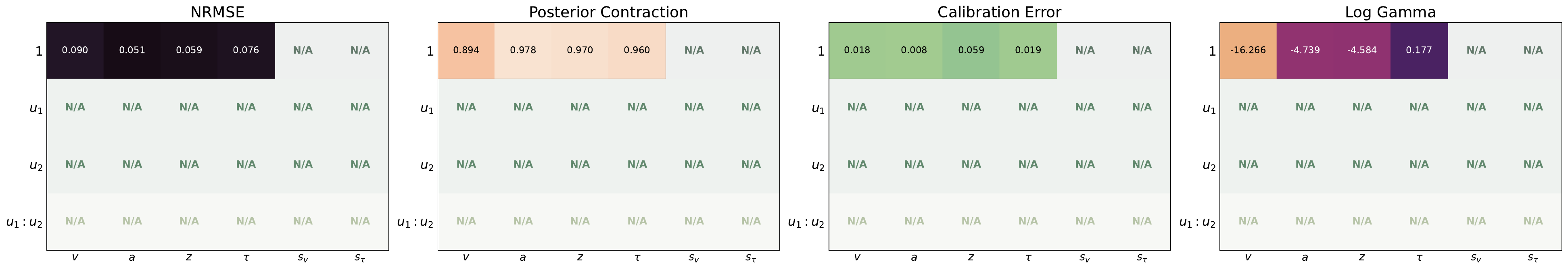}
    \caption{Parameter recovery (\emph{top}), calibration ECDF (\emph{middle}), and validation metrics (NRMSE, calibration error, and posterior contraction) for DDM baseline (Case \textbf{fixed}).}
    \label{fig:ddm-bf-fixed}
\end{figure}

\begin{figure}[!h]
    \centering
    \includegraphics[width=0.97\linewidth]{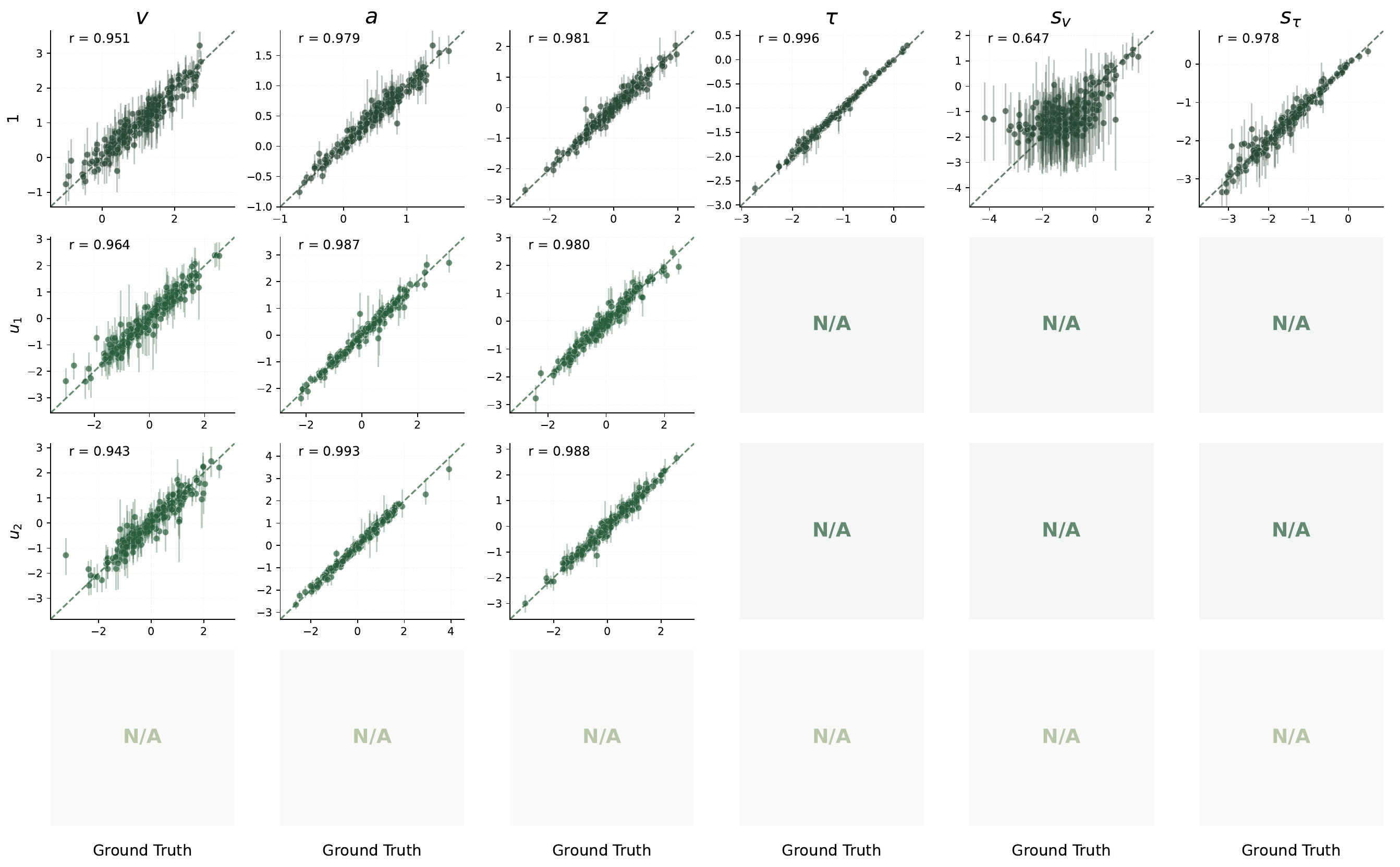}
    \includegraphics[width=0.97\linewidth]{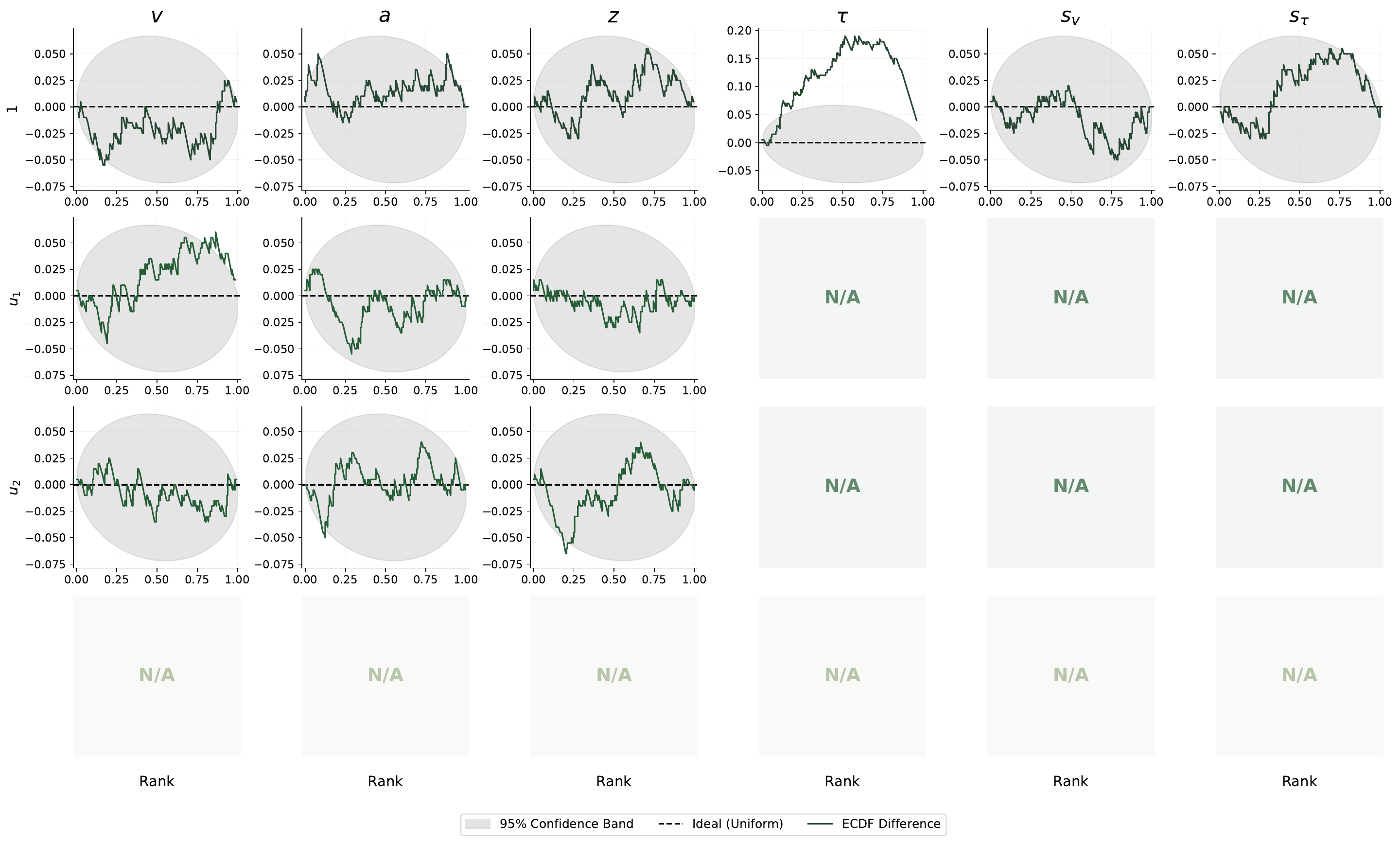}
    \includegraphics[width=0.97\linewidth]{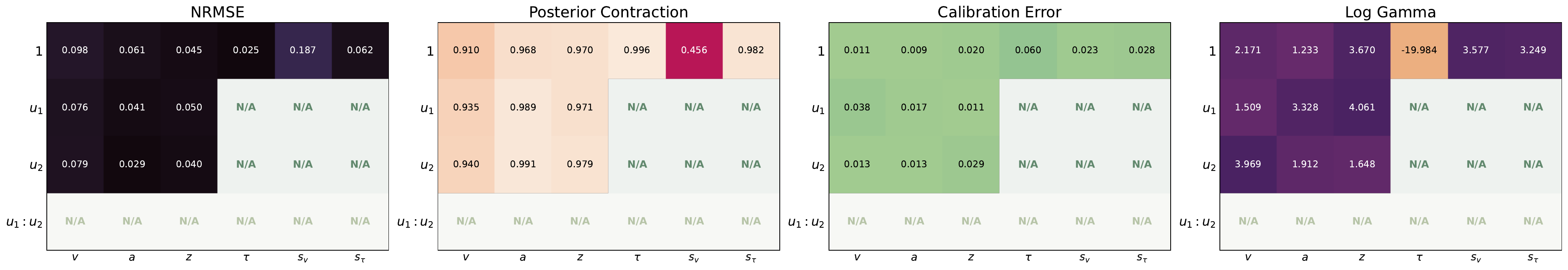}
    \caption{Parameter recovery (\emph{top}), calibration ECDF (\emph{middle}), and validation metrics (NRMSE, calibration error, and posterior contraction) for DDM baseline (Case \textbf{regressed}).}
    \label{fig:ddm-bf-regressed}
\end{figure}

\begin{figure}[!h]
    \centering
    \includegraphics[width=0.97\linewidth]{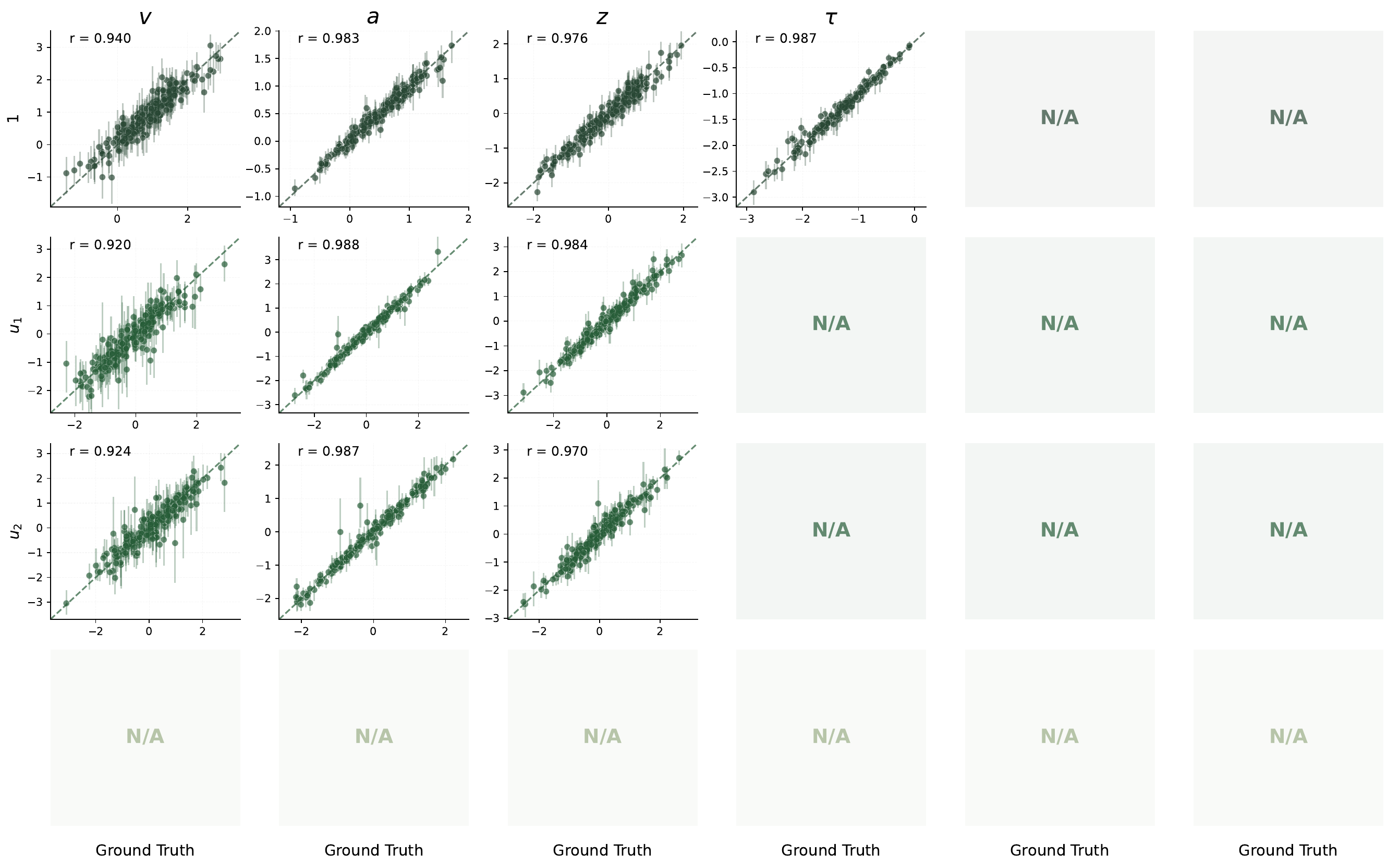}
    \includegraphics[width=0.97\linewidth]{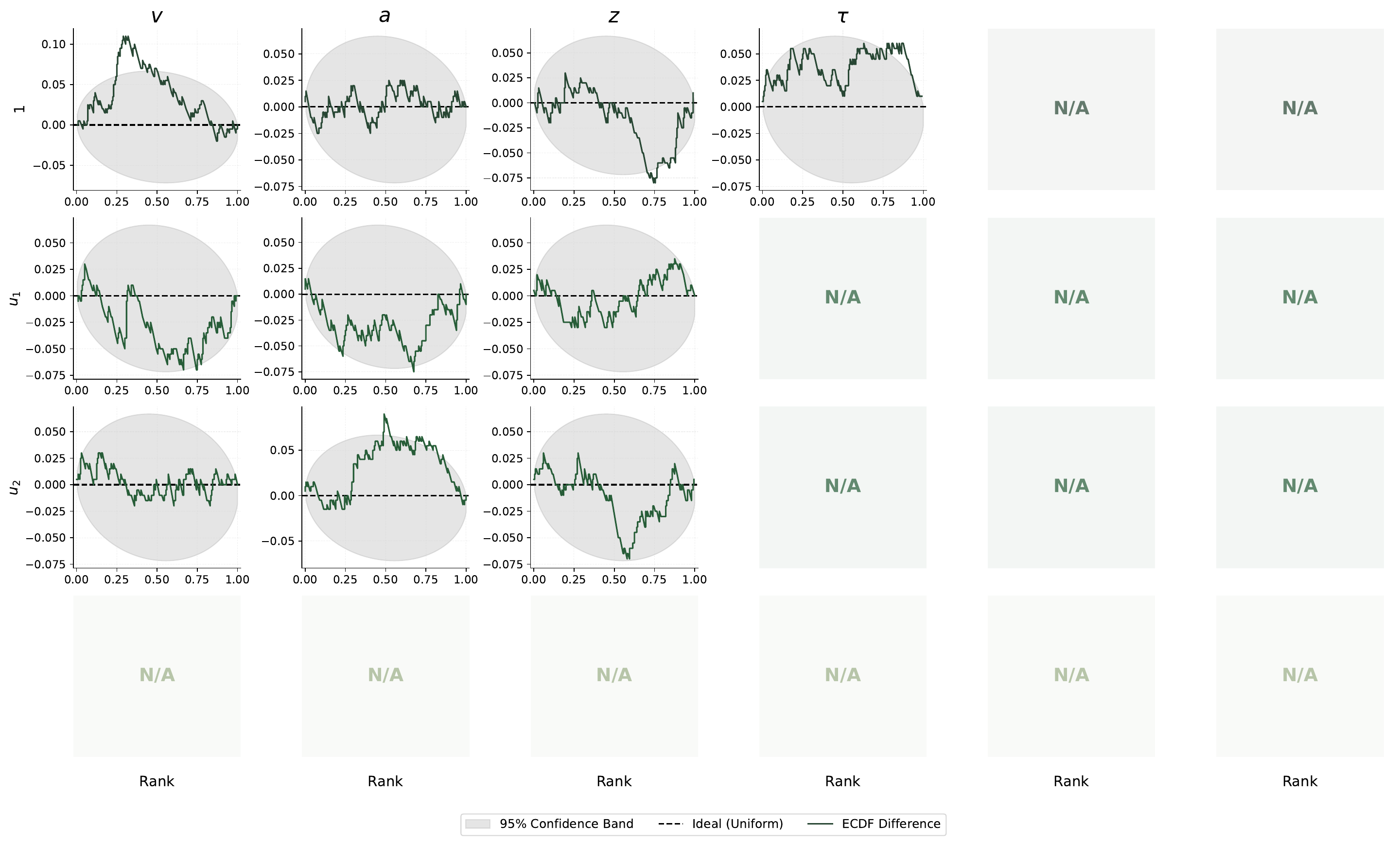}
    \includegraphics[width=0.97\linewidth]{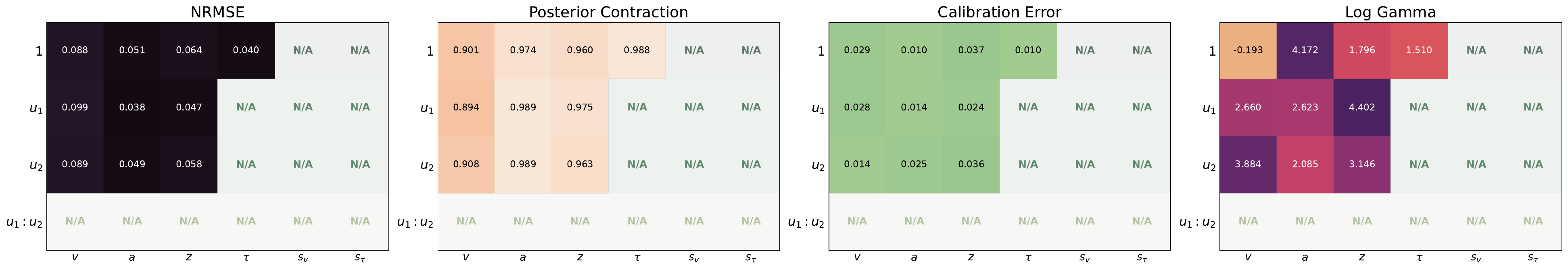}
    \caption{Parameter recovery (\emph{top}), calibration ECDF (\emph{middle}), and validation metrics (NRMSE, calibration error, and posterior contraction) for DDM baseline (Case \textbf{fixed\_regressed}).}
    \label{fig:ddm-bf-fixed-regressed}
\end{figure}

\begin{figure}[!h]
    \centering
    \includegraphics[width=0.97\linewidth]{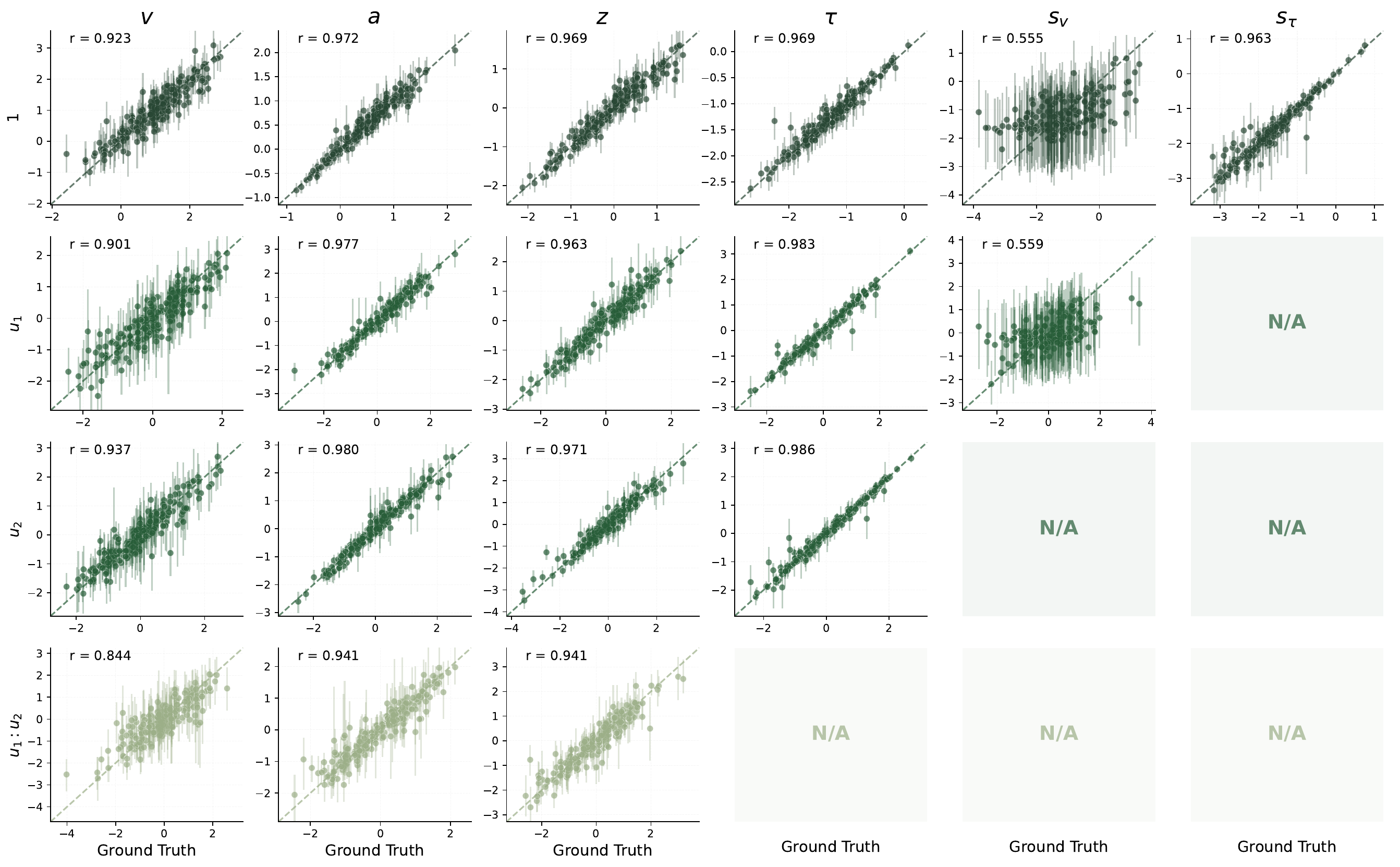}
    \includegraphics[width=0.97\linewidth]{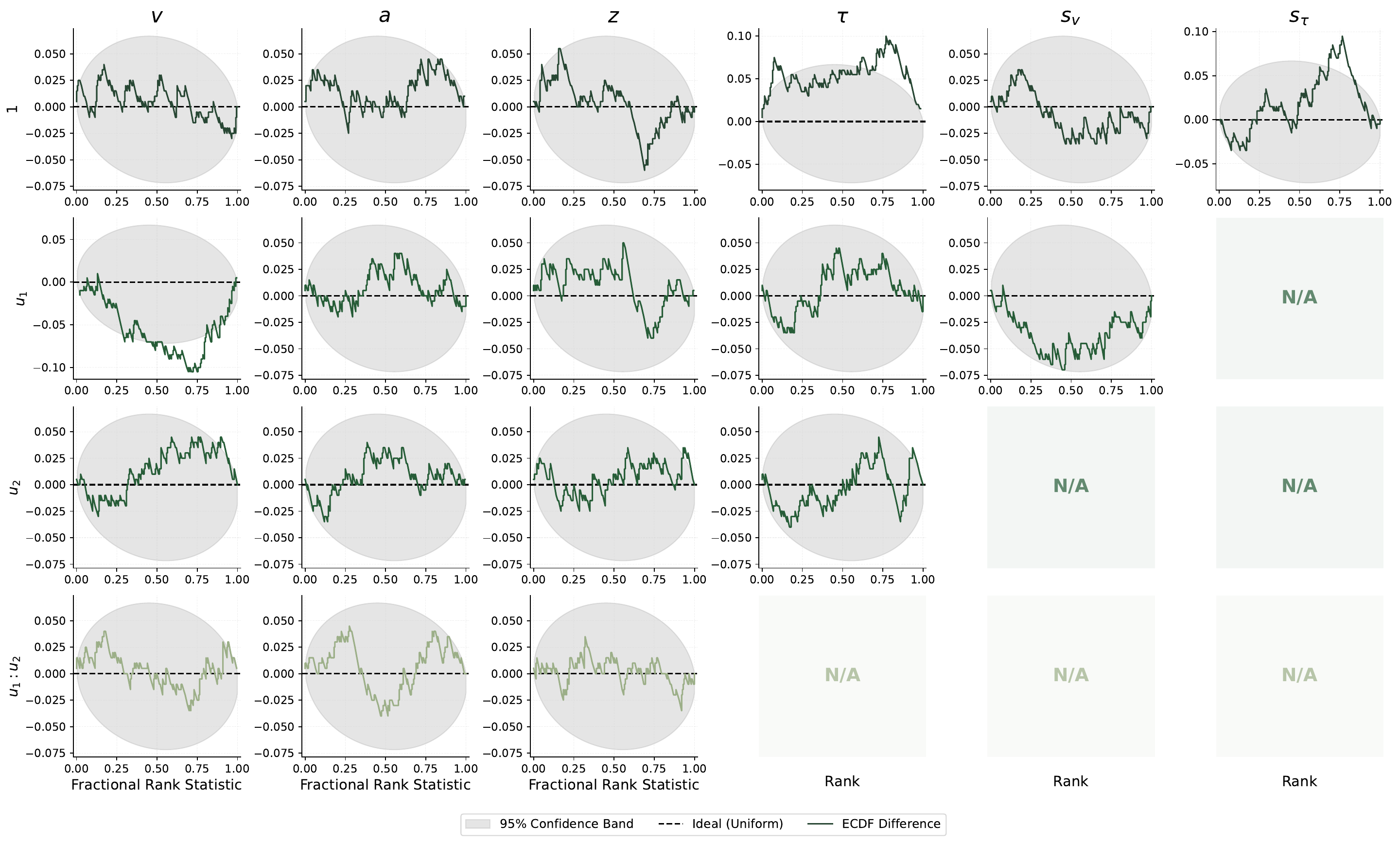}
    \includegraphics[width=0.97\linewidth]{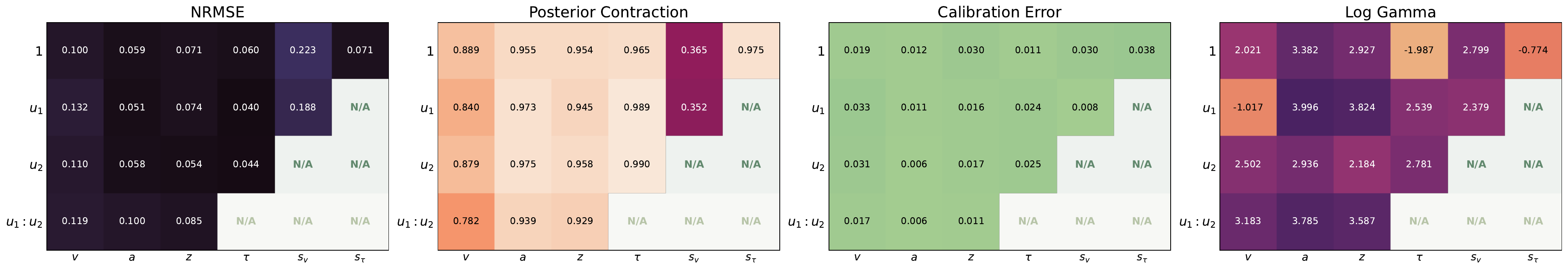}
    \caption{Parameter recovery (\emph{top}), calibration ECDF (\emph{middle}), and validation metrics (NRMSE, calibration error, and posterior contraction) for DDM baseline (Case \textbf{interaction}).}
    \label{fig:ddm-bf-interaction}
\end{figure}


\begin{figure}[!h]
    \centering
    \includegraphics[width=0.97\linewidth]{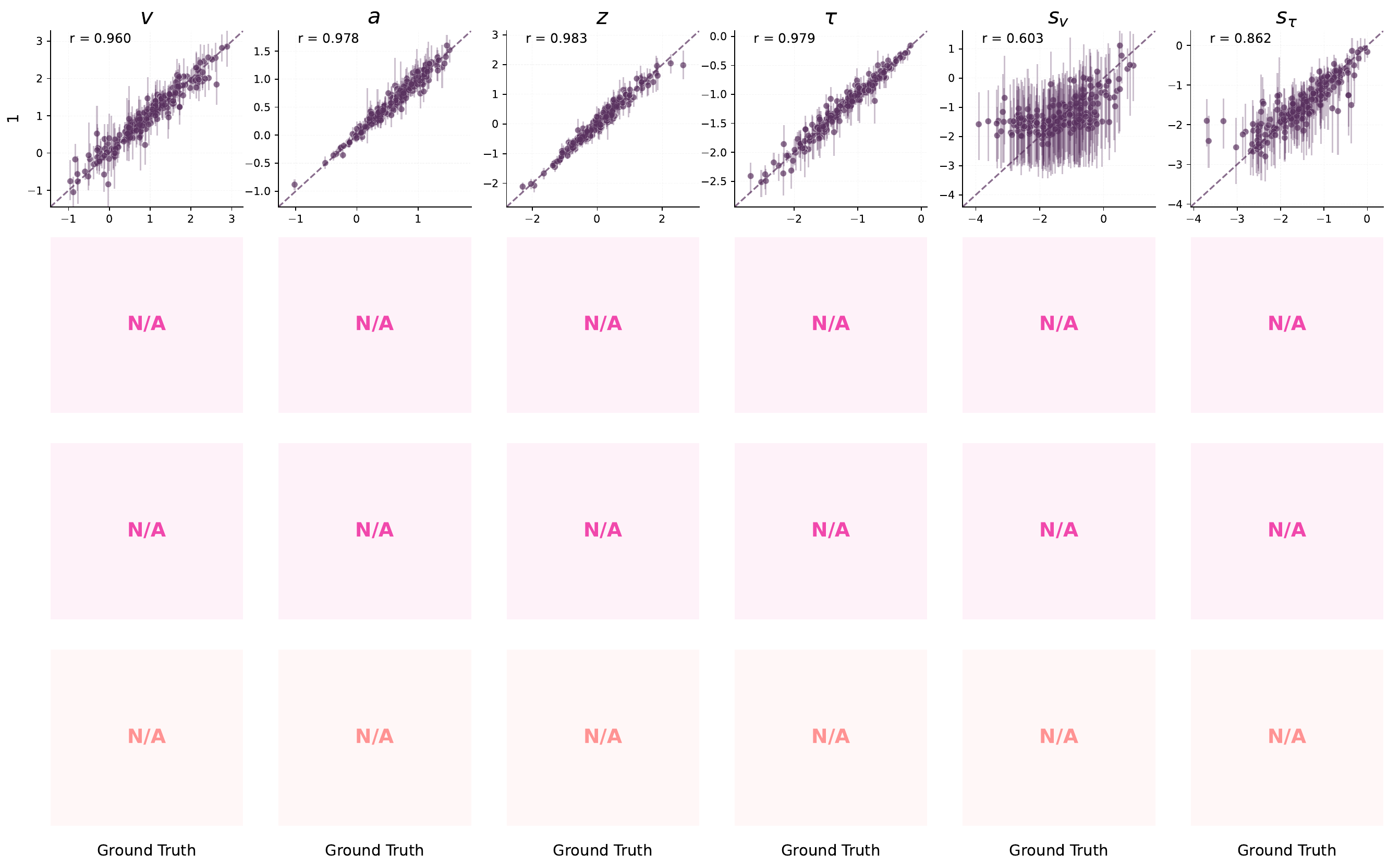}
    \includegraphics[width=0.97\linewidth]{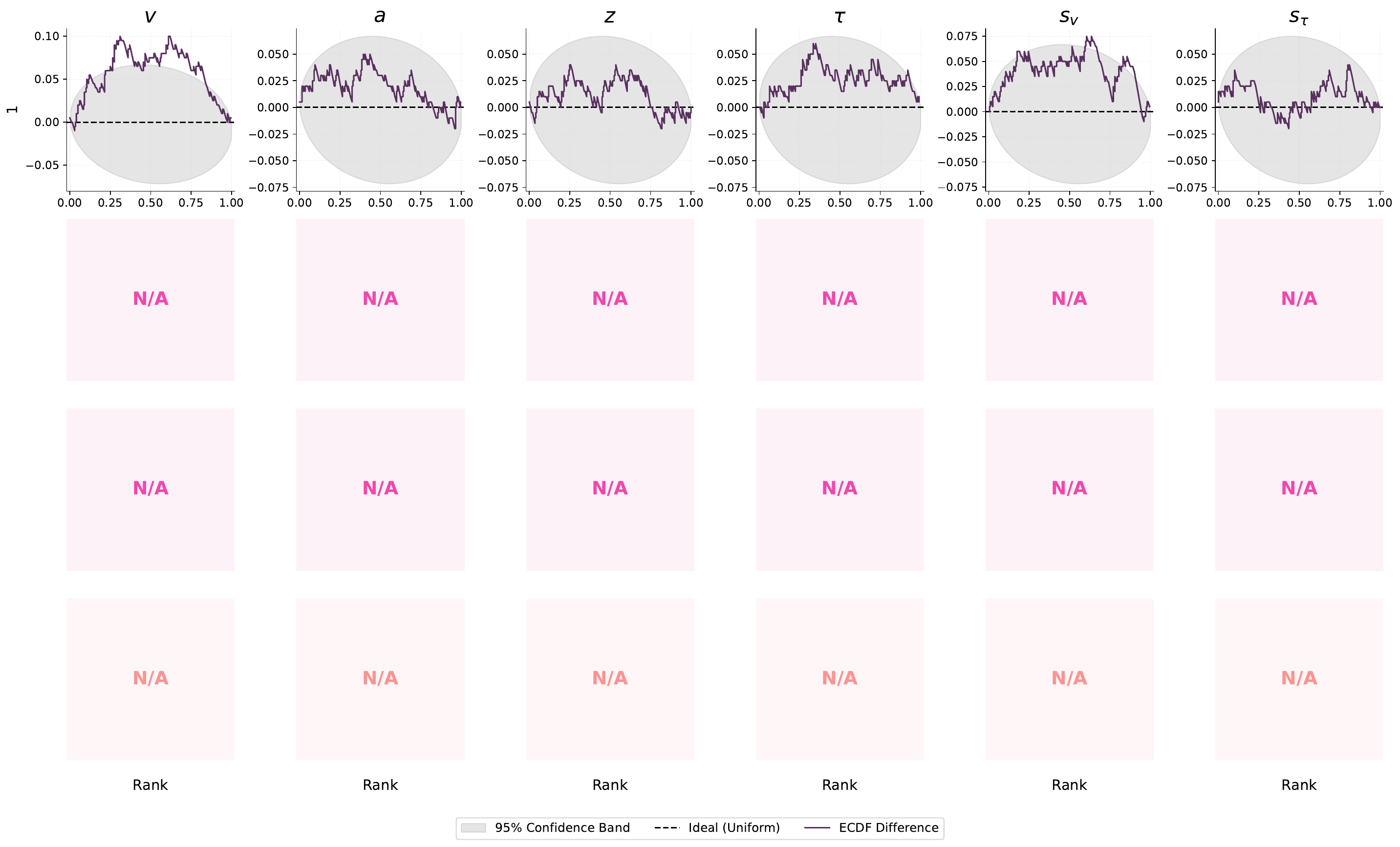}
    \includegraphics[width=0.97\linewidth]{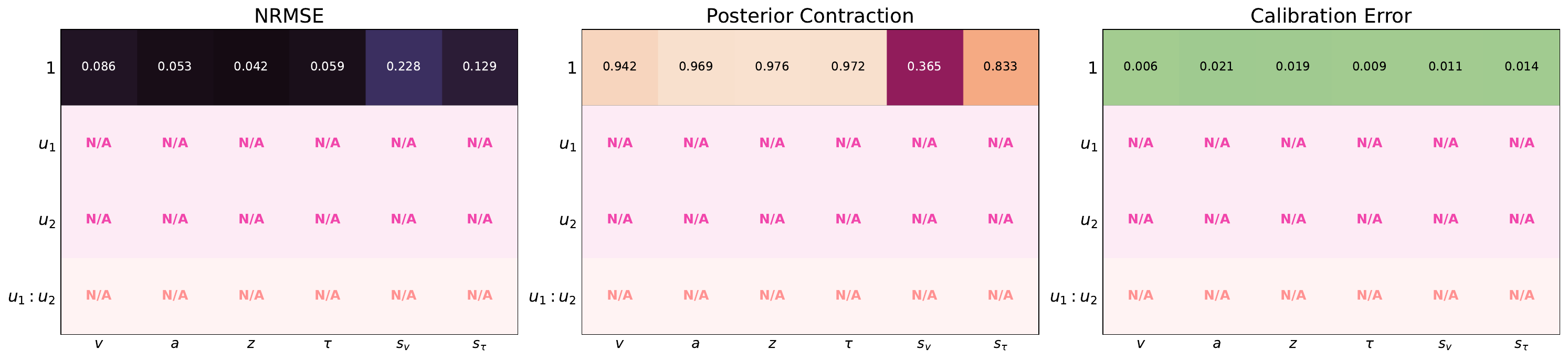}
    \caption{Parameter recovery (\emph{top}), calibration ECDF (\emph{middle}), and validation metrics (NRMSE, calibration error, and posterior contraction) for DDM model family (Case \textbf{intercept\_only}).}
    \label{fig:ddm-fm-intercept-only}
\end{figure}

\begin{figure}[!h]
    \centering
    \includegraphics[width=0.97\linewidth]{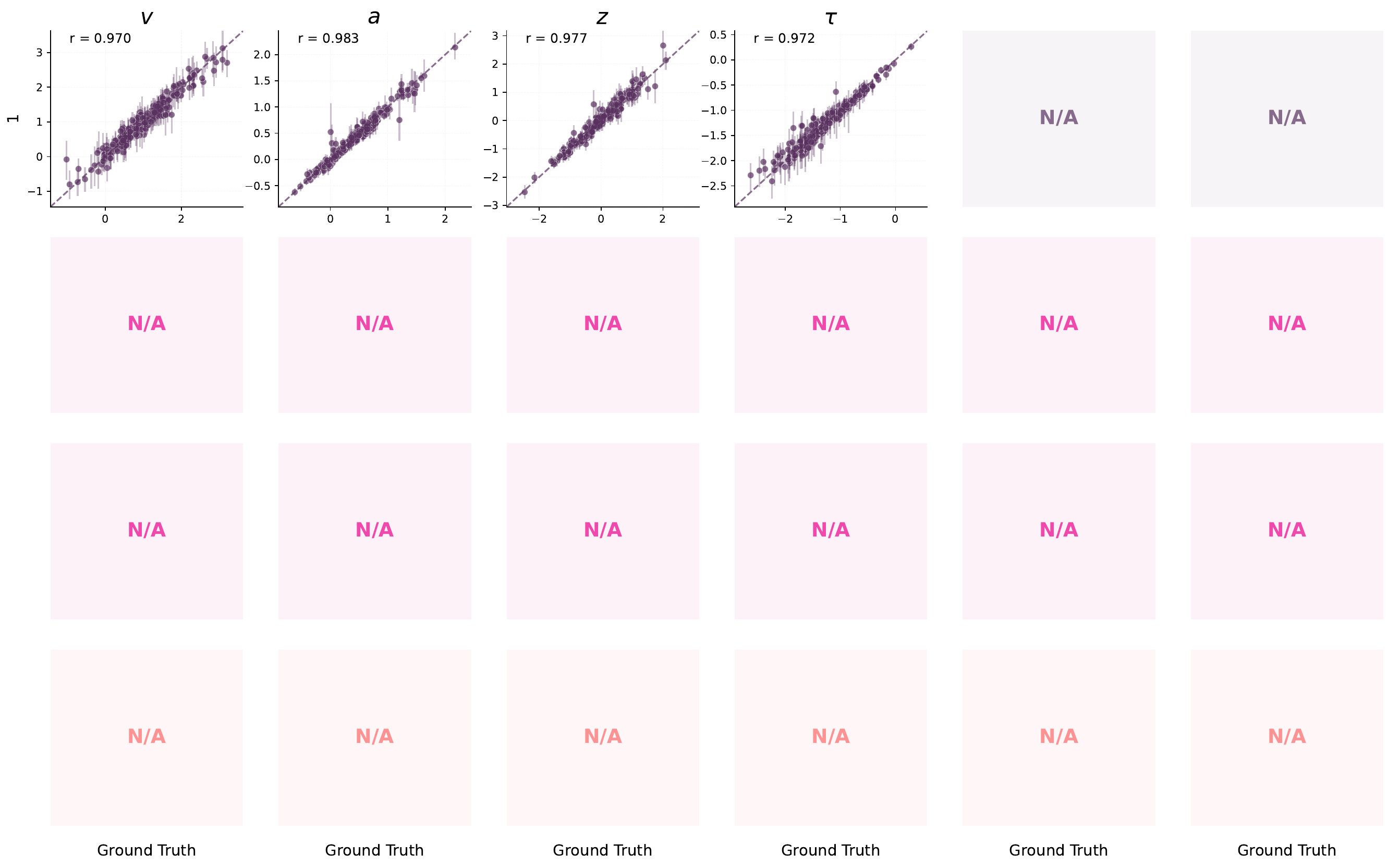}
    \includegraphics[width=0.97\linewidth]{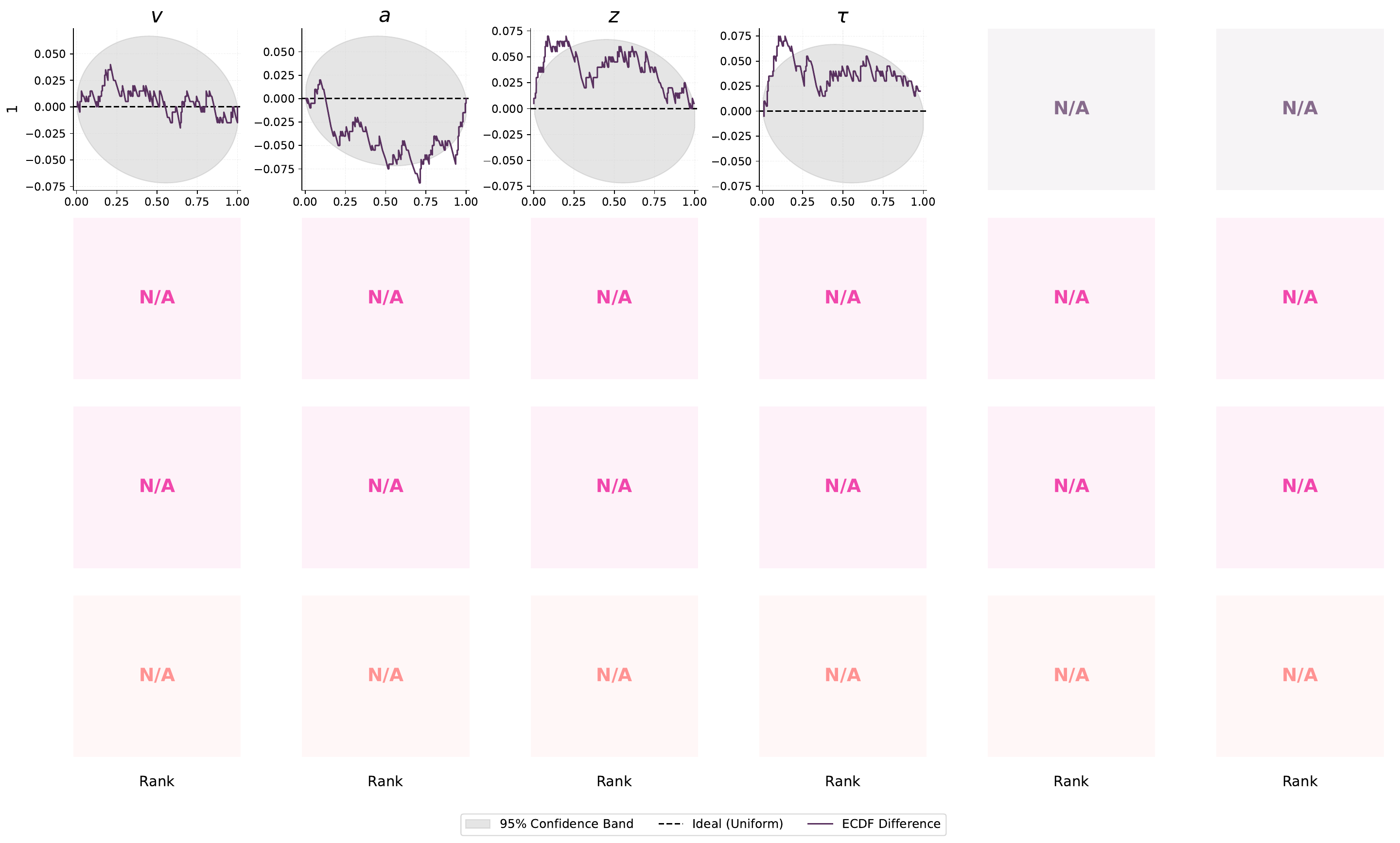}
    \includegraphics[width=0.97\linewidth]{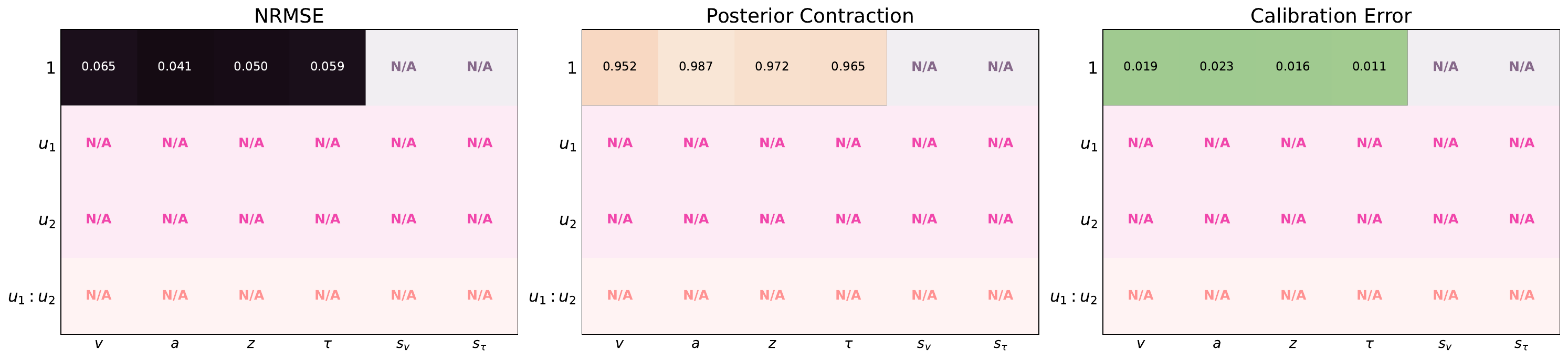}
    \caption{Parameter recovery (\emph{top}), calibration ECDF (\emph{middle}), and validation metrics (NRMSE, calibration error, and posterior contraction) for DDM model family (Case \textbf{fixed}).}
    \label{fig:ddm-fm-fixed}
\end{figure}

\begin{figure}[!h]
    \centering
    \includegraphics[width=0.97\linewidth]{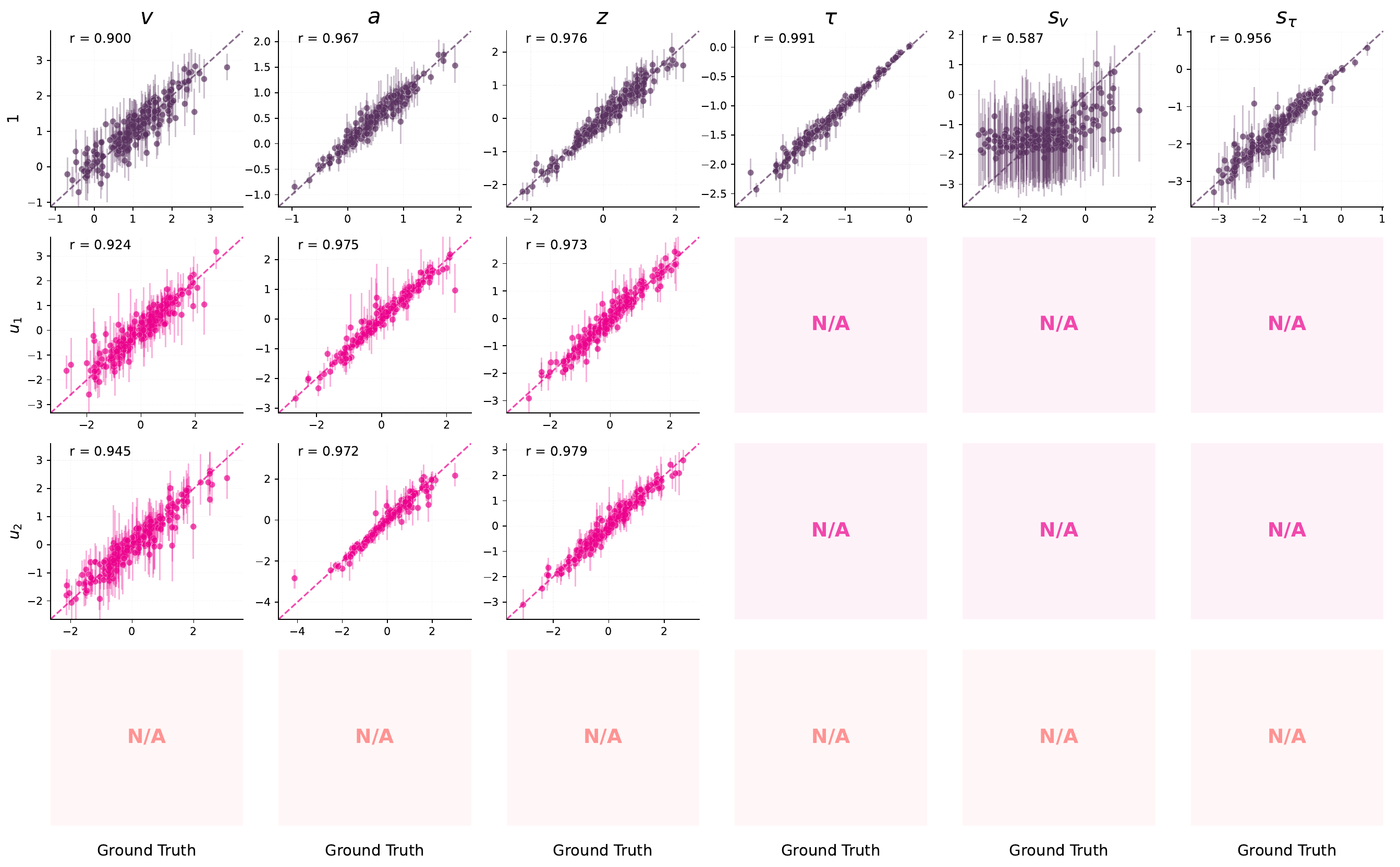}
    \includegraphics[width=0.97\linewidth]{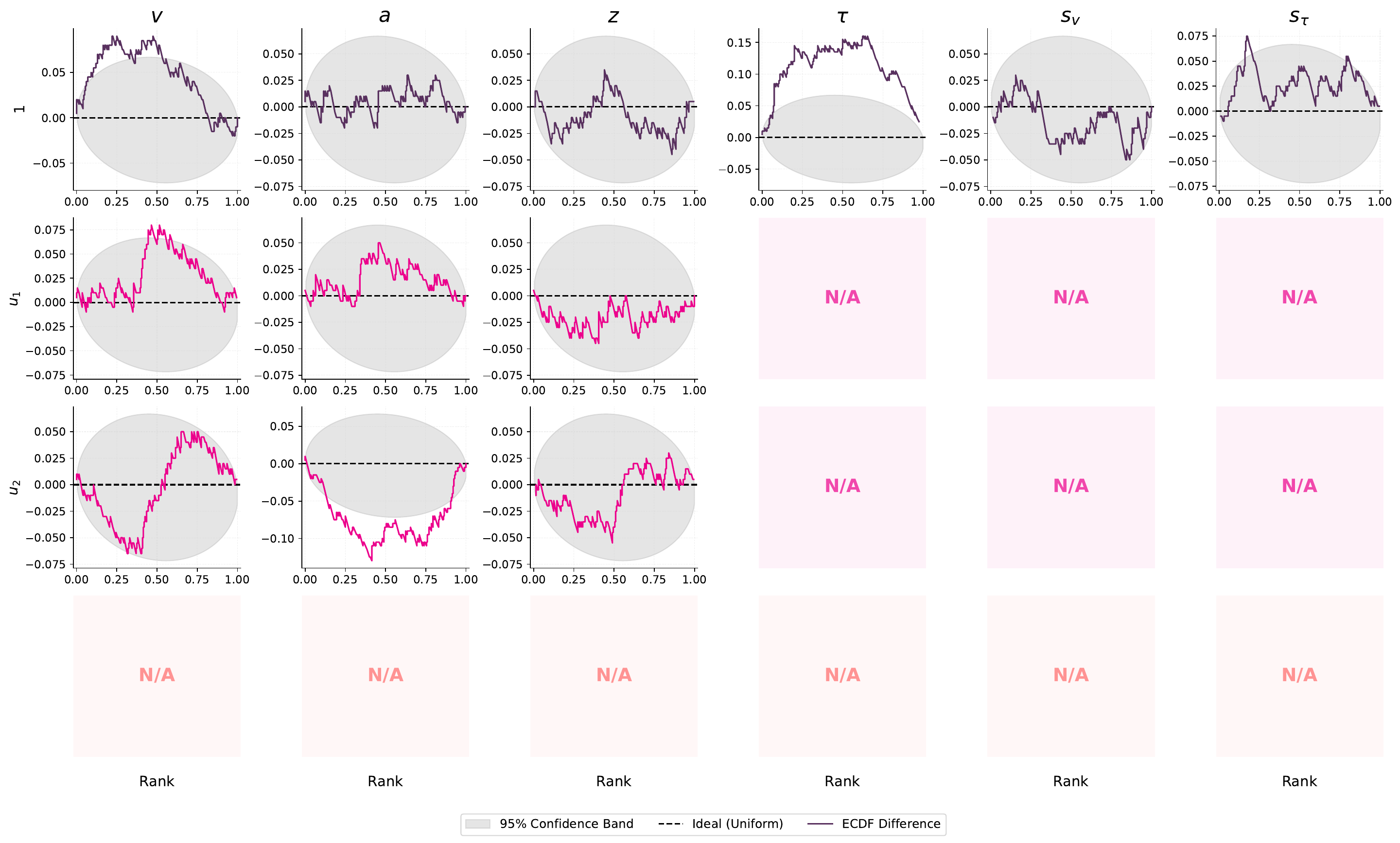}
    \includegraphics[width=0.97\linewidth]{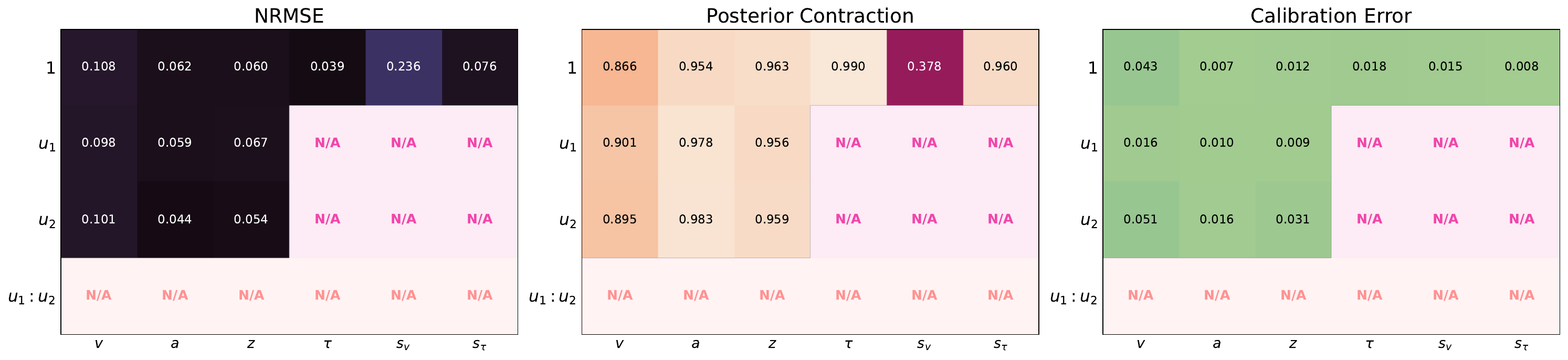}
    \caption{Parameter recovery (\emph{top}), calibration ECDF (\emph{middle}), and validation metrics (NRMSE, calibration error, and posterior contraction) for DDM model family (Case \textbf{regressed}).}
    \label{fig:ddm-fm-regressed}
\end{figure}

\begin{figure}[!h]
    \centering
    \includegraphics[width=0.97\linewidth]{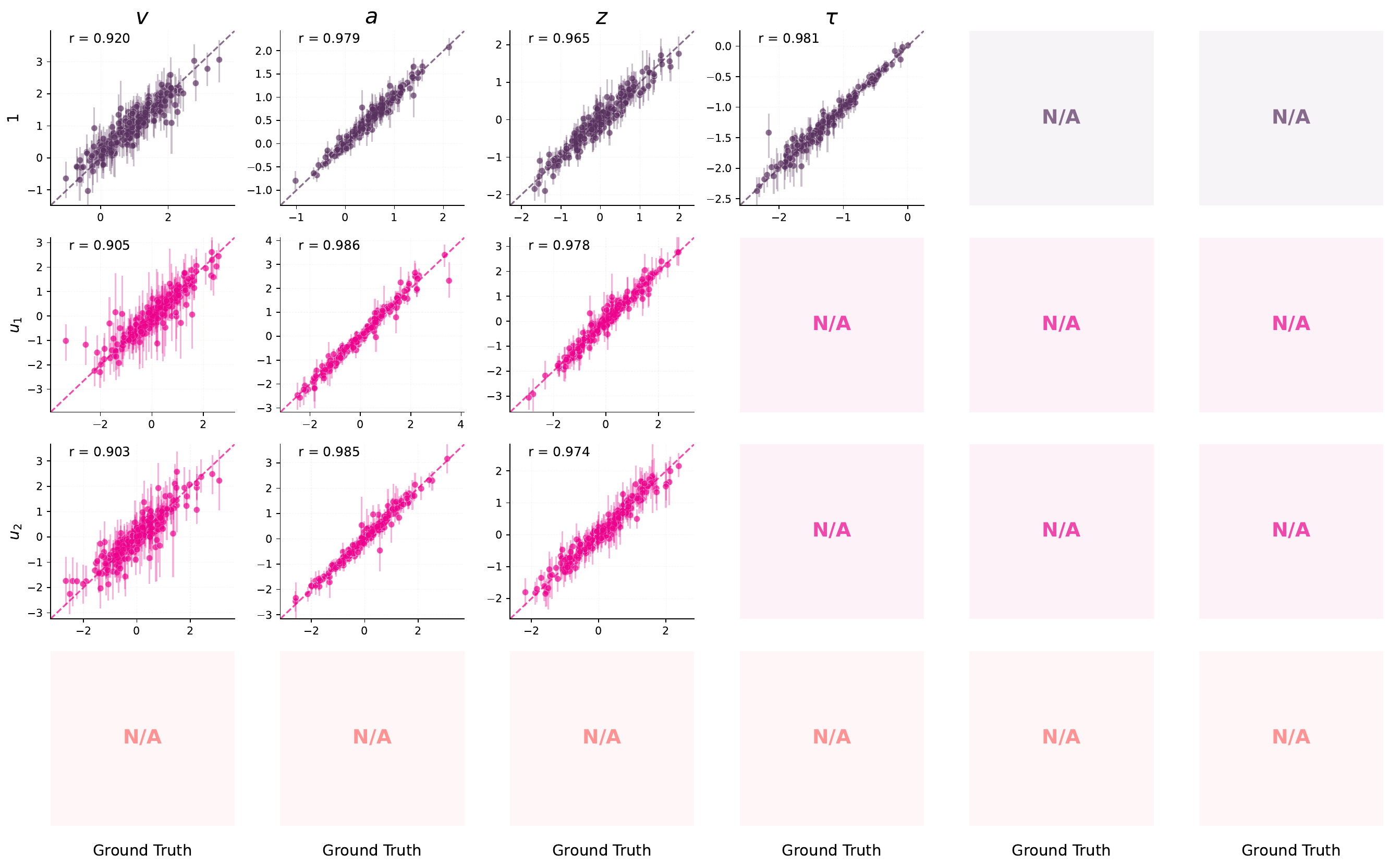}
    \includegraphics[width=0.97\linewidth]{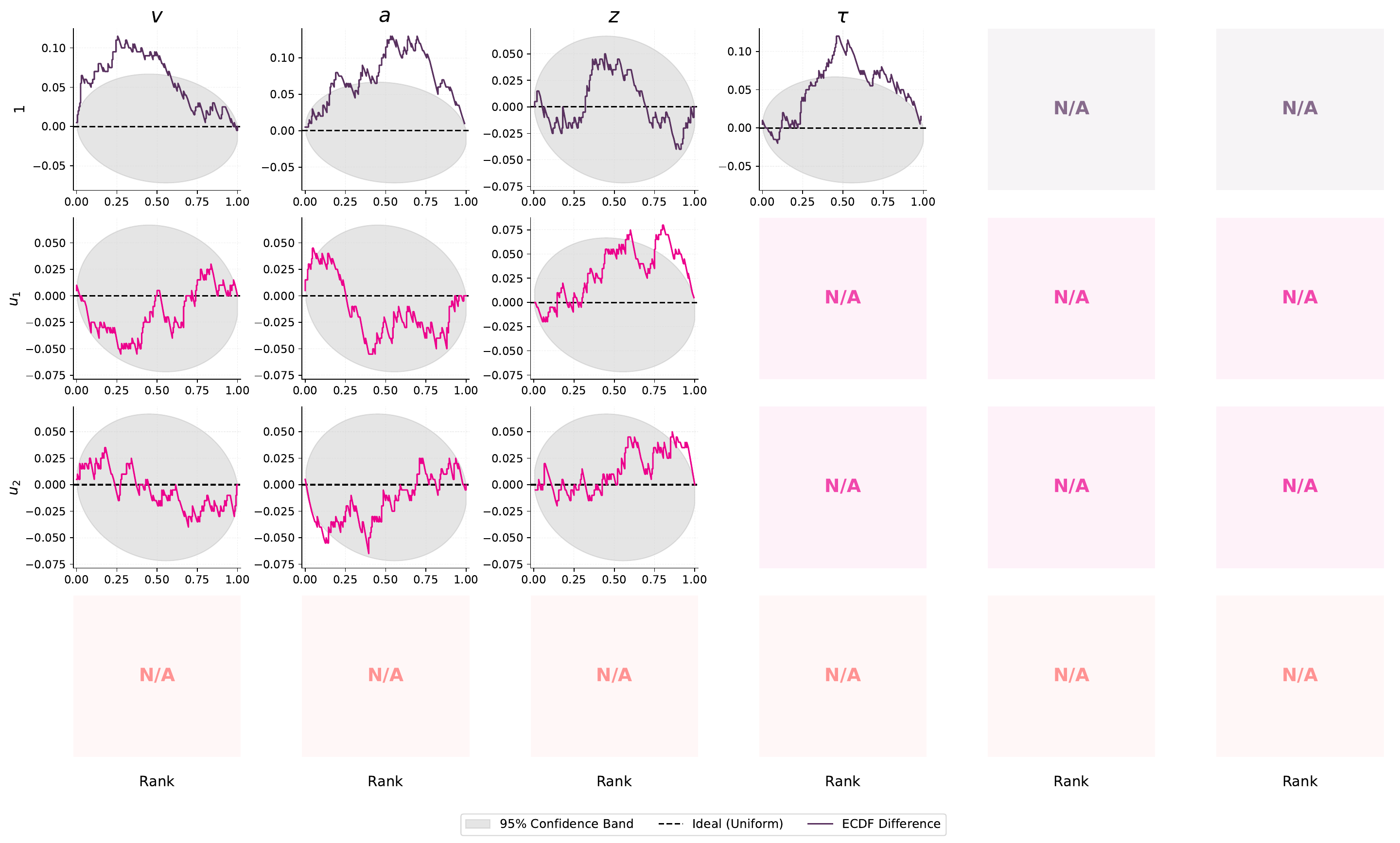}
    \includegraphics[width=0.97\linewidth]{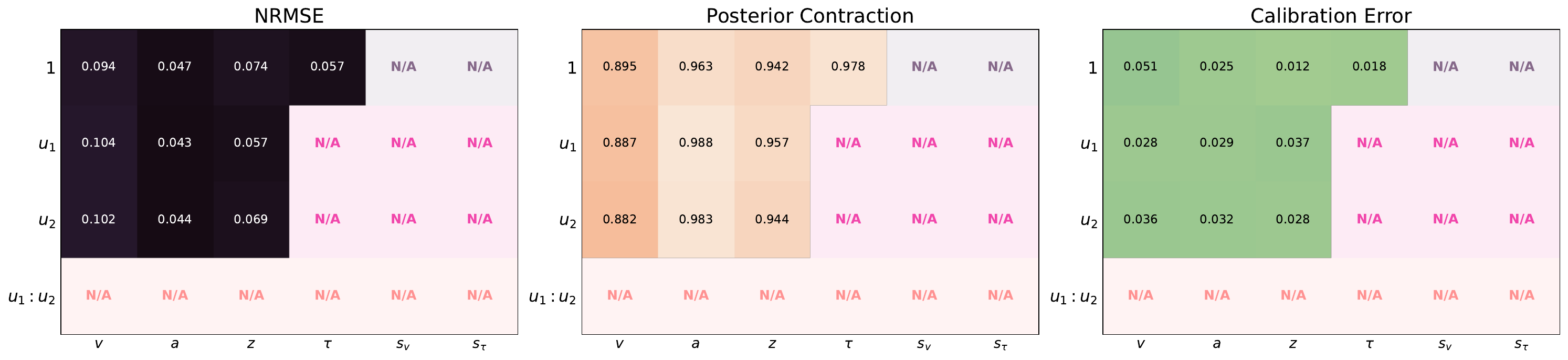}
    \caption{Parameter recovery (\emph{top}), calibration ECDF (\emph{middle}), and validation metrics (NRMSE, calibration error, and posterior contraction) for DDM model family (Case \textbf{fixed\_regressed}).}
    \label{fig:ddm-fm-fixed-regressed}
\end{figure}

\begin{figure}[!h]
    \centering
    \includegraphics[width=0.97\linewidth]{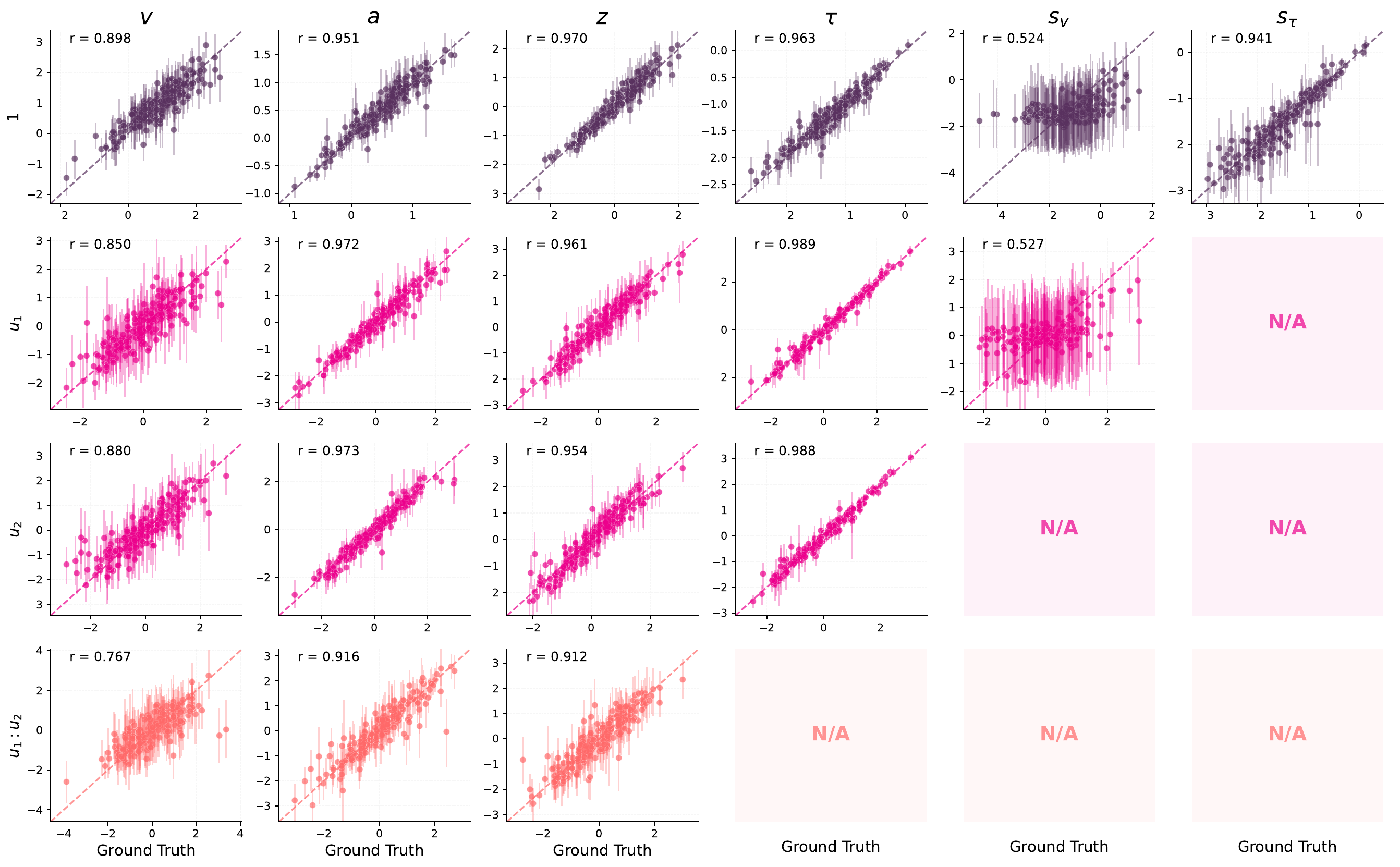}
    \includegraphics[width=0.97\linewidth]{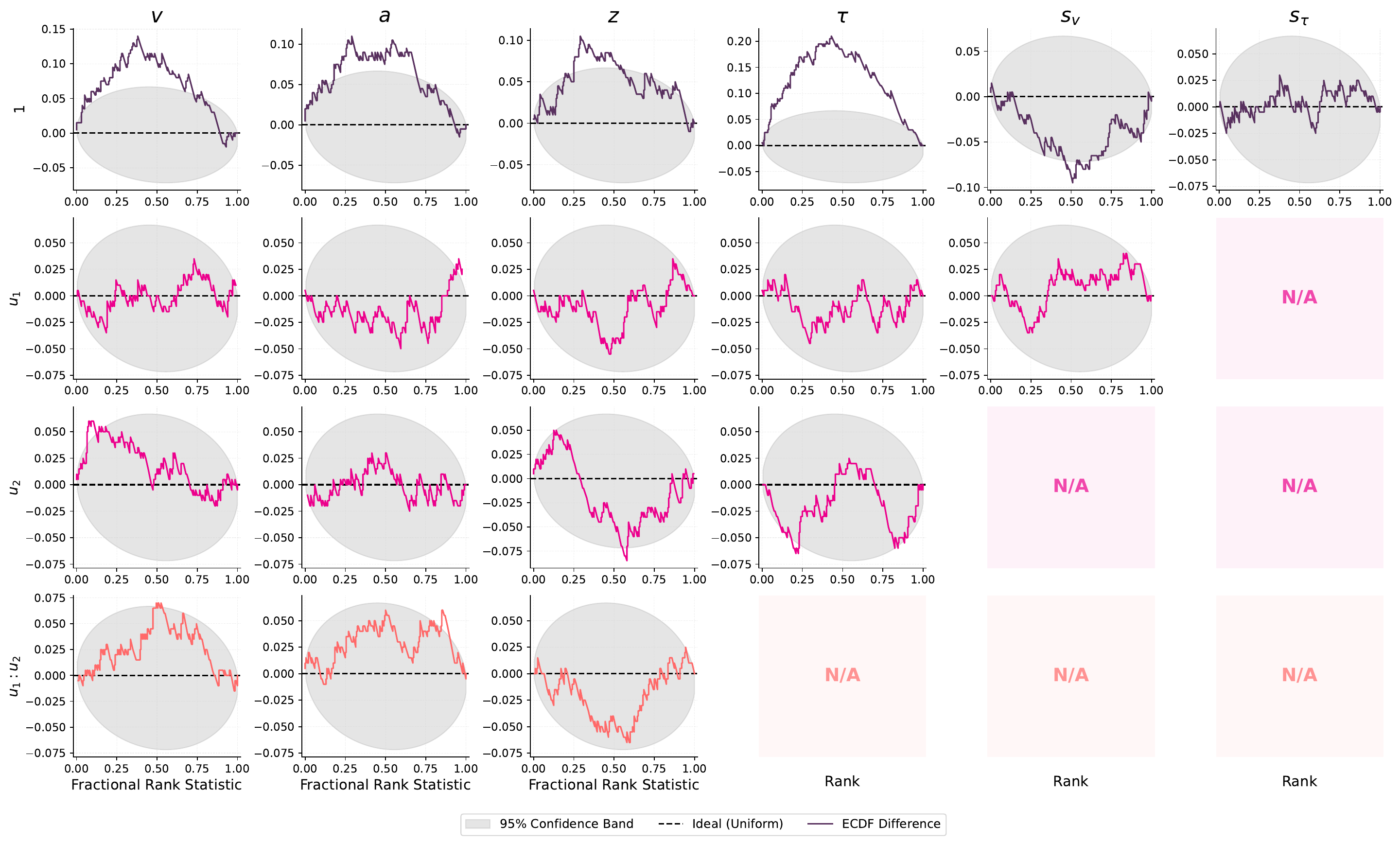}
    \includegraphics[width=0.97\linewidth]{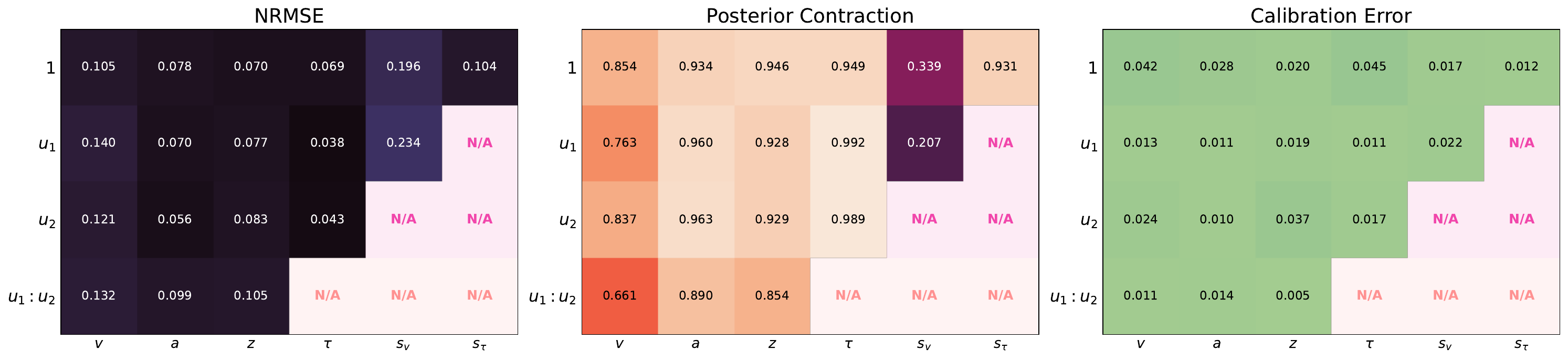}
    \caption{Parameter recovery (\emph{top}), calibration ECDF (\emph{middle}), and validation metrics (NRMSE, calibration error, and posterior contraction) for DDM model family (Case \textbf{interaction}).}
    \label{fig:ddm-fm-interaction}
\end{figure}


\begin{figure}[!h]
    \centering
    \includegraphics[width=0.97\linewidth]{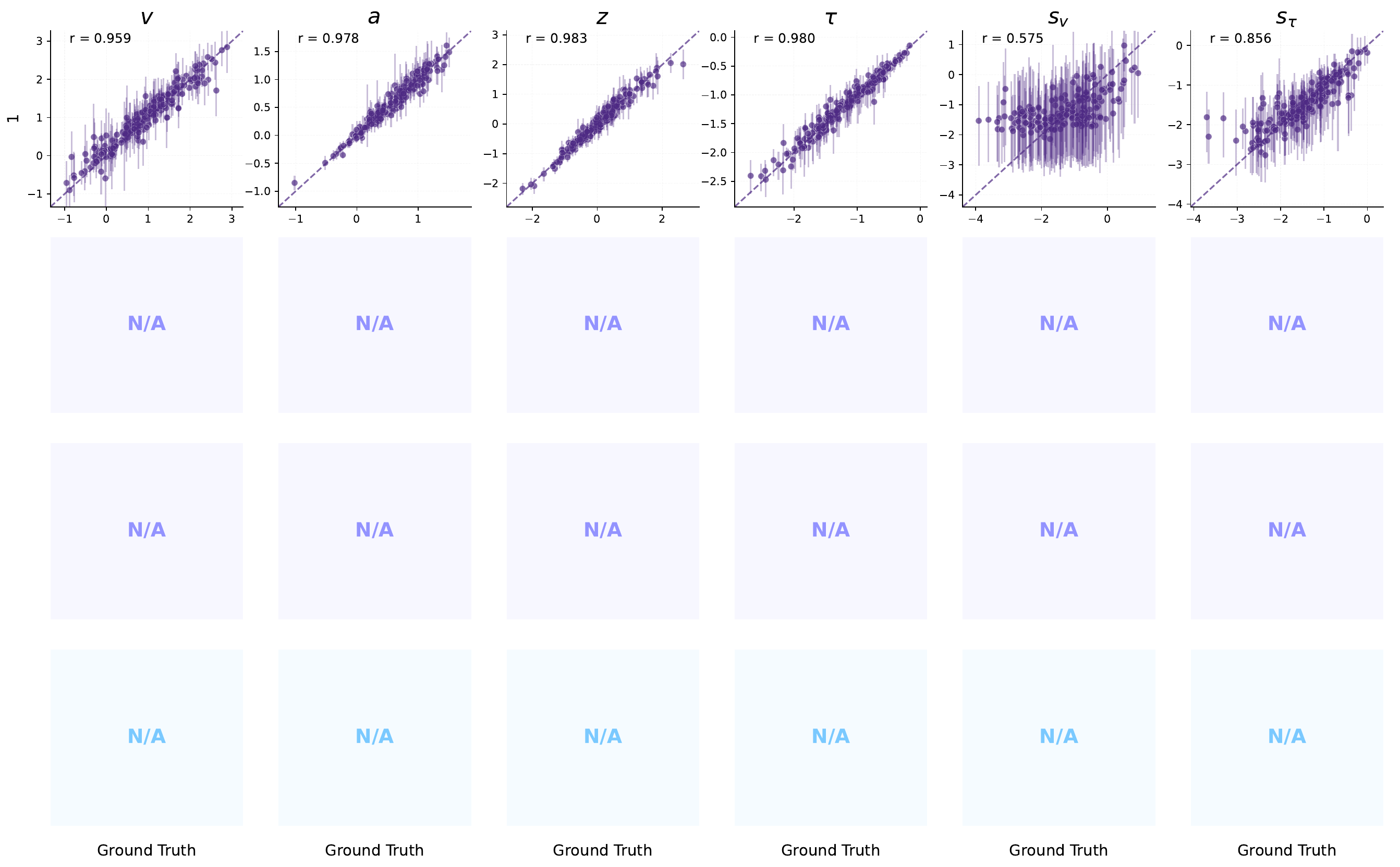}
    \includegraphics[width=0.97\linewidth]{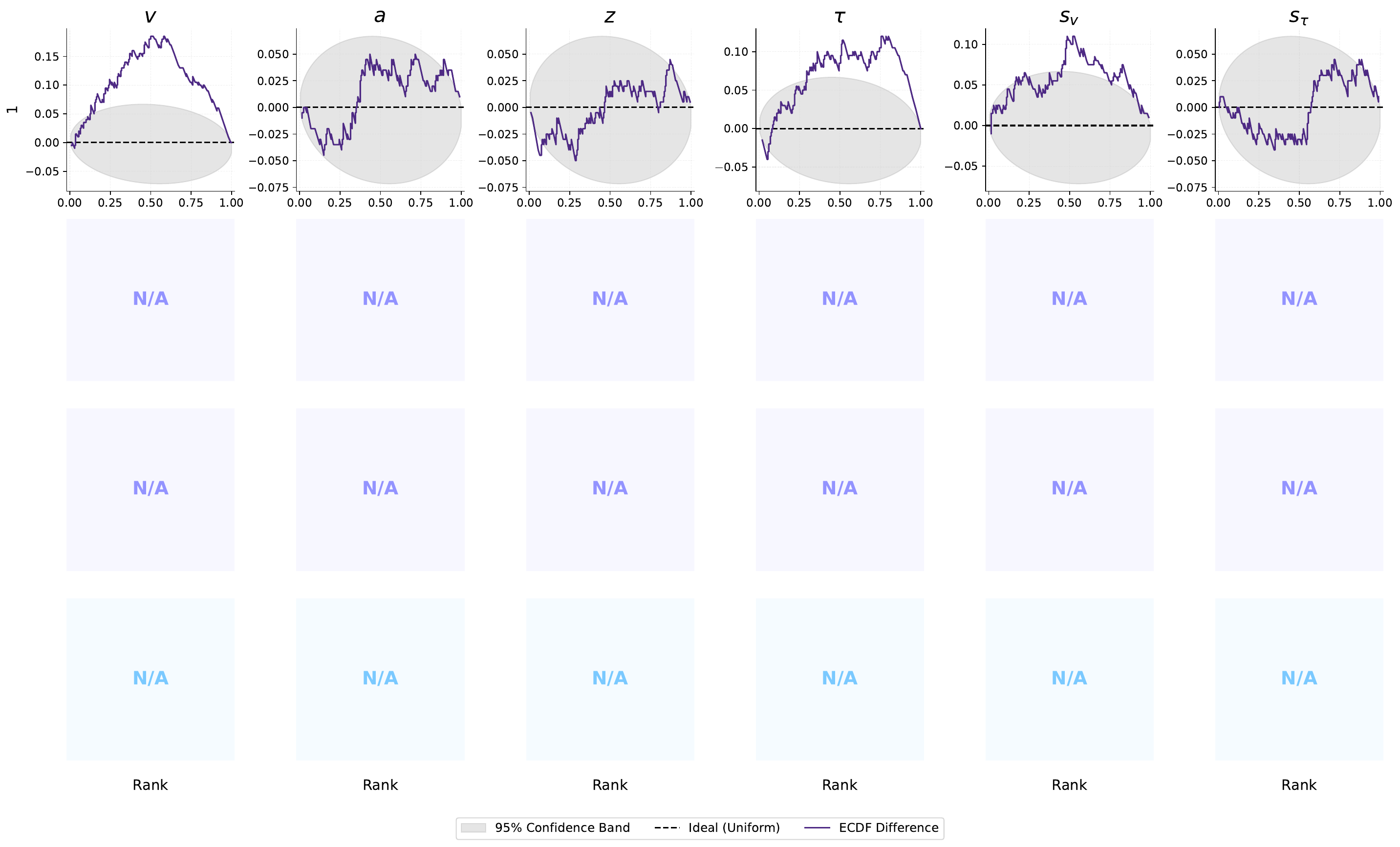}
    \includegraphics[width=0.97\linewidth]{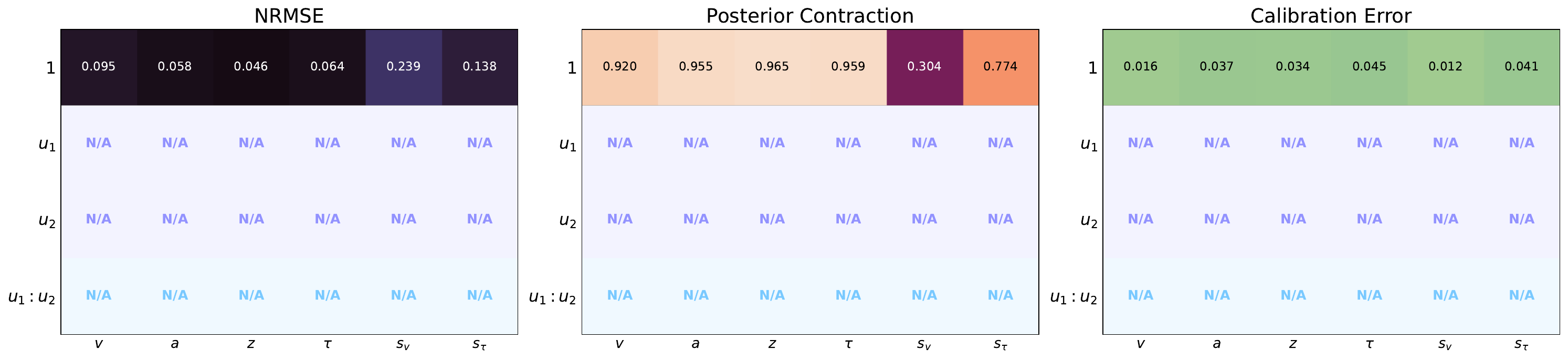}
    \caption{Parameter recovery (\emph{top}), calibration ECDF (\emph{middle}), and parameter-wise metrics (NRMSE, calibration error, and posterior contraction) for DDM model class (Case \textbf{intercept\_only}).}
    \label{fig:ddm-mc-intercept-only}
\end{figure}

\begin{figure}[!h]
    \centering
    \includegraphics[width=0.97\linewidth]{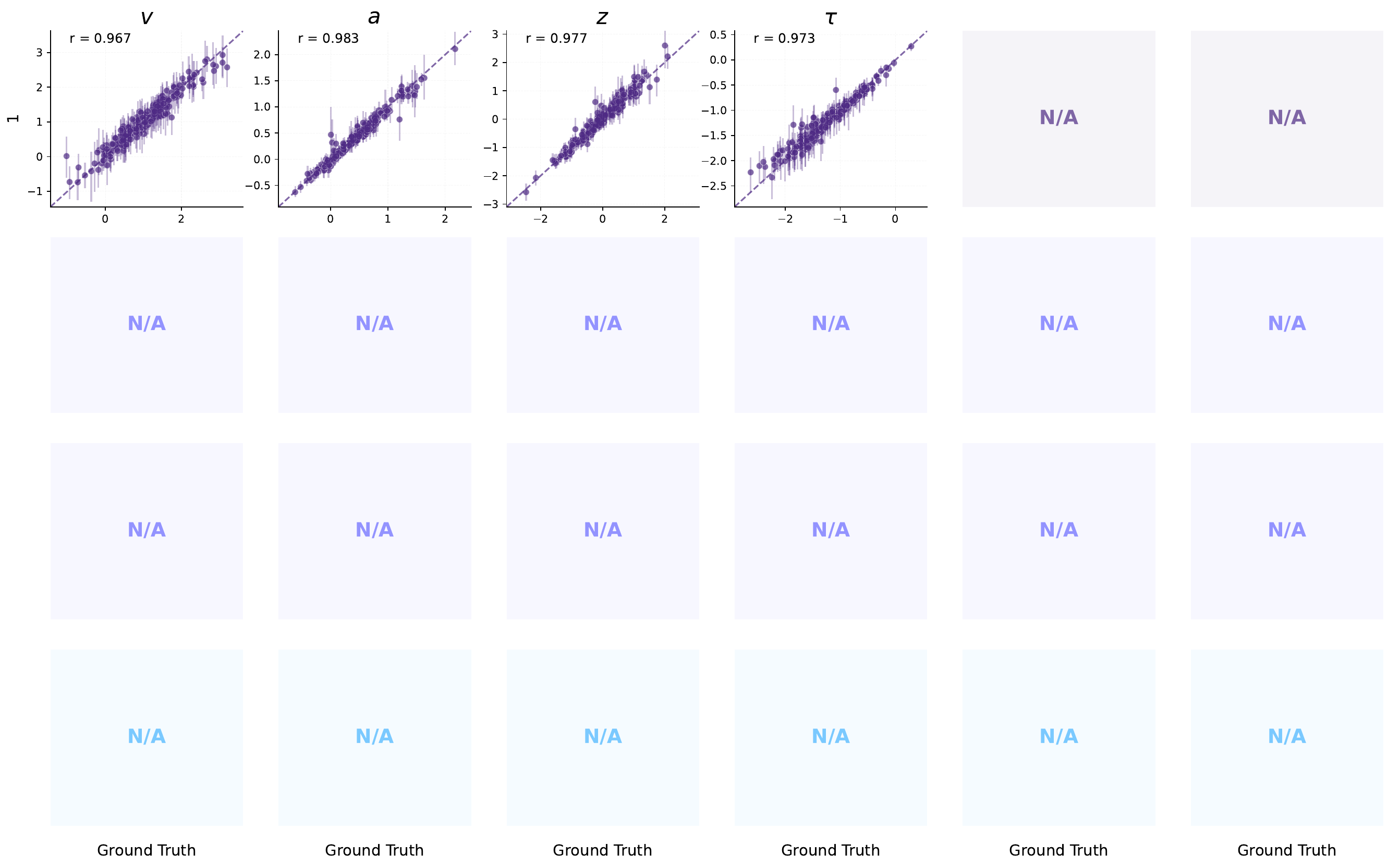}
    \includegraphics[width=0.97\linewidth]{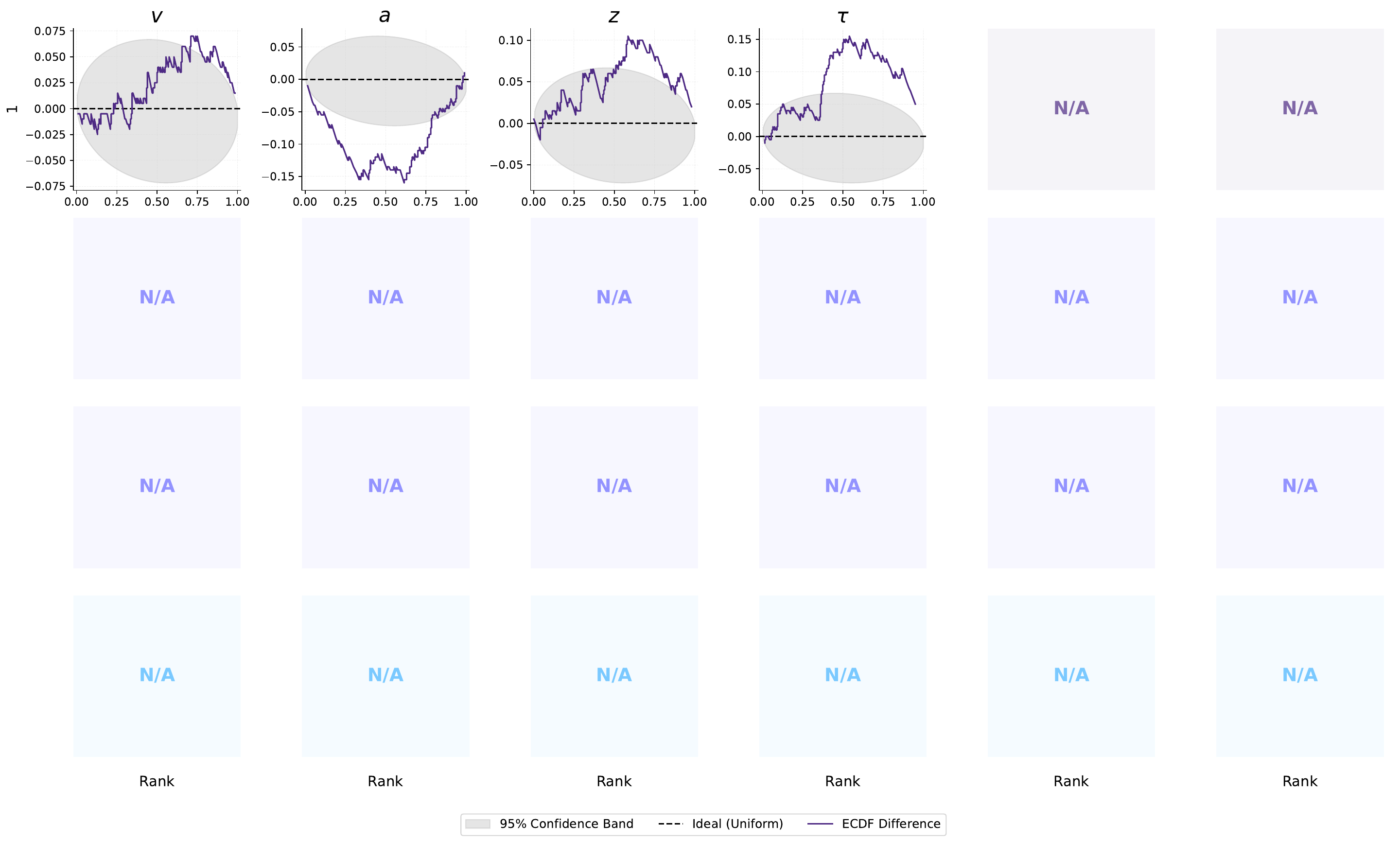}
    \includegraphics[width=0.97\linewidth]{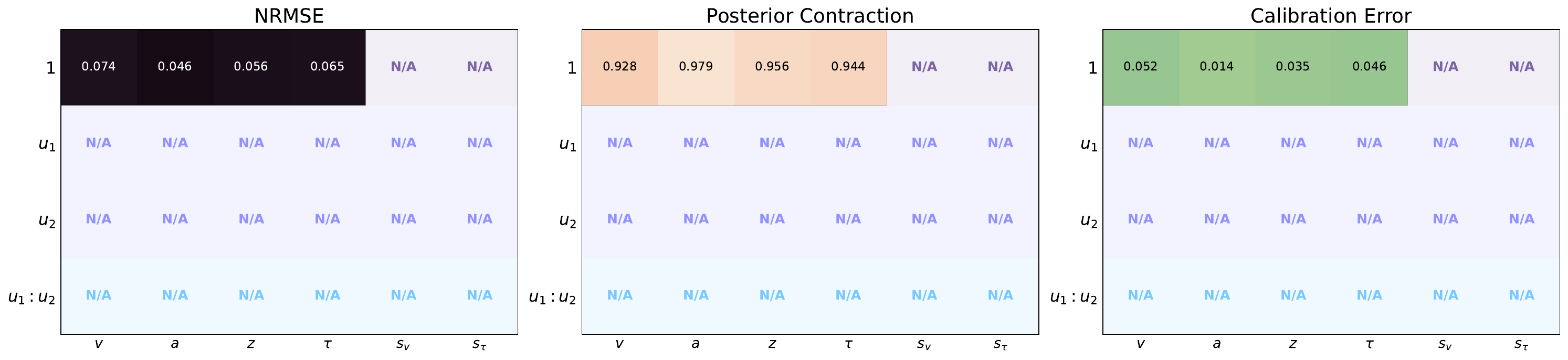}
    \caption{Parameter recovery (\emph{top}), calibration ECDF (\emph{middle}), and parameter-wise metrics (NRMSE, calibration error, and posterior contraction) for DDM model class (Case \textbf{fixed}).}
    \label{fig:ddm-mc-fixed}
\end{figure}

\begin{figure}[!h]
    \centering
    \includegraphics[width=0.97\linewidth]{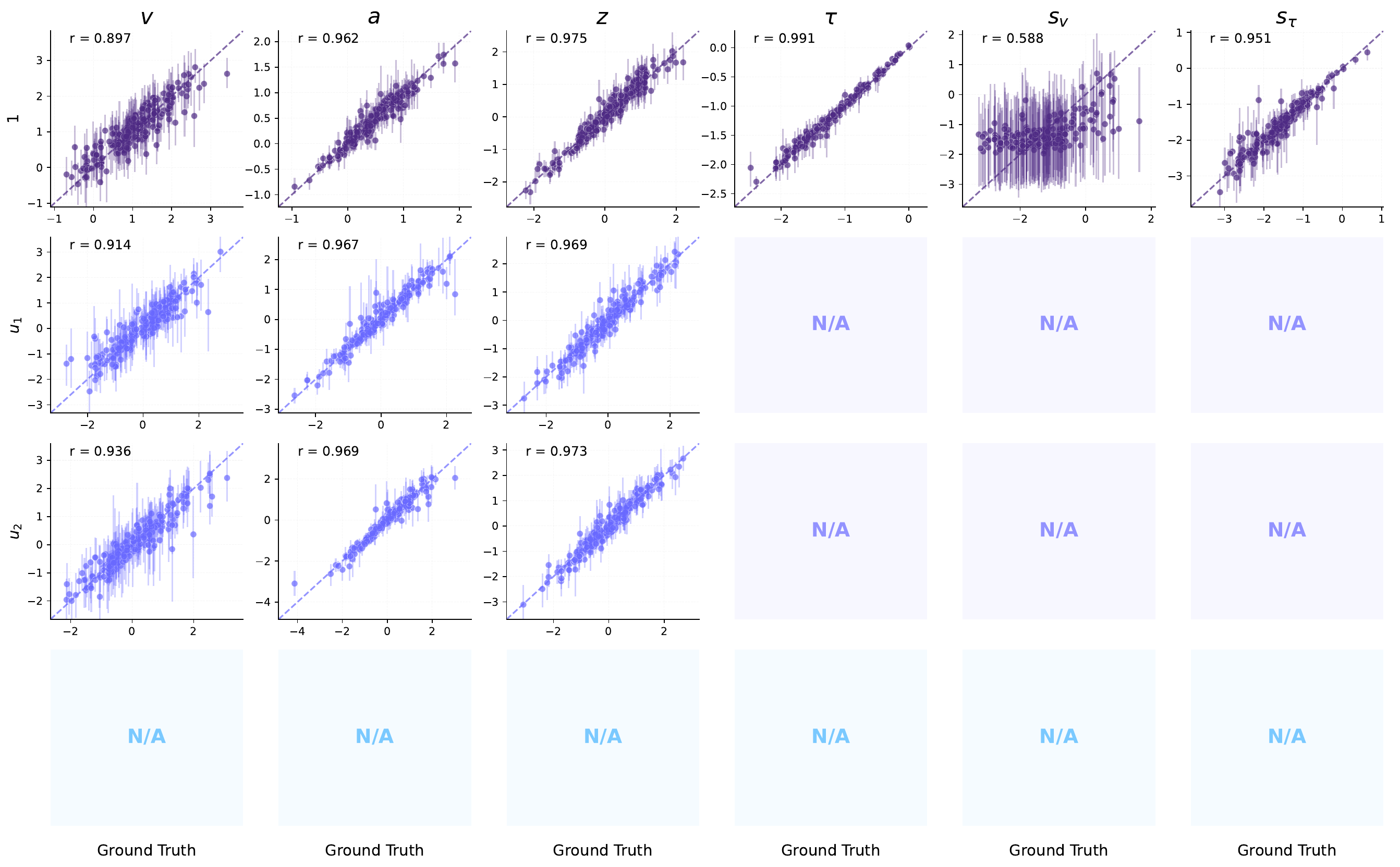}
    \includegraphics[width=0.97\linewidth]{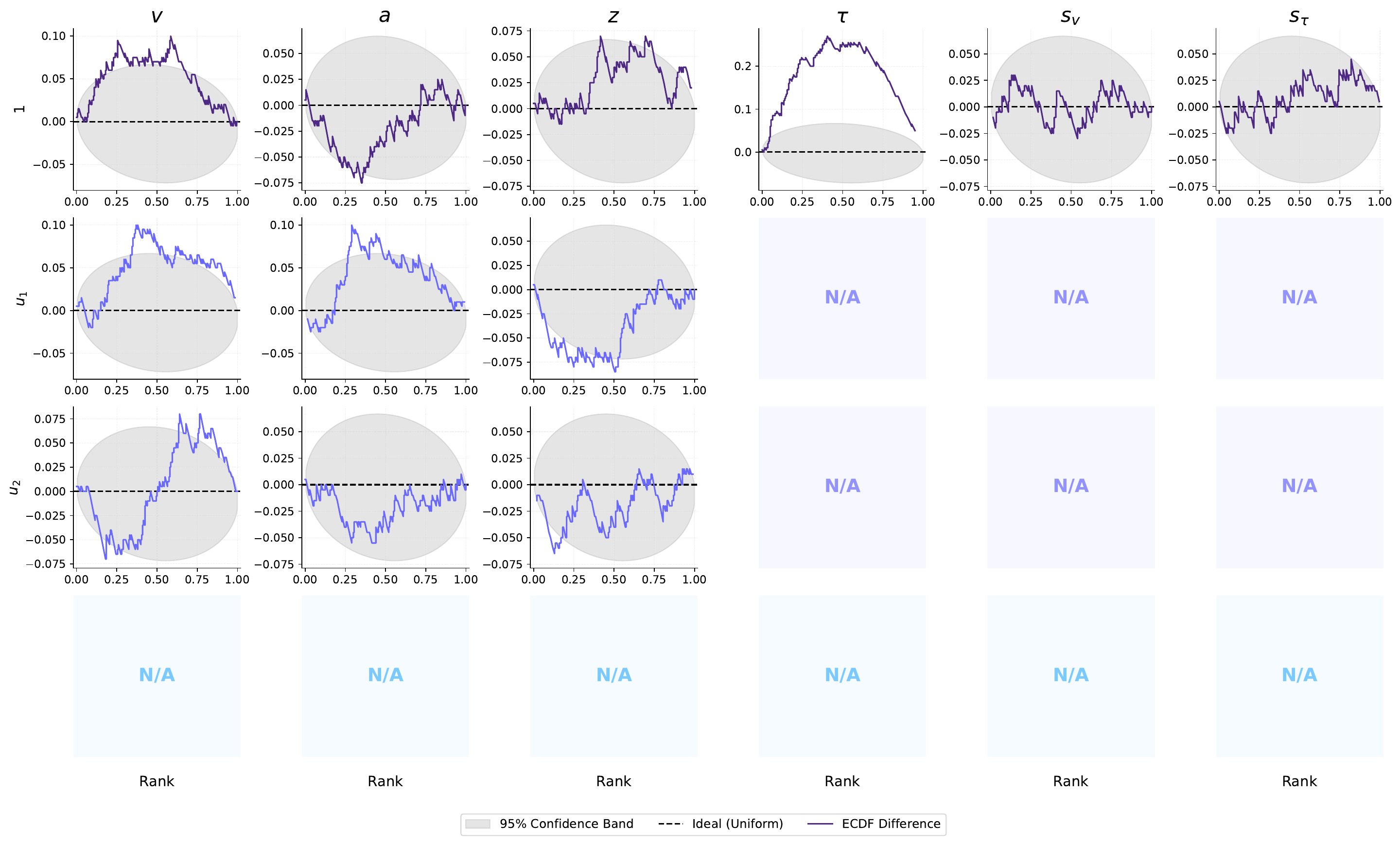}
    \includegraphics[width=0.97\linewidth]{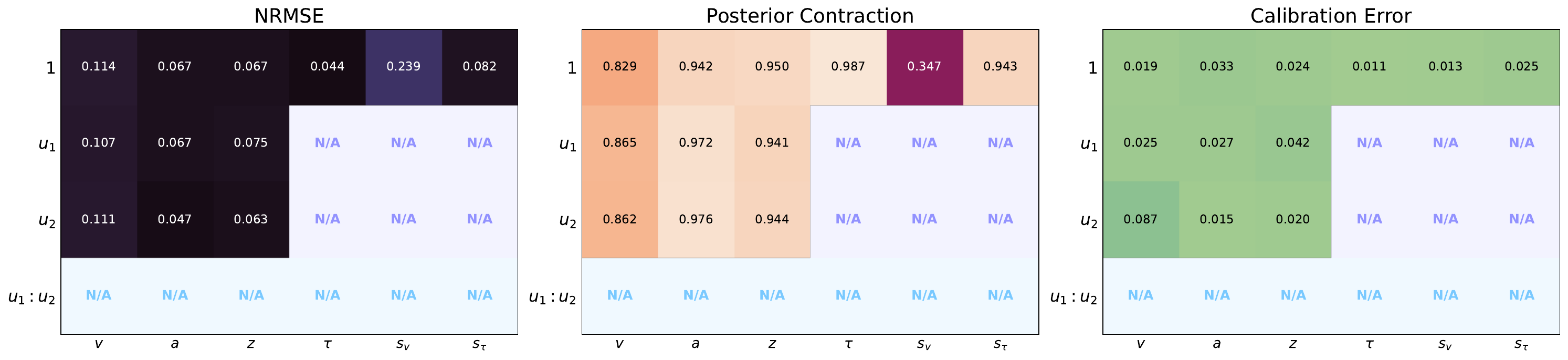}
    \caption{Parameter recovery (\emph{top}), calibration ECDF (\emph{middle}), and parameter-wise metrics (NRMSE, calibration error, and posterior contraction) for DDM model class (Case \textbf{regressed}).}
    \label{fig:ddm-mc-regressed}
\end{figure}

\begin{figure}[!h]
    \centering
    \includegraphics[width=0.97\linewidth]{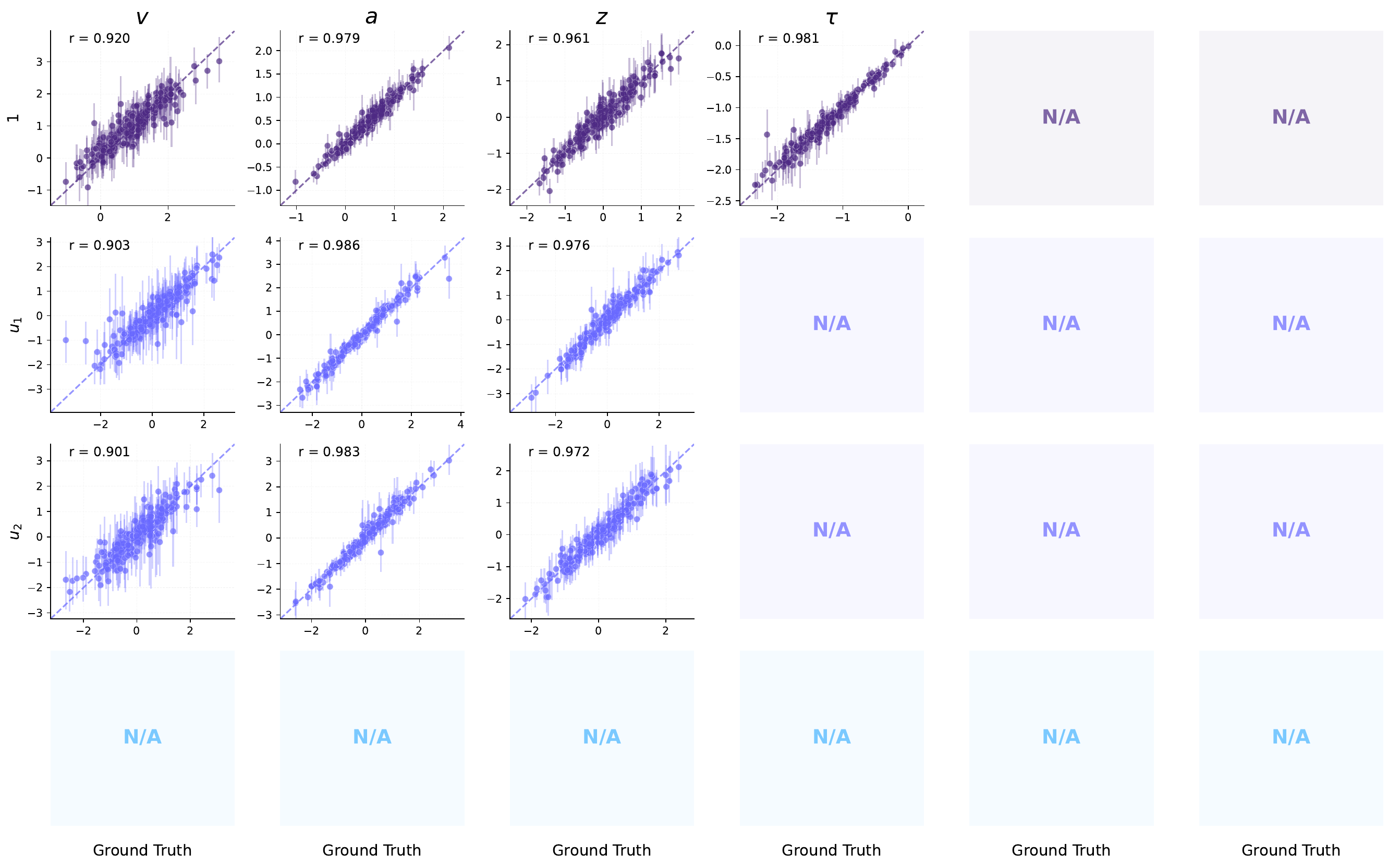}
    \includegraphics[width=0.97\linewidth]{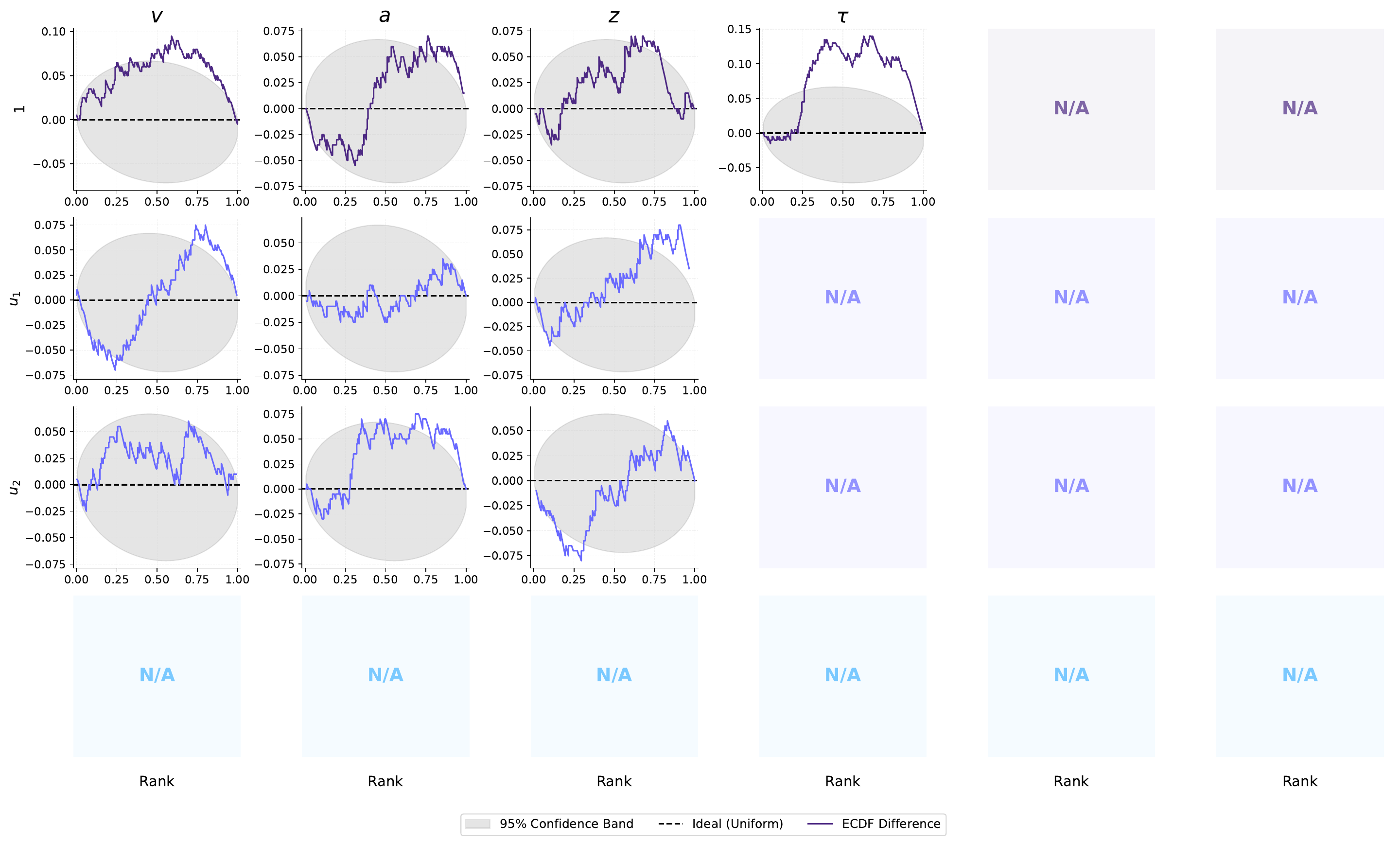}
    \includegraphics[width=0.97\linewidth]{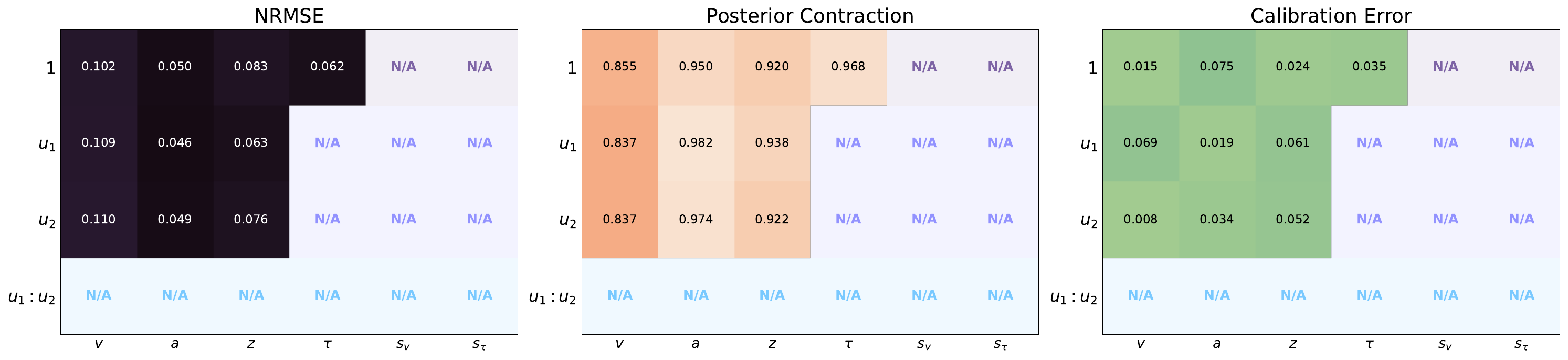}
    \caption{Parameter recovery (\emph{top}), calibration ECDF (\emph{middle}), and parameter-wise metrics (NRMSE, calibration error, and posterior contraction) for DDM model class (Case \textbf{fixed\_regressed}).}
    \label{fig:ddm-mc-fixed-regressed}
\end{figure}

\begin{figure}[!h]
    \centering
    \includegraphics[width=0.97\linewidth]{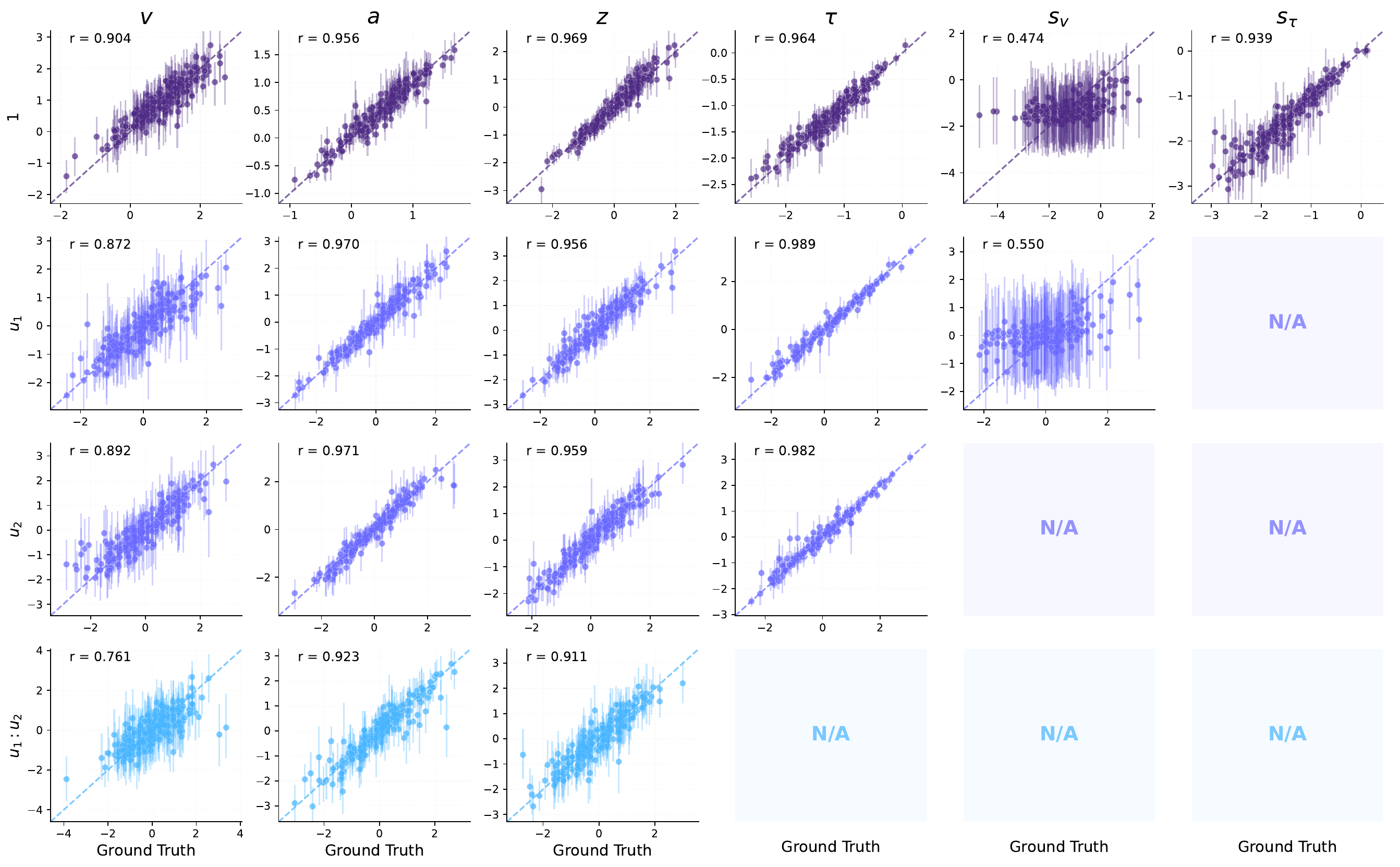}
    \includegraphics[width=0.97\linewidth]{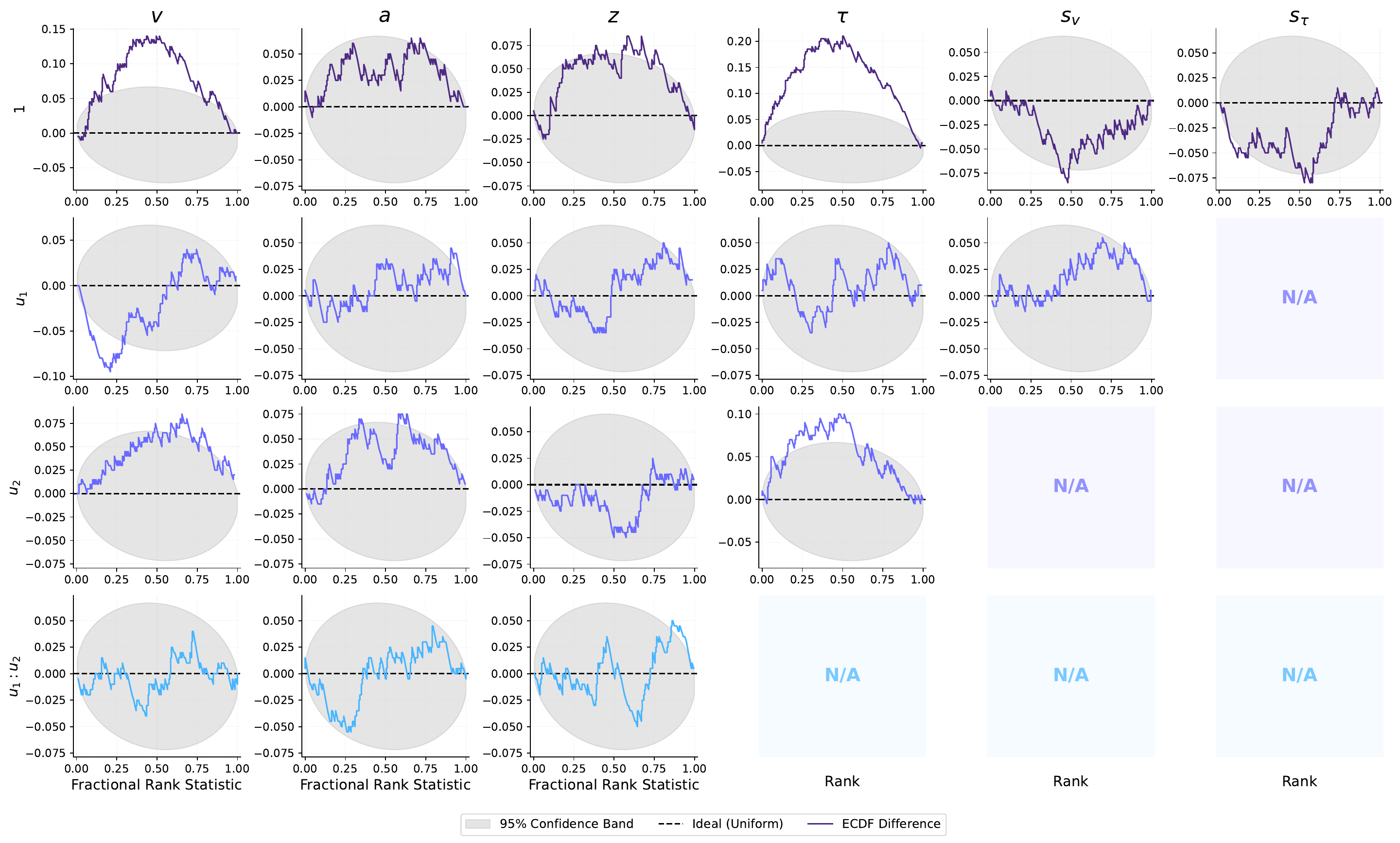}
    \includegraphics[width=0.97\linewidth]{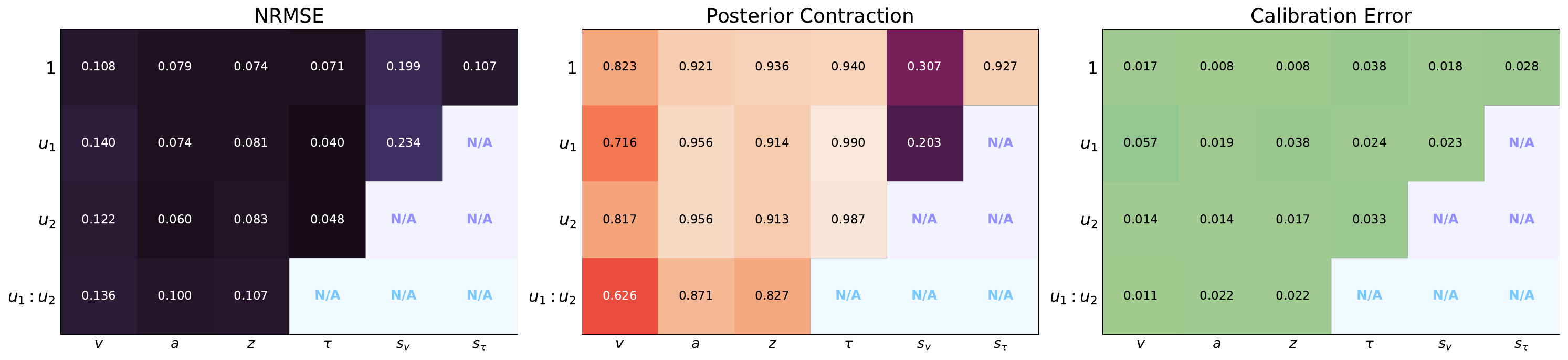}
    \caption{Parameter recovery (\emph{top}), calibration ECDF (\emph{middle}), and parameter-wise metrics (NRMSE, calibration error, and posterior contraction) for DDM model class (Case \textbf{interaction}).}
    \label{fig:ddm-mc-interaction}
\end{figure}


\begin{figure}[!h]
    \centering
    \includegraphics[width=0.97\linewidth]{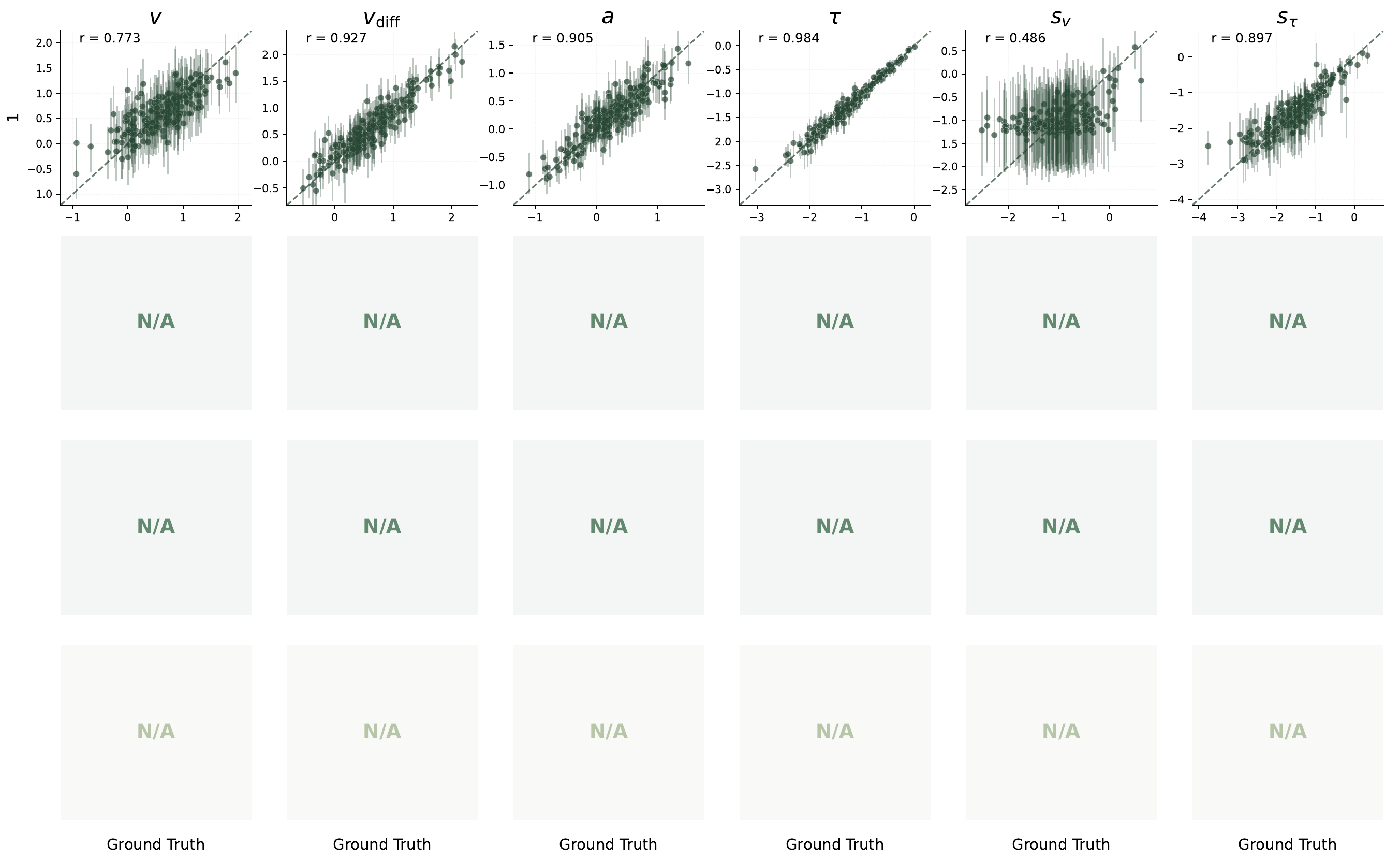}
    \includegraphics[width=0.97\linewidth]{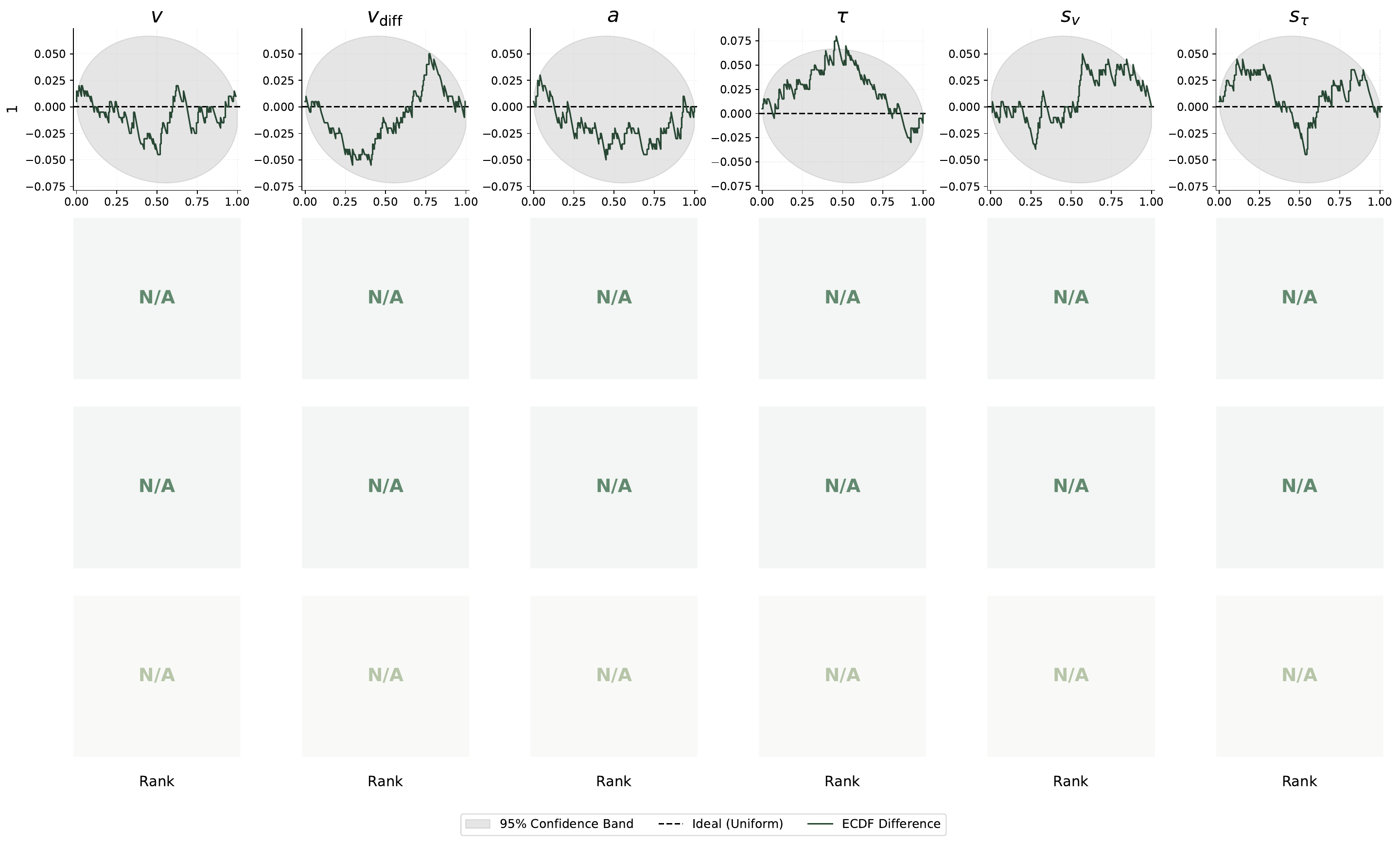}
    \includegraphics[width=0.97\linewidth]{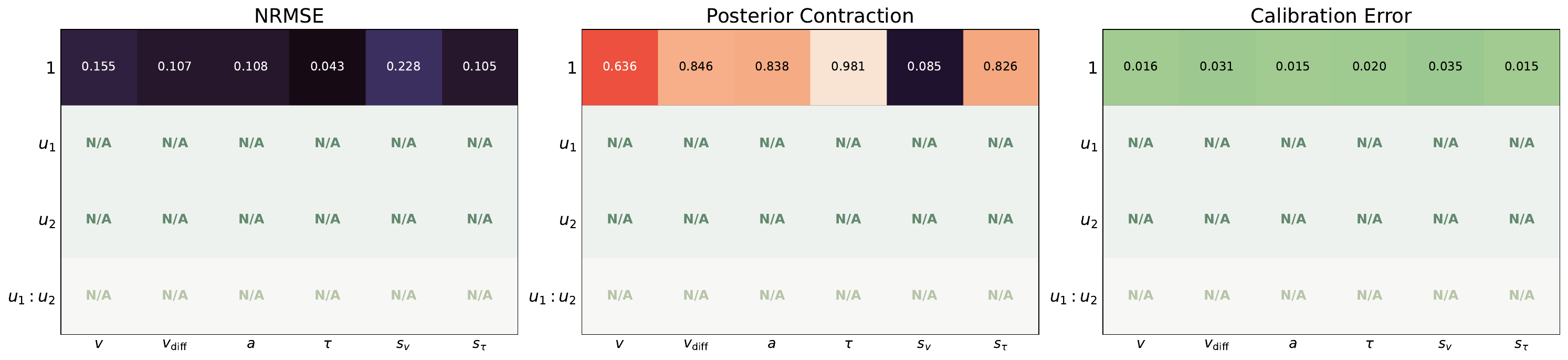}
    \caption{Parameter recovery (\emph{top}), calibration ECDF (\emph{middle}), and validation metrics (NRMSE, calibration error, and posterior contraction) for RDM baseline (Case \textbf{intercept\_only}).}
    \label{fig:rdm-bf-intercept-only}
\end{figure}

\begin{figure}[!h]
    \centering
    \includegraphics[width=0.97\linewidth]{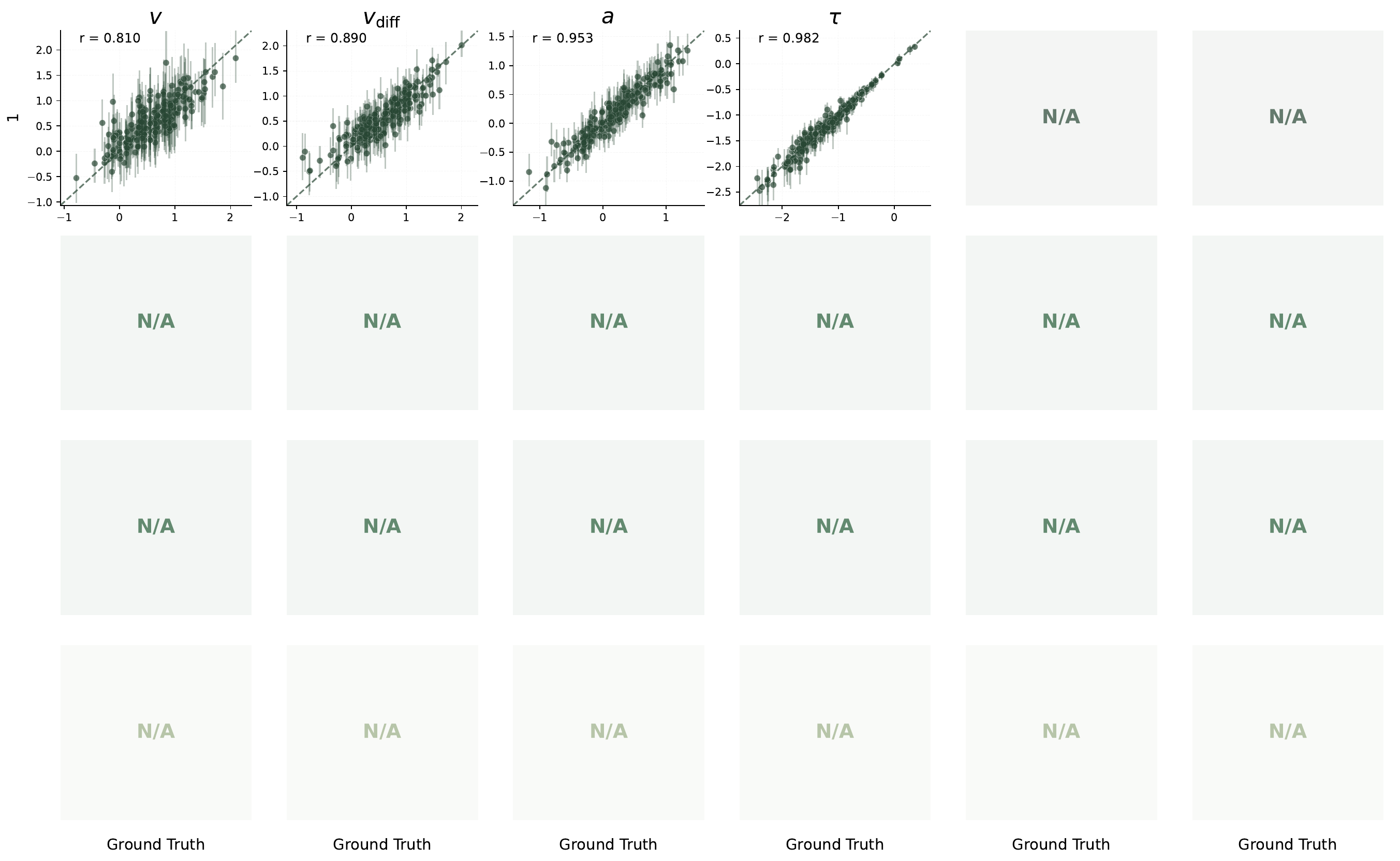}
    \includegraphics[width=0.97\linewidth]{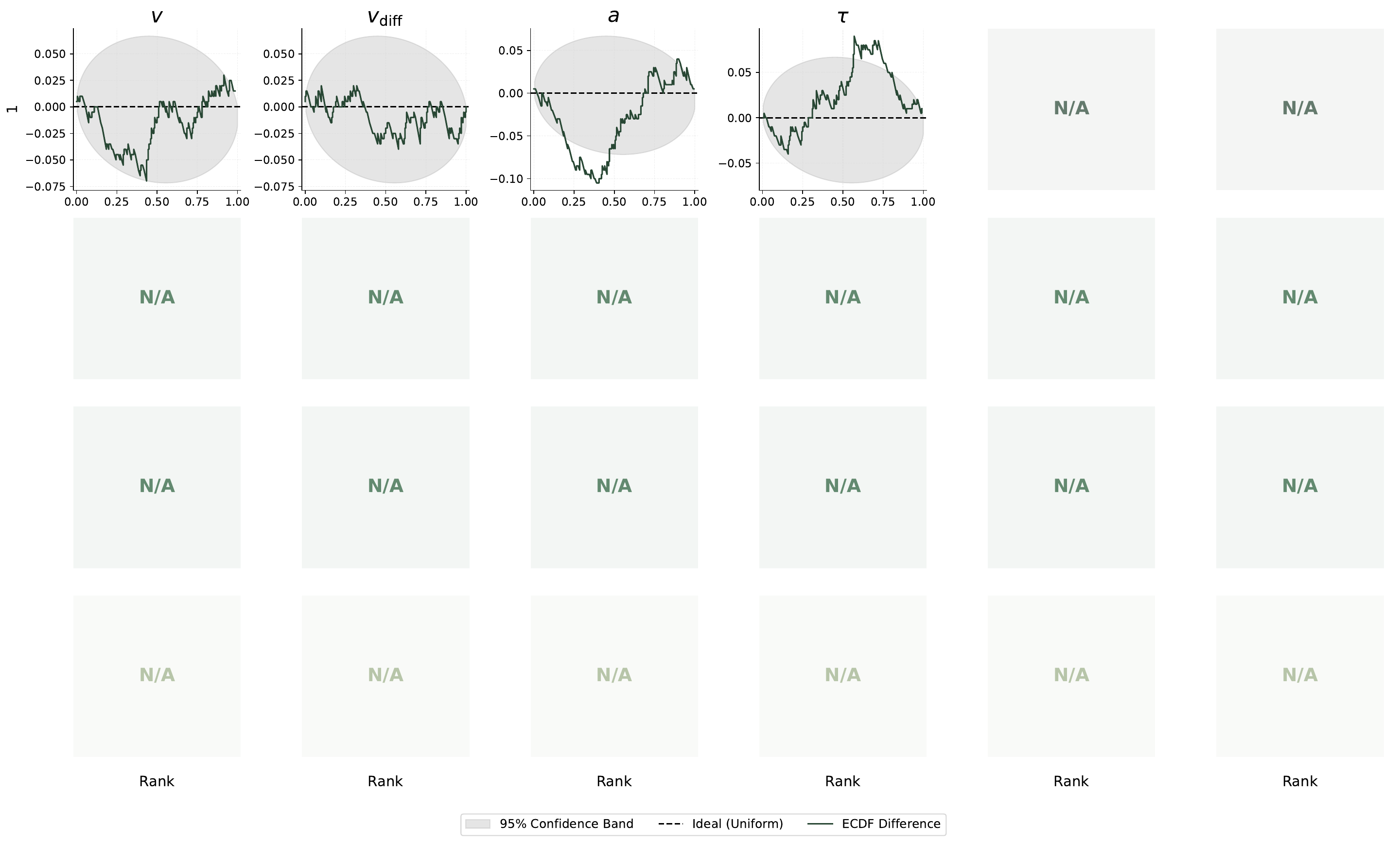}
    \includegraphics[width=0.97\linewidth]{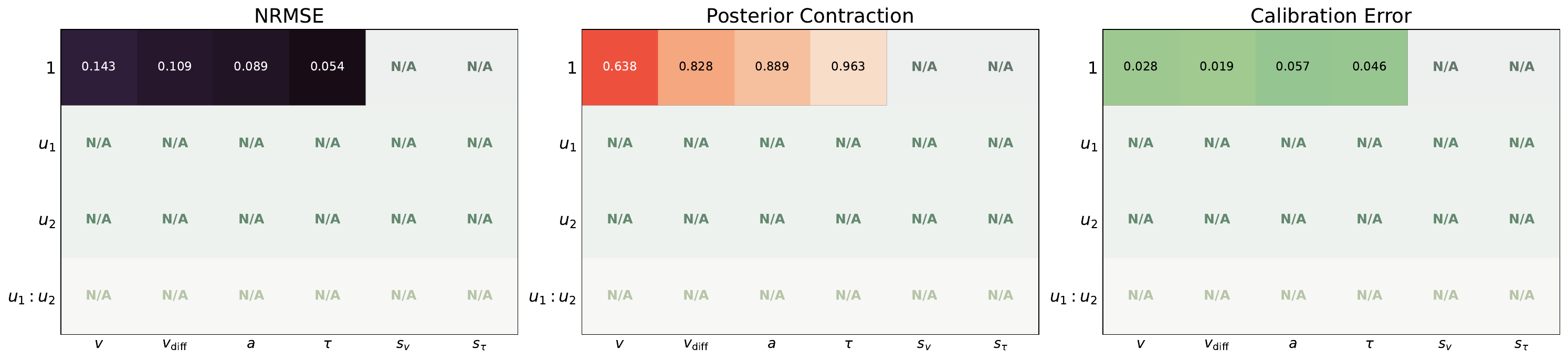}
    \caption{Parameter recovery (\emph{top}), calibration ECDF (\emph{middle}), and validation metrics (NRMSE, calibration error, and posterior contraction) for RDM baseline (Case \textbf{fixed}).}
    \label{fig:rdm-bf-fixed}
\end{figure}

\begin{figure}[!h]
    \centering
    \includegraphics[width=0.97\linewidth]{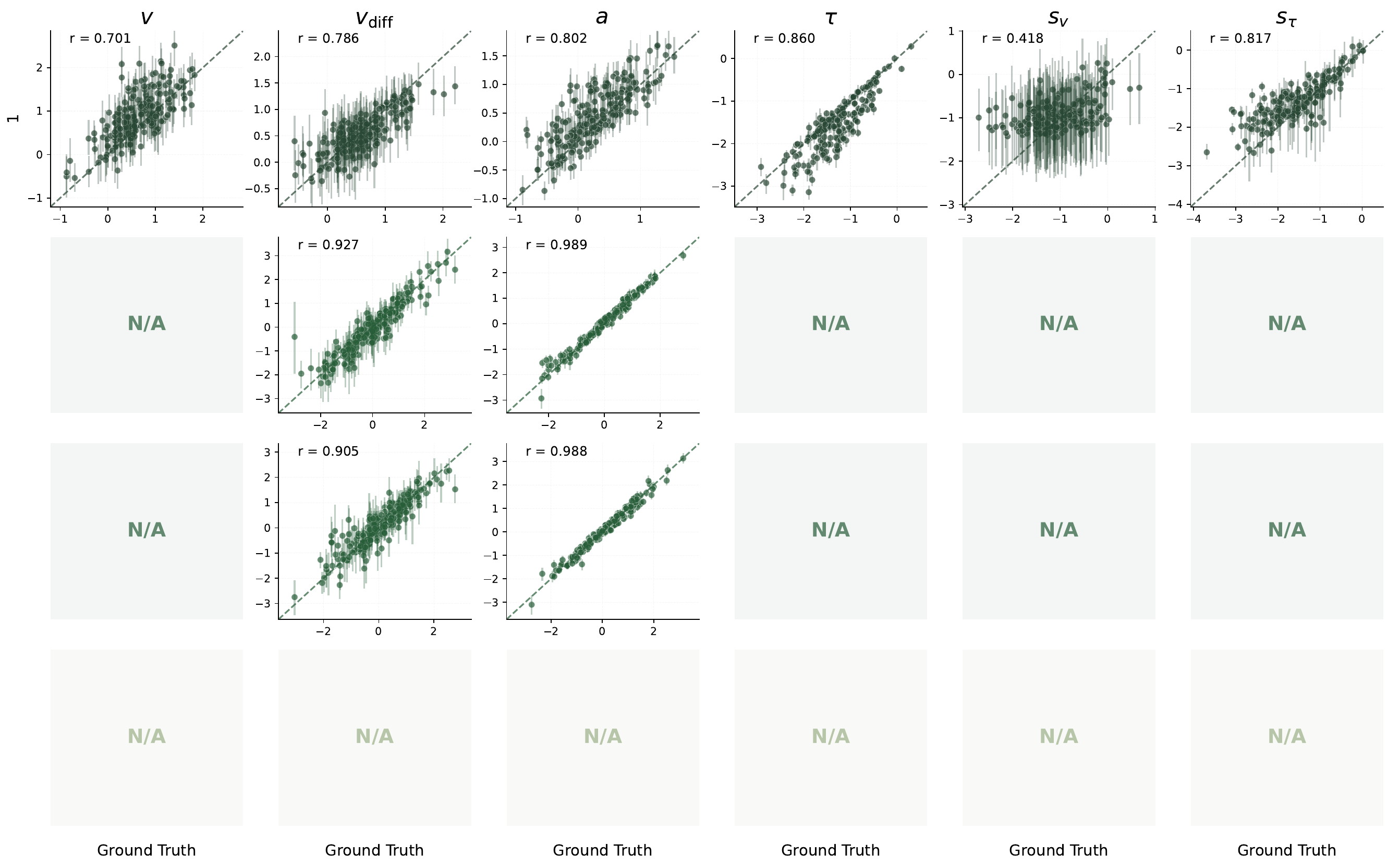}
    \includegraphics[width=0.97\linewidth]{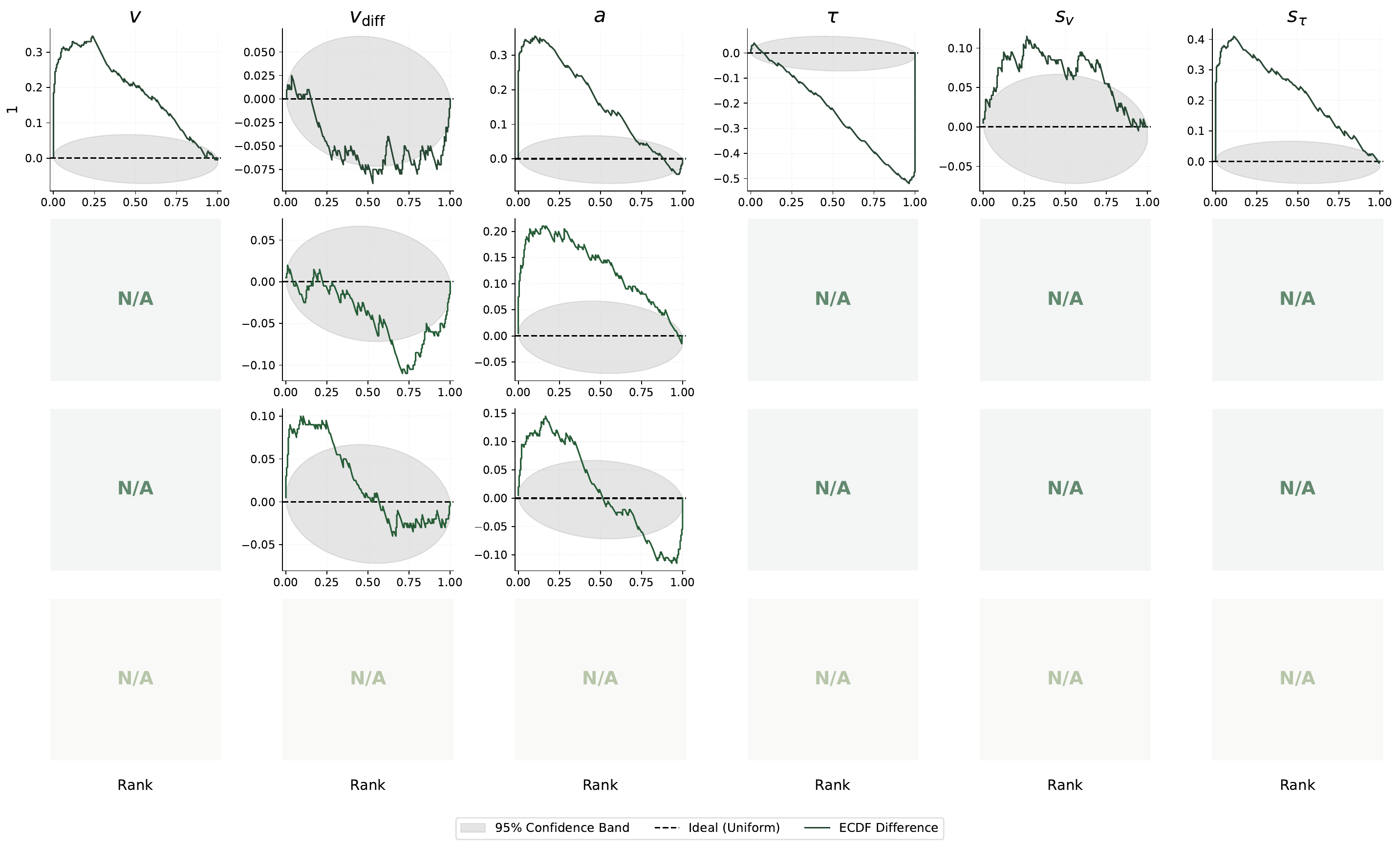}
    \includegraphics[width=0.97\linewidth]{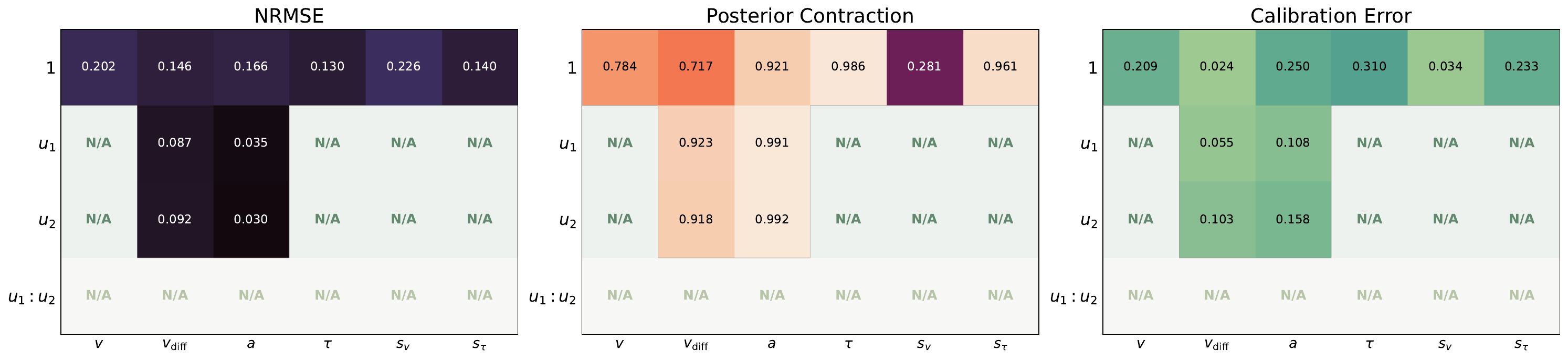}
    \caption{Parameter recovery (\emph{top}), calibration ECDF (\emph{middle}), and validation metrics (NRMSE, calibration error, and posterior contraction) for RDM baseline (Case \textbf{regressed}).}
    \label{fig:rdm-bf-regressed}
\end{figure}

\begin{figure}[!h]
    \centering
    \includegraphics[width=0.97\linewidth]{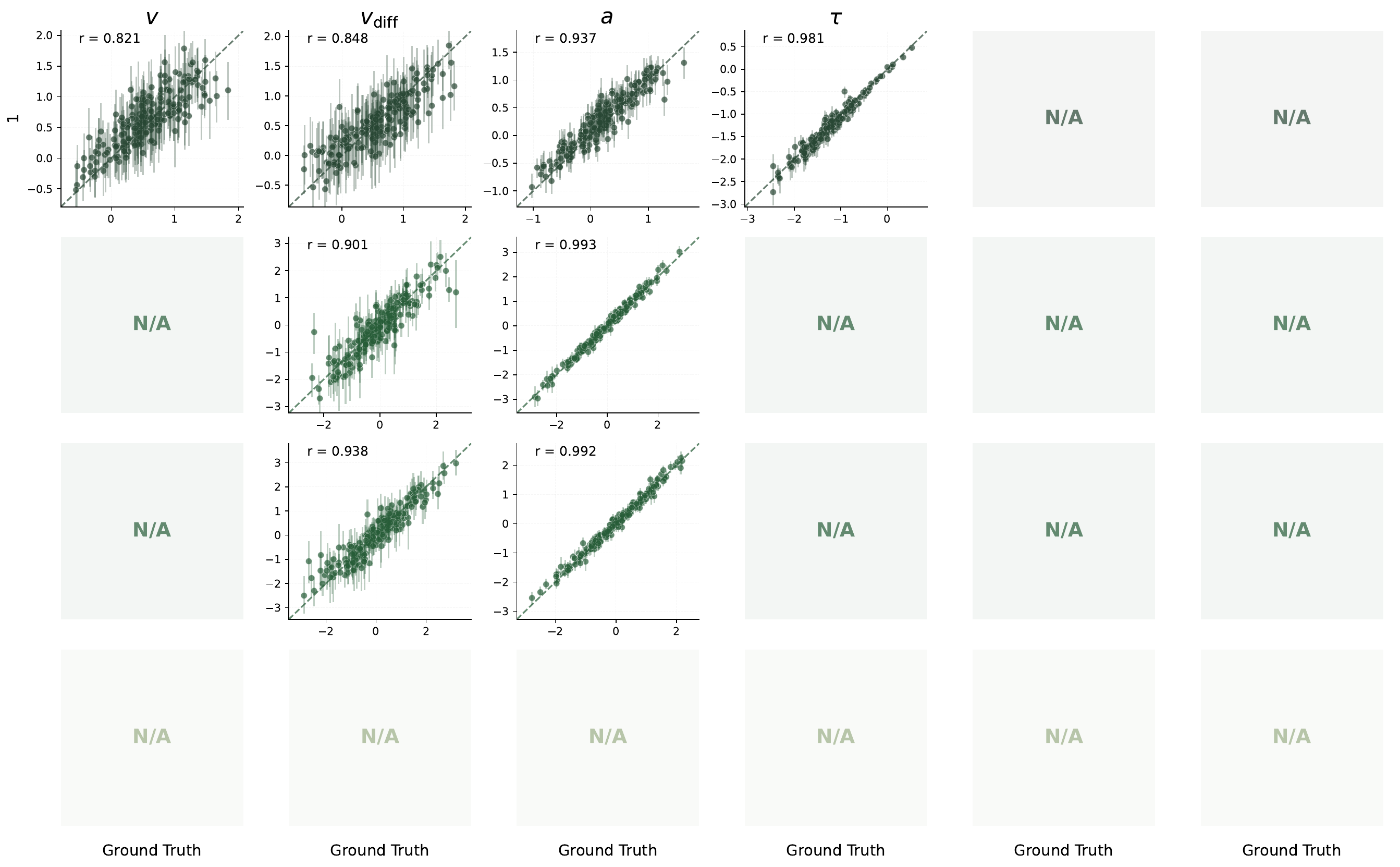}
    \includegraphics[width=0.97\linewidth]{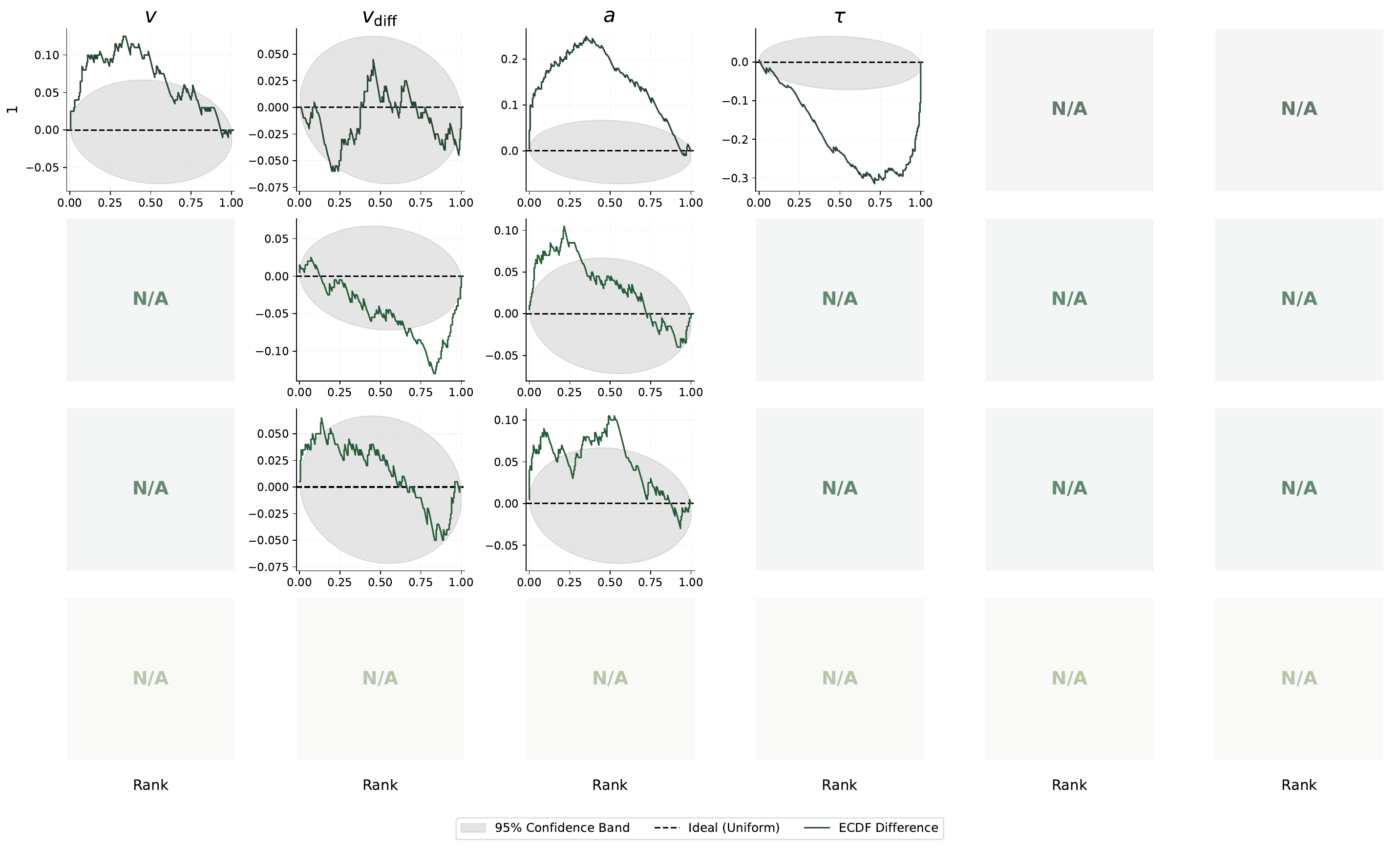}
    \includegraphics[width=0.97\linewidth]{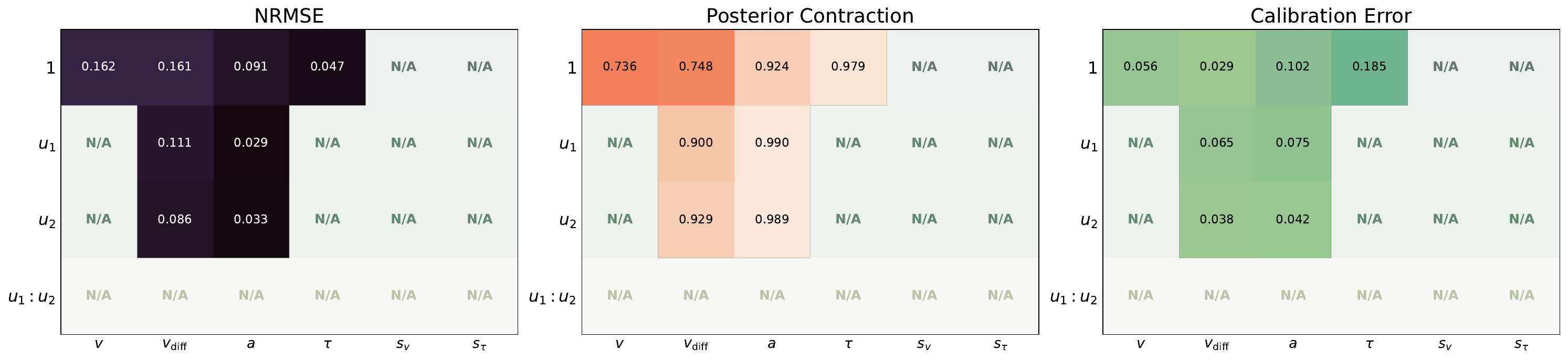}
    \caption{Parameter recovery (\emph{top}), calibration ECDF (\emph{middle}), and validation metrics (NRMSE, calibration error, and posterior contraction) for RDM baseline (Case \textbf{fixed\_regressed}).}
    \label{fig:rdm-bf-fixed-regressed}
\end{figure}

\begin{figure}[!h]
    \centering
    \includegraphics[width=0.97\linewidth]{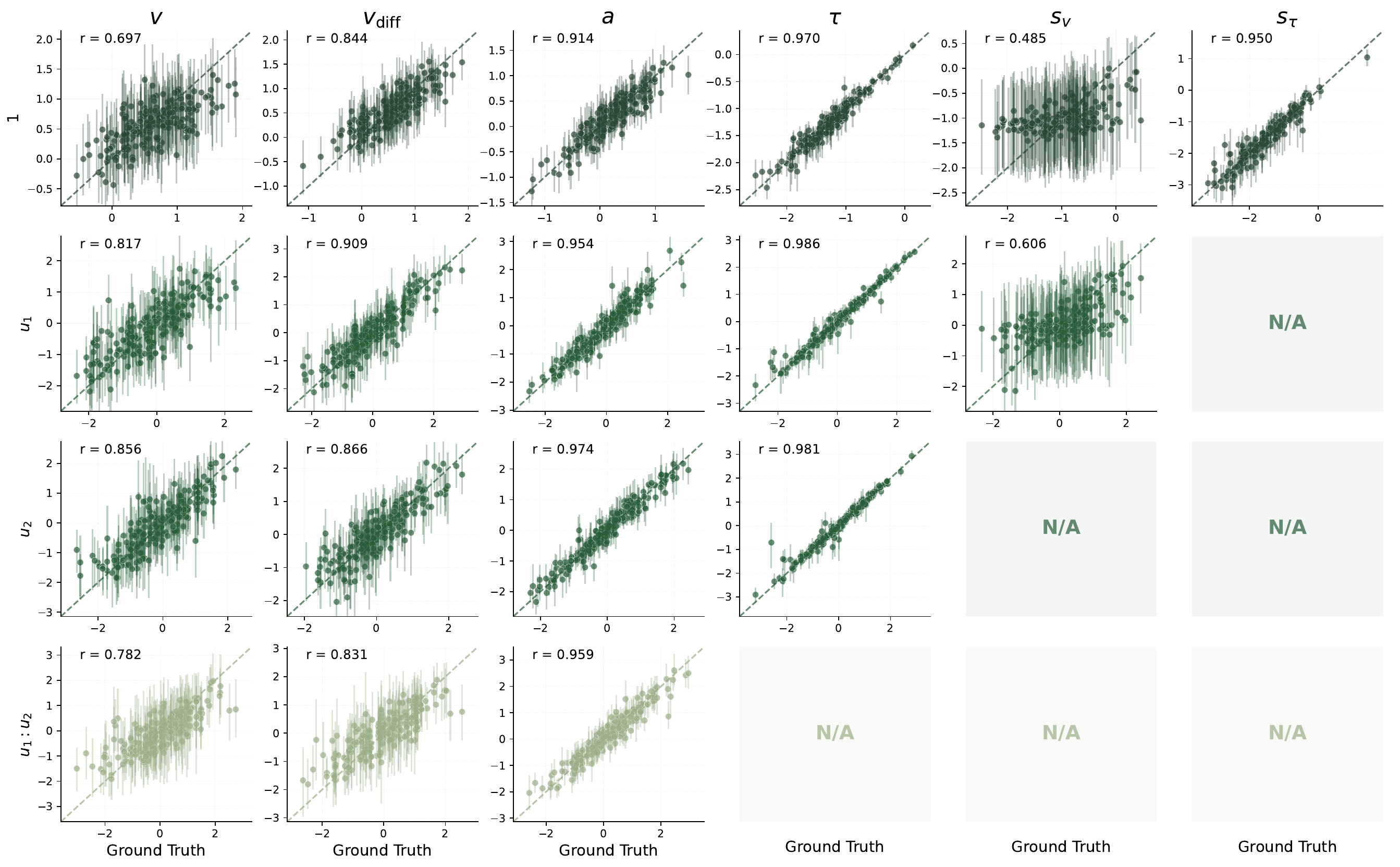}
    \includegraphics[width=0.97\linewidth]{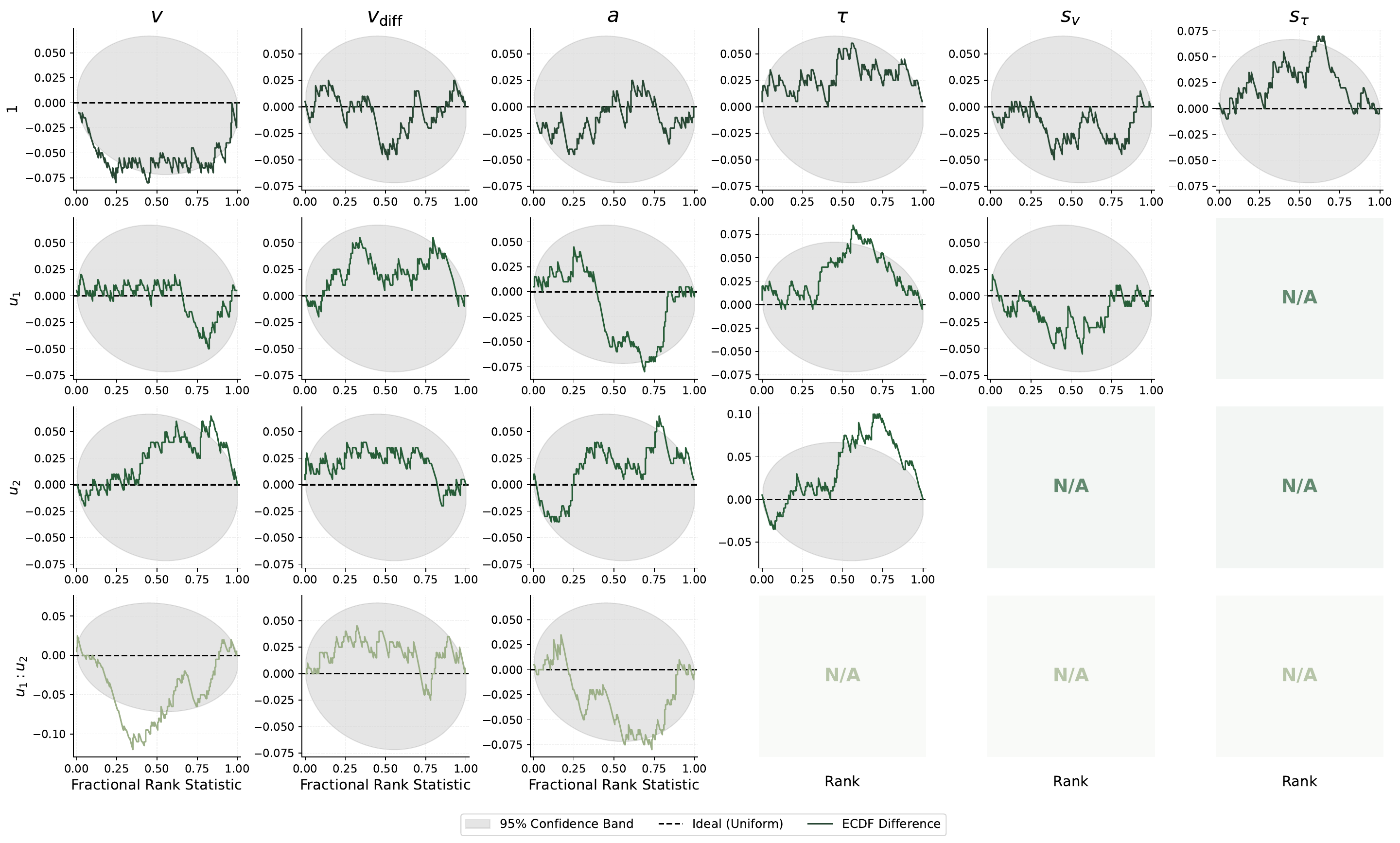}
    \includegraphics[width=0.97\linewidth]{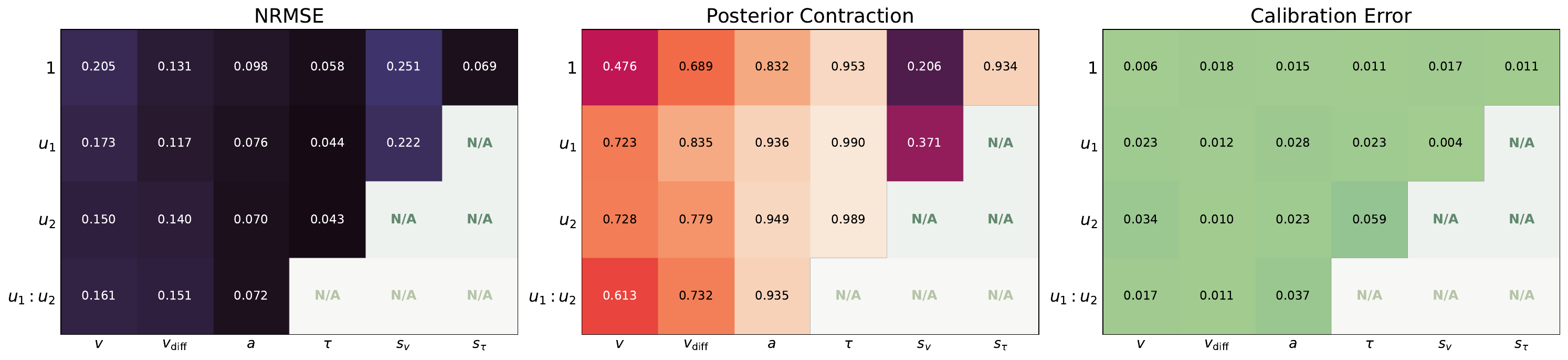}
    \caption{Parameter recovery (\emph{top}), calibration ECDF (\emph{middle}), and validation metrics (NRMSE, calibration error, and posterior contraction) for RDM baseline (Case \textbf{interaction}).}
    \label{fig:rdm-bf-interaction}
\end{figure}


\begin{figure}[!h]
    \centering
    \includegraphics[width=0.97\linewidth]{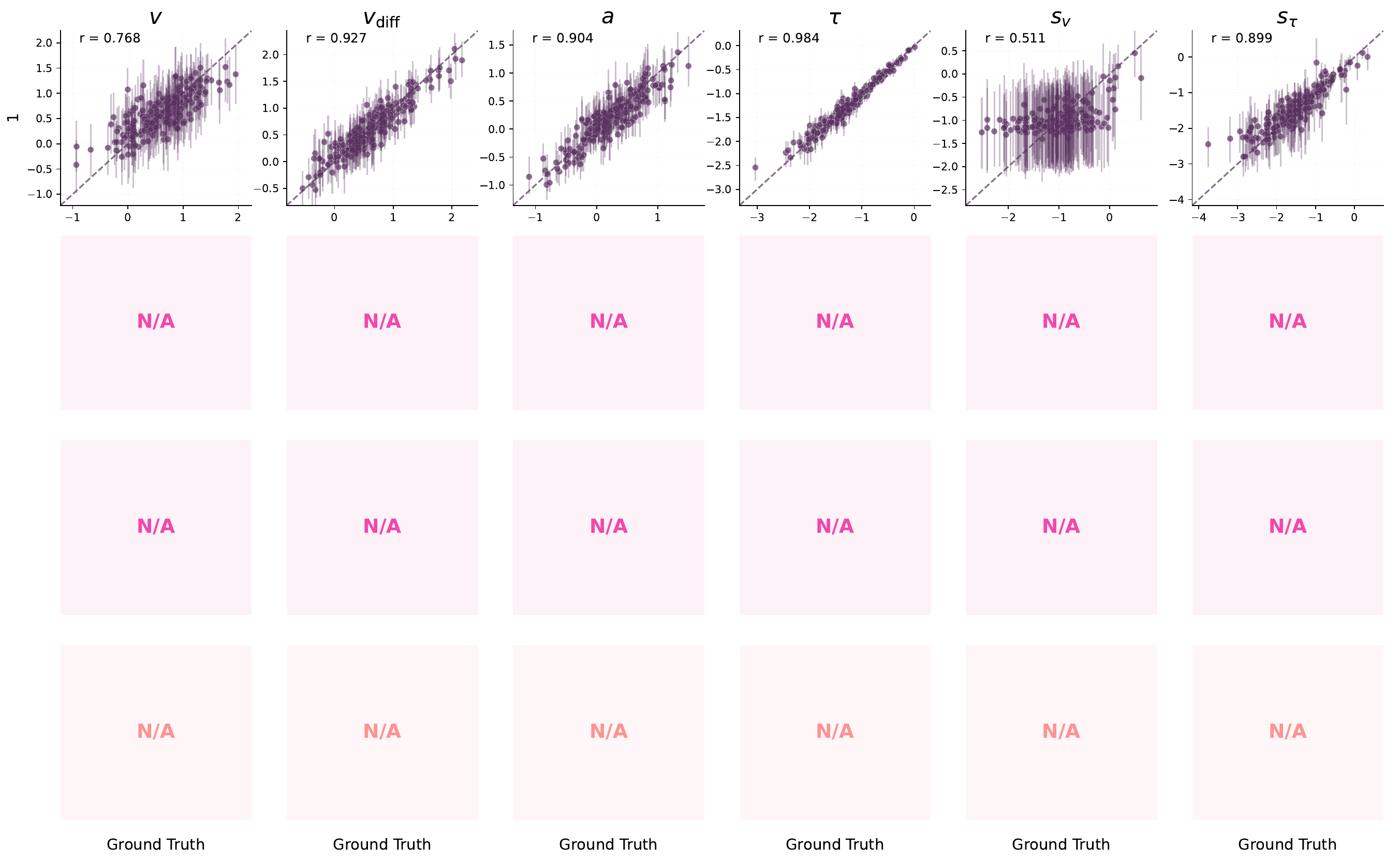}
    \includegraphics[width=0.97\linewidth]{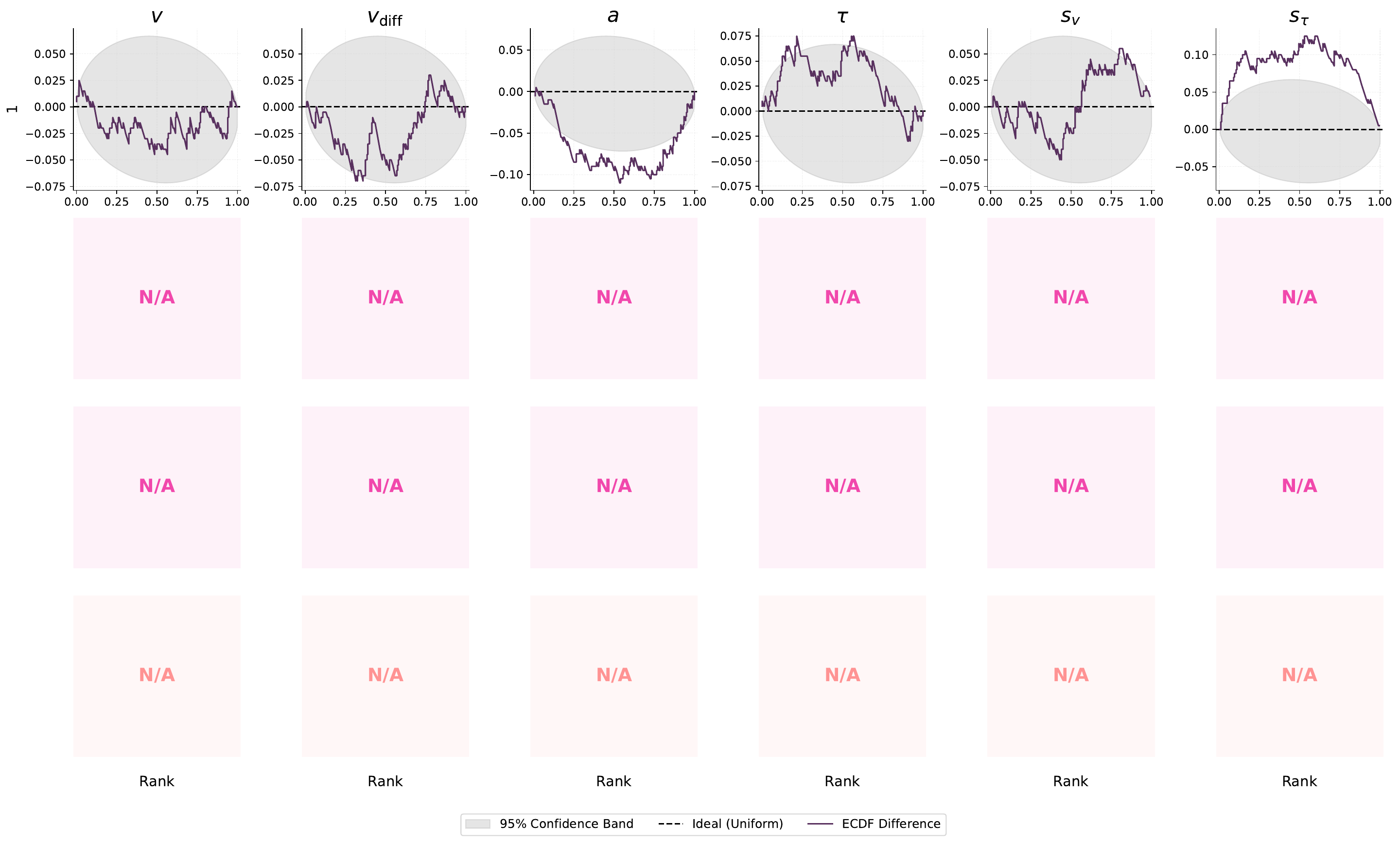}
    \includegraphics[width=0.97\linewidth]{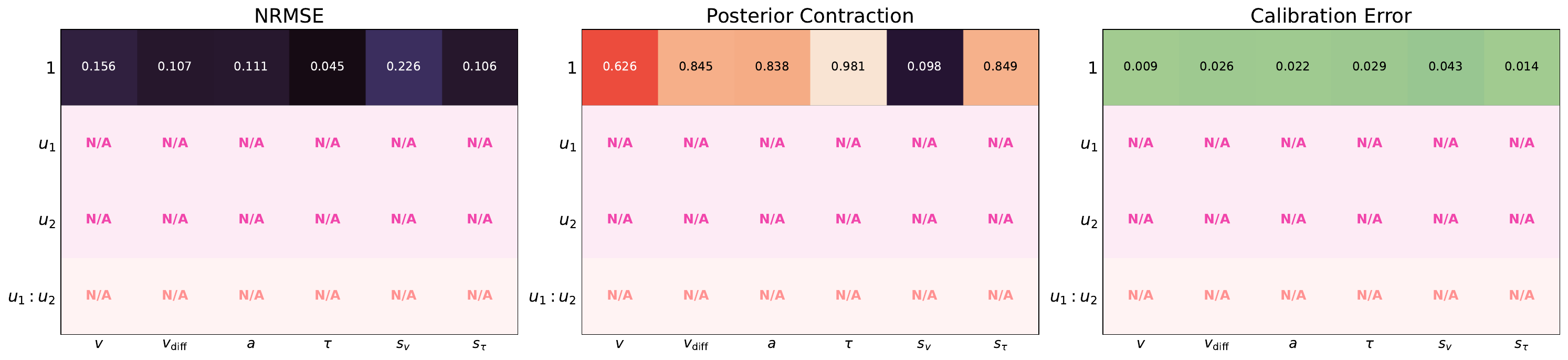}
    \caption{Parameter recovery (\emph{top}), calibration ECDF (\emph{middle}), and validation metrics (NRMSE, calibration error, and posterior contraction) for RDM model family (Case \textbf{intercept\_only}).}
    \label{fig:rdm-fm-intercept-only}
\end{figure}

\begin{figure}[!h]
    \centering
    \includegraphics[width=0.97\linewidth]{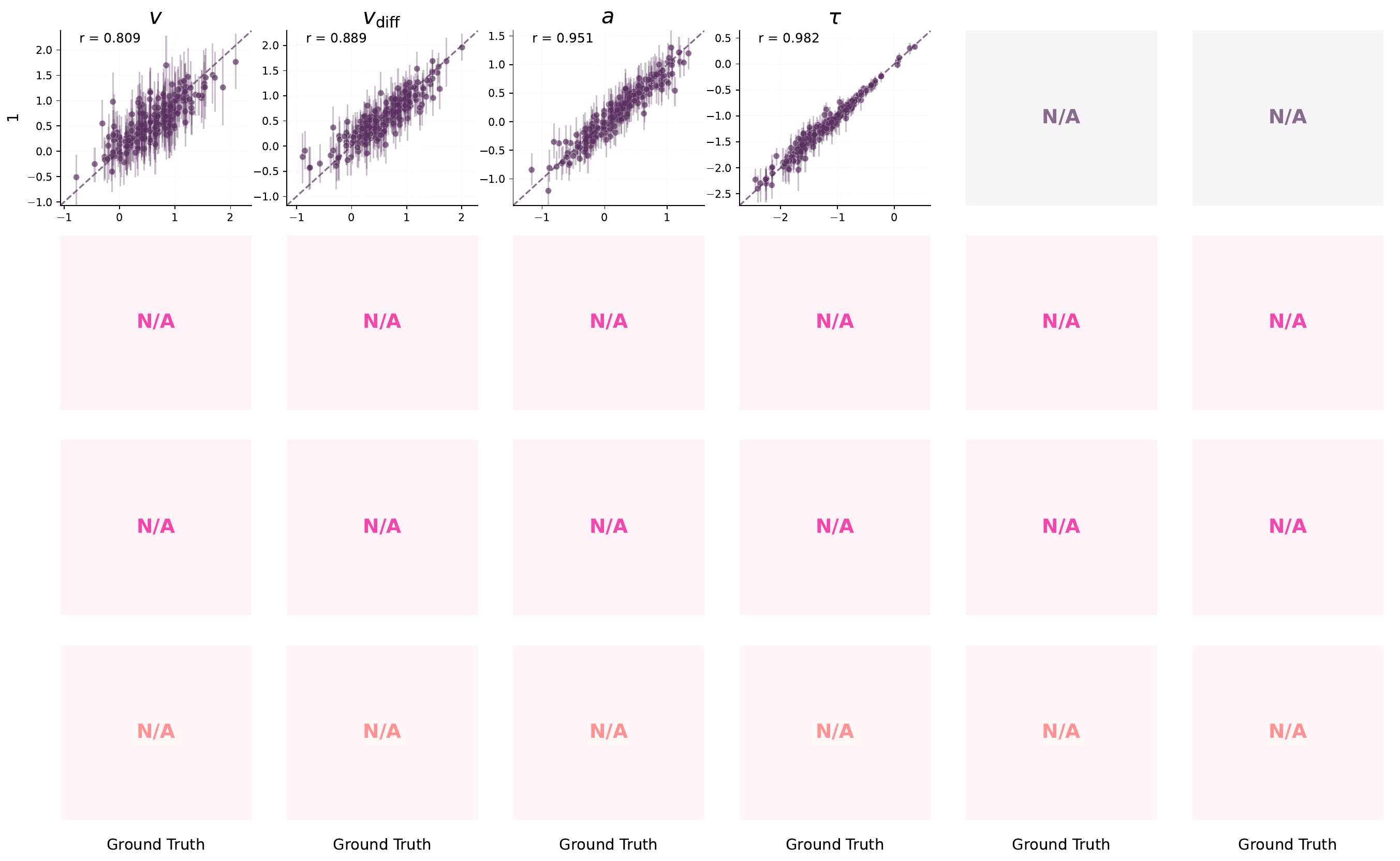}
    \includegraphics[width=0.97\linewidth]{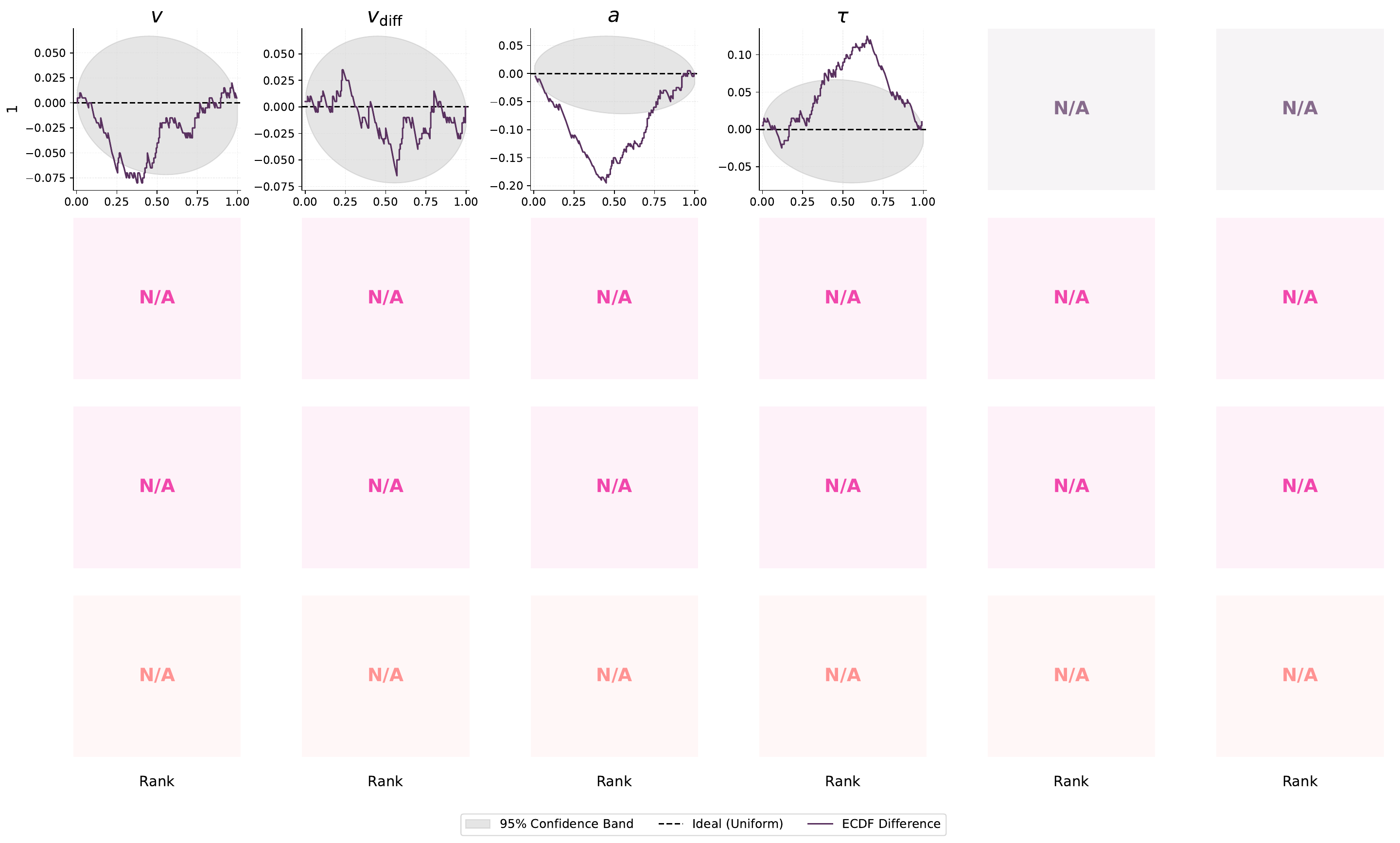}
    \includegraphics[width=0.97\linewidth]{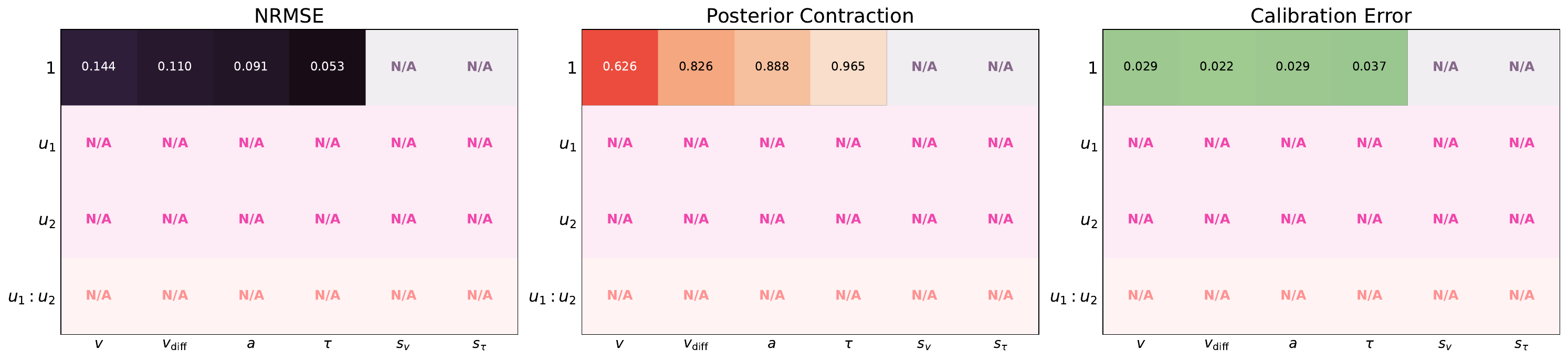}
    \caption{Parameter recovery (\emph{top}), calibration ECDF (\emph{middle}), and validation metrics (NRMSE, calibration error, and posterior contraction) for RDM model family (Case \textbf{fixed}).}
    \label{fig:rdm-fm-fixed}
\end{figure}

\begin{figure}[!h]
    \centering
    \includegraphics[width=0.97\linewidth]{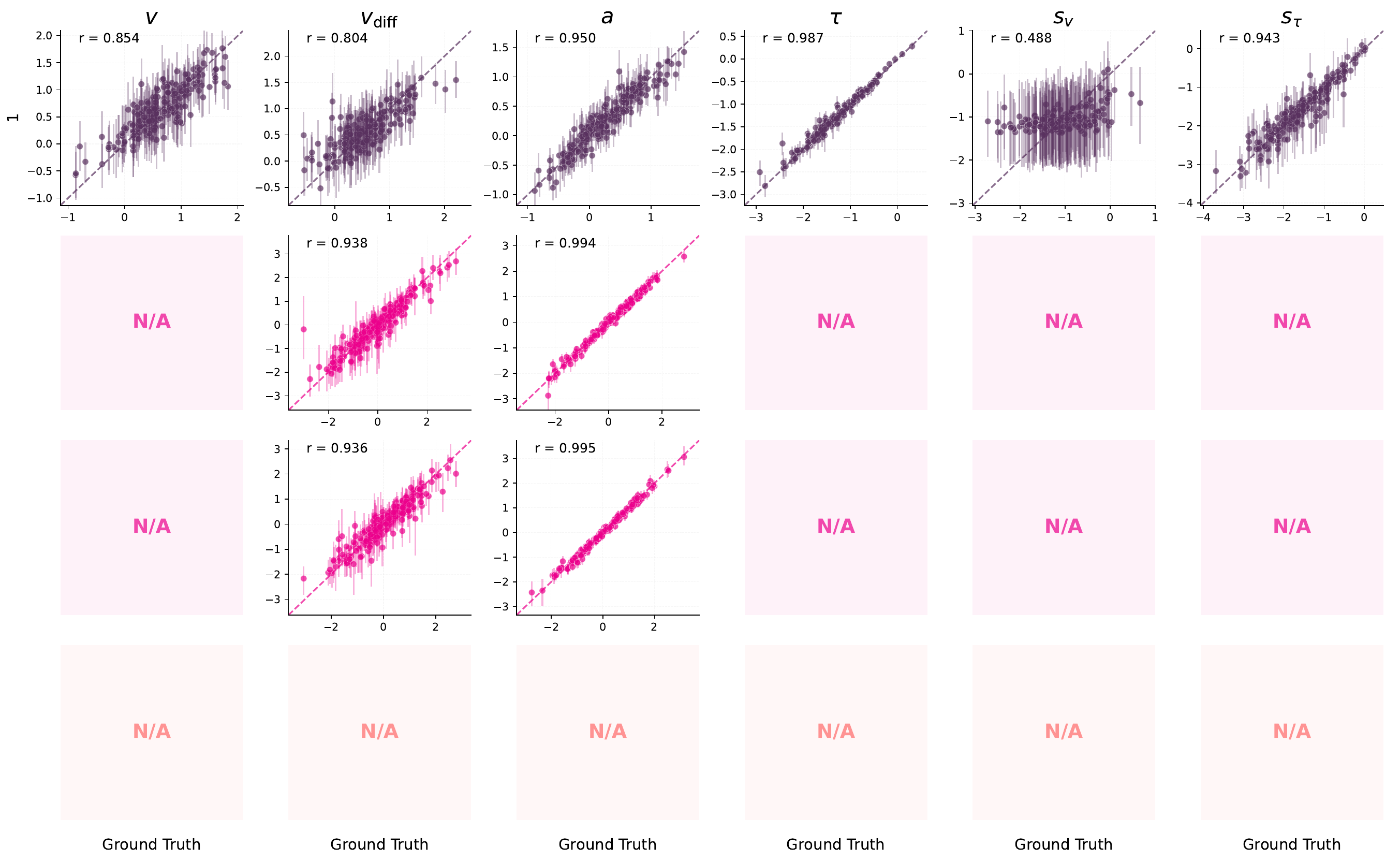}
    \includegraphics[width=0.97\linewidth]{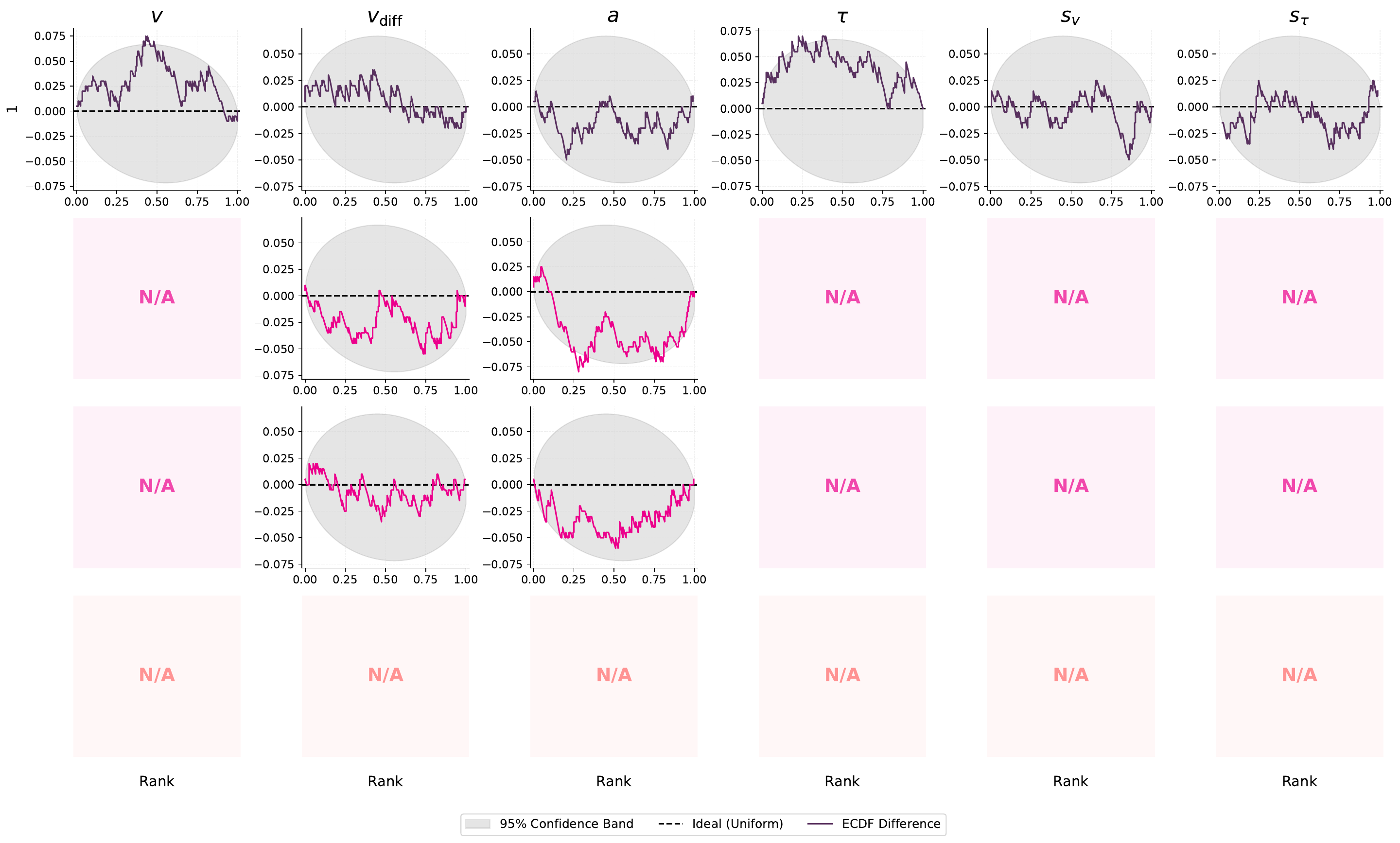}
    \includegraphics[width=0.97\linewidth]{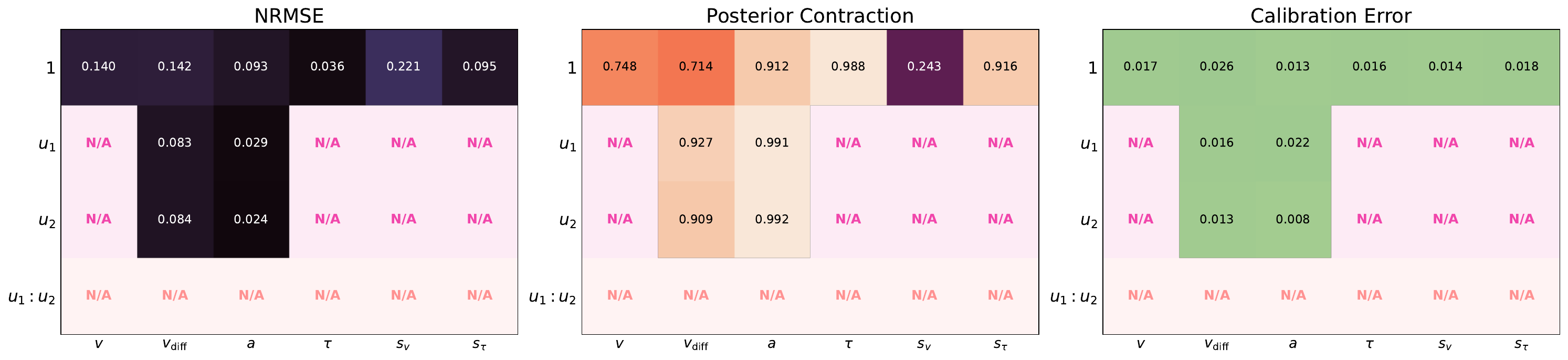}
    \caption{Parameter recovery (\emph{top}), calibration ECDF (\emph{middle}), and validation metrics (NRMSE, calibration error, and posterior contraction) for RDM model family (Case \textbf{regressed}).}
    \label{fig:rdm-fm-regressed}
\end{figure}

\begin{figure}[!h]
    \centering
    \includegraphics[width=0.97\linewidth]{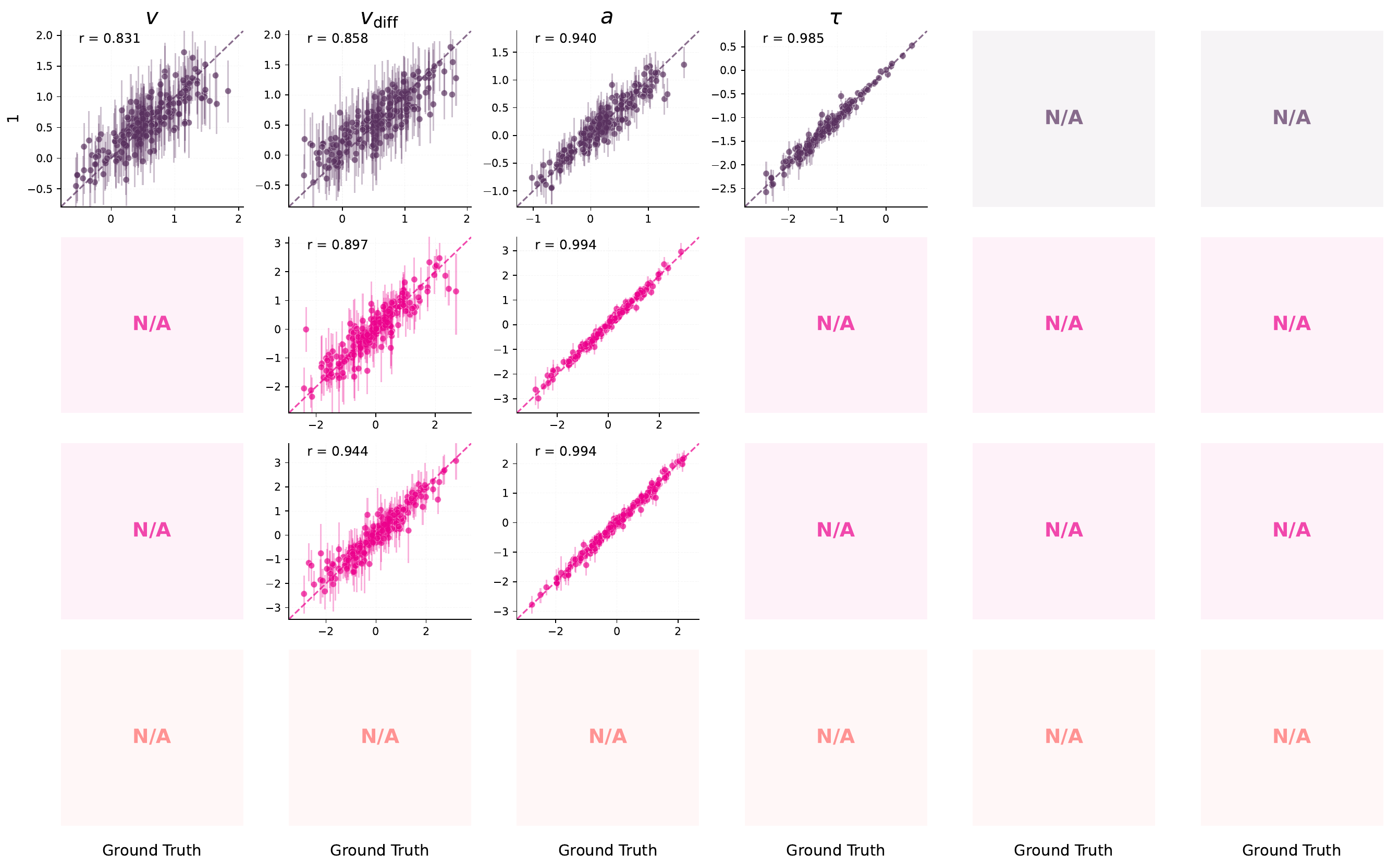}
    \includegraphics[width=0.97\linewidth]{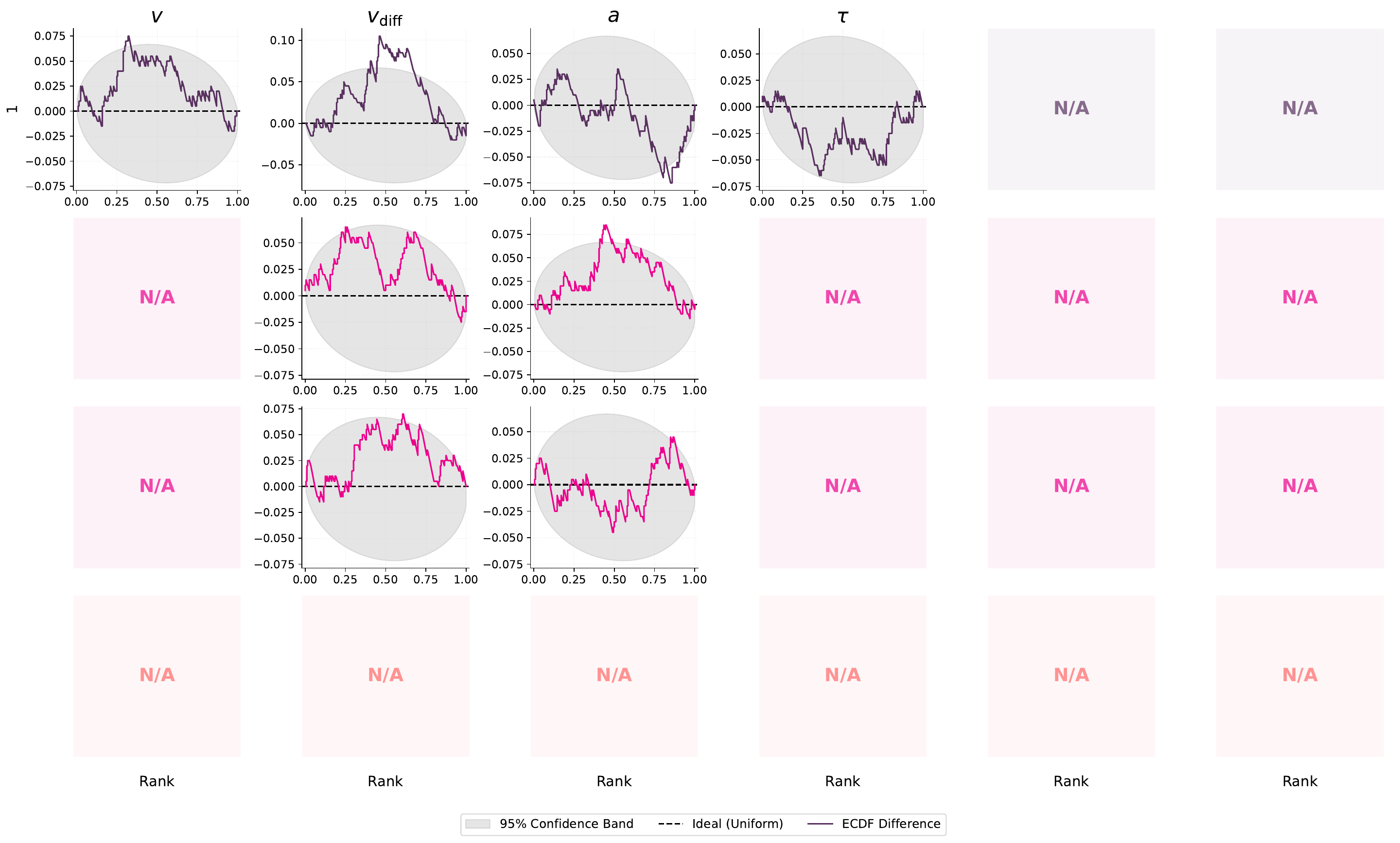}
    \includegraphics[width=0.97\linewidth]{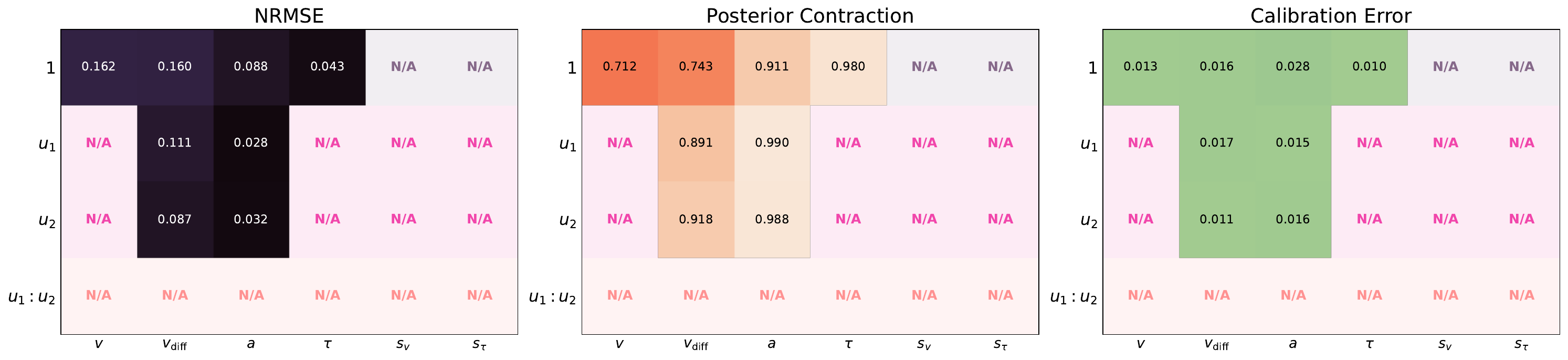}
    \caption{Parameter recovery (\emph{top}), calibration ECDF (\emph{middle}), and validation metrics (NRMSE, calibration error, and posterior contraction) for RDM model family (Case \textbf{fixed\_regressed}).}
    \label{fig:rdm-fm-fixed-regressed}
\end{figure}

\begin{figure}[!h]
    \centering
    \includegraphics[width=0.97\linewidth]{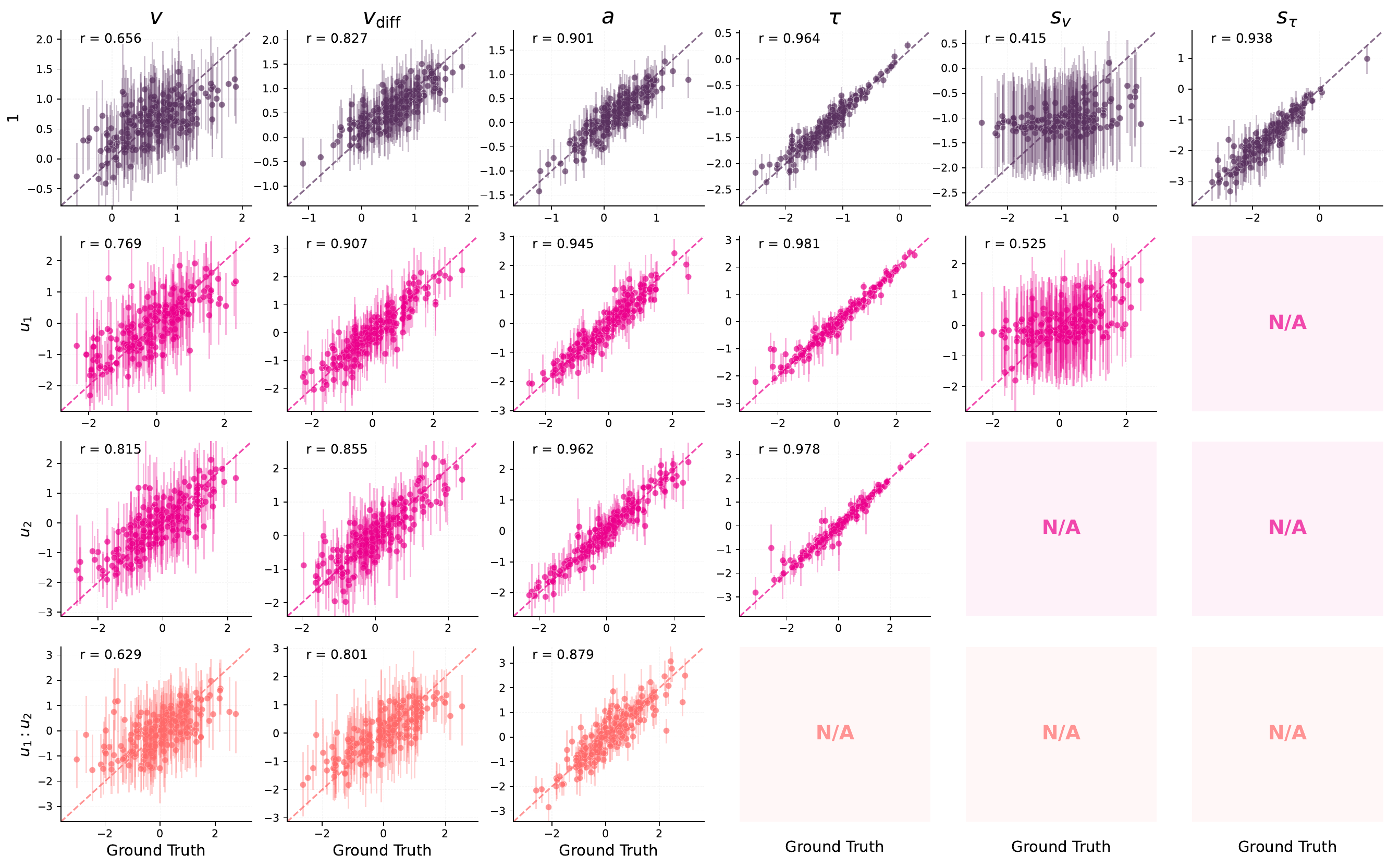}
    \includegraphics[width=0.97\linewidth]{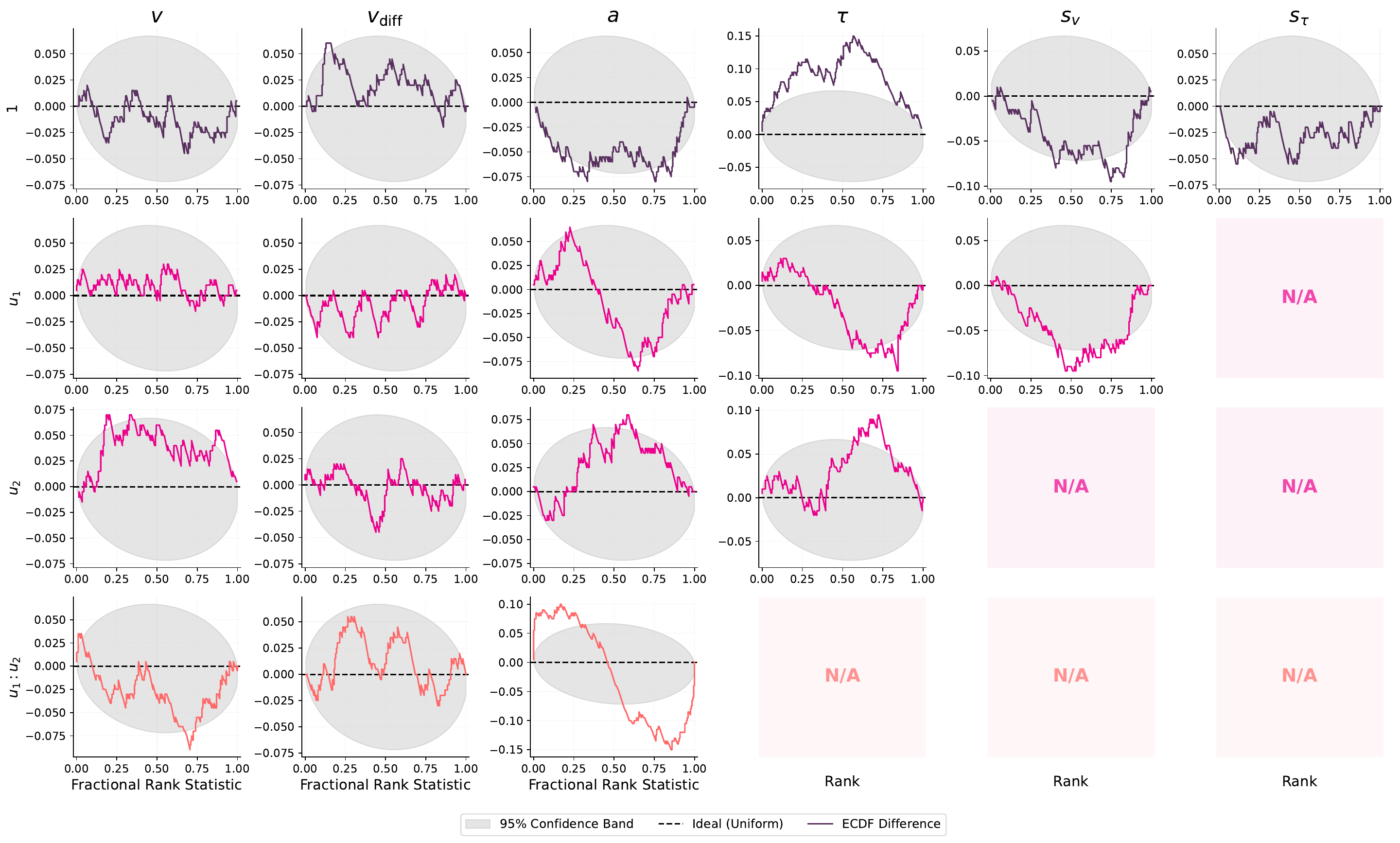}
    \includegraphics[width=0.97\linewidth]{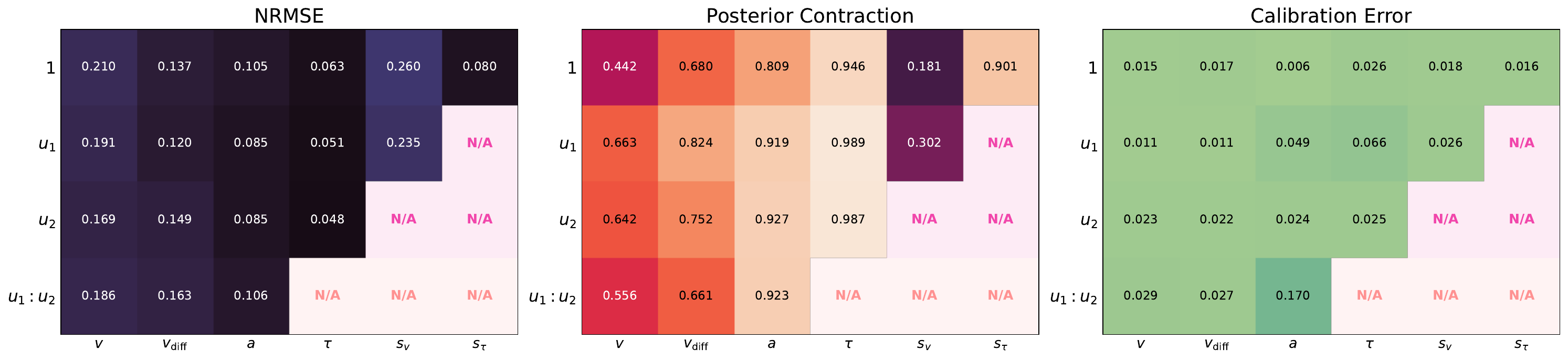}
    \caption{Parameter recovery (\emph{top}), calibration ECDF (\emph{middle}), and validation metrics (NRMSE, calibration error, and posterior contraction) for RDM model family (Case \textbf{interaction}).}
    \label{fig:rdm-fm-interaction}
\end{figure}


\begin{figure}[!h]
    \centering
    \includegraphics[width=0.97\linewidth]{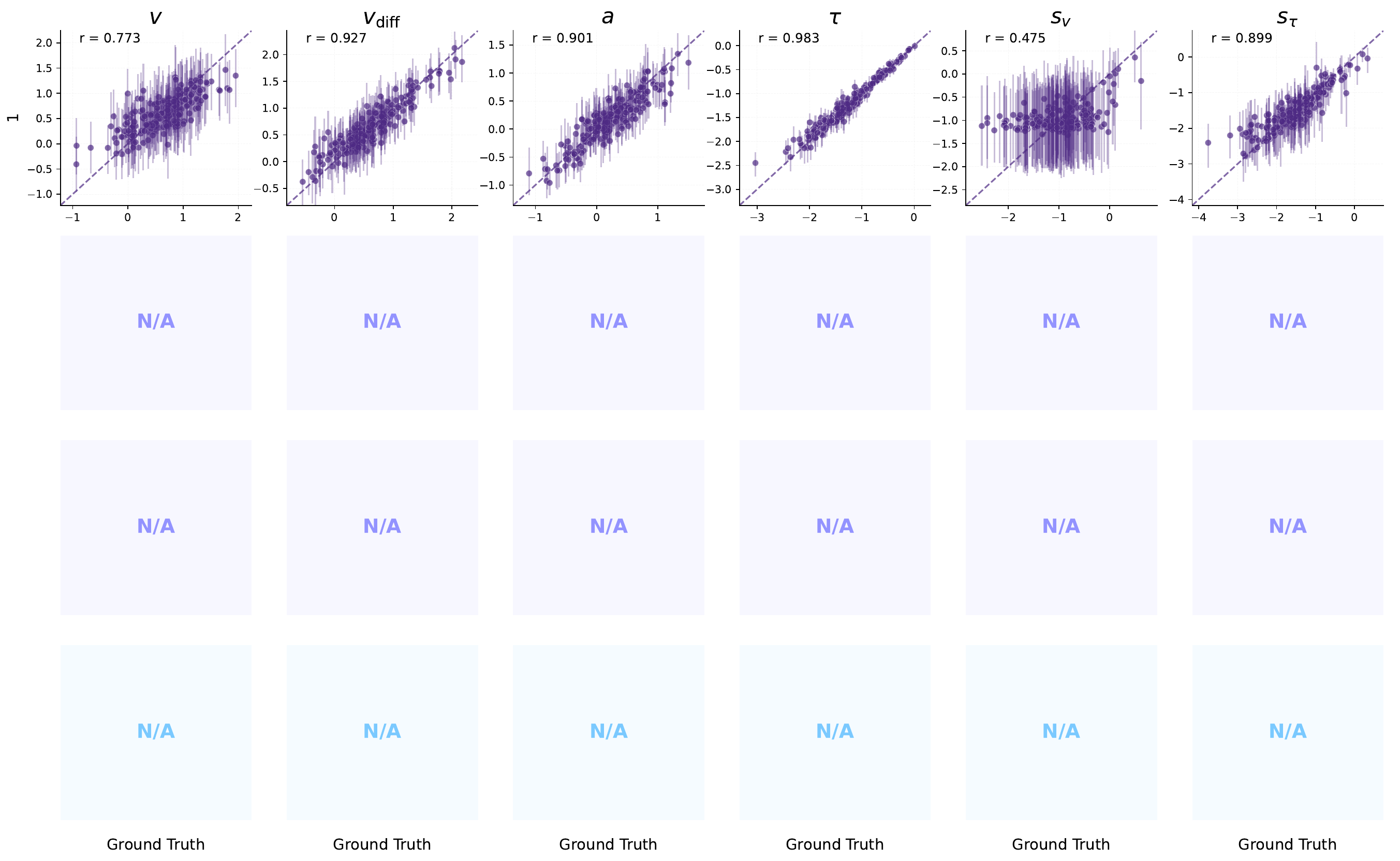}
    \includegraphics[width=0.97\linewidth]{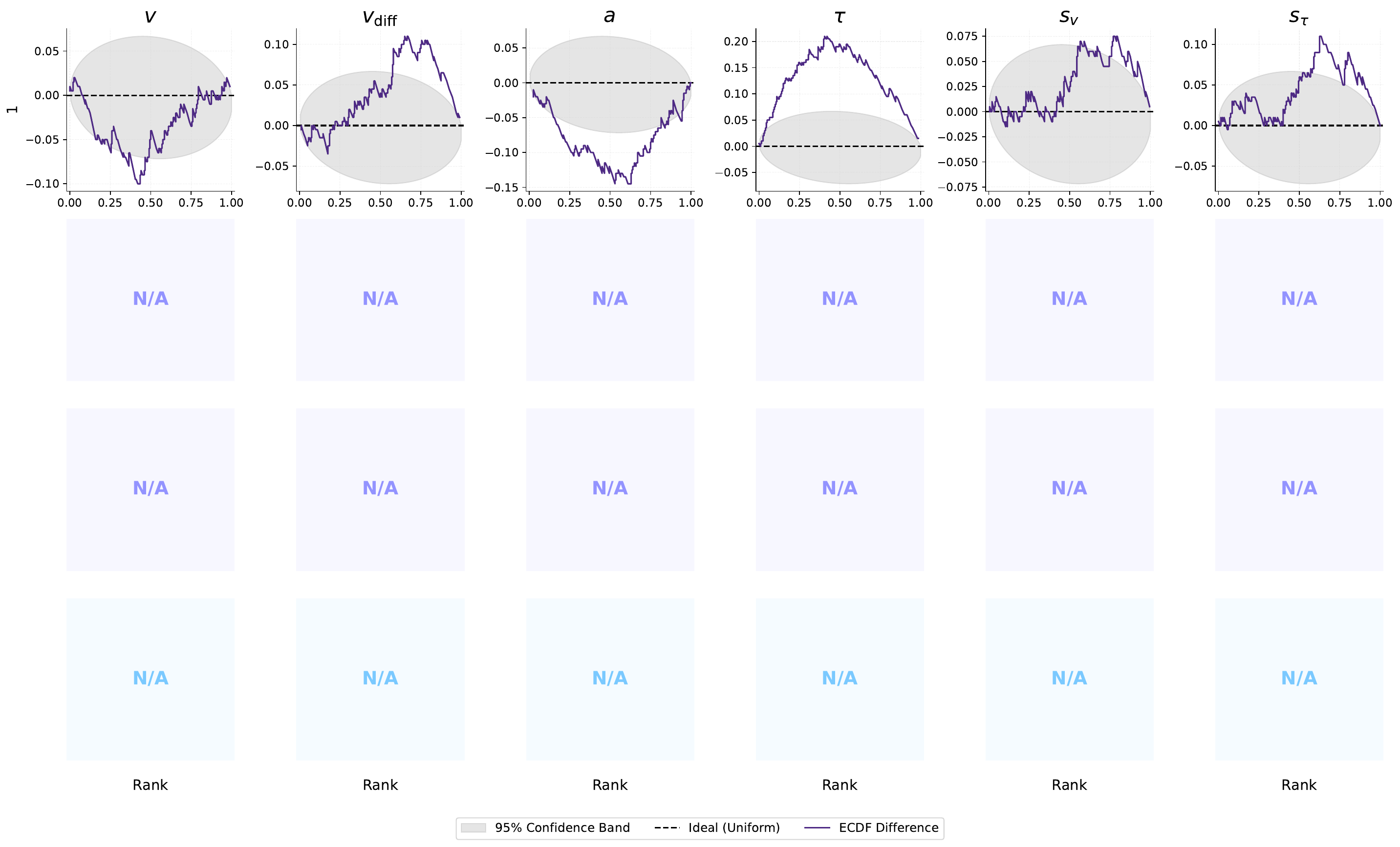}
    \includegraphics[width=0.97\linewidth]{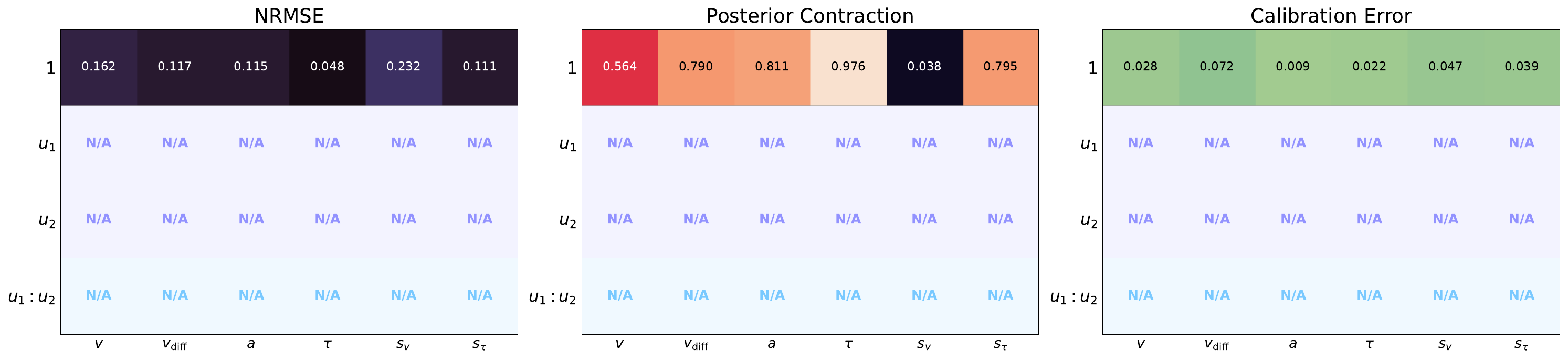}
    \caption{Parameter recovery (\emph{top}), calibration ECDF (\emph{middle}), and parameter-wise metrics (NRMSE, calibration error, and posterior contraction) for RDM model class (Case \textbf{intercept\_only}).}
    \label{fig:rdm-mc-intercept-only}
\end{figure}

\begin{figure}[!h]
    \centering
    \includegraphics[width=0.97\linewidth]{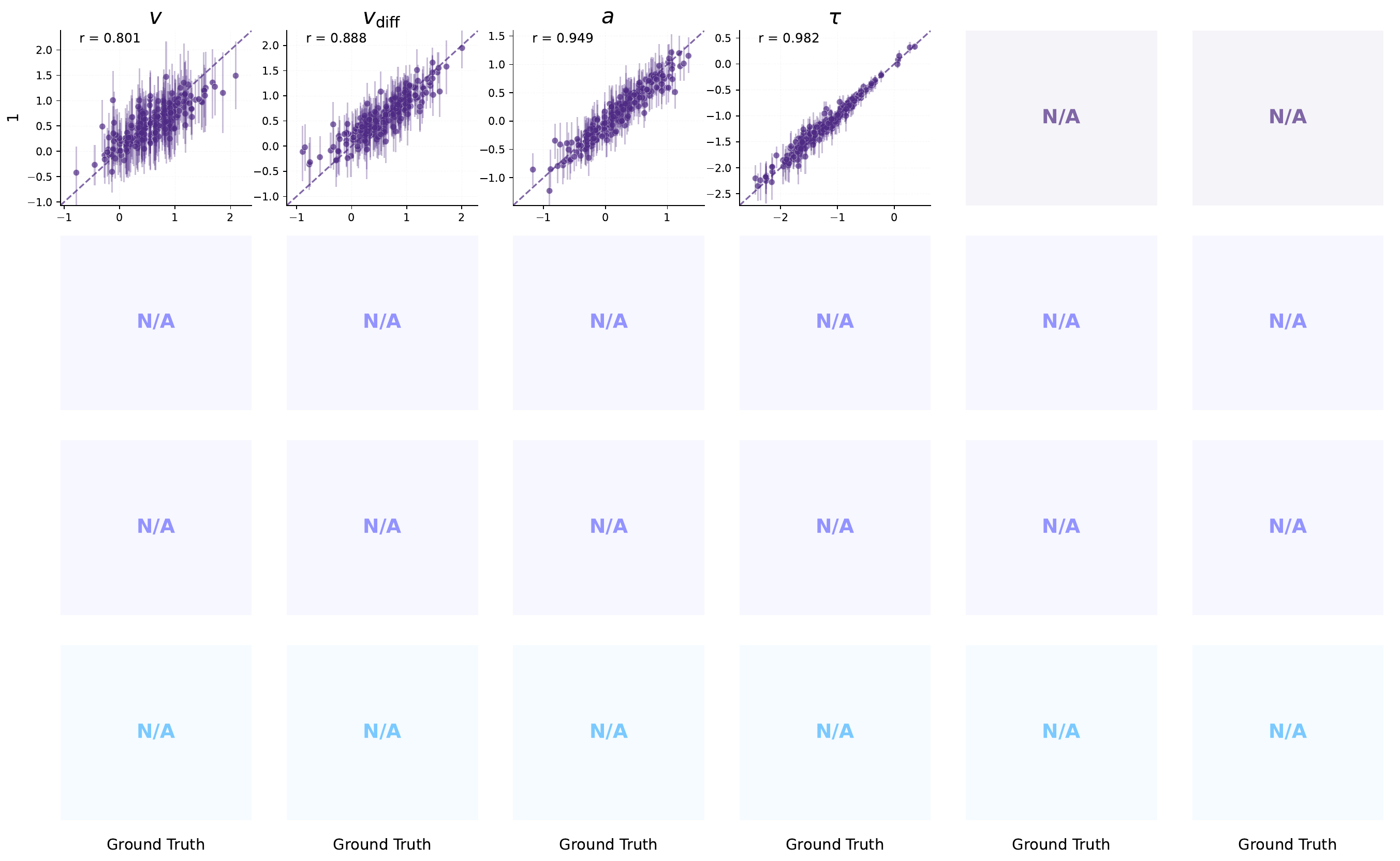}
    \includegraphics[width=0.97\linewidth]{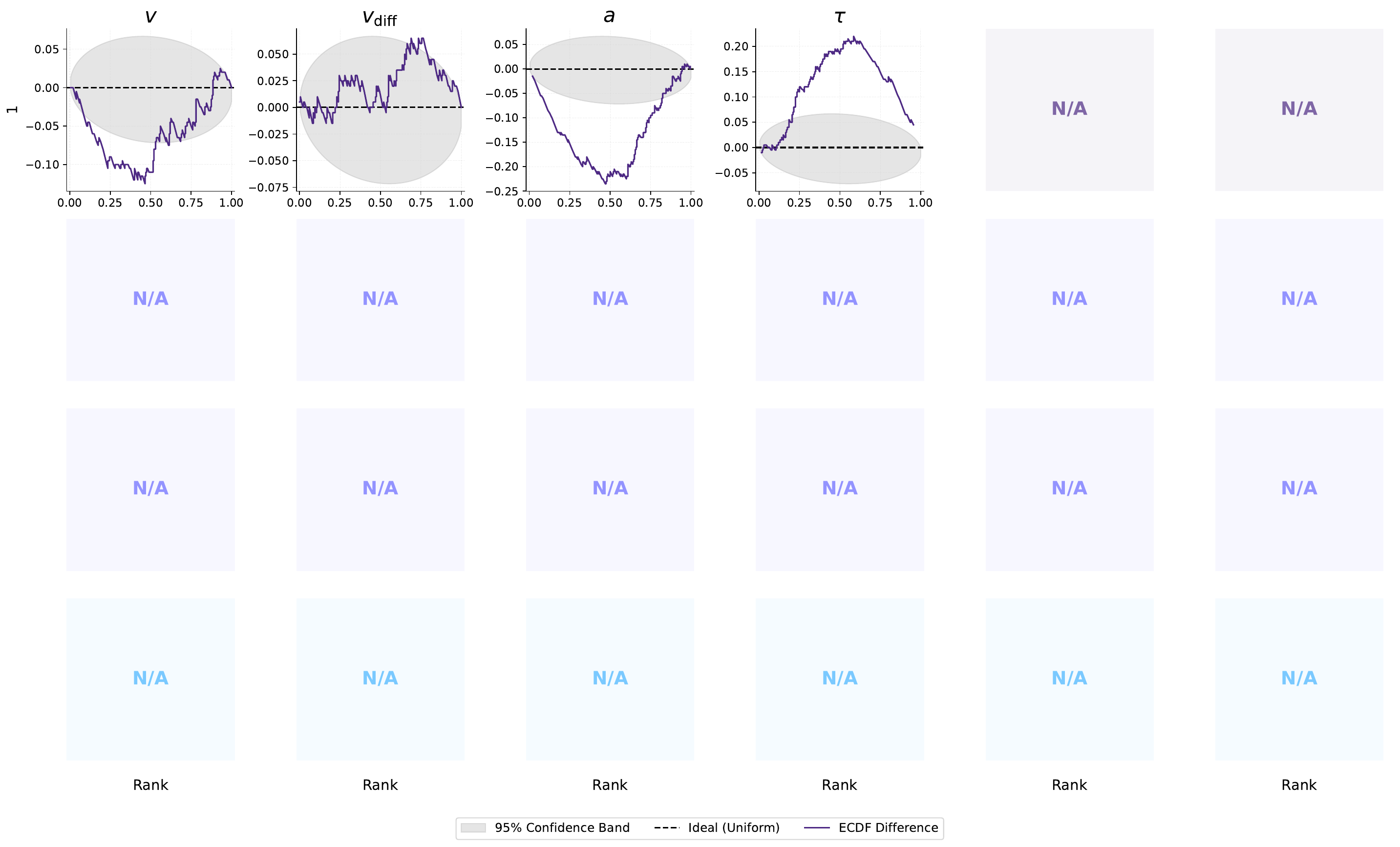}
    \includegraphics[width=0.97\linewidth]{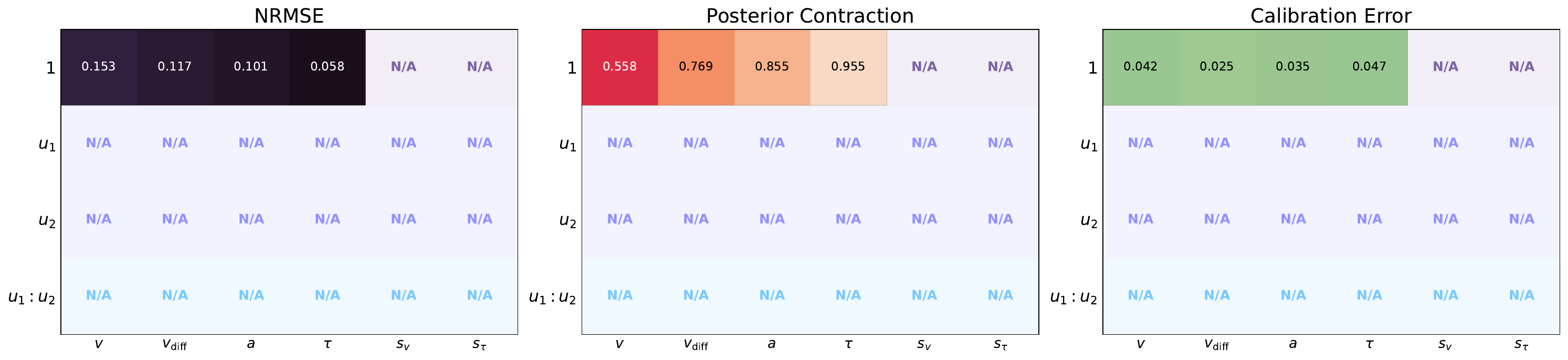}
    \caption{Parameter recovery (\emph{top}), calibration ECDF (\emph{middle}), and parameter-wise metrics (NRMSE, calibration error, and posterior contraction) for RDM model class (Case \textbf{fixed}).}
    \label{fig:rdm-mc-fixed}
\end{figure}

\begin{figure}[!h]
    \centering
    \includegraphics[width=0.97\linewidth]{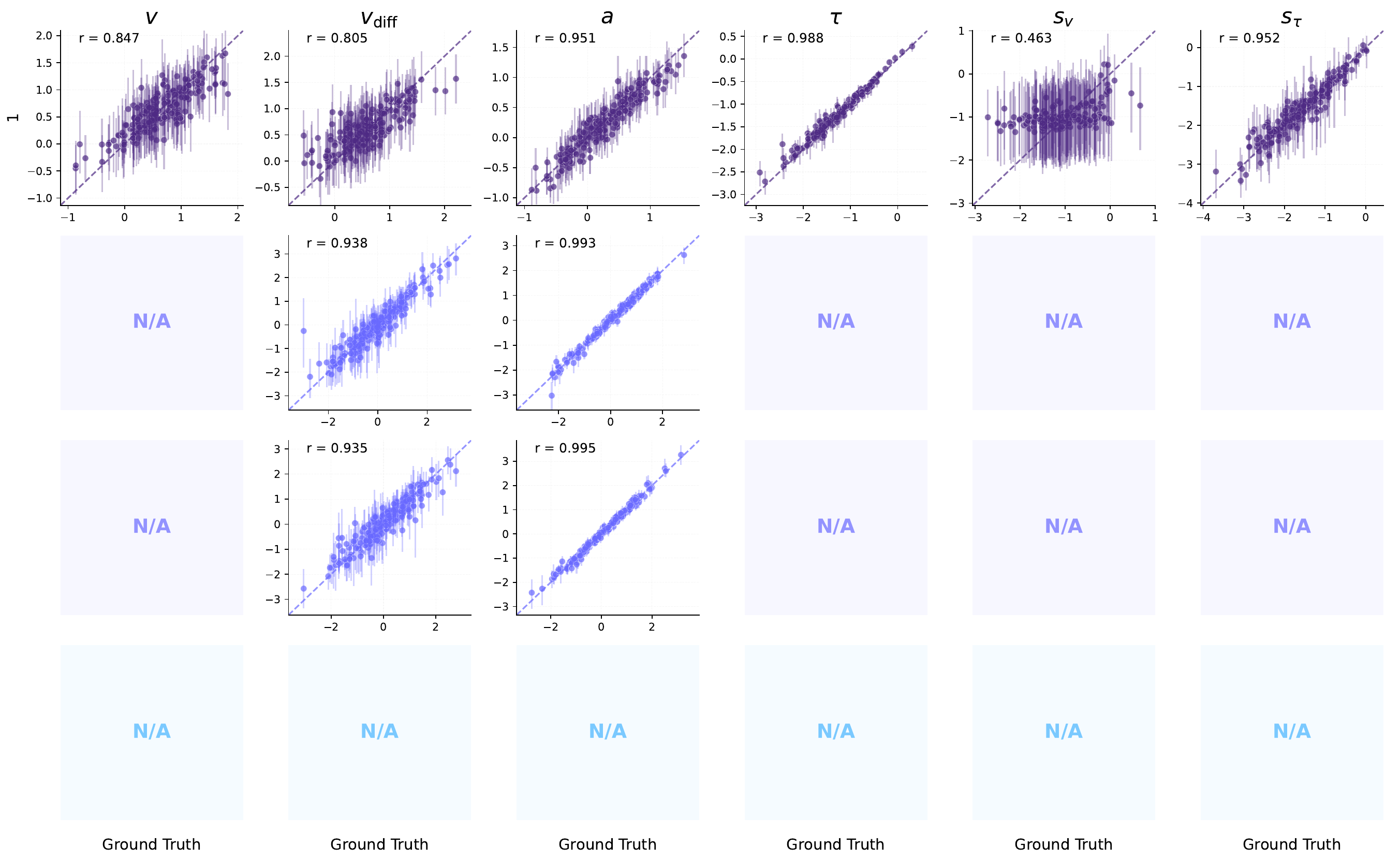}
    \includegraphics[width=0.97\linewidth]{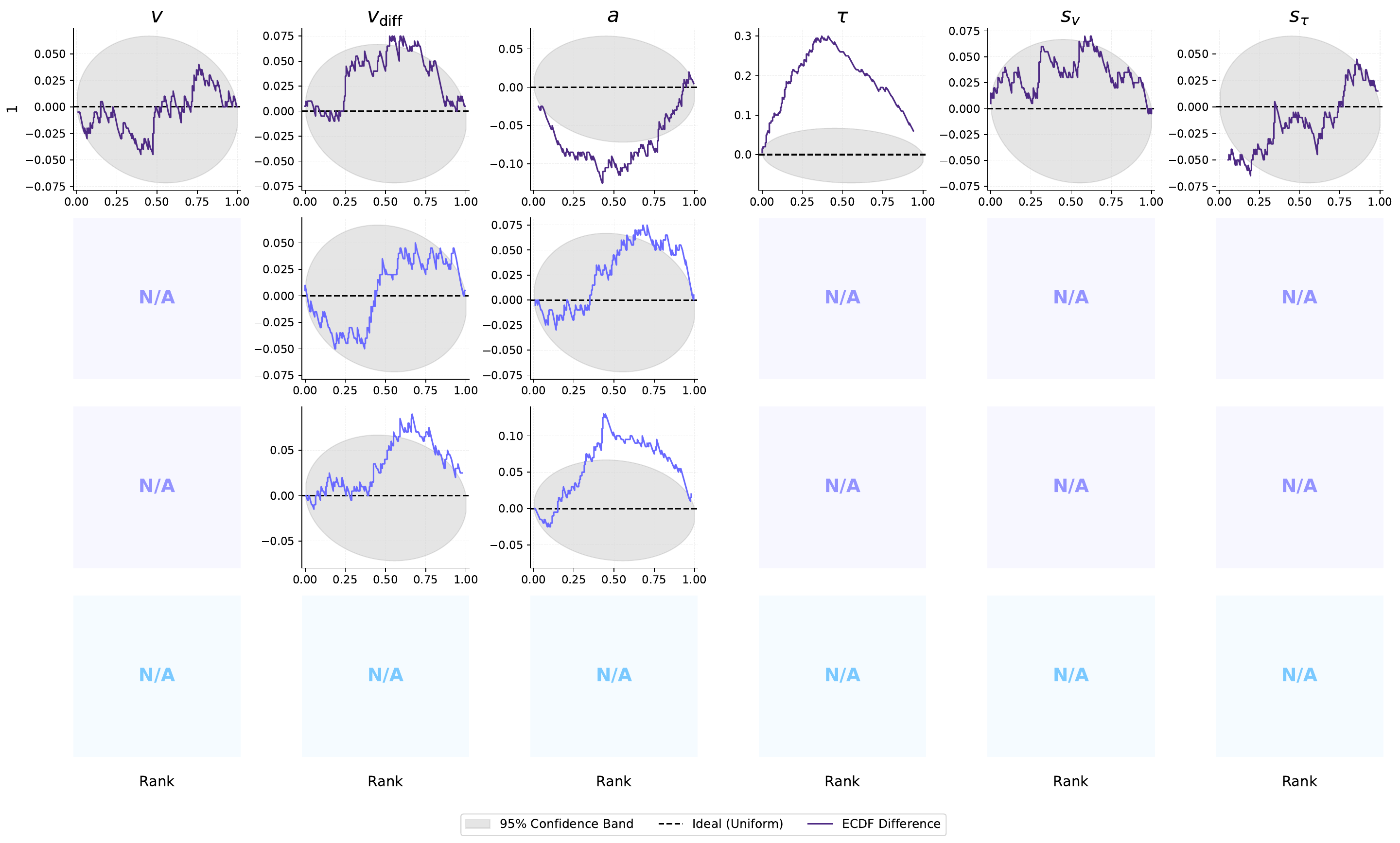}
    \includegraphics[width=0.97\linewidth]{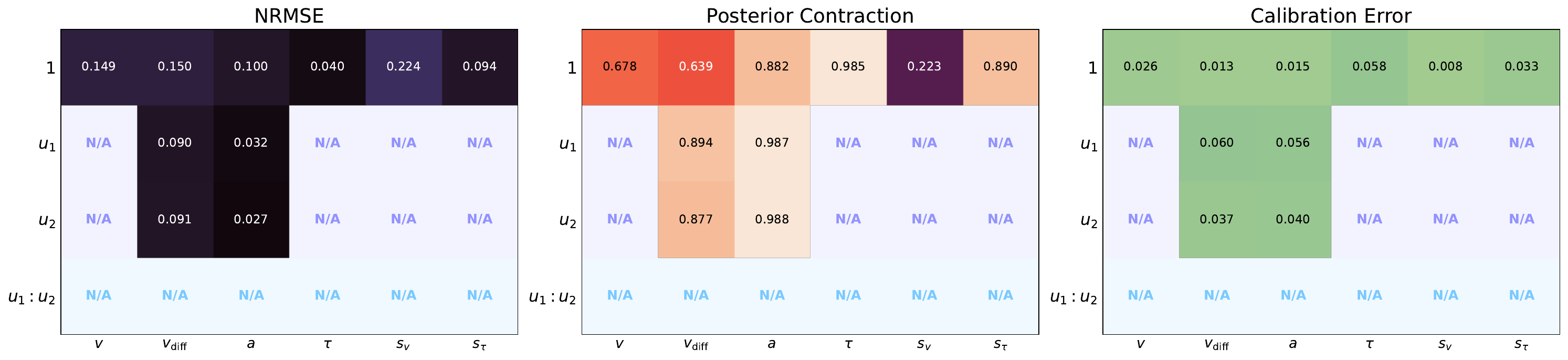}
    \caption{Parameter recovery (\emph{top}), calibration ECDF (\emph{middle}), and parameter-wise metrics (NRMSE, calibration error, and posterior contraction) for RDM model class (Case \textbf{regressed}).}
    \label{fig:rdm-mc-regressed}
\end{figure}

\begin{figure}[!h]
    \centering
    \includegraphics[width=0.97\linewidth]{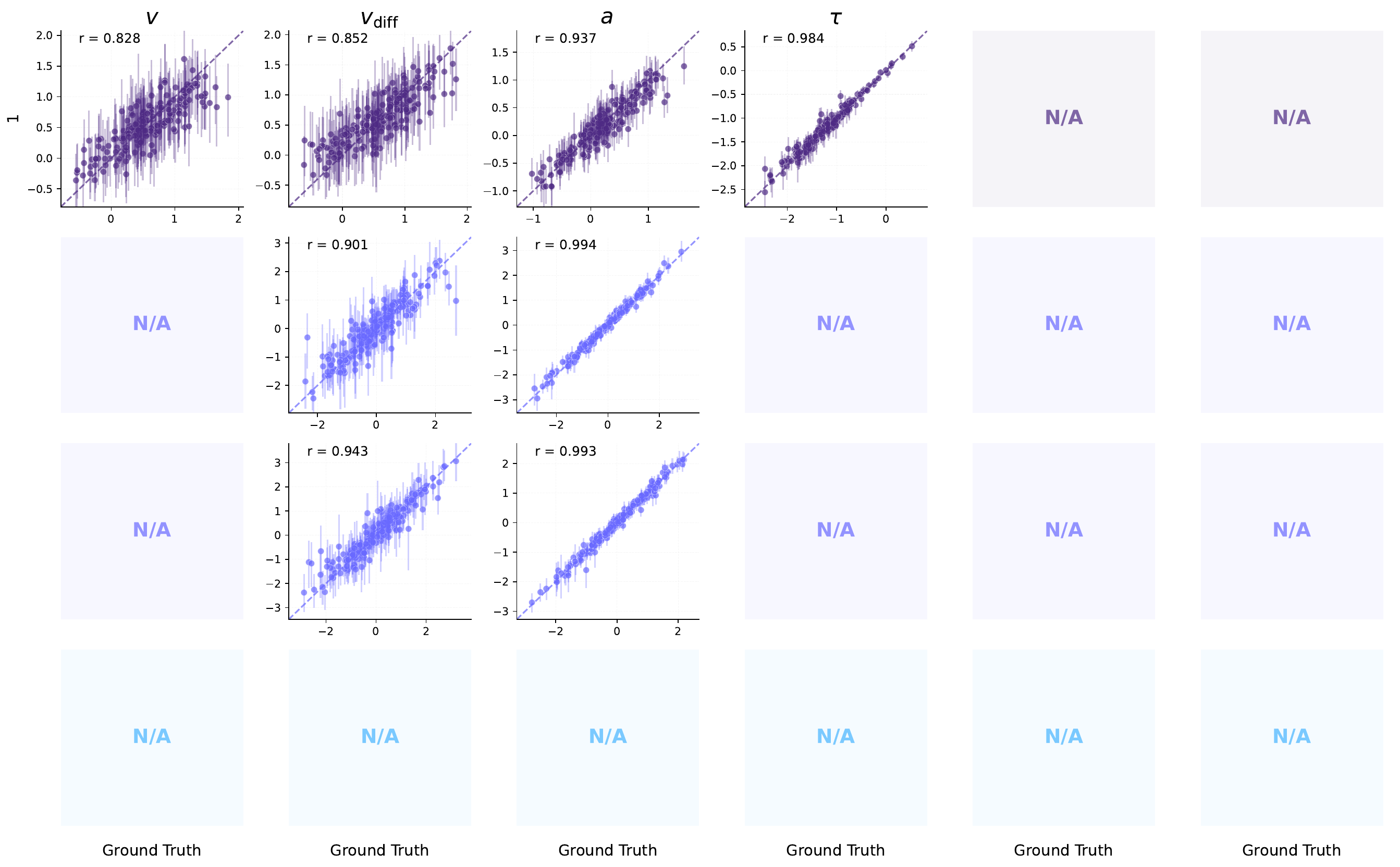}
    \includegraphics[width=0.97\linewidth]{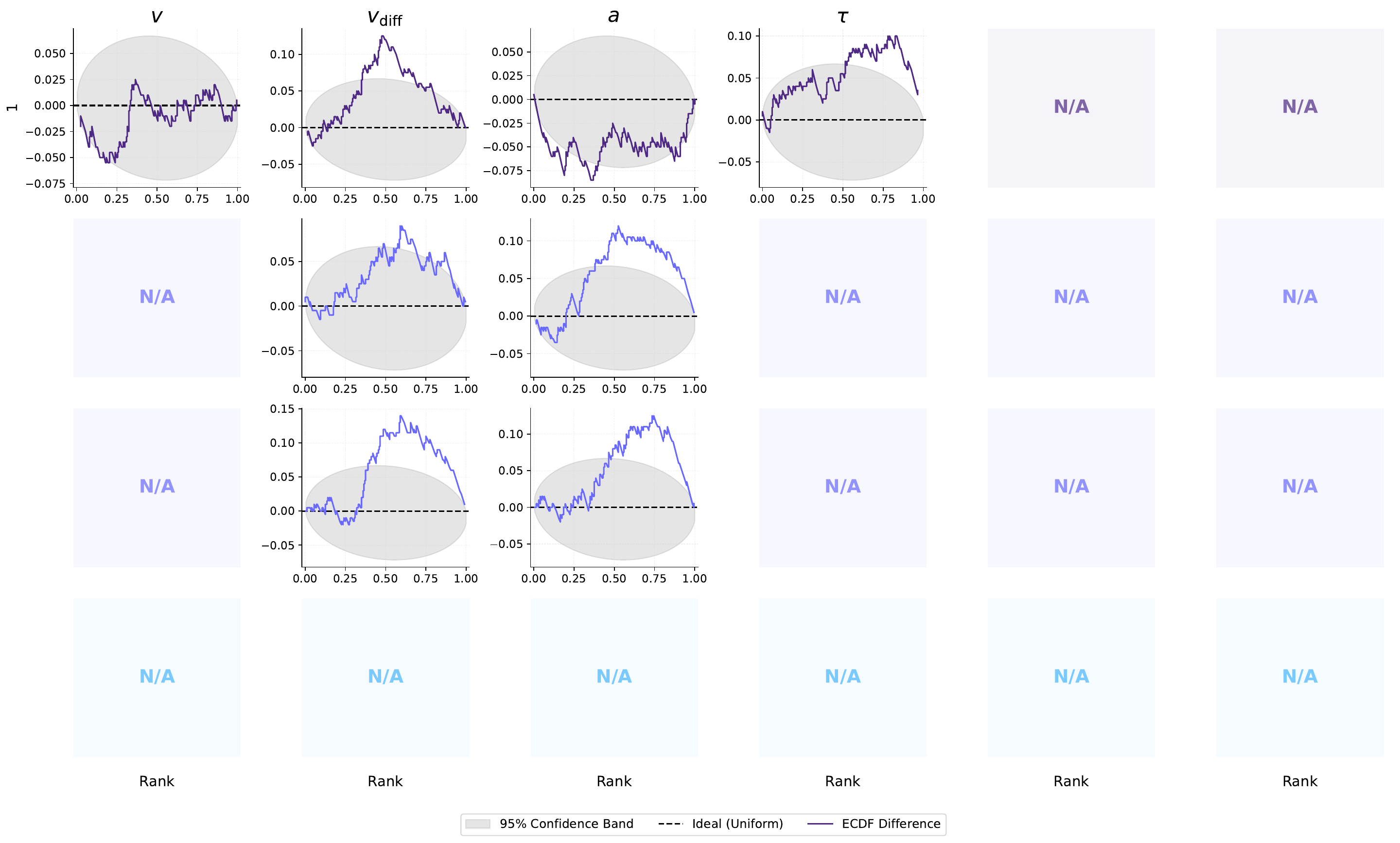}
    \includegraphics[width=0.97\linewidth]{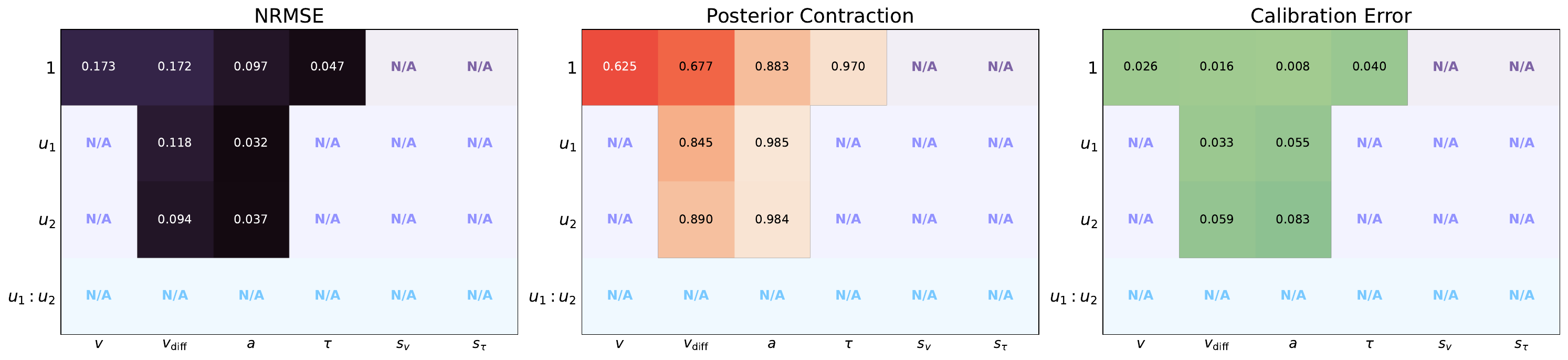}
    \caption{Parameter recovery (\emph{top}), calibration ECDF (\emph{middle}), and parameter-wise metrics (NRMSE, calibration error, and posterior contraction) for RDM model class (Case \textbf{fixed\_regressed}).}
    \label{fig:rdm-mc-fixed-regressed}
\end{figure}

\begin{figure}[!h]
    \centering
    \includegraphics[width=0.97\linewidth]{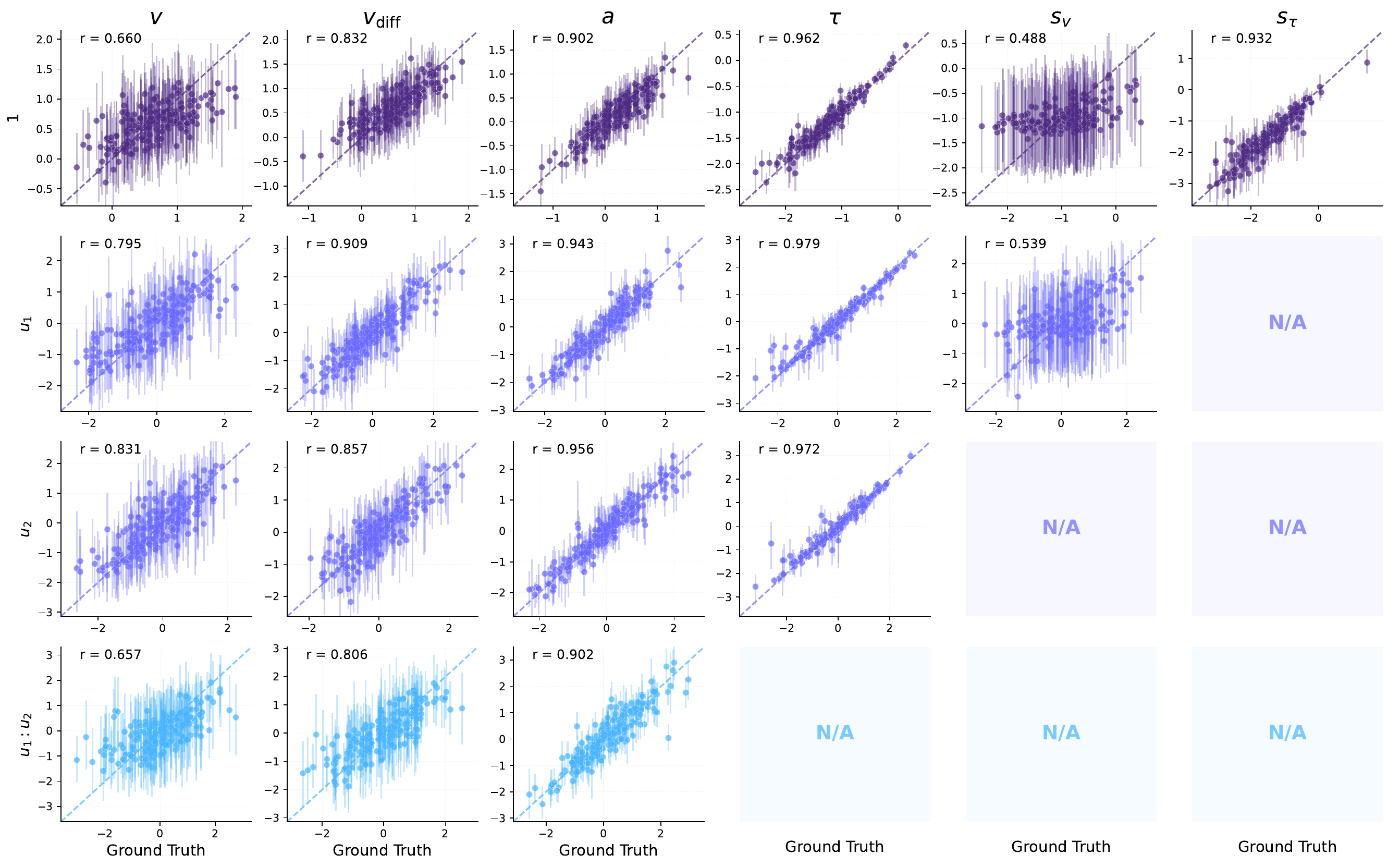}
    \includegraphics[width=0.97\linewidth]{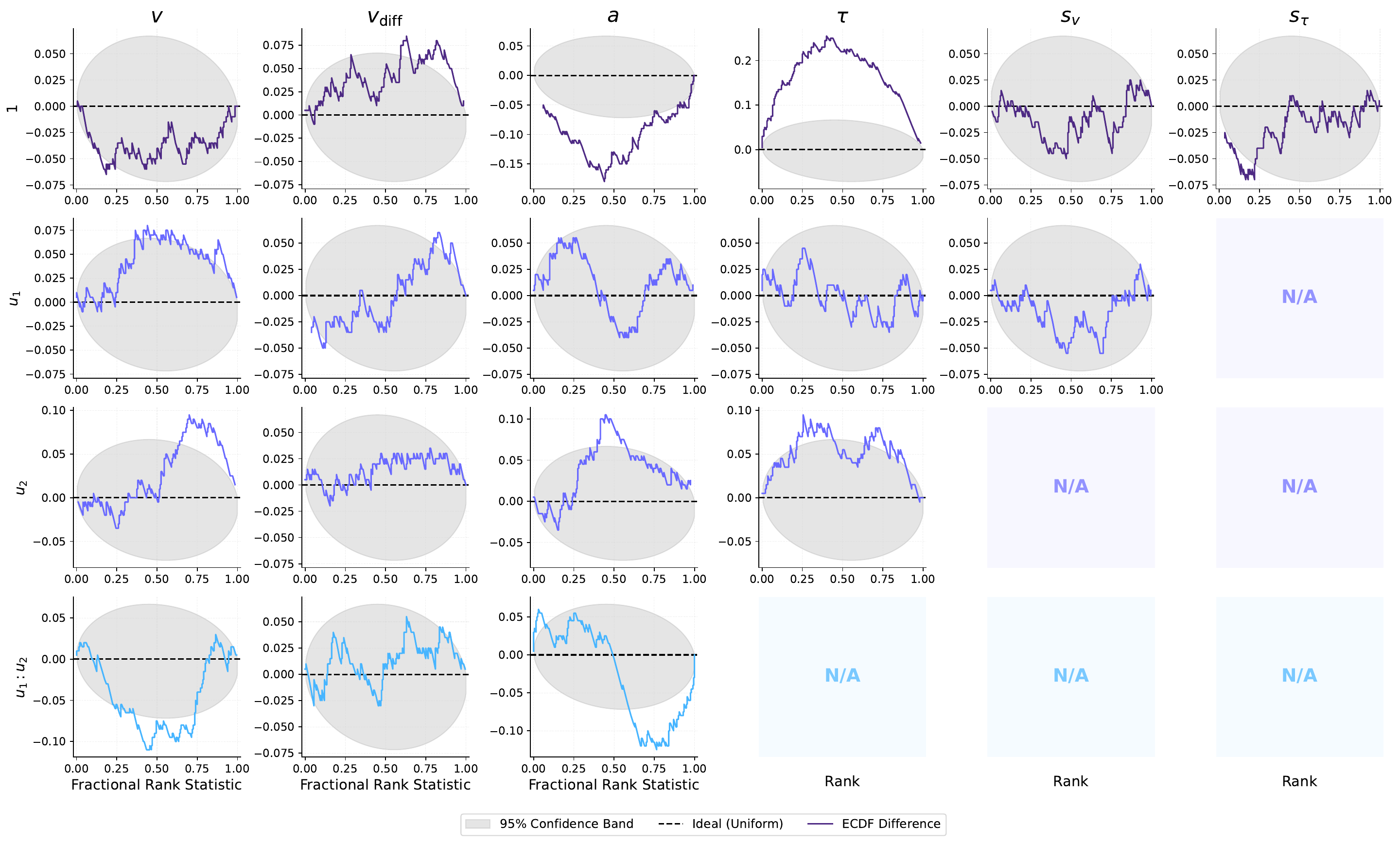}
    \includegraphics[width=0.97\linewidth]{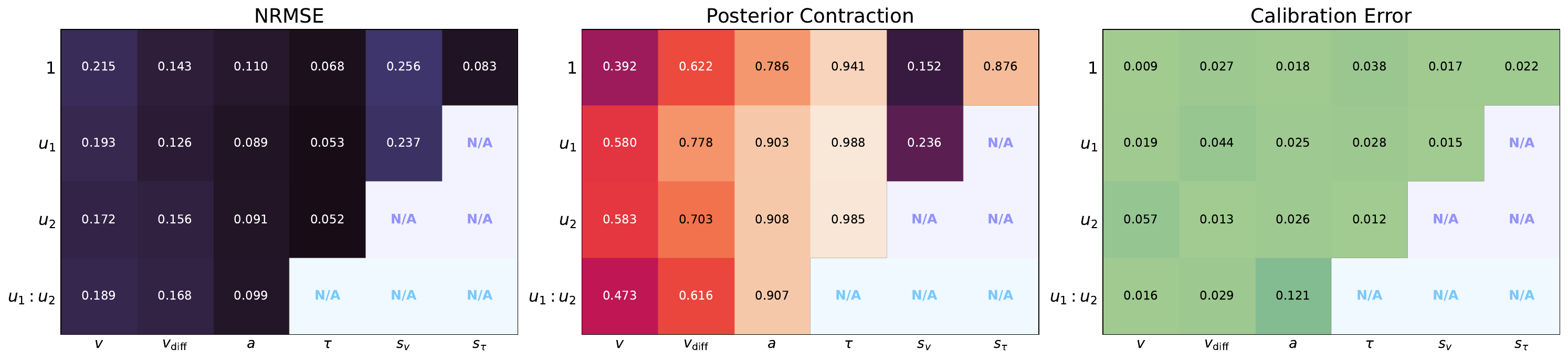}
    \caption{Parameter recovery (\emph{top}), calibration ECDF (\emph{middle}), and parameter-wise metrics (NRMSE, calibration error, and posterior contraction) for RDM model class (Case \textbf{interaction}).}
    \label{fig:rdm-mc-interaction}
\end{figure}


\begin{figure}
    \centering
    \includegraphics[width=0.97\linewidth]{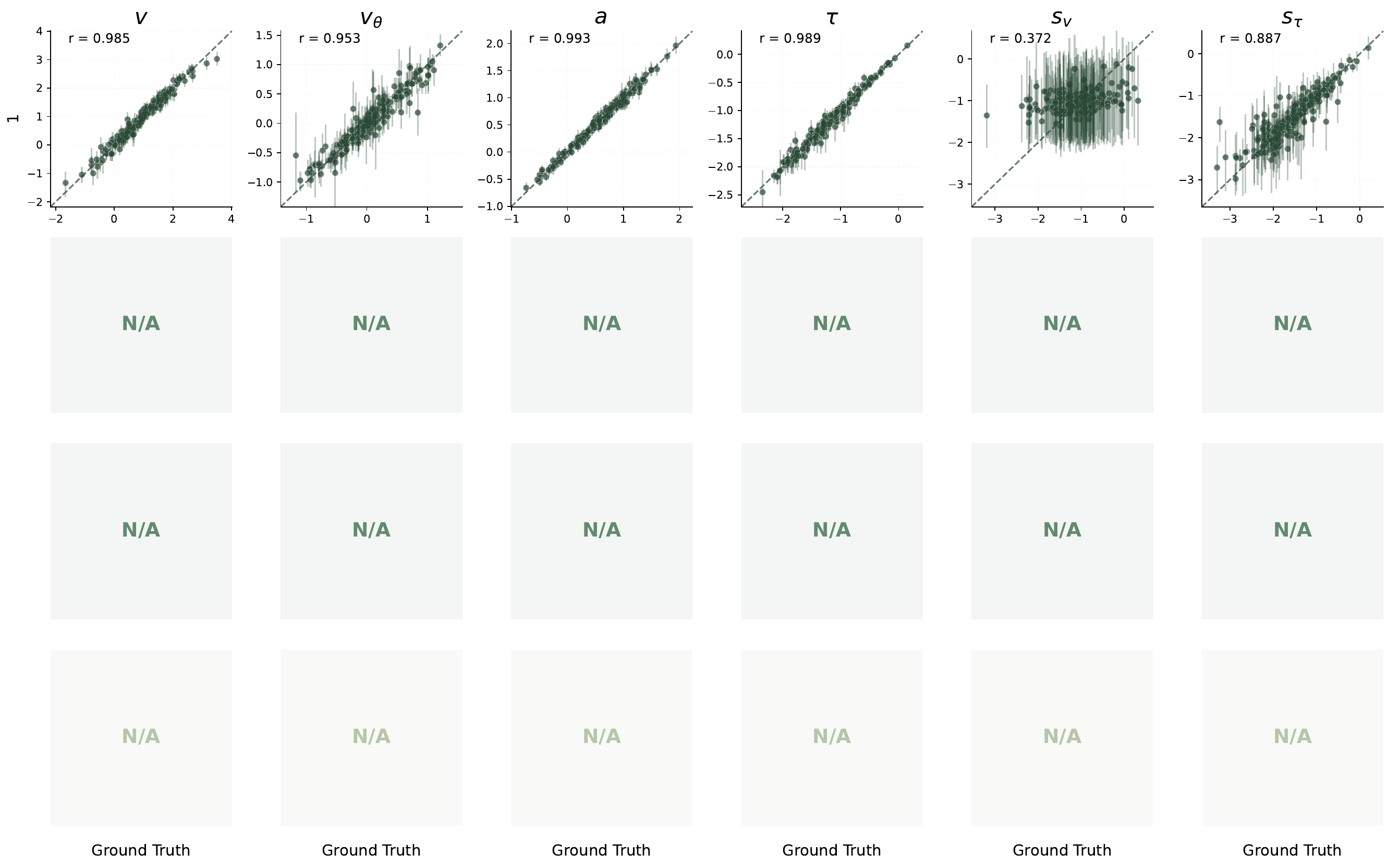}
    \includegraphics[width=0.97\linewidth]{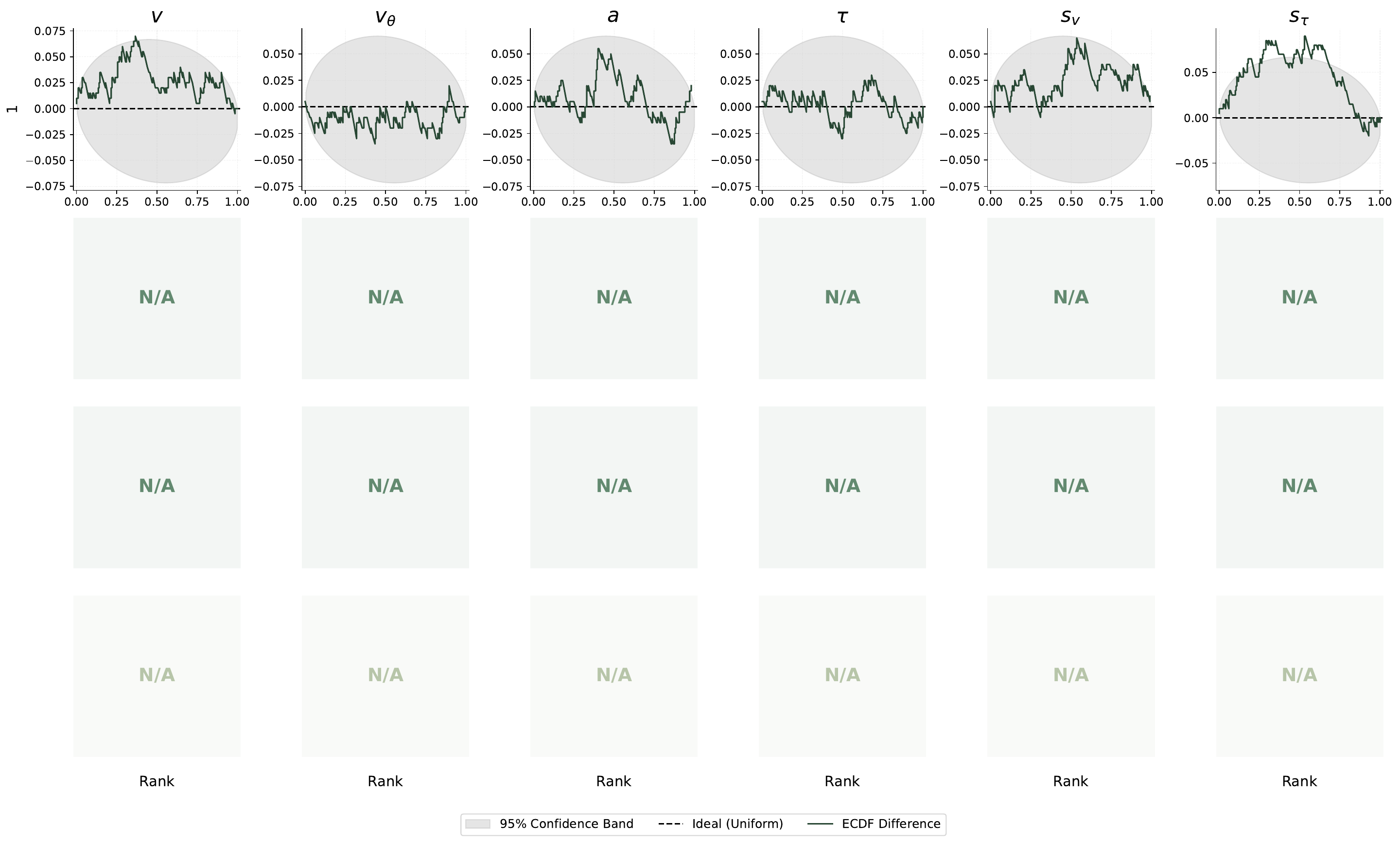}
    \includegraphics[width=0.97\linewidth]{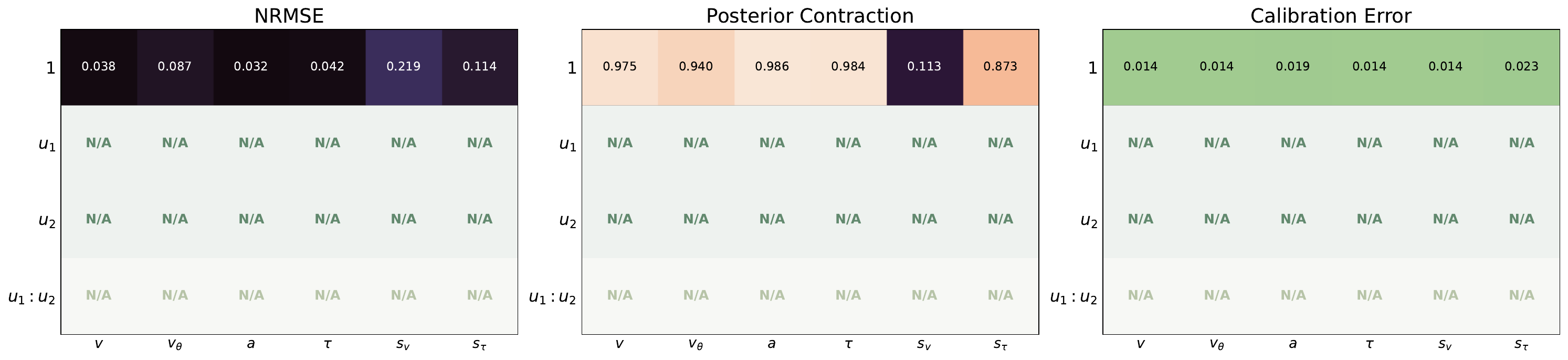}
    \caption{Parameter recovery (\emph{top}), calibration ECDF (\emph{middle}), and validation metrics (NRMSE, calibration error, and posterior contraction) for CDM baseline (Case \textbf{intercept\_only}).}
    \label{fig:cdm-bf-intercept-only}
\end{figure}

\begin{figure}
    \centering
    \includegraphics[width=0.97\linewidth]{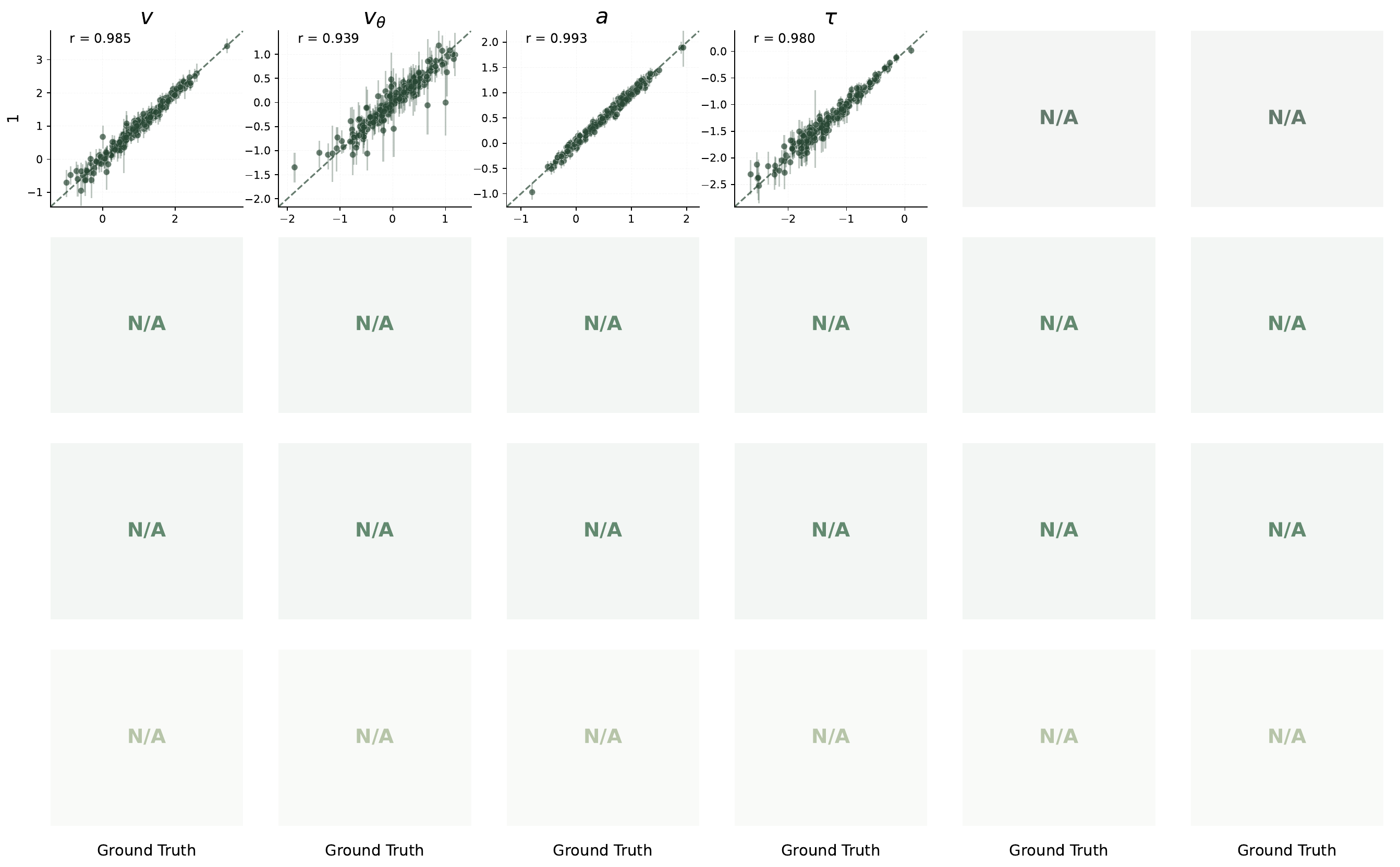}
    \includegraphics[width=0.97\linewidth]{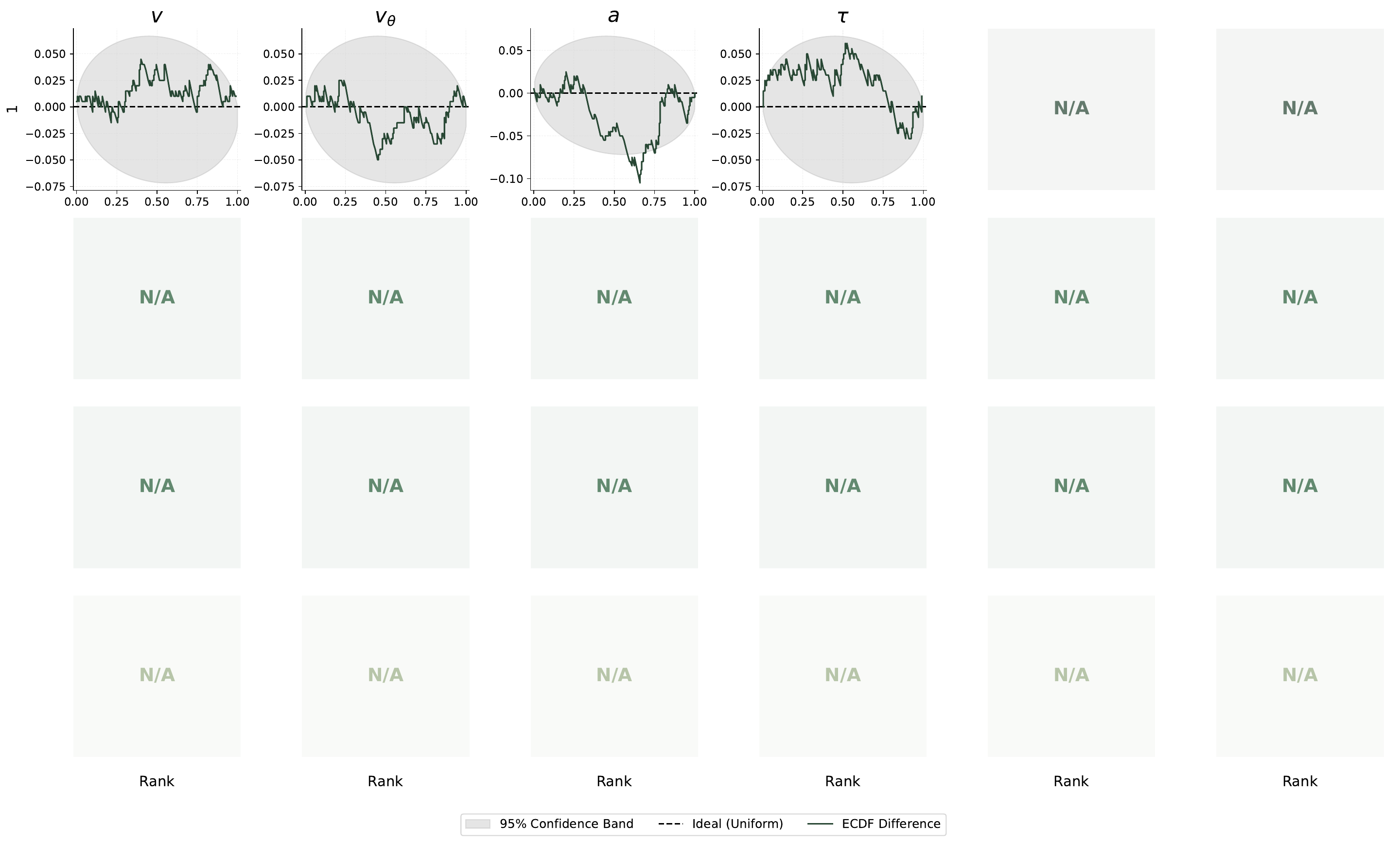}
    \includegraphics[width=0.97\linewidth]{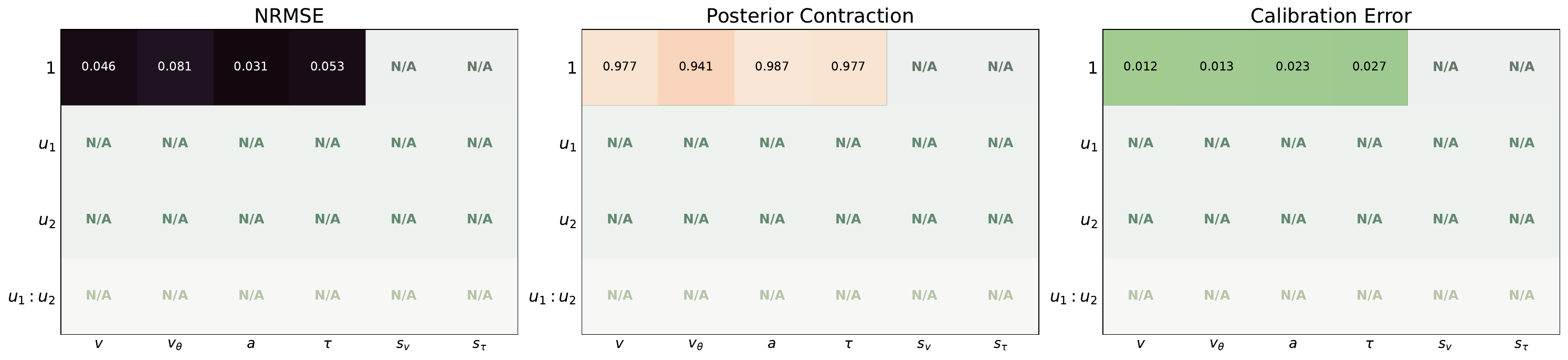}
    \caption{Parameter recovery (\emph{top}), calibration ECDF (\emph{middle}), and validation metrics (NRMSE, calibration error, and posterior contraction) for CDM baseline (Case \textbf{fixed}).}
    \label{fig:cdm-bf-fixed}
\end{figure}

\begin{figure}
    \centering
    \includegraphics[width=0.97\linewidth]{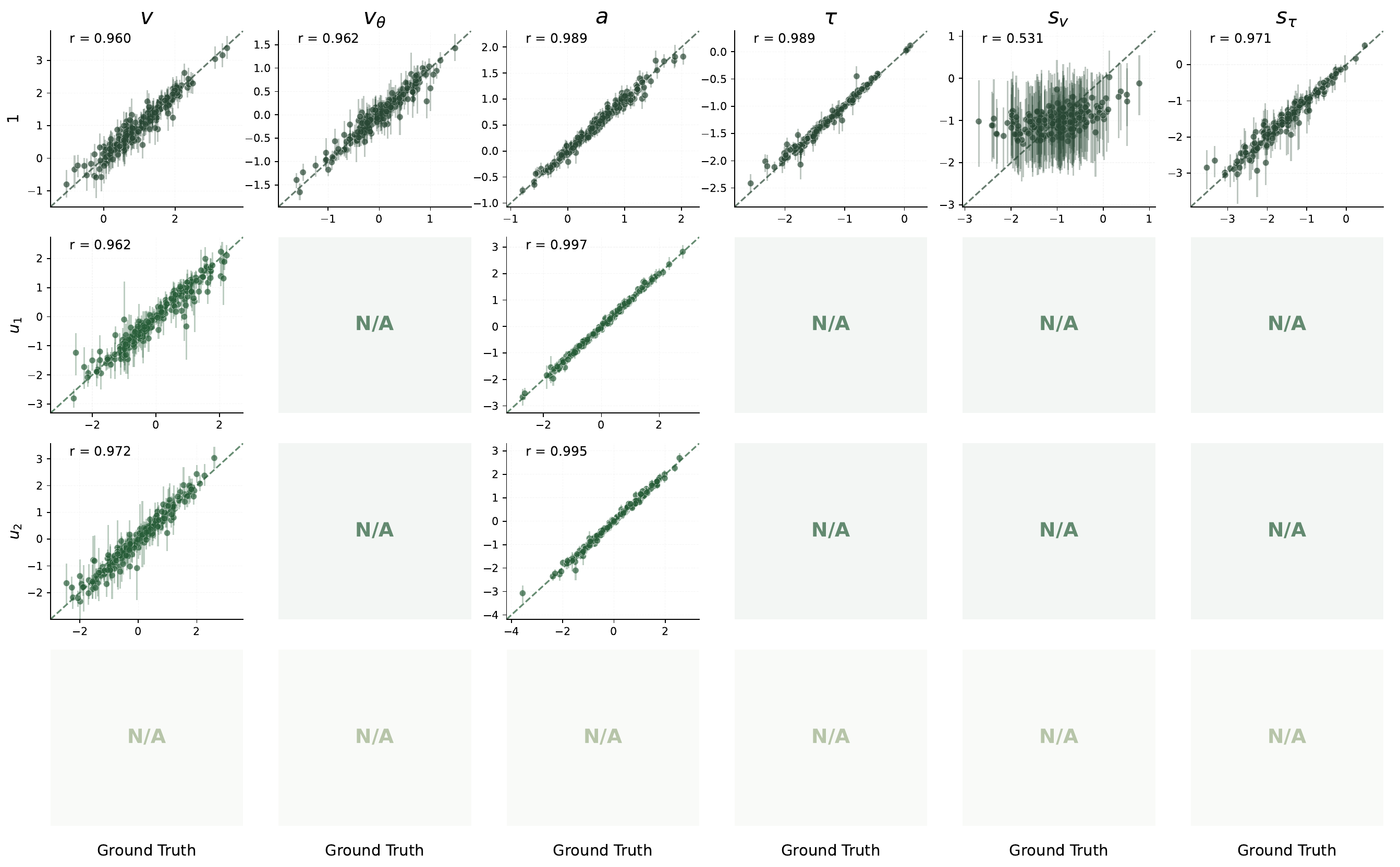}
    \includegraphics[width=0.97\linewidth]{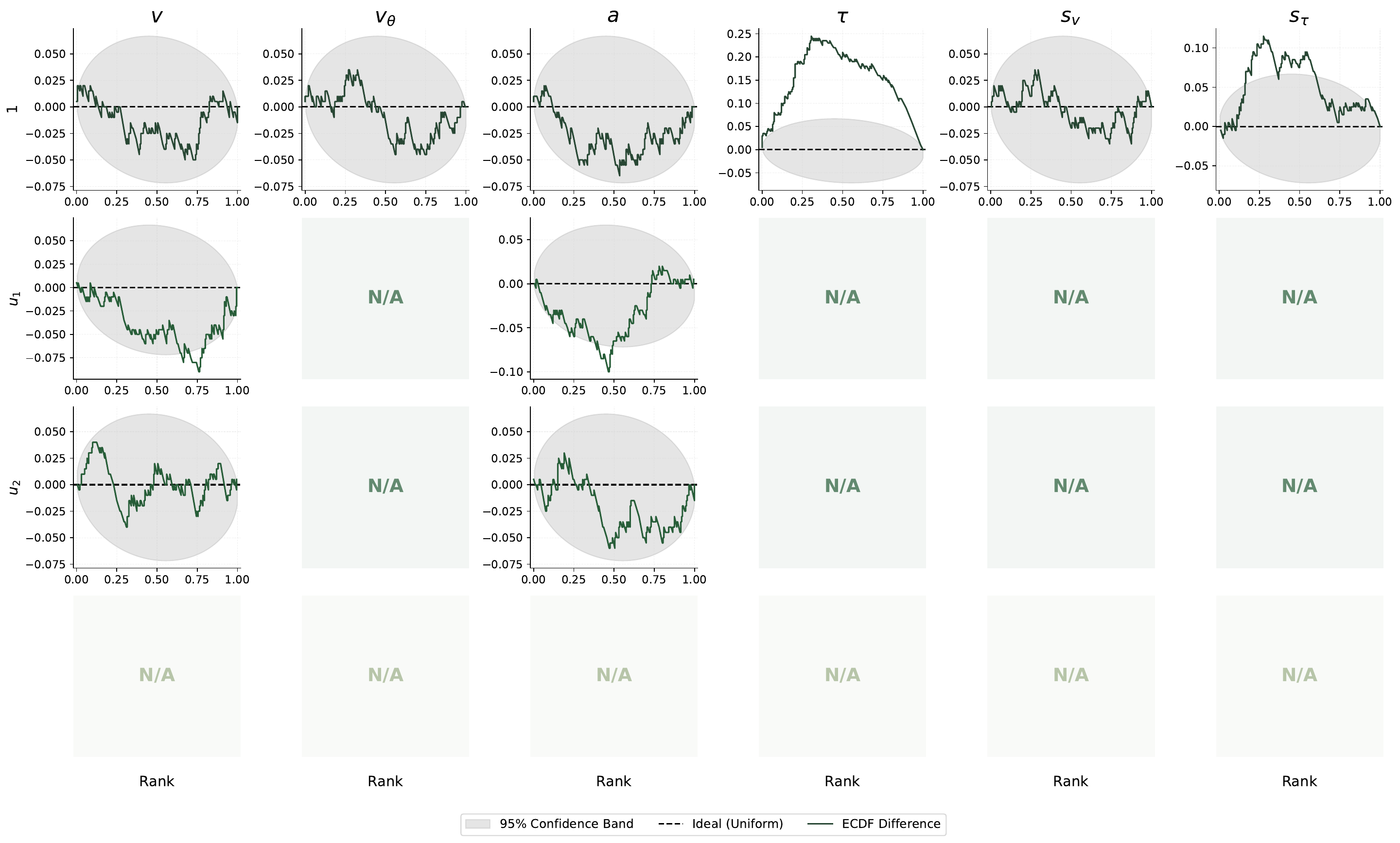}
    \includegraphics[width=0.97\linewidth]{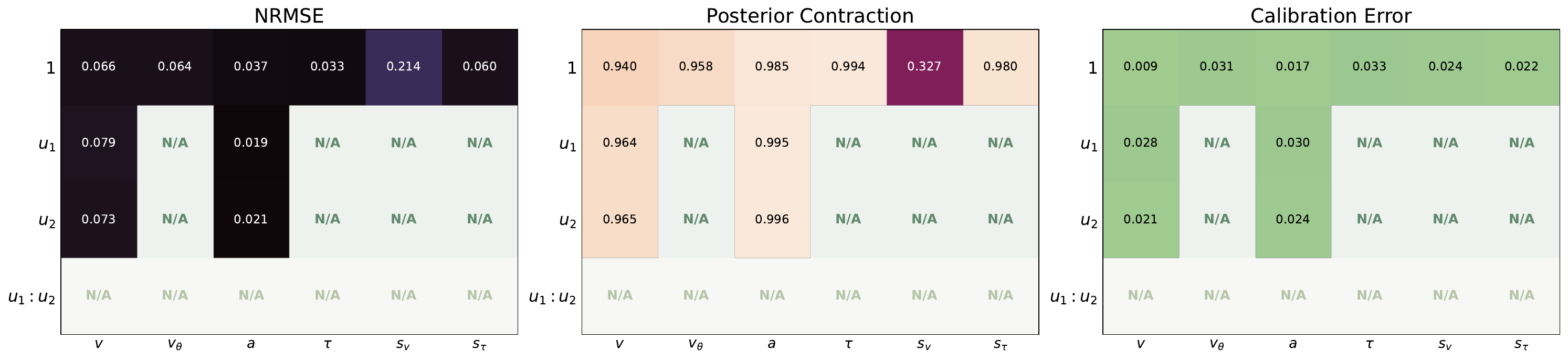}
    \caption{Parameter recovery (\emph{top}), calibration ECDF (\emph{middle}), and validation metrics (NRMSE, calibration error, and posterior contraction) for CDM baseline (Case \textbf{regressed}).}
    \label{fig:cdm-bf-regressed}
\end{figure}

\begin{figure}
    \centering
    \includegraphics[width=0.97\linewidth]{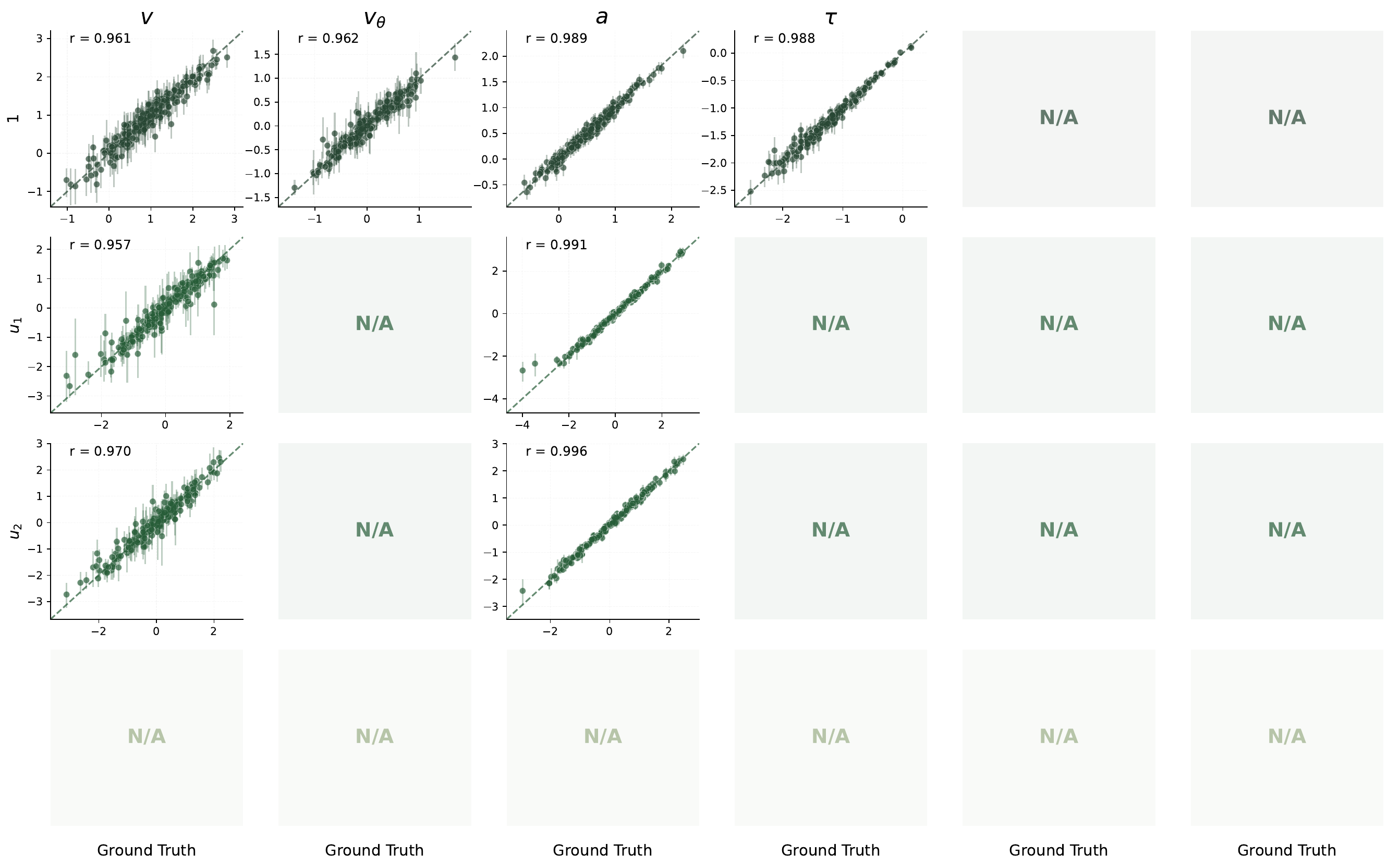}
    \includegraphics[width=0.97\linewidth]{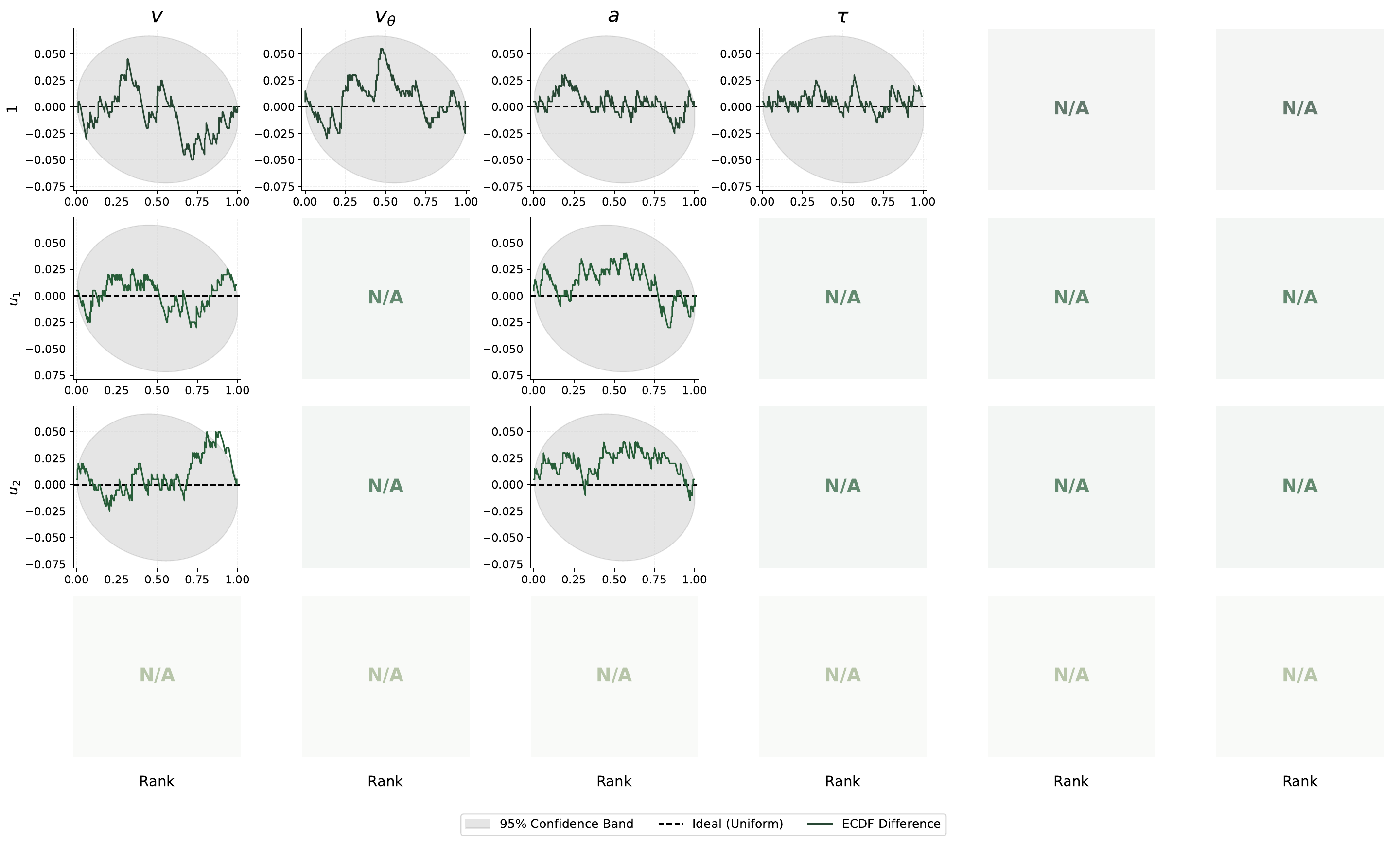}
    \includegraphics[width=0.97\linewidth]{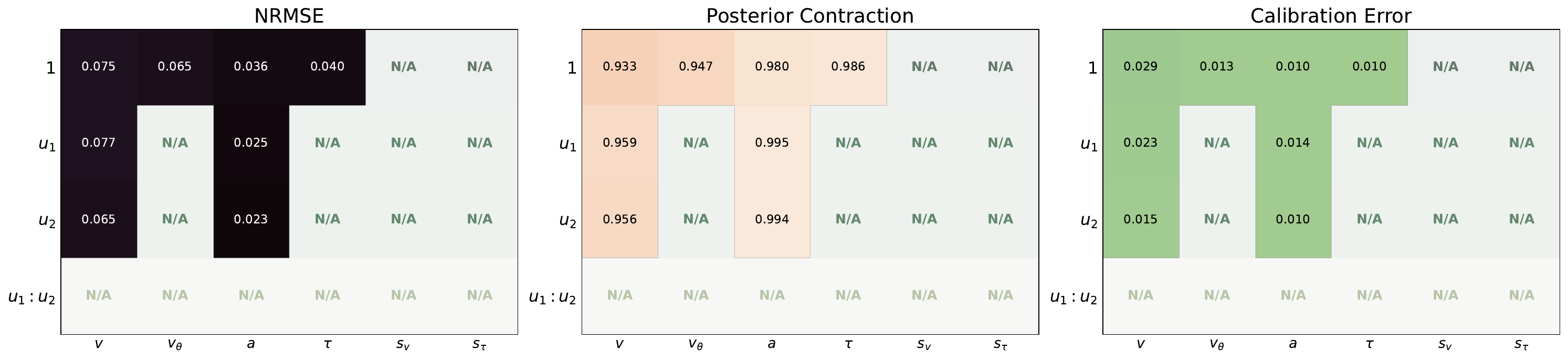}
    \caption{Parameter recovery (\emph{top}), calibration ECDF (\emph{middle}), and validation metrics (NRMSE, calibration error, and posterior contraction) for CDM baseline (Case \textbf{fixed\_regressed}).}
    \label{fig:cdm-bf-fixed-regressed}
\end{figure}

\begin{figure}
    \centering
    \includegraphics[width=0.97\linewidth]{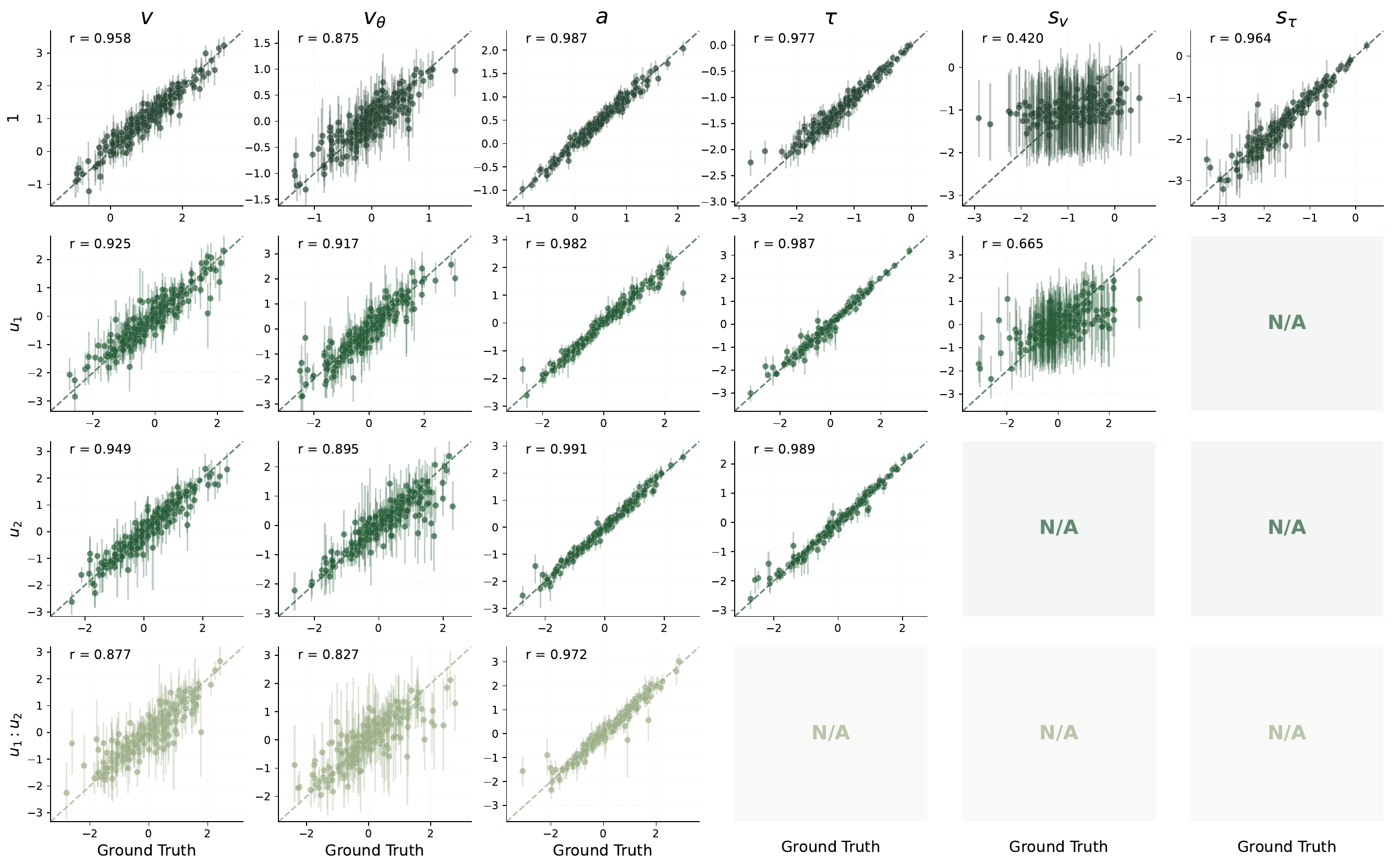}
    \includegraphics[width=0.97\linewidth]{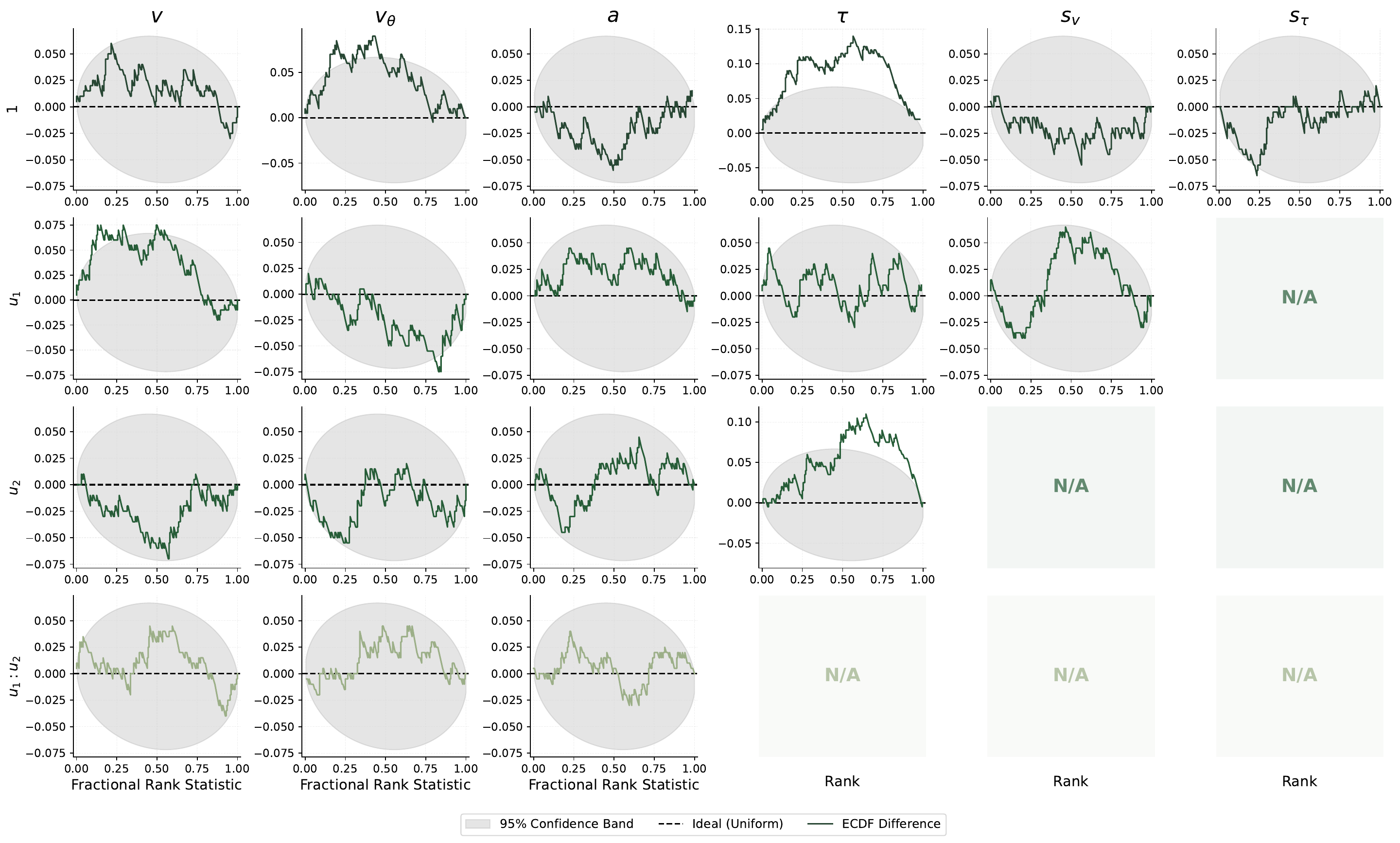}
    \includegraphics[width=0.97\linewidth]{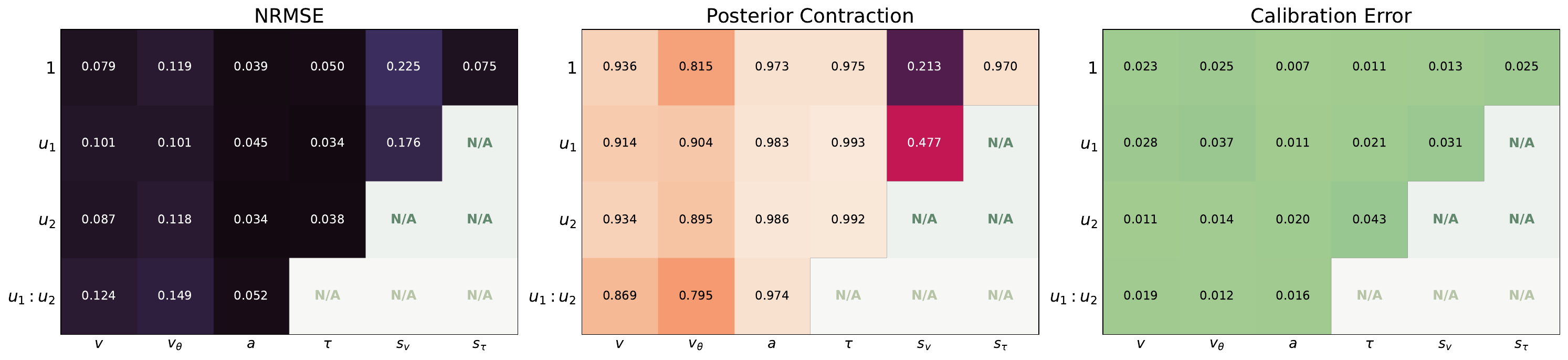}
    \caption{Parameter recovery (\emph{top}), calibration ECDF (\emph{middle}), and validation metrics (NRMSE, calibration error, and posterior contraction) for CDM baseline (Case \textbf{interaction}).}
    \label{fig:cdm-bf-interaction}
\end{figure}


\begin{figure}
    \centering
    \includegraphics[width=0.97\linewidth]{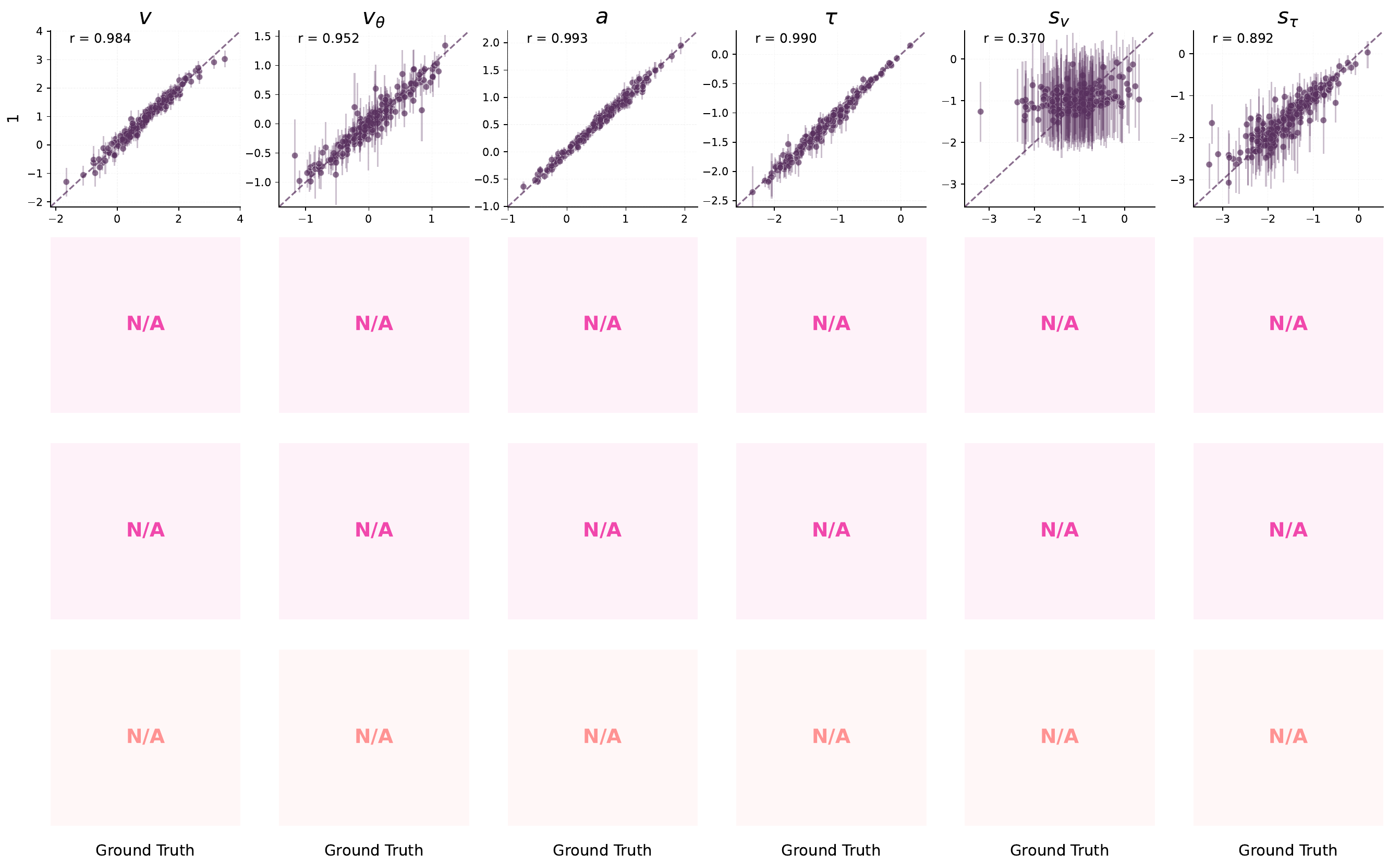}
    \includegraphics[width=0.97\linewidth]{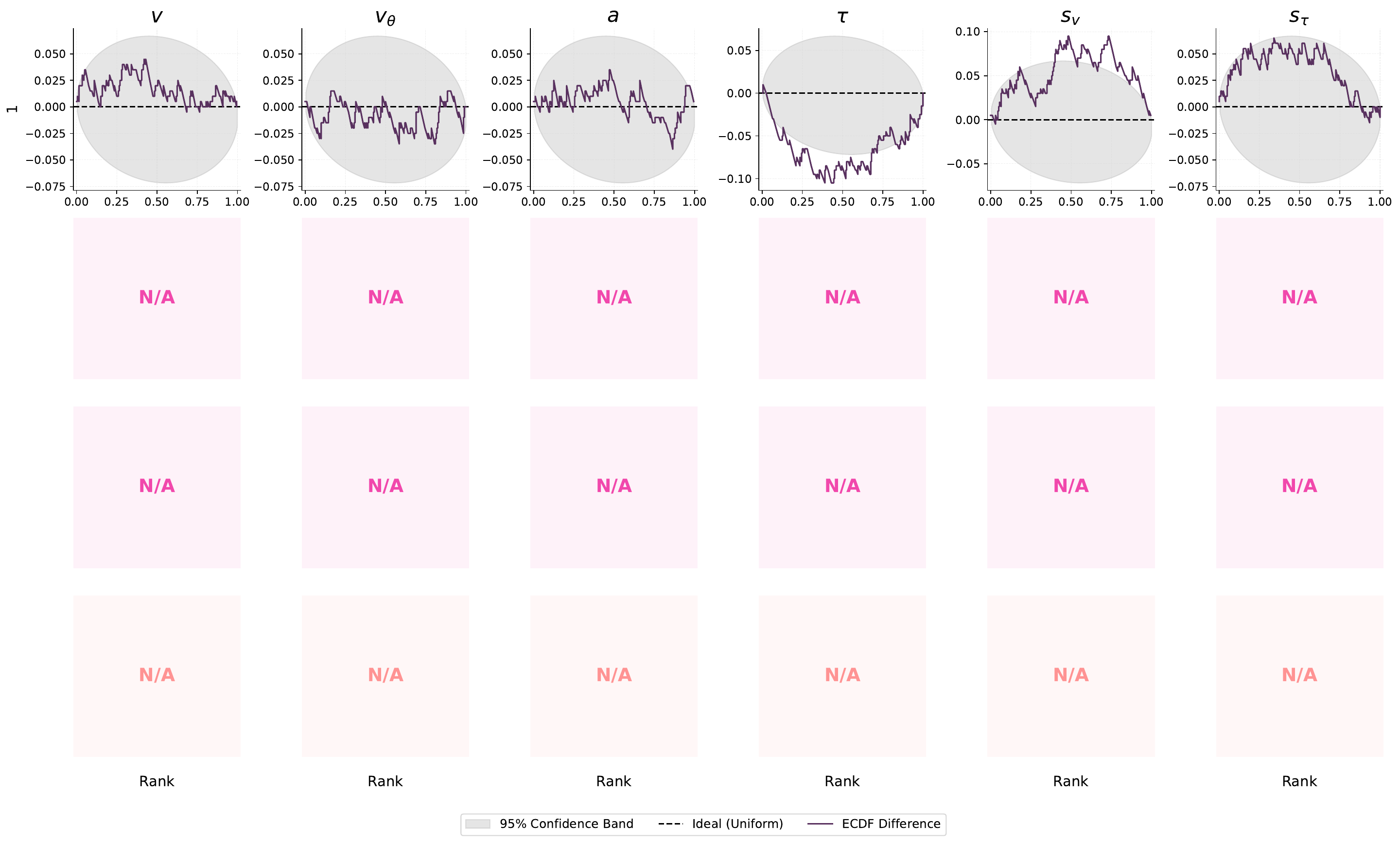}
    \includegraphics[width=0.97\linewidth]{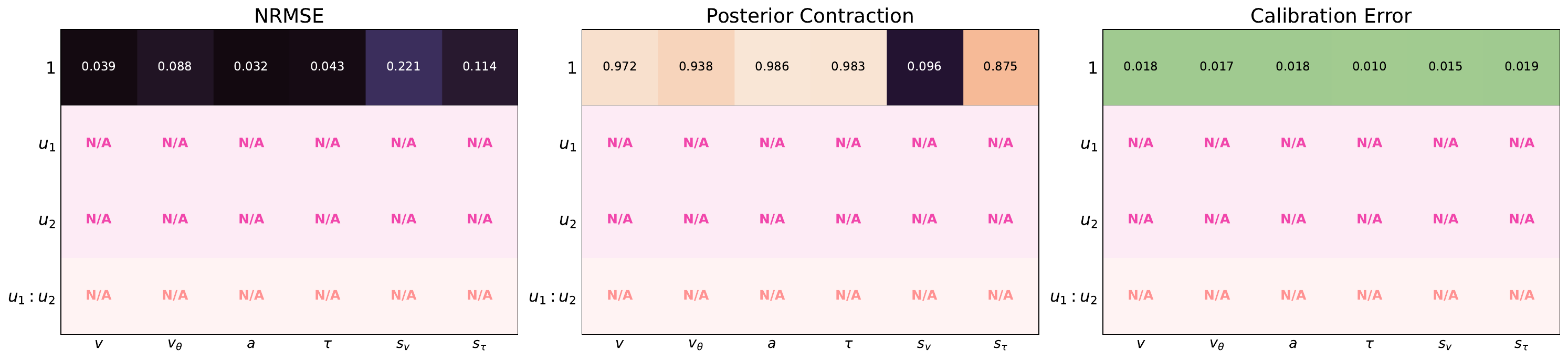}
    \caption{Parameter recovery (\emph{top}), calibration ECDF (\emph{middle}), and validation metrics (NRMSE, calibration error, and posterior contraction) for CDM model family (Case \textbf{intercept\_only}).}
    \label{fig:cdm-fm-intercept-only}
\end{figure}

\begin{figure}
    \centering
    \includegraphics[width=0.97\linewidth]{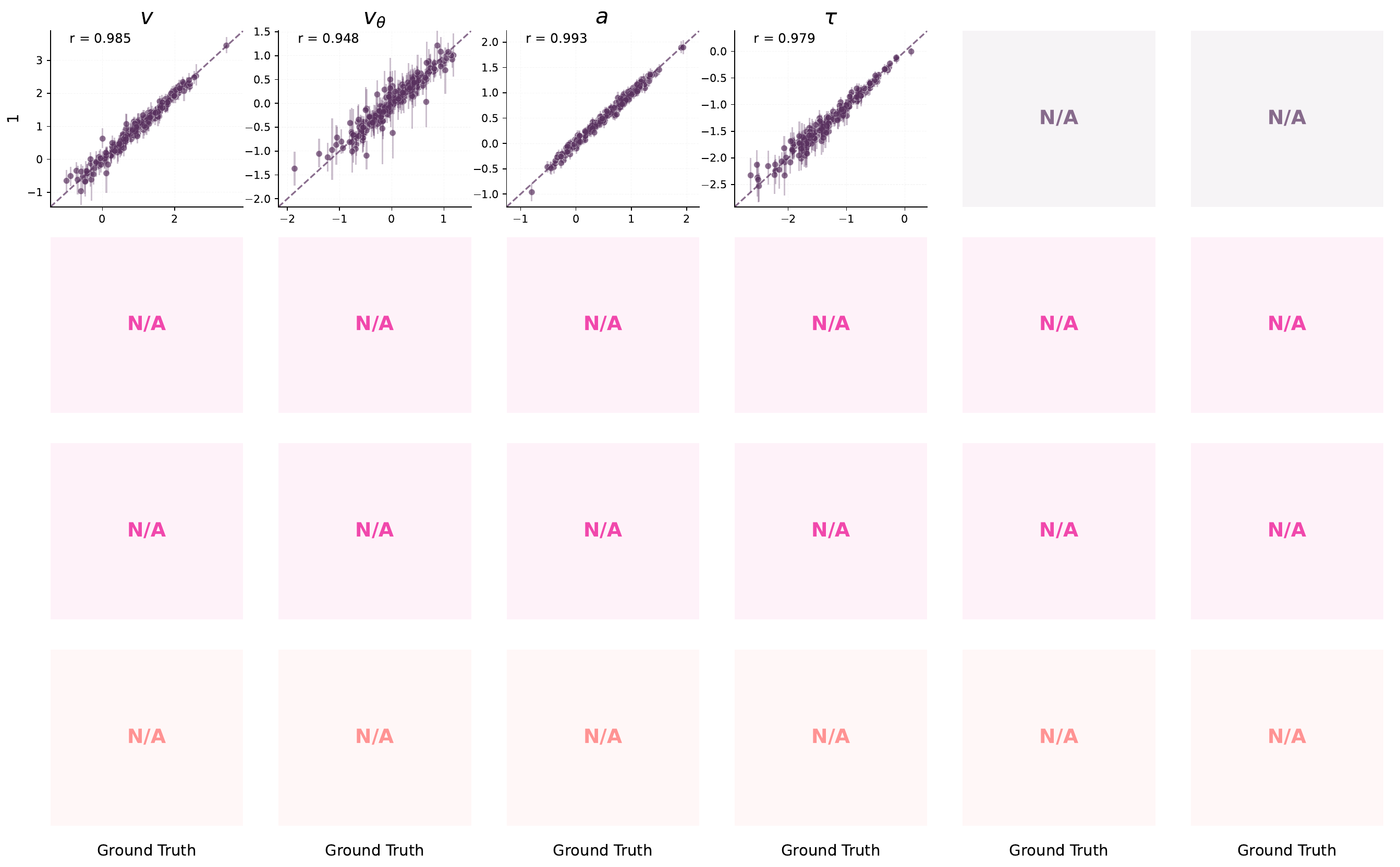}
    \includegraphics[width=0.97\linewidth]{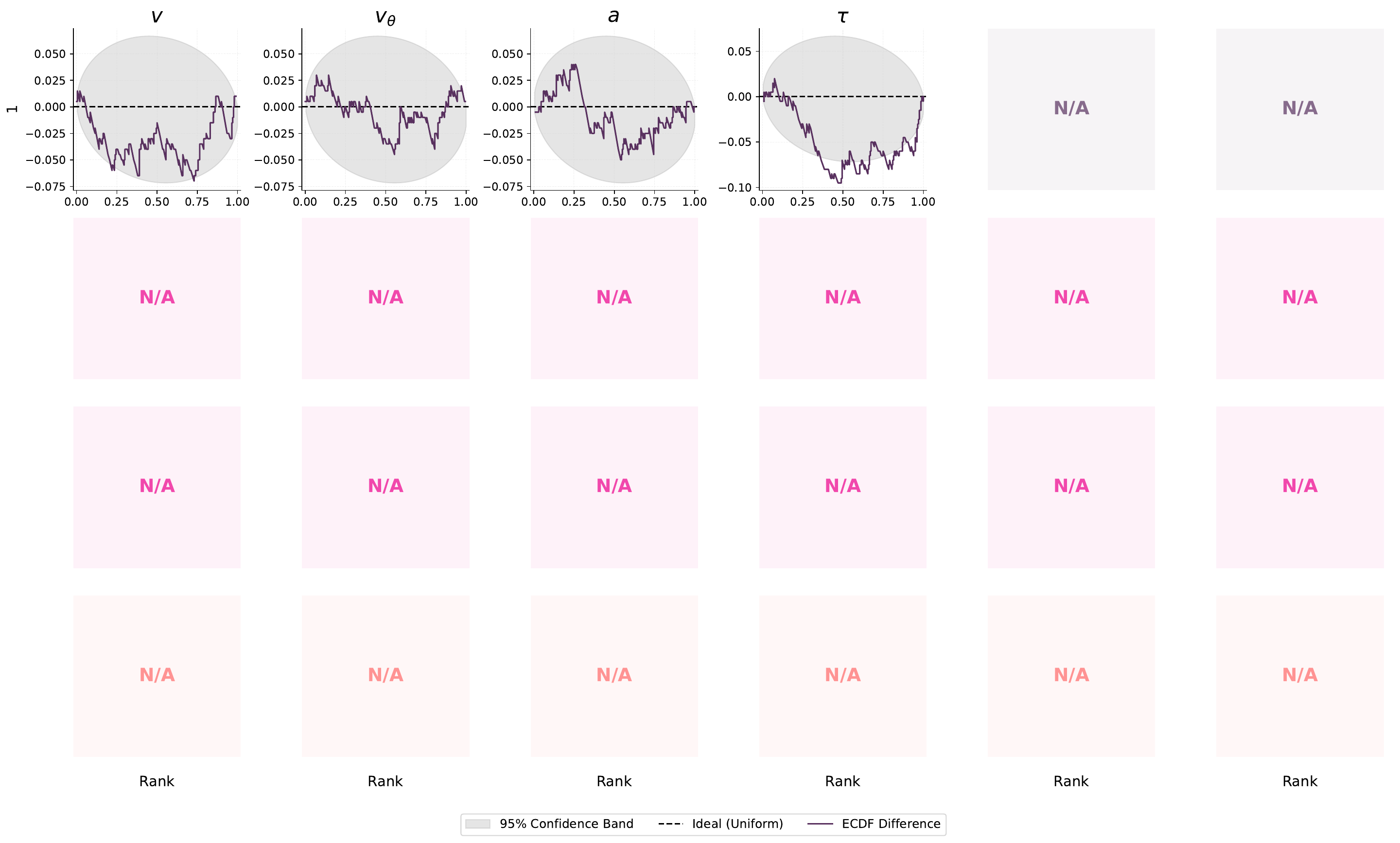}
    \includegraphics[width=0.97\linewidth]{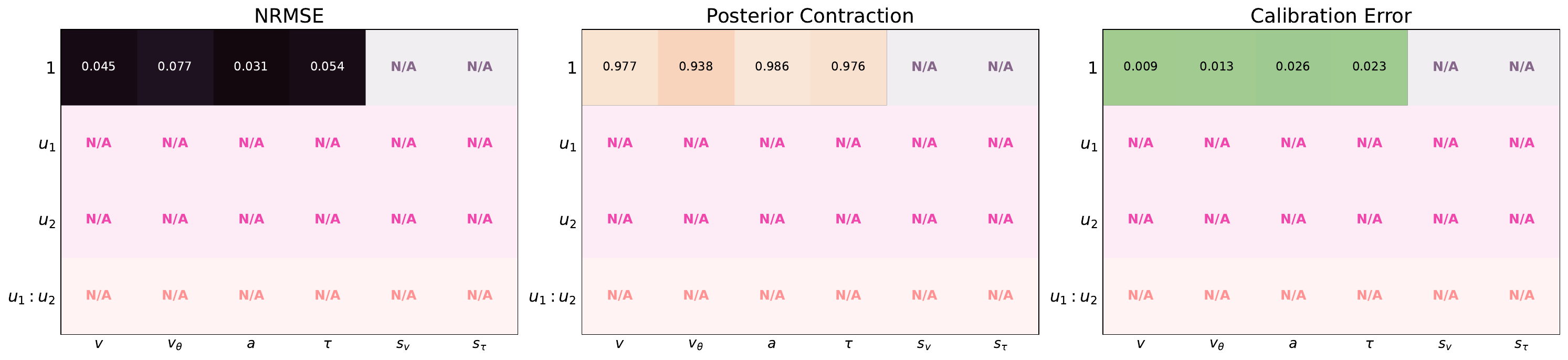}
    \caption{Parameter recovery (\emph{top}), calibration ECDF (\emph{middle}), and validation metrics (NRMSE, calibration error, and posterior contraction) for CDM model family (Case \textbf{fixed}).}
    \label{fig:cdm-fm-fixed}
\end{figure}

\begin{figure}
    \centering
    \includegraphics[width=0.97\linewidth]{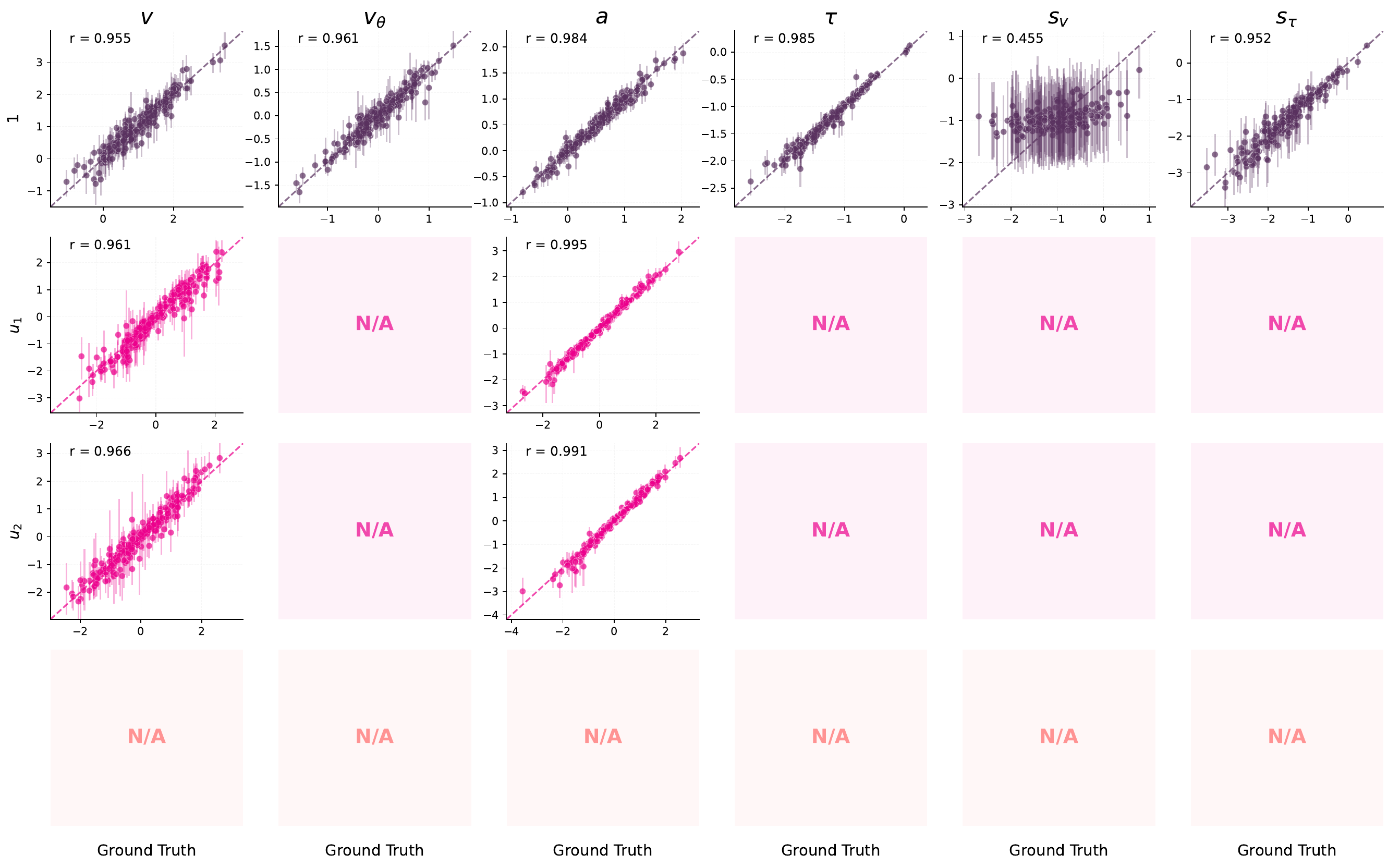}
    \includegraphics[width=0.97\linewidth]{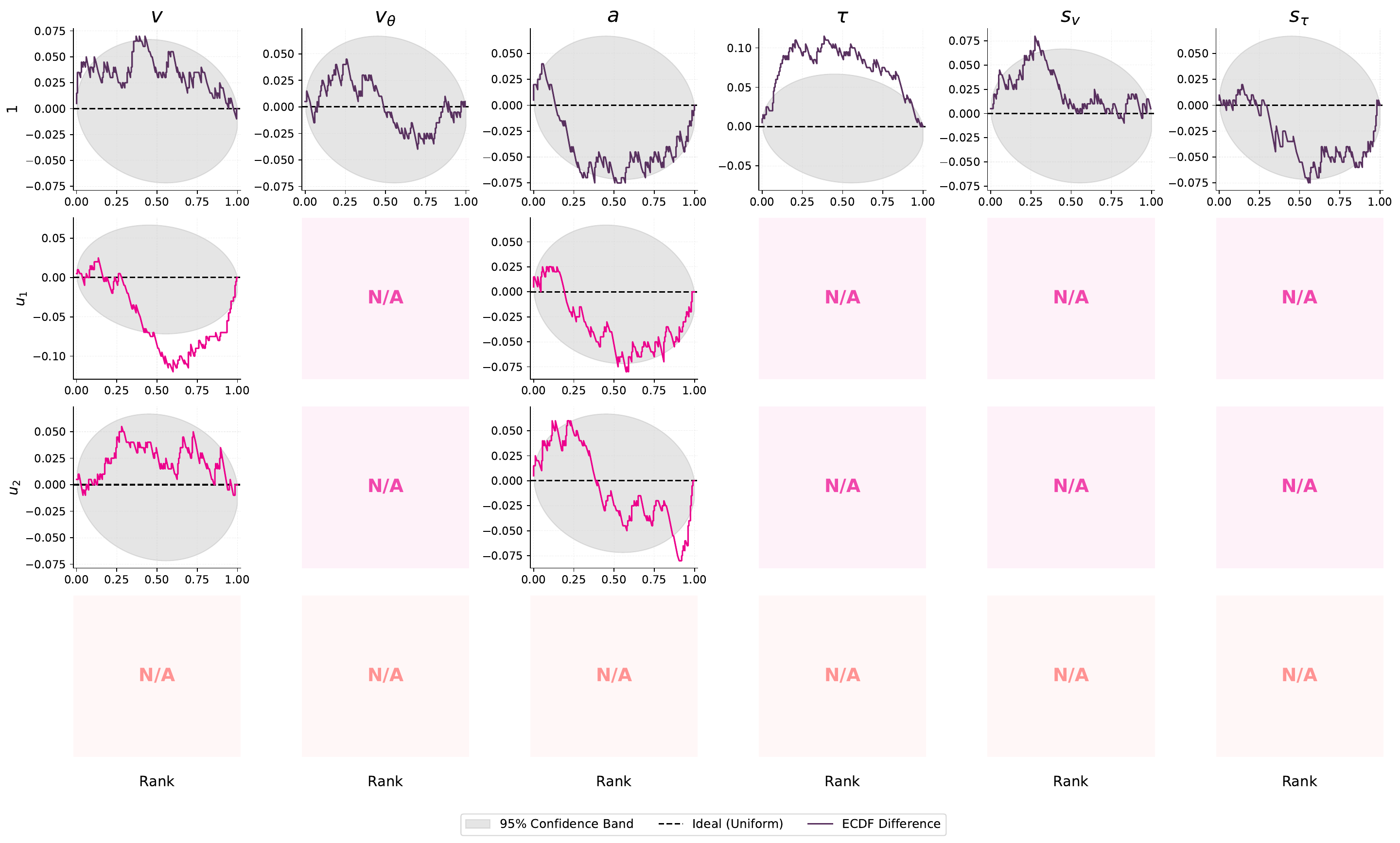}
    \includegraphics[width=0.97\linewidth]{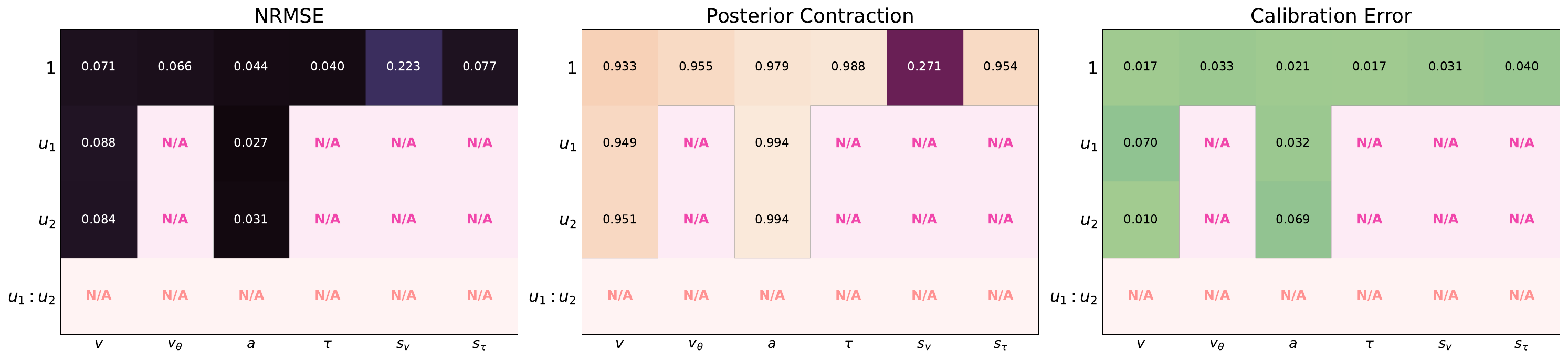}
    \caption{Parameter recovery (\emph{top}), calibration ECDF (\emph{middle}), and validation metrics (NRMSE, calibration error, and posterior contraction) for CDM model family (Case \textbf{regressed}).}
    \label{fig:cdm-fm-regressed}
\end{figure}

\begin{figure}
    \centering
    \includegraphics[width=0.97\linewidth]{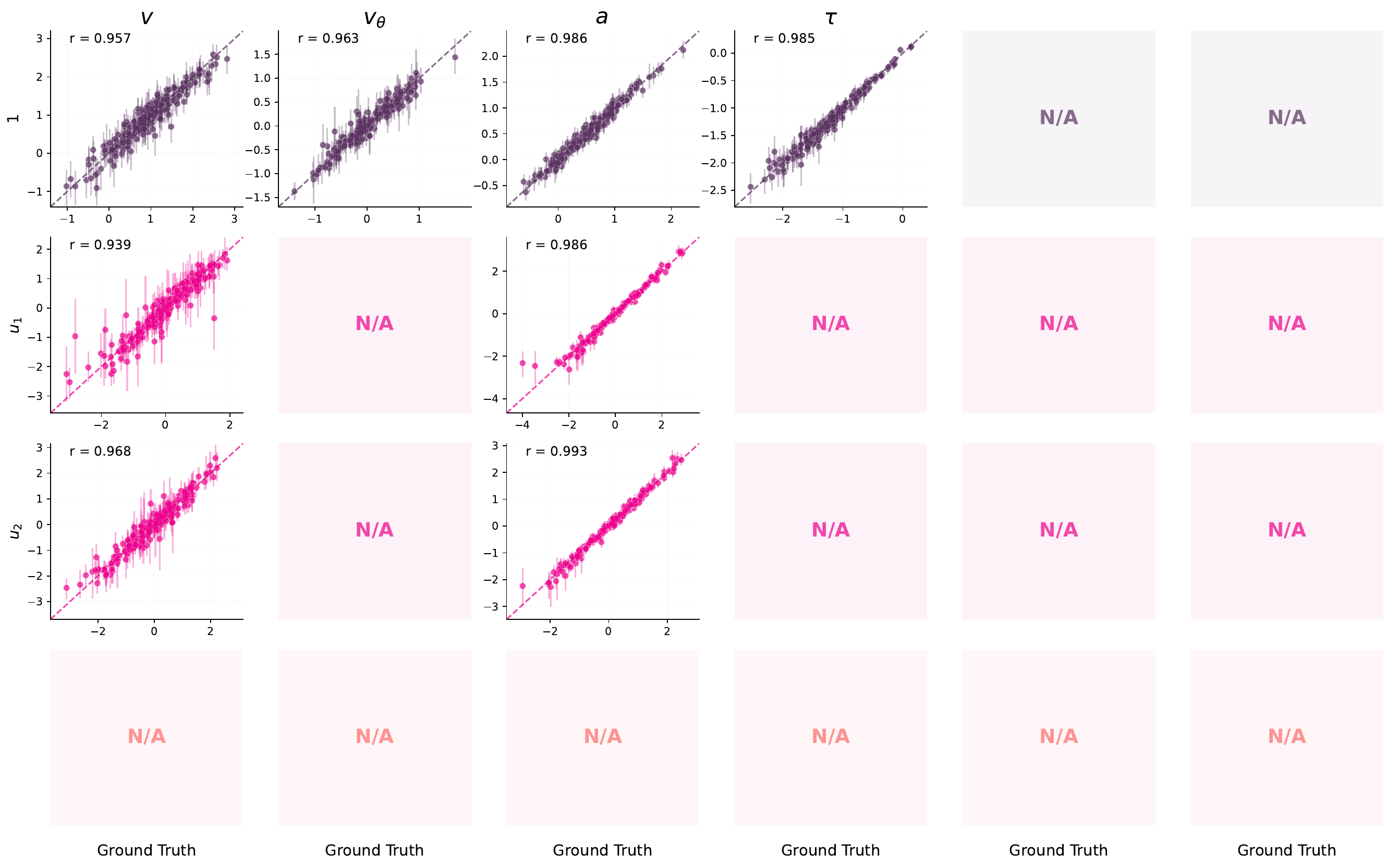}
    \includegraphics[width=0.97\linewidth]{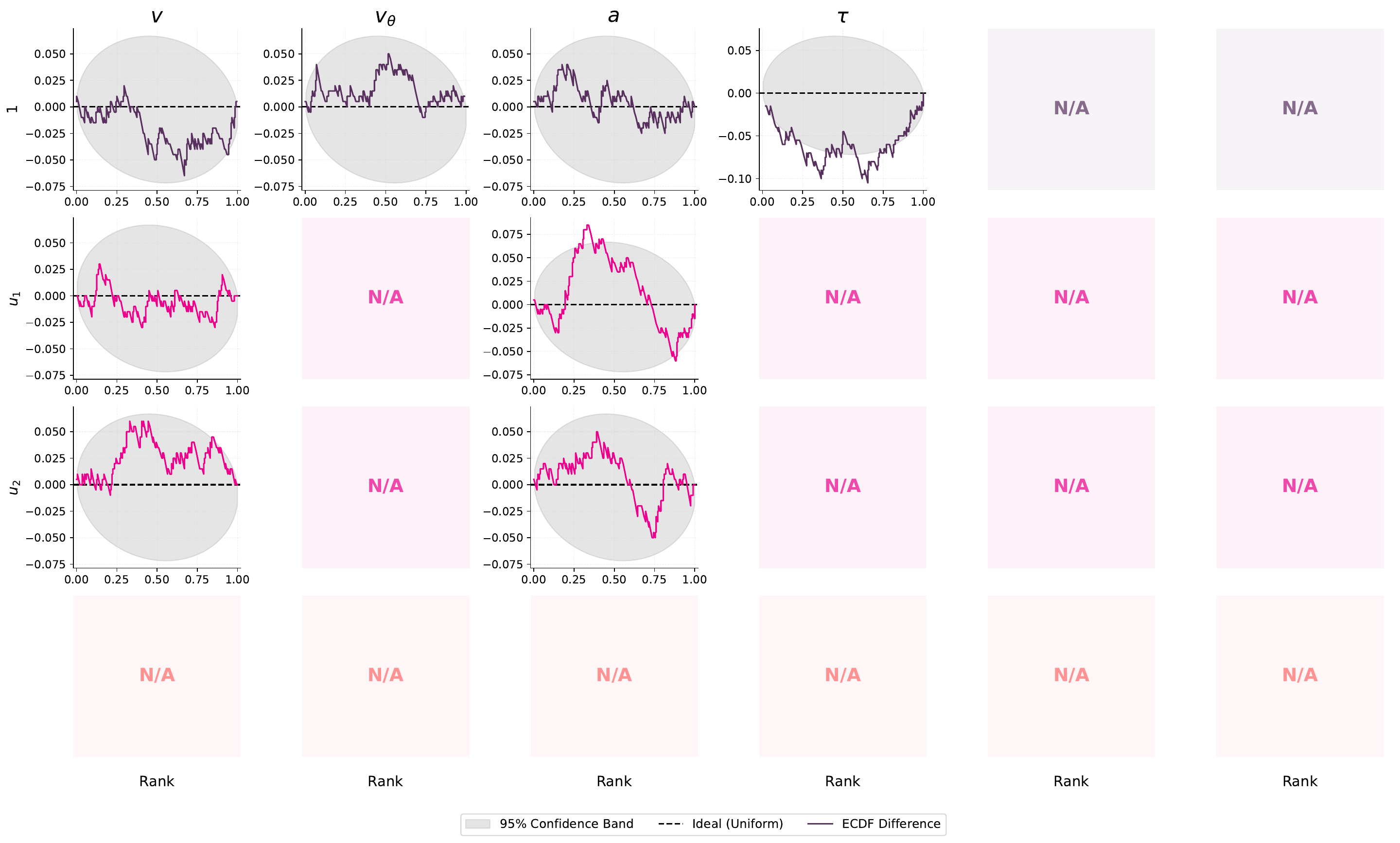}
    \includegraphics[width=0.97\linewidth]{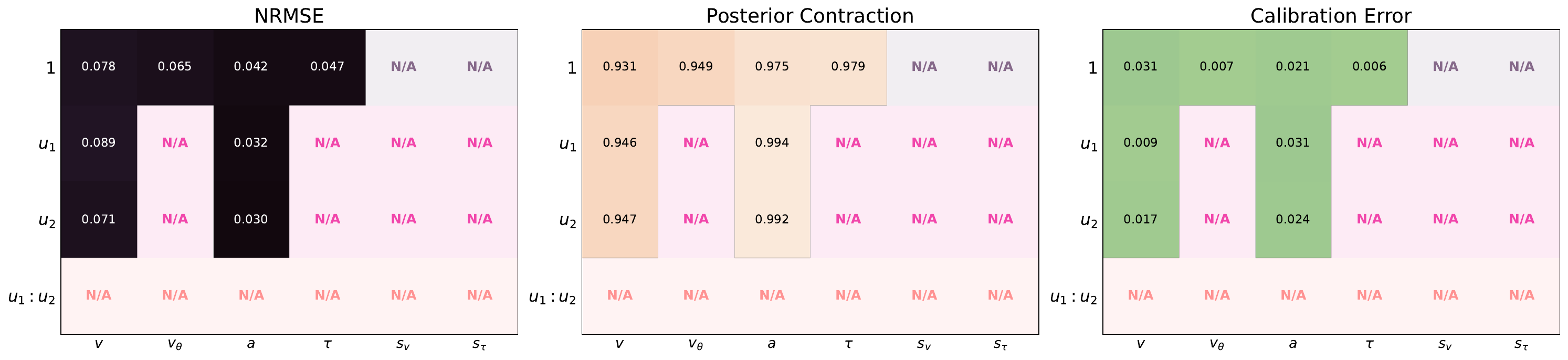}
    \caption{Parameter recovery (\emph{top}), calibration ECDF (\emph{middle}), and validation metrics (NRMSE, calibration error, and posterior contraction) for CDM model family (Case \textbf{fixed\_regressed}).}
    \label{fig:cdm-fm-fixed-regressed}
\end{figure}

\begin{figure}
    \centering
    \includegraphics[width=0.97\linewidth]{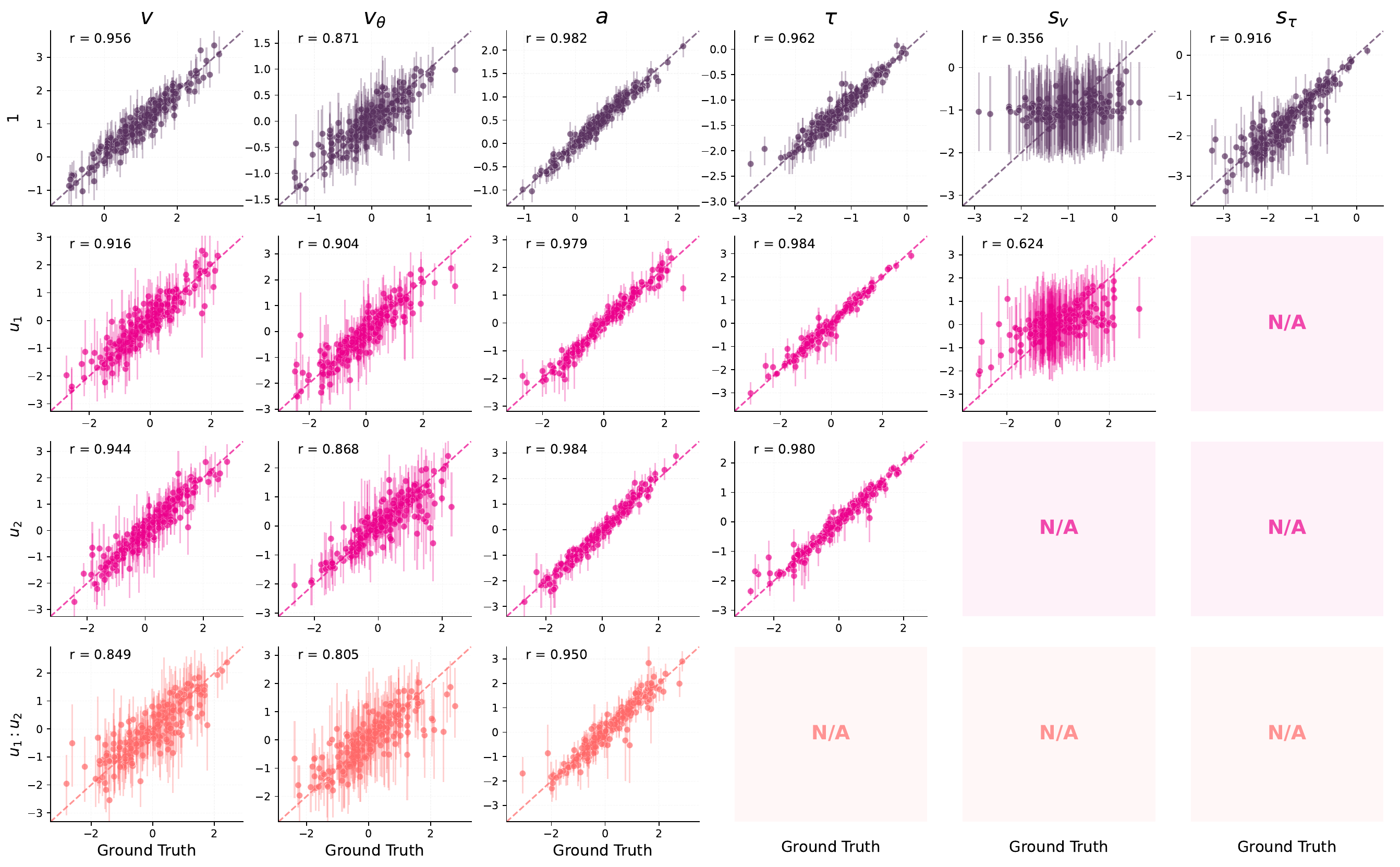}
    \includegraphics[width=0.97\linewidth]{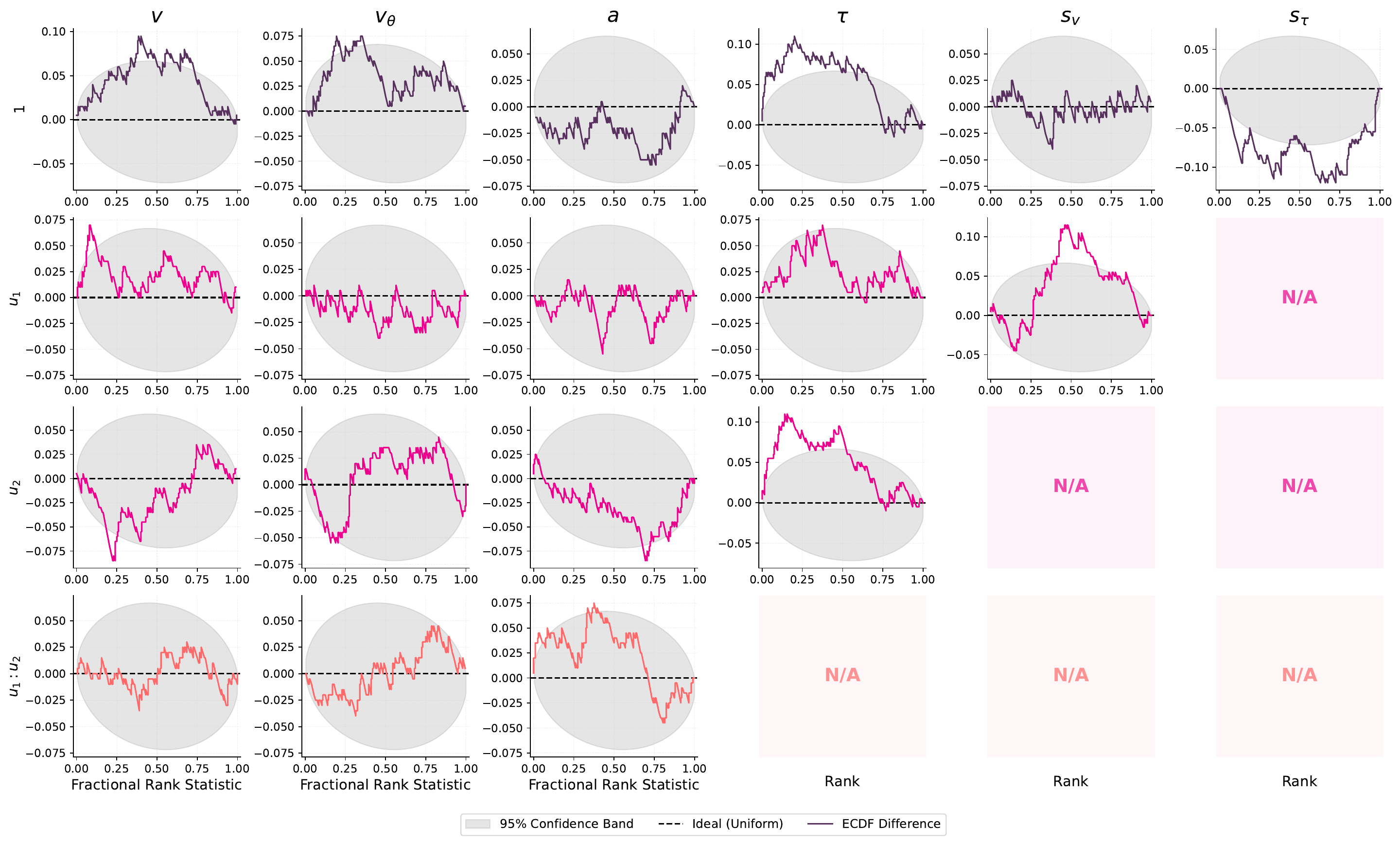}
    \includegraphics[width=0.97\linewidth]{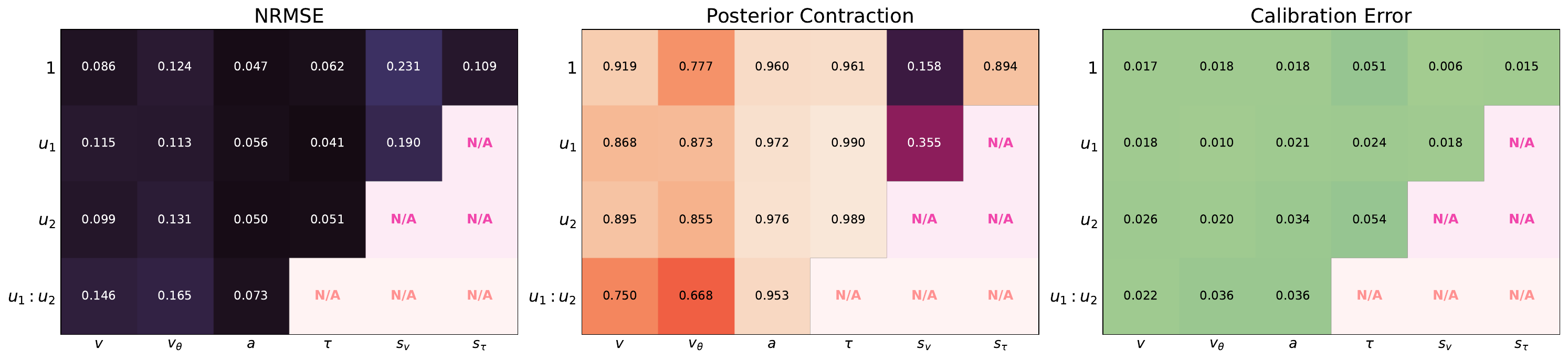}
    \caption{Parameter recovery (\emph{top}), calibration ECDF (\emph{middle}), and validation metrics (NRMSE, calibration error, and posterior contraction) for CDM model family (Case \textbf{interaction}).}
    \label{fig:cdm-fm-interaction}
\end{figure}


\begin{figure}
    \centering
    \includegraphics[width=0.97\linewidth]{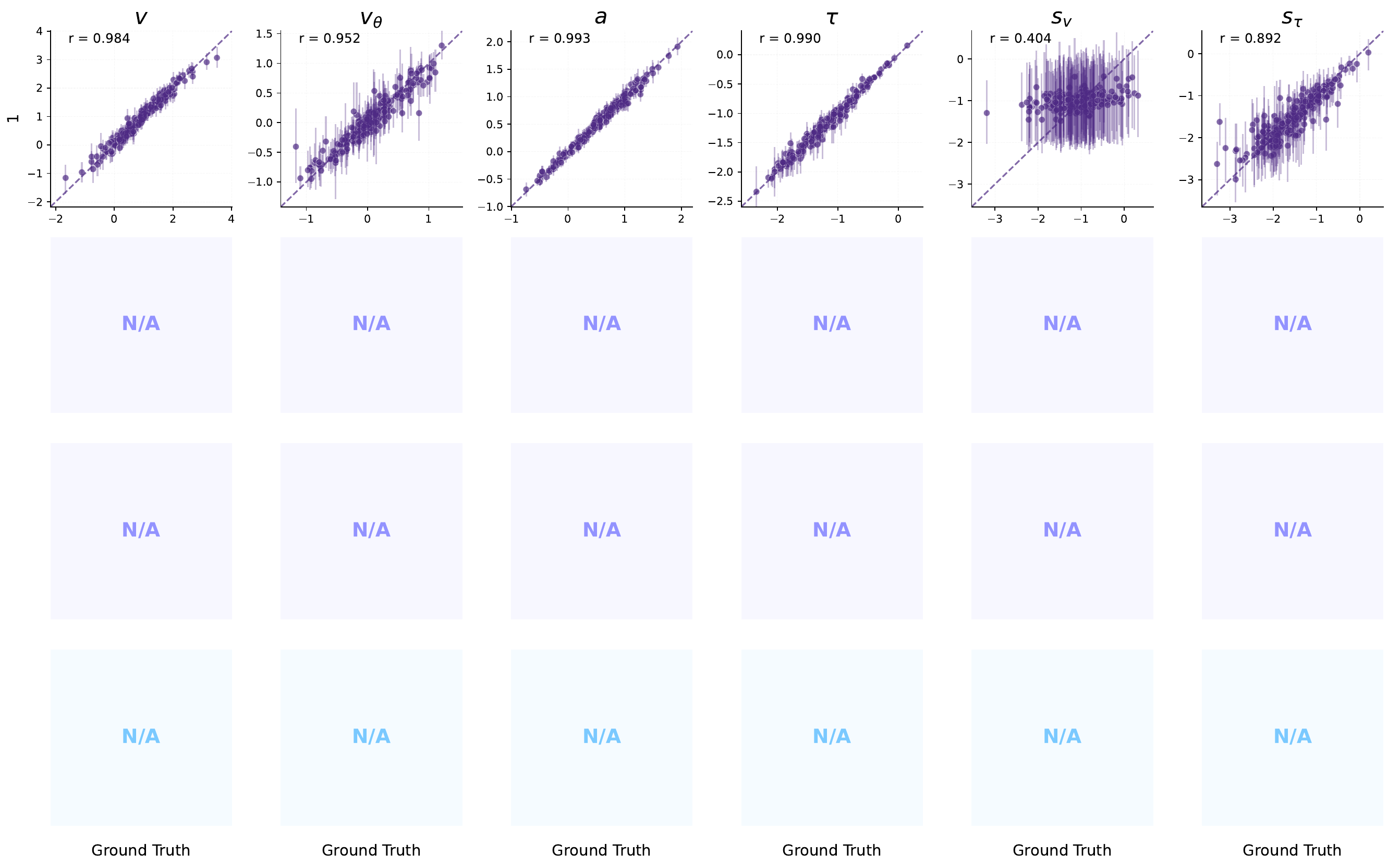}
    \includegraphics[width=0.97\linewidth]{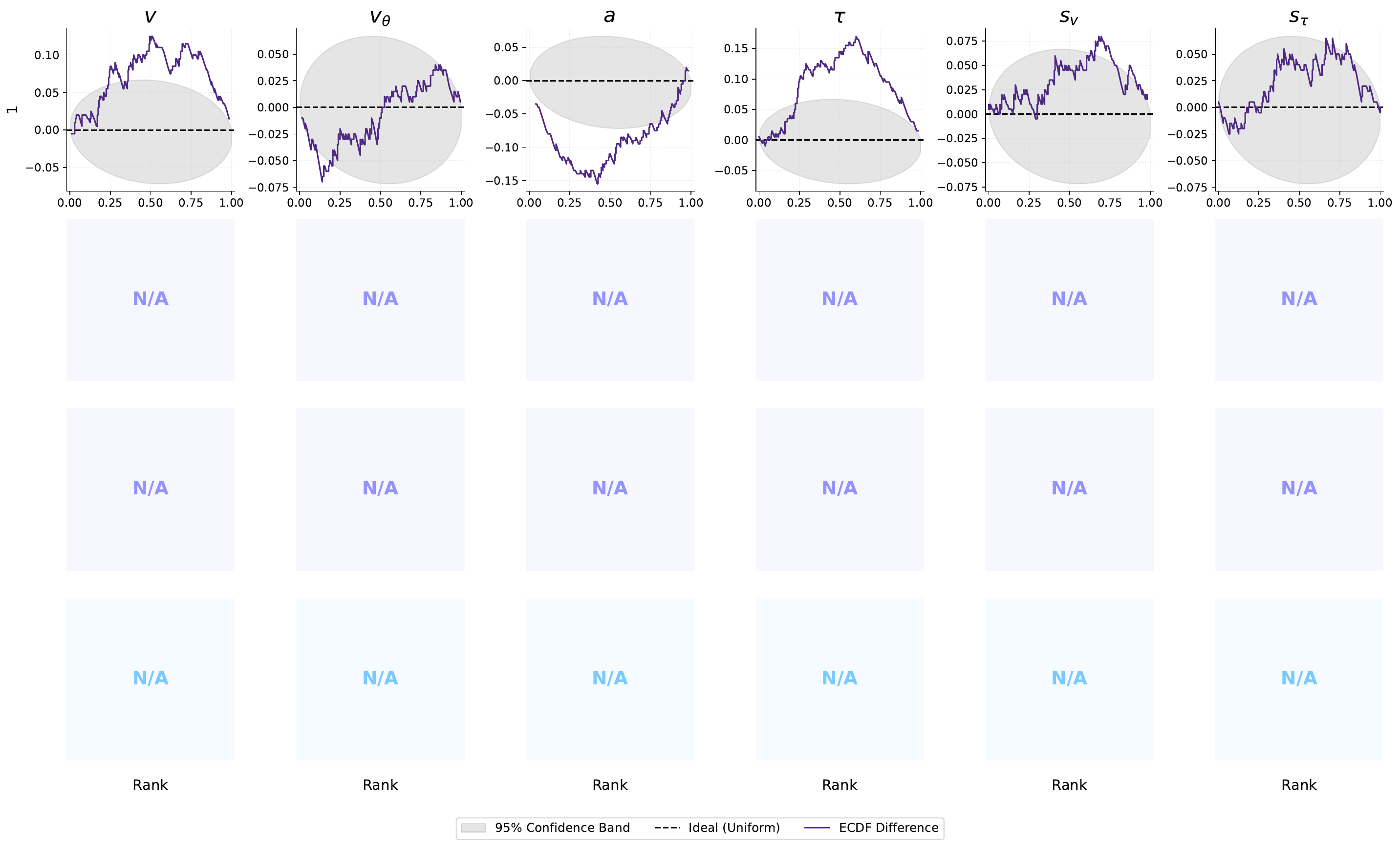}
    \includegraphics[width=0.97\linewidth]{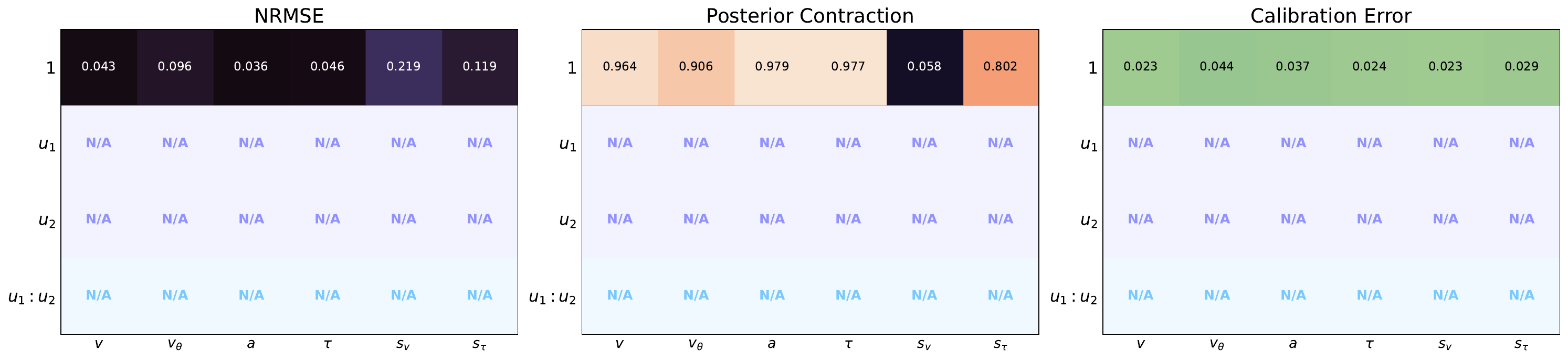}
    \caption{Parameter recovery (\emph{top}), calibration ECDF (\emph{middle}), and parameter-wise metrics (NRMSE, calibration error, and posterior contraction) for CDM model class (Case \textbf{intercept\_only}).}
    \label{fig:cdm-mc-intercept-only}
\end{figure}

\begin{figure}
    \centering
    \includegraphics[width=0.97\linewidth]{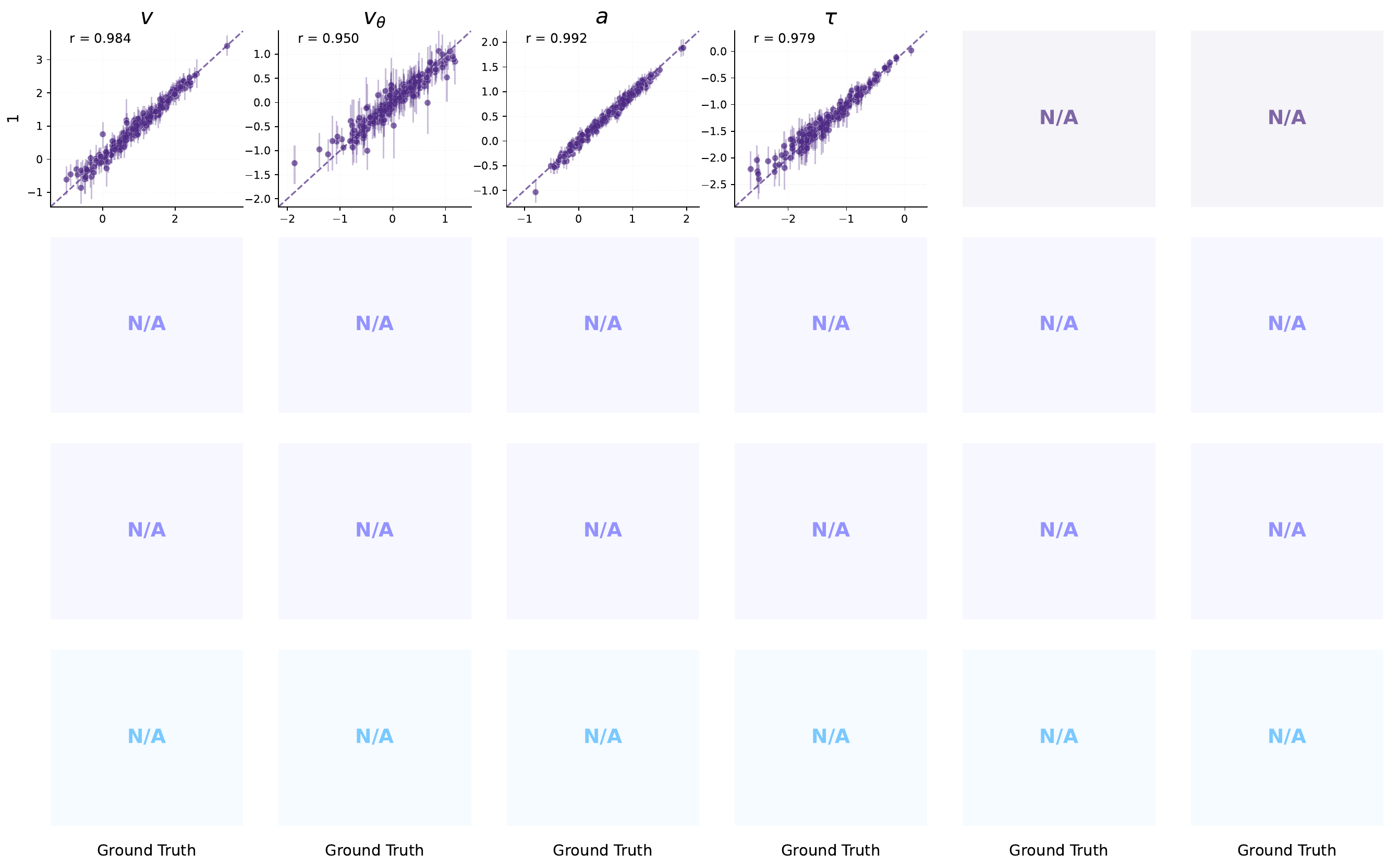}
    \includegraphics[width=0.97\linewidth]{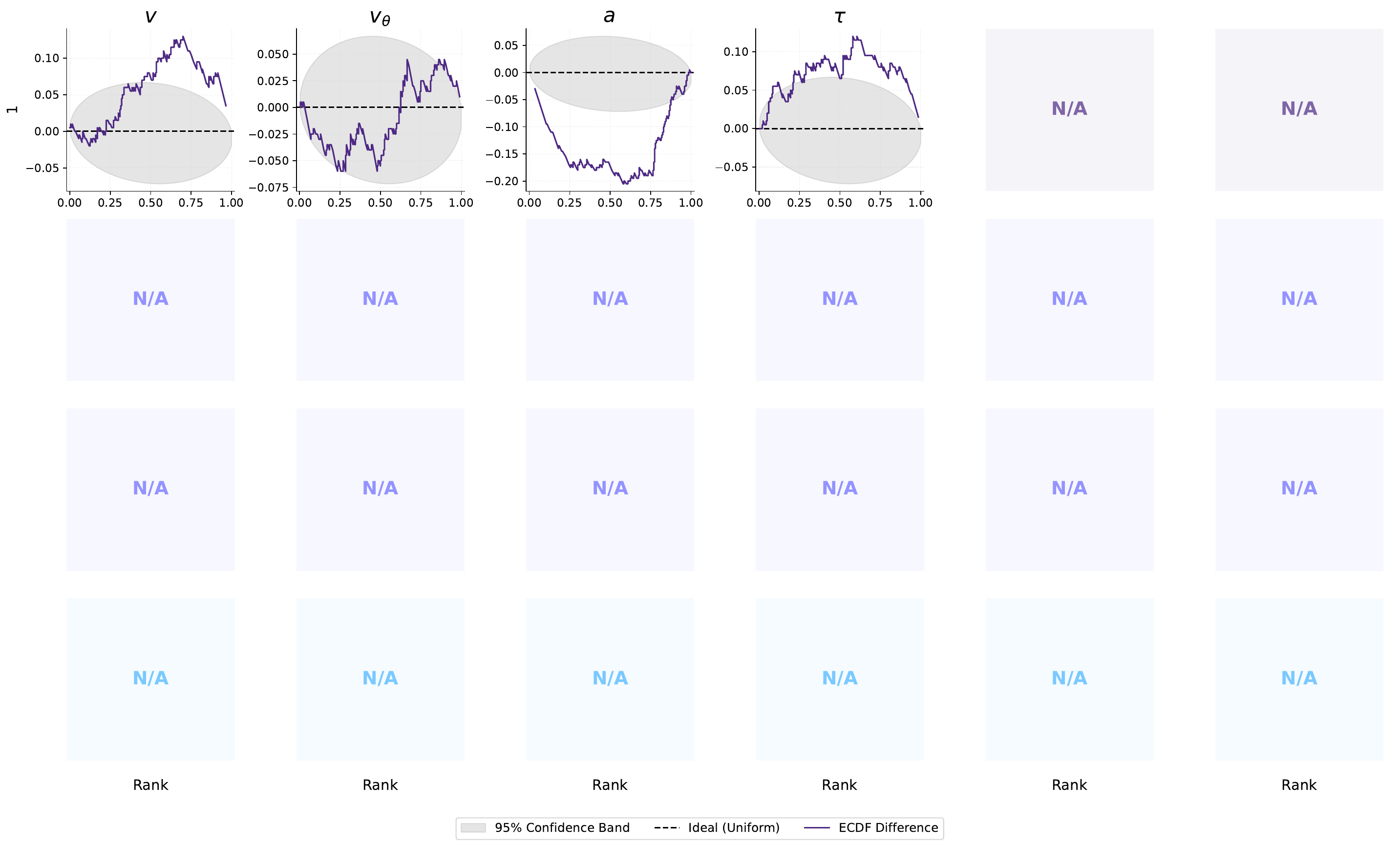}
    \includegraphics[width=0.97\linewidth]{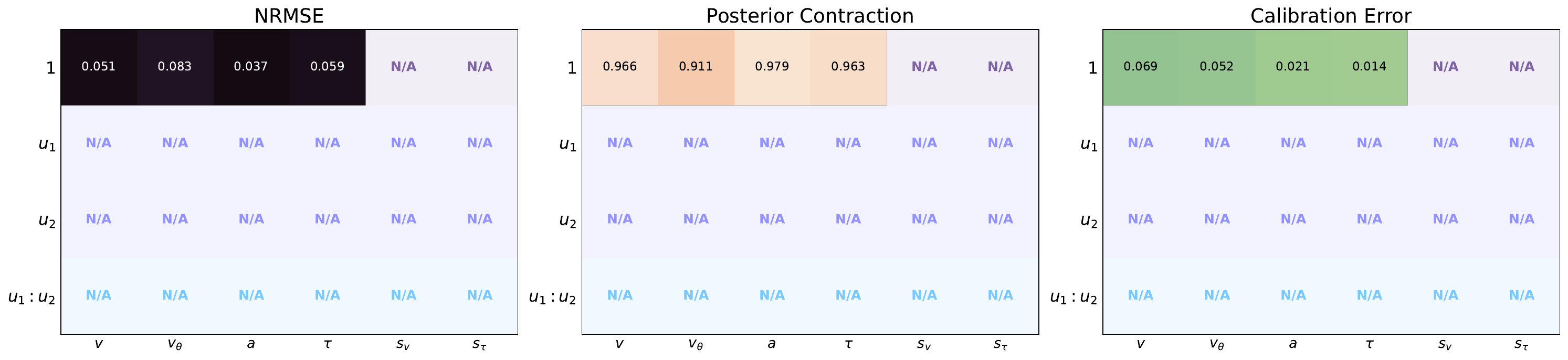}
    \caption{Parameter recovery (\emph{top}), calibration ECDF (\emph{middle}), and parameter-wise metrics (NRMSE, calibration error, and posterior contraction) for CDM model class (Case \textbf{fixed}).}
    \label{fig:cdm-mc-fixed}
\end{figure}

\begin{figure}
    \centering
    \includegraphics[width=0.97\linewidth]{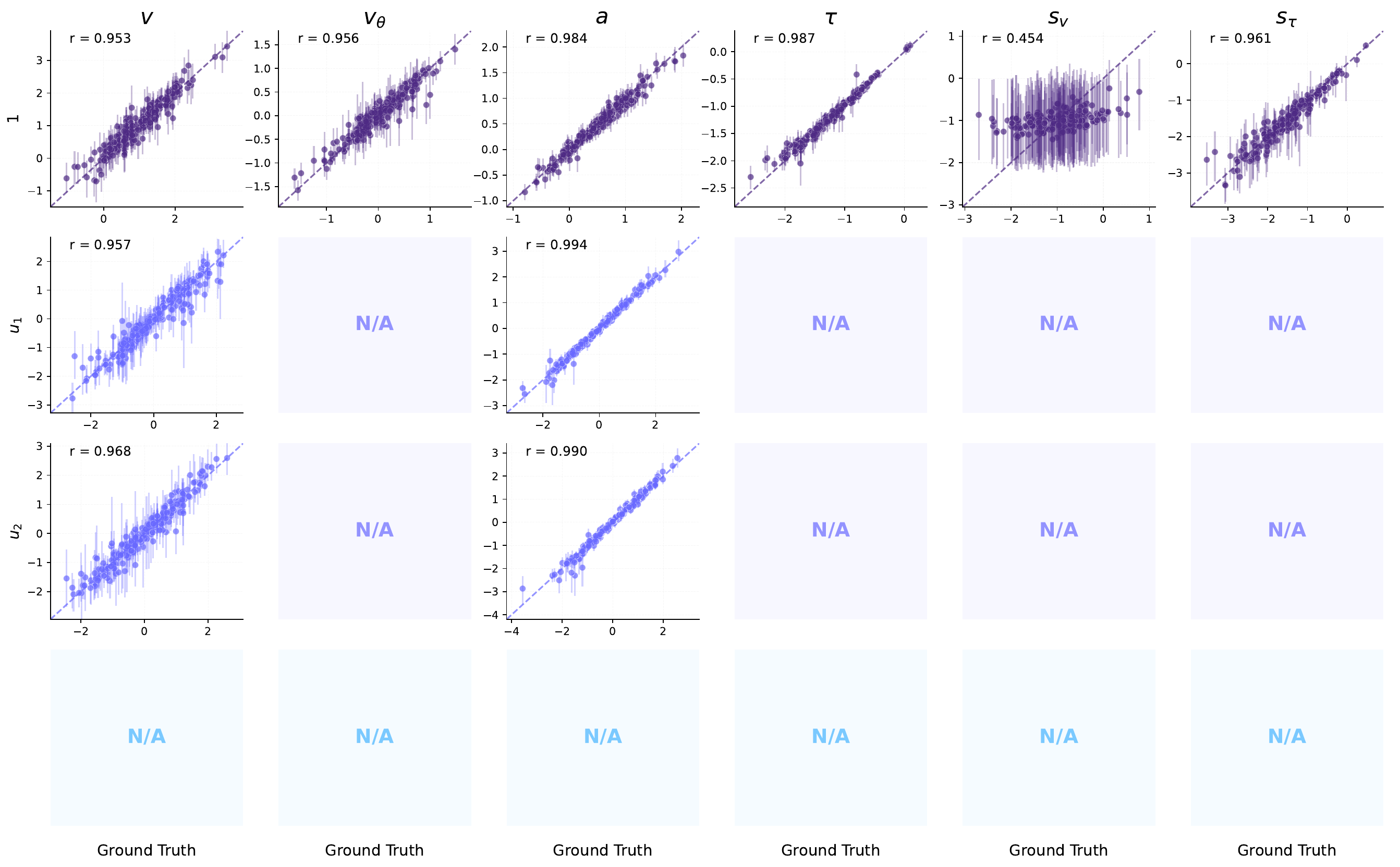}
    \includegraphics[width=0.97\linewidth]{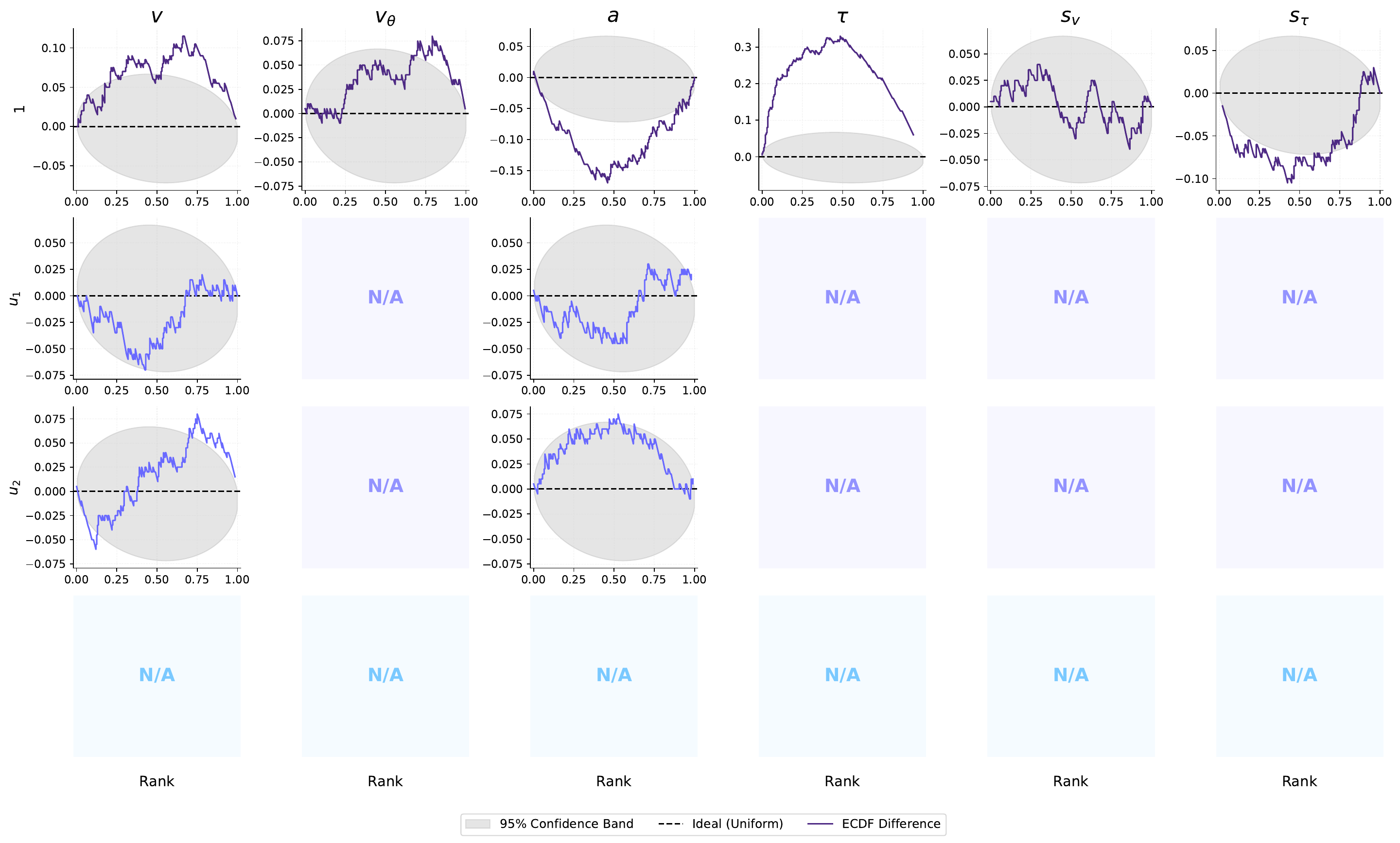}
    \includegraphics[width=0.97\linewidth]{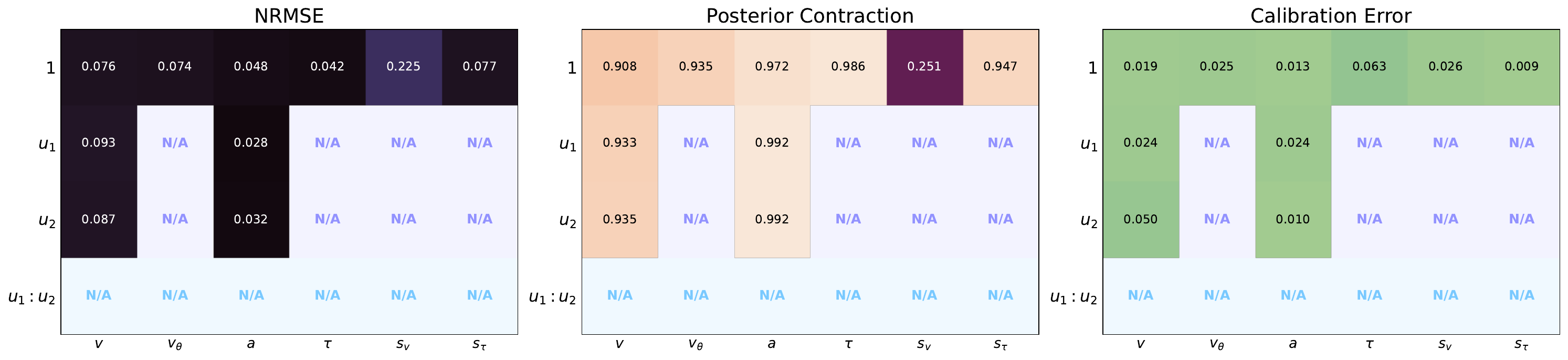}
    \caption{Parameter recovery (\emph{top}), calibration ECDF (\emph{middle}), and parameter-wise metrics (NRMSE, calibration error, and posterior contraction) for CDM model class (Case \textbf{regressed}).}
    \label{fig:cdm-mc-regressed}
\end{figure}

\begin{figure}
    \centering
    \includegraphics[width=0.97\linewidth]{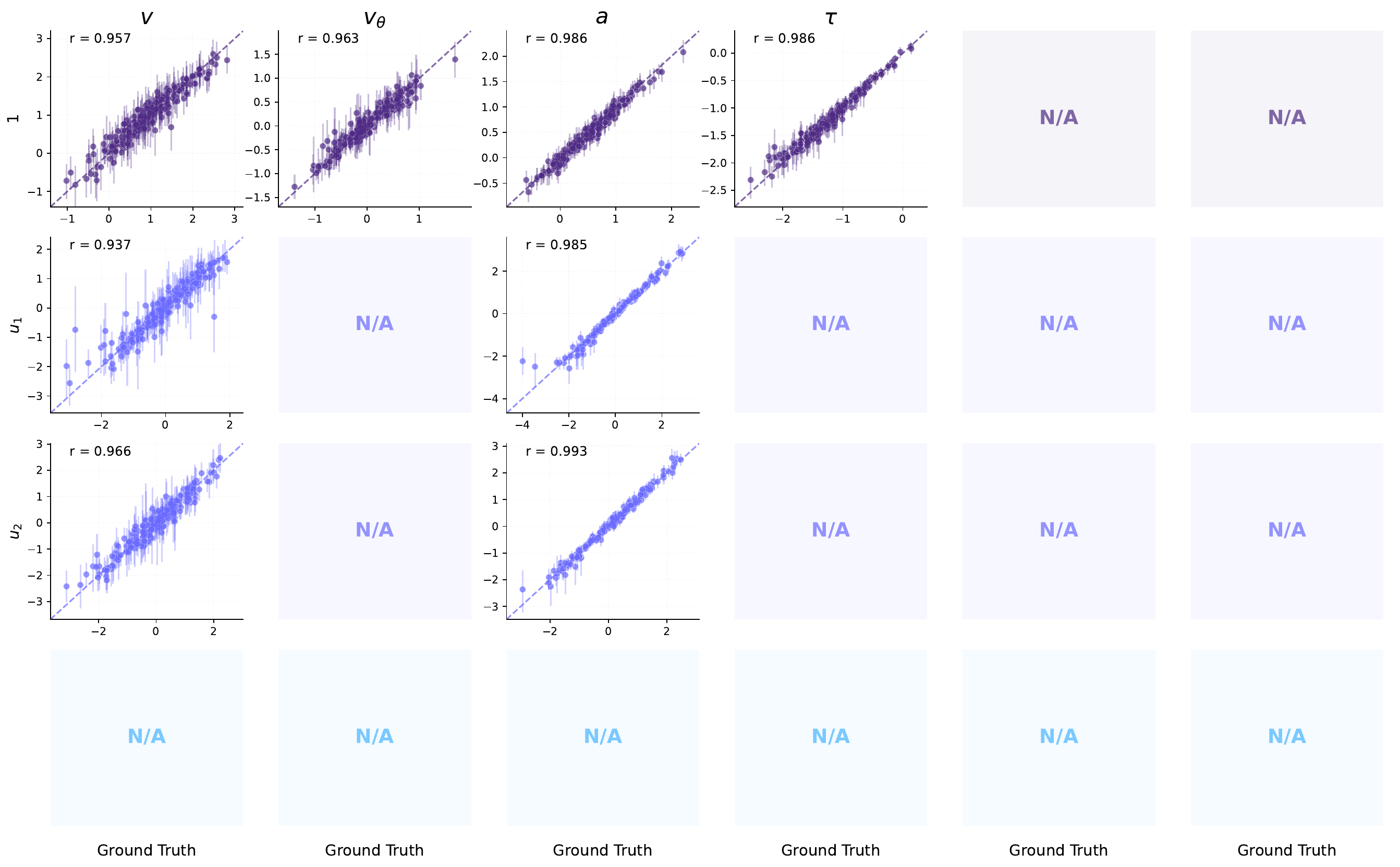}
    \includegraphics[width=0.97\linewidth]{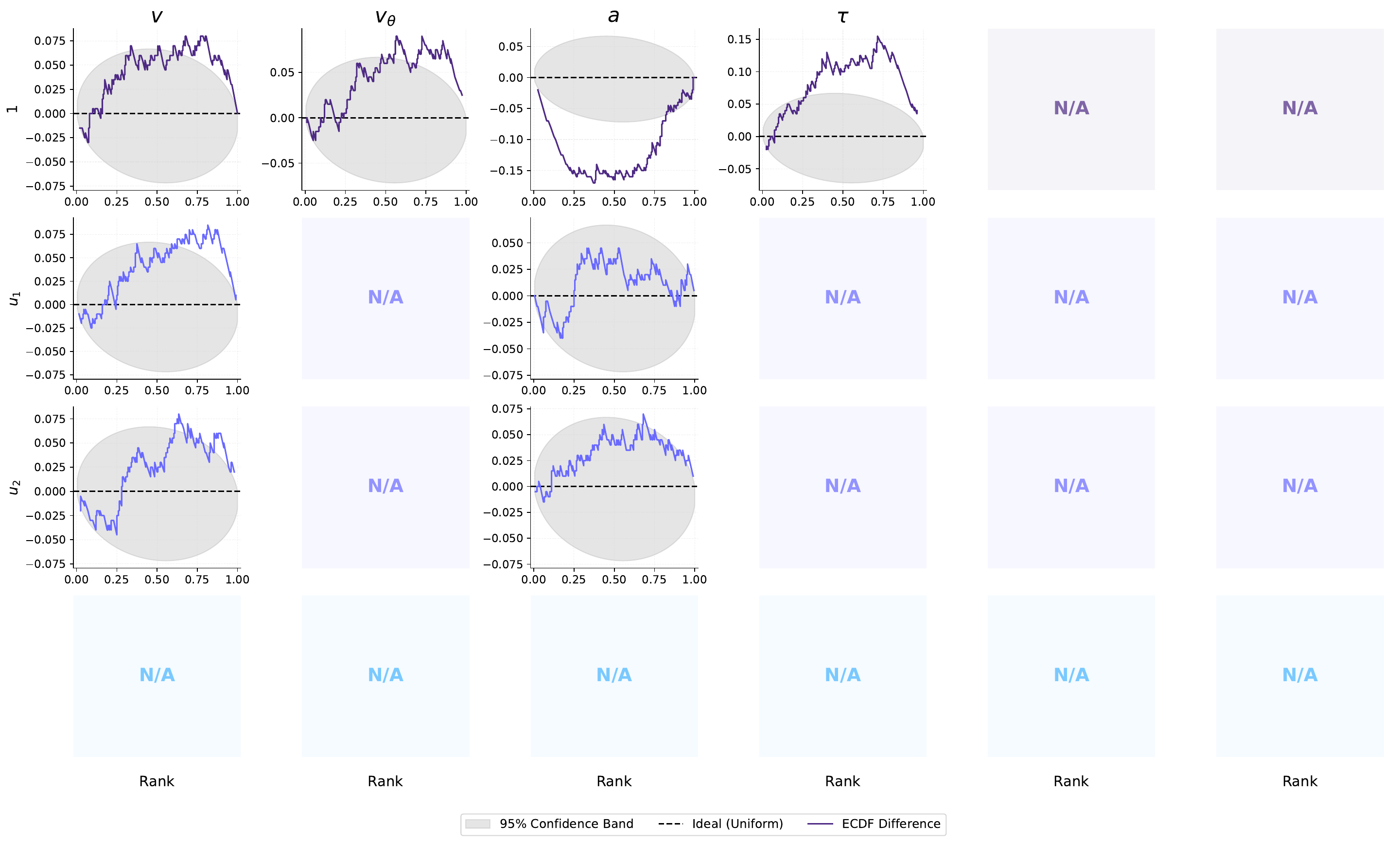}
    \includegraphics[width=0.97\linewidth]{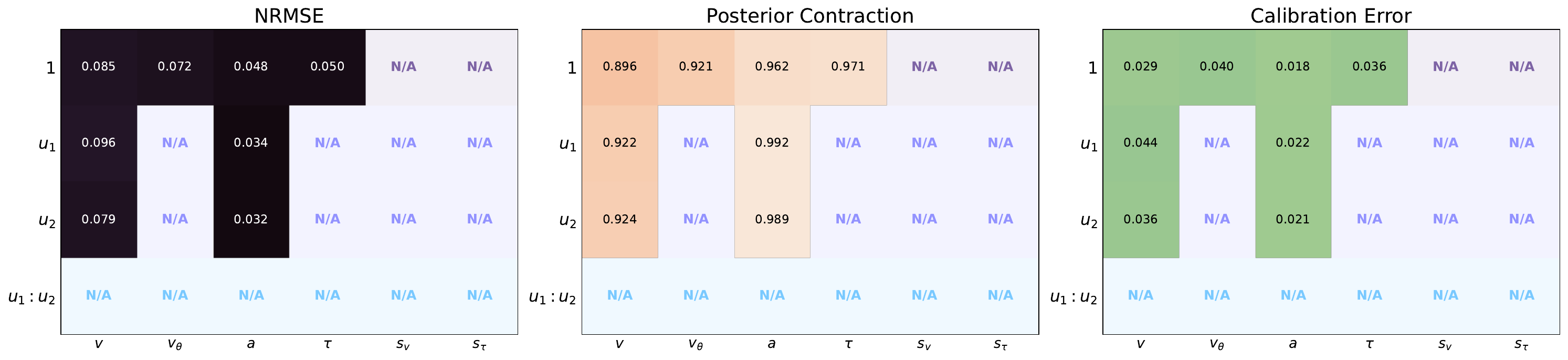}
    \caption{Parameter recovery (\emph{top}), calibration ECDF (\emph{middle}), and parameter-wise metrics (NRMSE, calibration error, and posterior contraction) for CDM model class (Case \textbf{fixed\_regressed}).}
    \label{fig:cdm-mc-fixed-regressed}
\end{figure}

\begin{figure}
    \centering
    \includegraphics[width=0.97\linewidth]{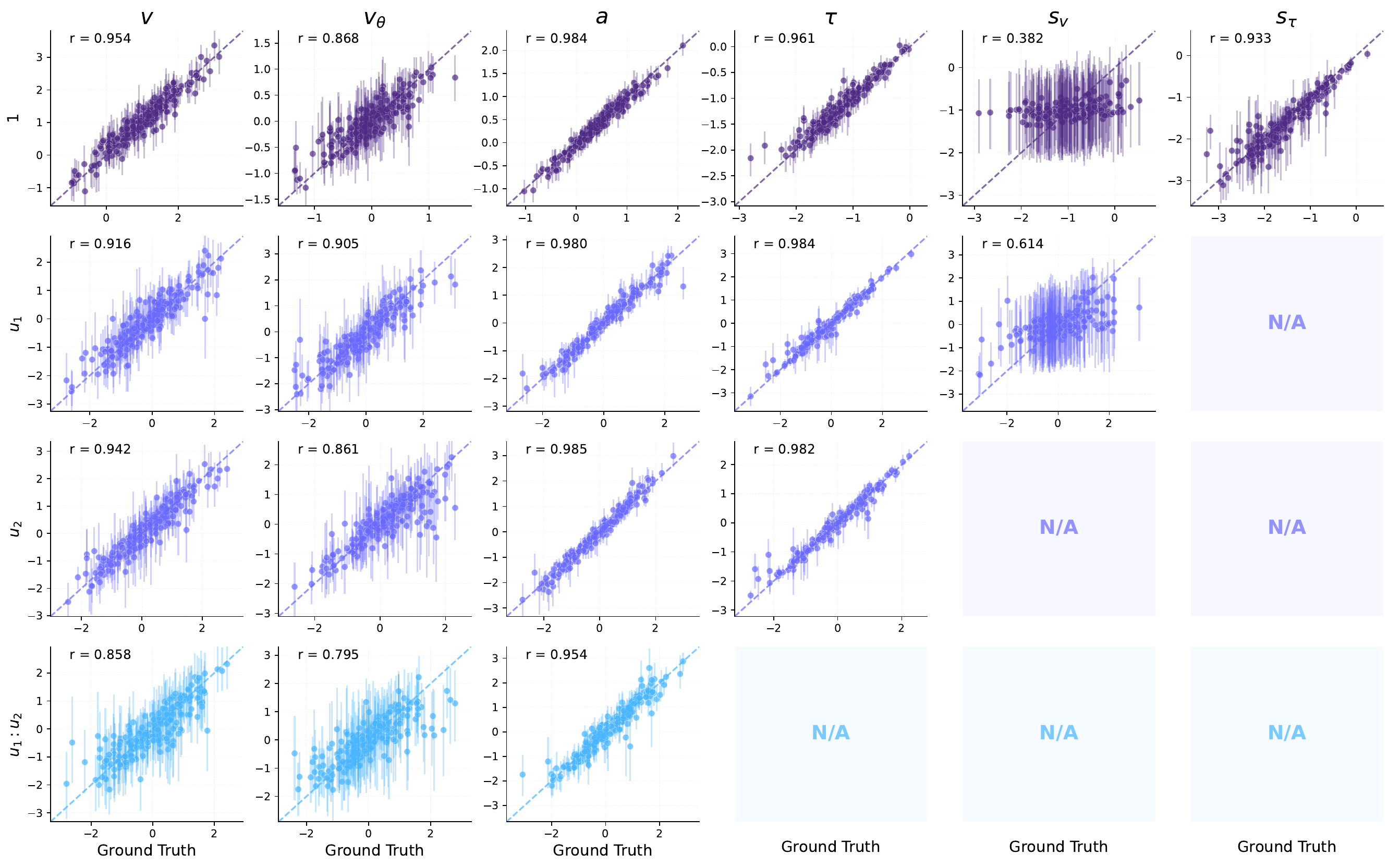}
    \includegraphics[width=0.97\linewidth]{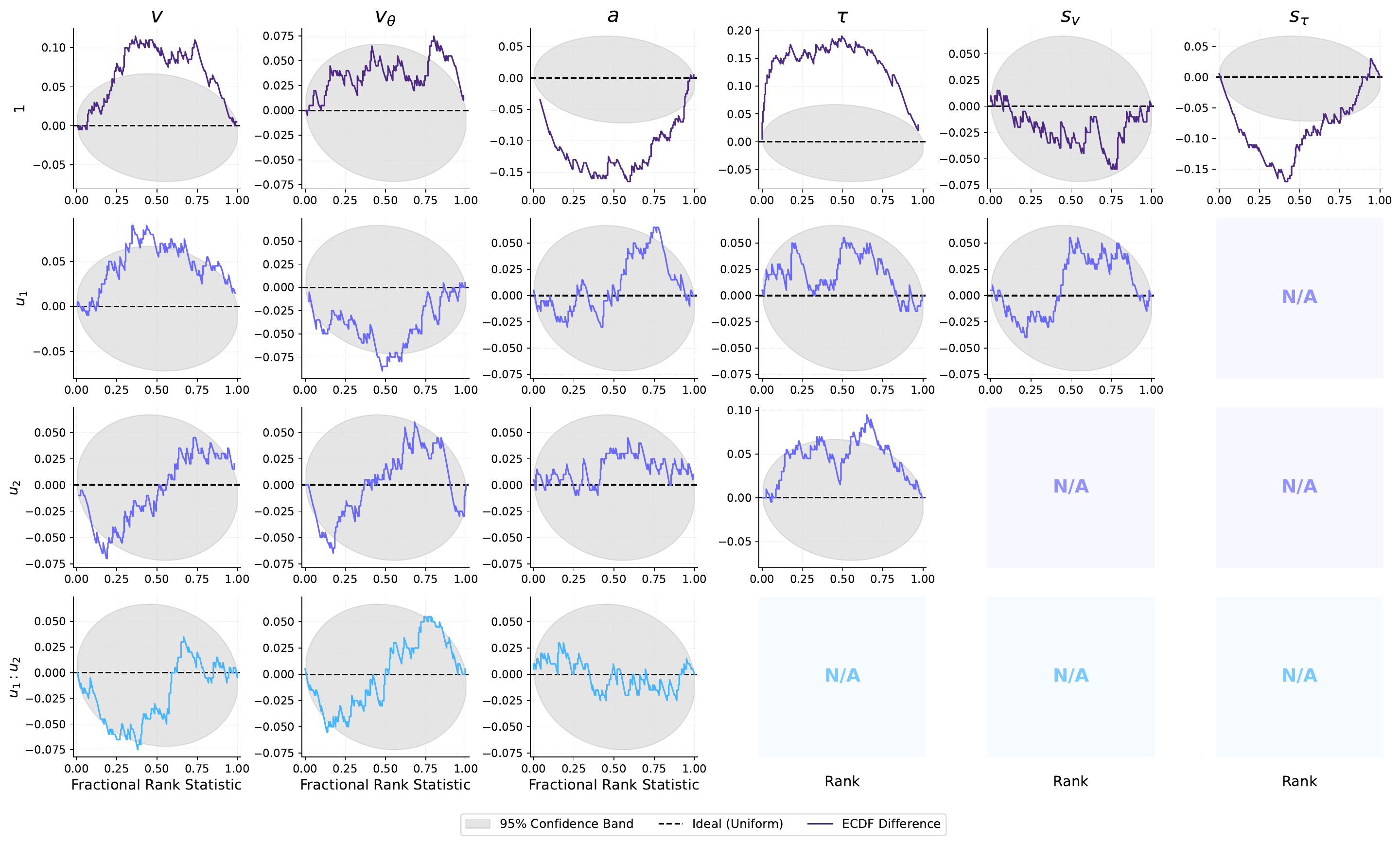}
    \includegraphics[width=0.97\linewidth]{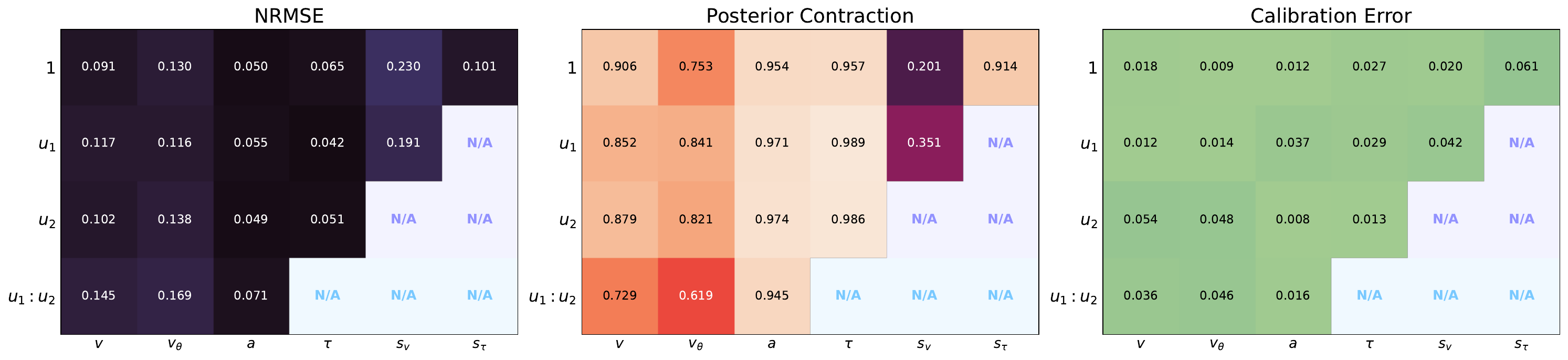}
    \caption{Parameter recovery (\emph{top}), calibration ECDF (\emph{middle}), and parameter-wise metrics (NRMSE, calibration error, and posterior contraction) for CDM model class (Case \textbf{interaction}).}
    \label{fig:cdm-mc-interaction}
\end{figure}

\end{document}